\pdfoutput=1
\documentclass[acmtog]{acmart}

\newcommand{\myparagraph}[1]{\paragraph{#1}}

\usepackage{booktabs}
\usepackage{float}
\usepackage{multirow}

\usepackage[ruled]{algorithm2e}

\SetAlFnt{\small}
\SetAlCapFnt{\small}
\SetAlCapNameFnt{\small}
\SetAlCapHSkip{0pt}

\usepackage{hyperref}
\usepackage{xcolor}
\usepackage[capitalize]{cleveref}
\Crefname{section}{Section}{Sections}
\crefname{section}{Sec.}{Secs.}
\Crefname{appendix}{Appendix}{Appendices}
\crefname{appendix}{App.}{Apps.}
\Crefname{table}{Table}{Tables}
\crefname{table}{Tab.}{Tabs.}
\Crefname{figure}{Figure}{Figures}
\crefname{figure}{Fig.}{Figs.}
\Crefname{equation}{Eq.}{Eqs.}
\crefname{equation}{}{}

\usepackage{pifont}

\newcommand{\losst}[1]{\mathcal{L}_\text{#1}}
\newcommand{\ray}{\mathbf{r}}

\newcommand{\slot}{\mathcal{S}}
\newcommand{\windowL}{t_l}
\newcommand{\windowR}{t_r}
\newcommand{\windowM}{t_m}
\newcommand{\temp}{\tau}
\newcommand{\tempk}{\temp_k}
\newcommand{\numslots}{K}

\newcommand{\mean}{\mu}
\newcommand{\cov}{\Sigma}
\newcommand{\opacity}{\alpha}
\newcommand{\radiance}{\mathbf{c}}
\newcommand{\logits}{\ell}
\newcommand{\stateprob}{\mathbf{p}}
\newcommand{\visibility}{\omega}
\newcommand{\transmittance}{T}

\usepackage{colortbl}
\usepackage{graphicx,array}
\usepackage{tabularx}

\definecolor{oursbg}{RGB}{242,246,250}
\definecolor{bestcolor}{RGB}{0,76,153}
\definecolor{secondcolor}{RGB}{90,130,170}

\newlength{\scenecolwidth}
\newcolumntype{S}{>{\centering\arraybackslash}m{\scenecolwidth}}
\newcolumntype{Z}{>{\centering\arraybackslash}X}
\newcommand{\tabbest}[1]{\textbf{#1}}

\definecolor{abgreen}{RGB}{35,135,70}
\definecolor{abred}{RGB}{255,0,0}

\newcommand{\cmark}{\textcolor{abgreen}{\ding{51}}}
\newcommand{\xmark}{\textcolor{abred}{\ding{55}}}

\usepackage[most]{tcolorbox}
\usepackage{listings}

\lstdefinestyle{vlmprompt}{
  basicstyle=\ttfamily\fontsize{6.4pt}{7.2pt}\selectfont,
  columns=fullflexible,
  keepspaces=true,
  showstringspaces=false,
  breaklines=true,
  breakatwhitespace=true,
  tabsize=2,
  aboveskip=0pt,
  belowskip=0pt,
}
 
\acmJournal{TOG}

\copyrightyear{2026}
\acmYear{2026}
\setcopyright{cc}
\setcctype{by}
\acmConference[SA Conference Papers '26]
  {SIGGRAPH Asia 2026 Conference Papers}
  {December 01--04, 2026}
  {Kuala Lumpur, Malaysia}
\acmBooktitle{SIGGRAPH Asia 2026 Conference Papers
  (SA Conference Papers '26), December 01--04, 2026,
  Kuala Lumpur, Malaysia}
\acmDOI{10.1145/3829340.3842325}
\acmISBN{979-8-4007-2842-6/2026/12}

\begin{document}
\title{SwitchSplat: Learning Discrete Object States from Casually Captured Scenes}

\author{Yizheng Xie}
\orcid{0009-0003-8670-6003}
\authornote{These authors contributed equally to this work.}
\authornote{Corresponding author.}
\affiliation{\institution{Simon Fraser University}
  \city{Burnaby}
  \country{Canada}}
\email{yizheng\_xie@sfu.ca}

\author{Yalda Foroutan}
\orcid{0009-0007-3462-2405}
\authornotemark[1]
\affiliation{\institution{Simon Fraser University}
  \city{Burnaby}
  \country{Canada}}
\email{yalda\_foroutan@sfu.ca}

\author{George Shramko}
\orcid{0009-0001-4664-4368}
\affiliation{\institution{Simon Fraser University}
  \city{Burnaby}
  \country{Canada}}
\email{george\_shramko@sfu.ca}

\author{Daniel Rebain}
\orcid{0000-0003-4691-7909}
\affiliation{\institution{University of British Columbia}
  \city{Vancouver}
  \country{Canada}}
\affiliation{\institution{Wayve Labs}
  \city{Vancouver}
  \country{Canada}}
\email{danielrebain08@gmail.com}

\author{Shrisudhan Govindarajan}
\orcid{0000-0002-3546-8223}
\authornote{These authors contributed equally to advising this work.}
\affiliation{\institution{Simon Fraser University}
  \city{Burnaby}
  \country{Canada}}
\affiliation{\institution{Wayve Labs}
  \city{Vancouver}
  \country{Canada}}
\email{shrisudhan\_govindarajan@sfu.ca}

\author{Andrea Tagliasacchi}
\orcid{0000-0002-2209-7187}
\authornotemark[3]
\affiliation{\institution{Simon Fraser University}
  \city{Burnaby}
  \country{Canada}}
\affiliation{\institution{University of Toronto}
  \city{Toronto}
  \country{Canada}}
\affiliation{\institution{Wayve Labs}
  \city{Vancouver}
  \country{Canada}}
\email{taiya+sigpc@sfu.ca}

\renewcommand{\shortauthors}{Xie, Foroutan, Shramko, Rebain, Govindarajan, and Tagliasacchi}

\begin{abstract}
3D Gaussian Splatting has made high-fidelity 3D reconstruction fast and accessible, yet existing representations assume a static world.
This assumption breaks down in casual captures, where a user naturally interacts with the environment -- opening doors, toggling switches, or moving objects -- introducing discrete state changes that no current framework can model.
Time-conditioned dynamic methods such as 4DGS conflate \emph{when} a state was observed with \emph{what} state the scene is in, limiting them to replaying seen sequences rather than composing arbitrary object configurations.
We present SwitchSplat, the first 3DGS representation to address this gap via a base$+$slot decomposition: a shared static base captures the invariant scene, while each slot independently holds the two binary states (on/off) of one interactable object.
Because the assignment of Gaussians to slots is fully differentiable, the model discovers the geometry of each state directly from video, without manual annotation.
At inference, the user can freely combine slot states and render the resulting scene from any viewpoint.
We validate SwitchSplat on a new casual-capture dataset and show that it outperforms static and dynamic baselines in novel-view synthesis quality, while uniquely supporting rendering of state combinations never seen during capture.
Project page: \url{https://switchsplat.github.io}.
\end{abstract} 


\begin{teaserfigure}
    \hfill
    \begingroup
    \setlength{\unitlength}{.0098\textwidth}
    \begin{picture}(100,31)
        \put(0,0){\makebox(0,0)[bl]{\includegraphics[width=.98\textwidth]{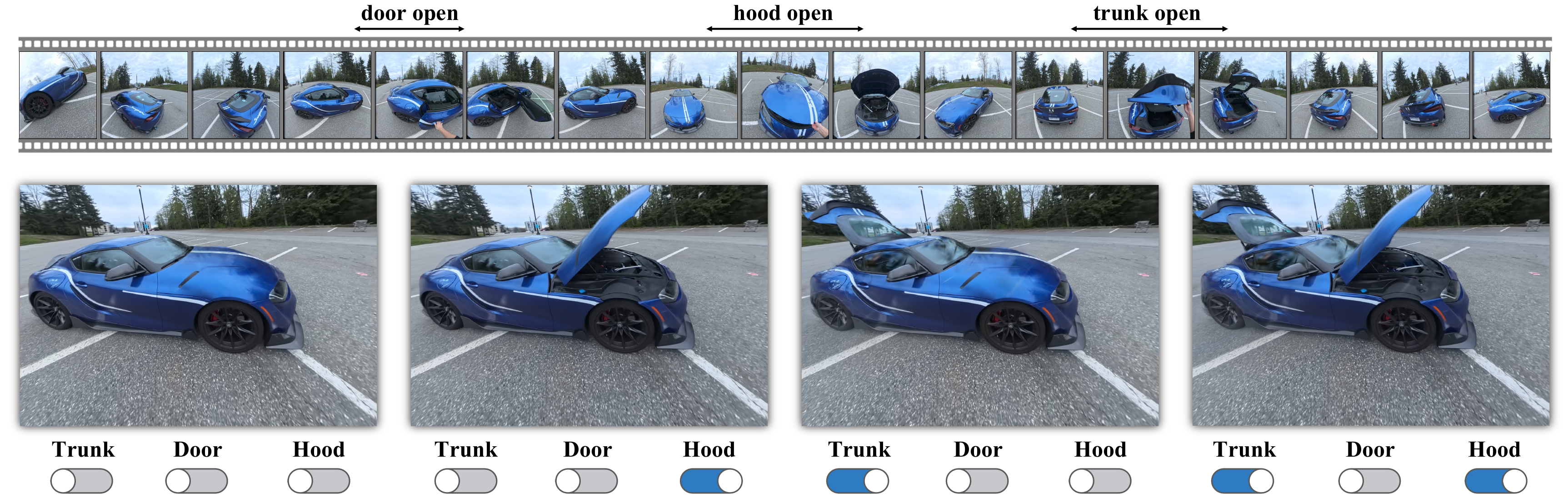}}}
\put(-.5, 25.5){\rotatebox{90}{\makebox[0pt][c]{\small Video}}}
        \put(-.5, 12.5){\rotatebox{90}{\makebox[0pt][c]{\small Novel View Synthesis}}}
    \end{picture}
    \endgroup
    \caption{\textbf{SwitchSplat} learns a state-aware scene representation from a casual capture (\emph{top}: input video frames showing the user interacting with the car).
        The scene is decomposed into a static base and a set of independently switchable slots, one per interactable object (here: right door, hood, trunk).
        Each slot models two binary states, enabling SwitchSplat to render any combination of object configurations from novel viewpoints at inference time (\emph{bottom}: four of the \(2^3=8\) possible slot configurations, labeled by their on/off state vector).
}
    \Description{The top row is a filmstrip of a camera moving around a blue car while a person successively opens its side door, hood, and trunk. The bottom row contains four novel-view renderings of the car, each paired with labeled switches for the trunk, door, and hood. The examples show all three parts closed, only the hood open, only the trunk open, and both the trunk and hood open, illustrating independent composition of the three binary object states.}
\label{fig:teaser}
\end{teaserfigure}
 \maketitle
\section{Introduction}
\label{sec:intro}

In inverse rendering, Gaussian Splatting (3DGS)~\cite{3dgs} has emerged as the leading representation for real-time, high-fidelity novel-view synthesis, enabling a new generation of interactive applications.
Its combination of fast training and efficient rendering has driven broad adoption across academia as well as industry, from Matterport's digital twins and Niantic Scaniverse's mobile reconstruction to Meta's Hyperscape\footnote{\url{https://www.meta.com/experiences/meta-horizon-hyperscape-capture-beta/8798130056953686/}}.

While early 3DGS pipelines required meticulous data acquisition~(fixed exposure and white balance, controlled lighting, and a perfectly static scene), recent works like 3DGRT~\cite{3dgrt} and FullCircle~\cite{fullcircle} have unlocked \emph{casual} capture: viewpoints need not be carefully curated -- made possible by $360^{\circ}$ cameras held in a single hand -- and the capturer's pixels are automatically masked out to satisfy the static-scene assumption.
This freedom naturally invites users to \emph{interact} with the environment during capture -- opening doors, toggling switches, or repositioning objects -- turning the operator into both an active agent \textit{and} a dynamic distractor within the scene.

Yet no existing reconstruction framework is equipped to handle these interactions.
Many contemporary 3D representations enforce a strict static-world assumption: when an operator triggers a discrete state change -- opening a door, flipping a switch, relocating an object -- the model has no mechanism to account for the resulting geometric variation.
As a result, even state-of-the-art pipelines can only represent a single, frozen scene configuration, and capturing multiple stable states requires re-recording the scene from scratch.

One could argue that the natural candidate to address this limitation is the family of 4D Gaussian Splatting methods~\cite{4dgs}, which extend 3DGS to dynamic scenes by conditioning the scene representation on time.
However, this design choice fundamentally conflates \emph{when} a state was observed with \emph{what} state the scene is in: the model learns to replay a specific sequence of observations, not to understand independent object configurations.

As a result, 4D methods are largely limited to novel-viewpoint synthesis of seen sequences; composing an unseen combination of object states---e.g., door open and light on---is not directly supported, since no single timestamp in the training data corresponds to that configuration.
What is needed instead is a representation that \emph{de-correlates} scene state from time, allowing the user to freely specify both the viewpoint and the desired configuration of each object independently.

To address these challenges, we propose \texttt{SwitchSplat}, a method that learns ``state-aware'' representations from casual captures featuring human-object interactions; see \Cref{fig:teaser}.
SwitchSplat organizes the scene into a \emph{static base} -- the parts that never change throughout the capture -- and a set of \emph{slots}, one per interactable object.
Each slot holds two 3D representations corresponding to the binary states of that object (``on'' and ``off'').
For example, if a room was captured while a closet door was closed for part of the recording and open for another, a single slot would hold both the ``off'' state (door closed, interior hidden) and the ``on'' state (door open, interior visible), while the walls, floor, and furniture -- unchanged throughout -- are handled by the static base.
To assign each Gaussian primitive to the right part of the scene, we attach a small learnable score vector that controls how strongly the primitive belongs to the base, or to either state of each slot.

This decomposition avoids data redundancy: without a shared static base, one would need to duplicate the full scene geometry for every slot, with no reusable structure across states.
Instead, SwitchSplat shares the base across all configurations and only stores the geometry that actually differs per state.\footnote{This is structurally analogous to PCA-based morphable models~\cite{blanz2023morphable}: the static base plays the role of the mean face, each slot's on/off representations correspond to the eigenvectors, and the state activations act as the blend-shape coefficients.}
Because the full pipeline is differentiable end-to-end, the model learns simultaneously which Gaussians belong to which slot and when each slot is active -- directly from the input video, without any manual annotation.
At inference, the user can query any combination of slot states independently -- e.g., closet door open and desk lamp on -- and render the resulting scene from any viewpoint.

To summarize our contributions:
(1) We \underline{\smash[b]{identify a novel prob}}\-\underline{\smash[b]{lem}} in the 3DGS literature: existing representations cannot model scenes with discrete, user-triggered state changes (e.g., objects that can be opened, switched, or moved), and we show that na\"{i}vely applying dynamic methods such as 4DGS fails to address this.
(2) We propose SwitchSplat, the first 3DGS representation to address this problem via a \underline{\smash[b]{base$+$slot decomposition}}, where a shared static base captures the invariant scene and each slot independently models the binary states of one interactable object.
(3) We demonstrate on a \underline{\smash[b]{new casual-capture dataset}} that SwitchSplat improves novel-view synthesis quality over static and dynamic baselines, and uniquely supports rendering arbitrary combinations of object states from unseen viewpoints.
 \section{Related Work}
\label{sec:related}

3D Gaussian Splatting (3DGS)~\cite{3dgs} has become the dominant representation for novel-view synthesis, optimizing a set of anisotropic Gaussian primitives that render at real-time rates.
Extensions beyond the standard perspective-camera model are particularly relevant to our setting: 3DGRT~\cite{3dgrt} reformulates rendering via ray tracing, and 3DGUT~\cite{3dgut} adds support for highly distorted cameras and secondary rays, enabling casual capture with wide field-of-view sensors.
FullCircle~\cite{fullcircle} further shows that raw dual-fisheye $360^{\circ}$ video suffices for high-quality reconstruction, automatically masking the camera operator.
Despite these advances, all of these pipelines represent the scene as a single, fixed configuration.
When multiple object states are present in the capture---a door open in some frames and closed in others---the model either conflates them into averaged geometry and ghosting artifacts, or learns to discard the minority state as a distractor~\cite{sabour2024sls}.
Either way, only one configuration survives in the final model; there is no mechanism to retain, switch between, or compose multiple stable states at inference time.

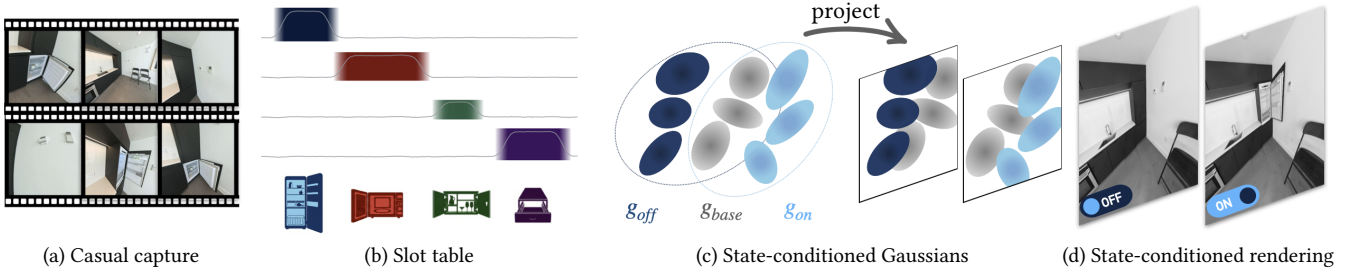
\begin{figure*}[t]
    \centering
    \begingroup
    \setlength{\unitlength}{.01\textwidth}
    \begin{picture}(100,19)
        \put(0,0){\makebox(0,0)[bl]{\includegraphics[width=\textwidth]{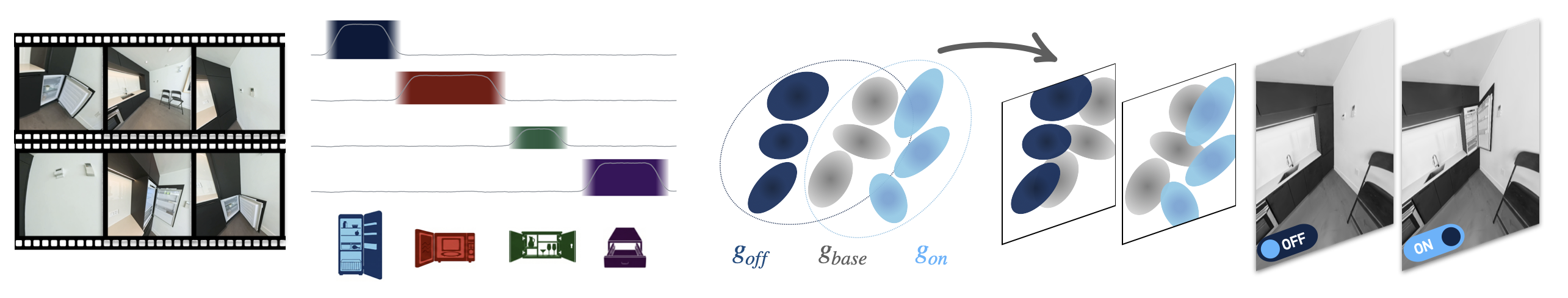}}}
        \put(9.5,  -1){\makebox[0pt][c]{\small (a) Casual capture}}
        \put(31.5, -1){\makebox[0pt][c]{\small (b) Slot table}}
        \put(63, 17){\makebox[0pt][c]{project}}
        \put(62, -1){\makebox[0pt][c]{\small (c) State-conditioned Gaussians}}
        \put(89, -1){\makebox[0pt][c]{\small (d) State-conditioned rendering}}
    \end{picture}
    \endgroup
\caption{\textbf{Pipeline --}
        From a casual capture in which the operator interacts with objects (a), we fit a differentiable \emph{slot table} (b) whose learnable endpoints $(\windowL, \windowR)$ and temperature $\tempk$ define a per-frame interaction probability $\pi_k(t)$ for each object.
        Each 3D Gaussian primitive carries a state-logit vector $\logits$ over a static \emph{base} channel together with an \emph{off} and an \emph{on} channel per slot, yielding per-state probabilities (c).
        At render time (d), these probabilities are fused with $\pi_k(t)$ into a frame-conditional opacity $\tilde{\opacity}(t)$ that drives standard volumetric composition.
        All Gaussians, state logits, and slot-table parameters are optimized end-to-end under photometric losses, producing a state-aware scene that renders arbitrary object configurations from novel viewpoints.
    }
    \Description{From left to right, the diagram shows filmstrips from a casual kitchen capture; four colored temporal slot windows aligned with icons for different interactable objects; Gaussian ellipses separated into dark-blue off, gray base, and light-blue on groups; projected Gaussian layers selected by the requested state; and kitchen renderings with an appliance closed or open. Arrows indicate that the slot probabilities and categorical Gaussian assignments determine state-conditioned opacity and the final rendering.}
    \label{fig:pipeline}
\end{figure*}
 
\myparagraph{Dynamic Gaussian Splatting}
A large family of methods extends 3DGS to dynamic scenes by conditioning the representation on time.
Dynamic 3D Gaussians~\cite{dynamic} tracks persistent primitives across frames, while Deformable 3D Gaussians~\cite{deformable} and 4DGS~\cite{4dgaussiansplatting} deform a canonical scene over time, with 4DGS using a learned space-time encoding.
Spacetime Gaussians~\cite{spacetime} further augments primitives with temporal opacity and parametric motion.
Subsequent work improves efficiency, temporal consistency, or geometric accuracy via depth priors~\cite{4dgs,3dgstream,4drotorgs,st4dgs,modgs}.
Differently from our method, all of these methods are parameterized by time, not by object state: recovering a specific scene configuration requires locating the right moment in the recording, and rendering a novel combination of states---e.g., two objects in configurations never co-observed---is not supported.

\myparagraph{Decomposed Gaussian Splatting}
More recent dynamic Gaussian methods, such as SDD-4DGS~\cite{sdd}, observe that applying deformation models to the entire scene is often inefficient, since most real-world scenes are largely static.
This motivates methods that decompose scenes into static and dynamic components: GauFRe~\cite{gaufre} combines a static Gaussian component with deformable dynamic Gaussians, while Swift4D~\cite{swift4d} improves efficiency by passing only dynamic Gaussians through the deformation model.
DeGauss~\cite{degauss} further uses dynamic-static decomposition to improve robustness to transient distractors.
This observation is closely aligned with our setting, since object interactions usually affect only a small part of a $360^{\circ}$ environment.
In contrast to our method, existing decomposition methods still treat the dynamic component as a temporally varying foreground: they separate dynamic regions from a static background, but provide no mechanism to independently switch or compose object states at inference time.

\myparagraph{Controllable Scene Representation}
A parallel line of work studies controllable and articulated 3D representations.
Pre-Gaussian methods such as Watch It Move~\cite{watchitmove} discover articulated structure from object motion; PARIS~\cite{paris} goes further by decoupling static and movable parts from two multi-view captures of the same scene.
Gaussian-based approaches extend this to photorealistic settings: SC-GS~\cite{scgs} enables motion editing via sparse control points, Template-free Articulated Gaussian Splatting~\cite{template} supports re-posing with super-points, and RigGS~\cite{riggs} adds rig-based editing and interpolation.
ArtGS~\cite{artgs} and ArticulatedGS~\cite{articulatedgs} further focus on part-level modeling of individual objects.
Crucially, all of these methods aim to \emph{encode} object motion or articulation---they model how parts move, rather than discarding motion entirely to focus on the discrete stable configurations of the scene.
In contrast, our method treats object interactions as a nuisance to be factored out, retaining only the clean on/off states and ignoring the transient motion between them.

\myparagraph{State-aware Scene Reconstruction}
Closest to our setting are methods that reconstruct scenes across discrete object states.
IGFuse~\cite{igfuse} and RecurGS~\cite{recurgs} address this by using semantic cues to detect state changes and aligning them via rigid transformations.
GaussianCut~\cite{gaussiancut} and DecoupledGaussian~\cite{decoupledgaussian} contribute related tools for object segmentation and scene decomposition.
However, all of these methods assume that state changes can be described as rigid motions of segmented object entities---an assumption that breaks down for casual capture, where interaction boundaries are fuzzy and deformations are non-rigid.
In contrast, SwitchSplat learns which Gaussians belong to each state directly from photometric supervision, without requiring object masks, or rigid-body assumptions.
 \section{Method}
SwitchSplat learns a state-aware 3D representation from casual captures by decomposing the scene into controllable, state-conditioned components.
In the following subsections, we detail our methodology by first formulating our state-conditioned Gaussian primitives and describing their integration within volume rendering~(\cref{sec:rendering}). 
We then introduce the parameterization of the differentiable slot-table~(\cref{sec:slot-table}), followed by our slot initialization procedure~(\cref{sec:init}). 
Finally, we describe our training pipeline and objectives that facilitate granular, state-controlled visualization during inference~(\cref{sec:optim,sec:train_obj}).

\myparagraph{Terminology} 
A \textit{slot} corresponds to one interactable object and holds two 3D representations for its binary states: ``off'' (e.g., door closed) and ``on'' (e.g., door open).
A \textit{slot activation} is a per-frame binary variable indicating which state is currently observed.
All parts of the scene that remain constant throughout the capture are handled by the \textit{static base}, which is shared across all configurations.

\subsection{State-aware Gaussian Splat}
\label{sec:rendering}

\begin{figure}[t]
    \centering

\includegraphics[width=\linewidth]{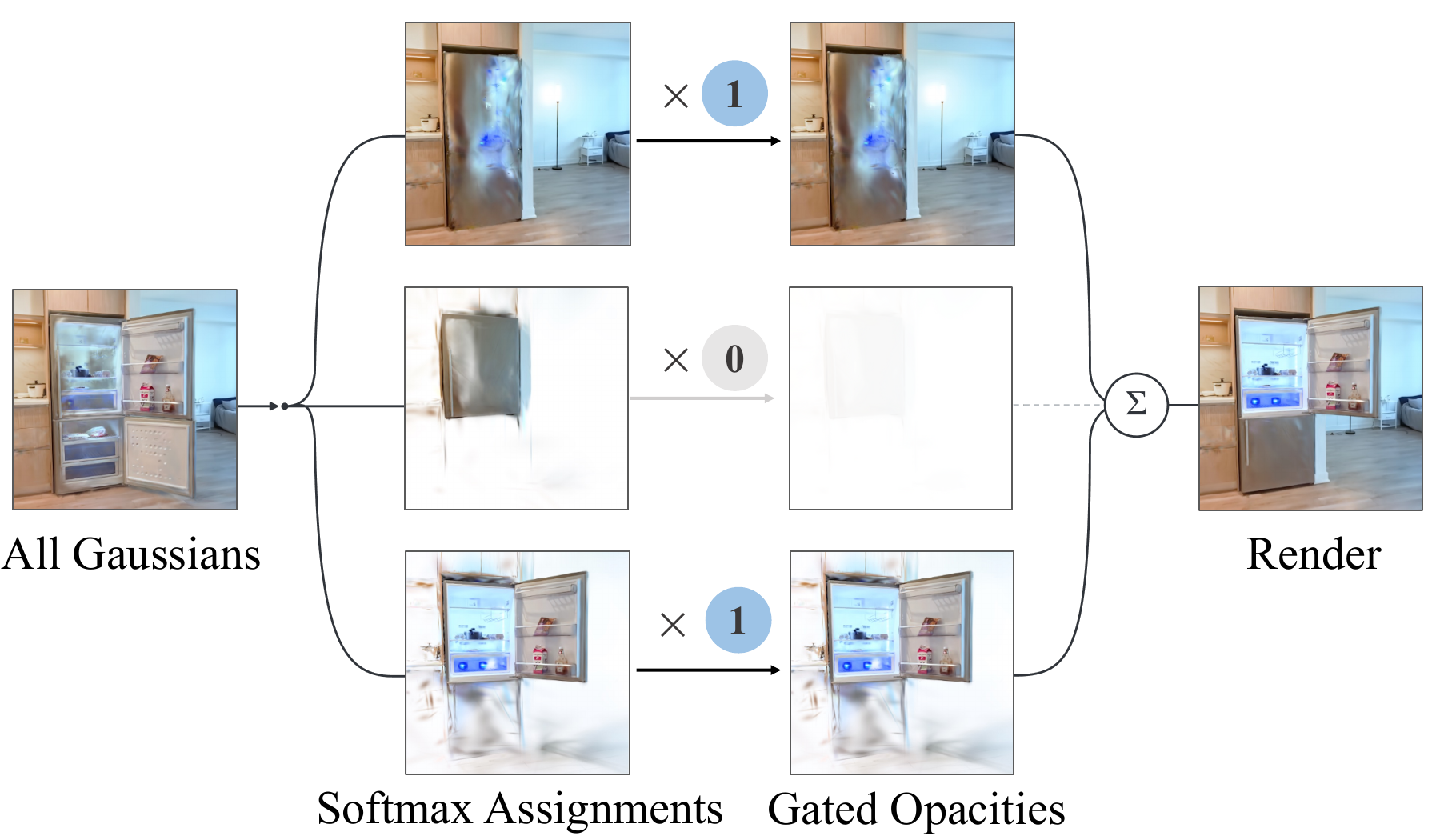}
\caption{
\textbf{State-aware Gaussian -- } \textbf{(left)} Each Gaussian carries a learnable logit vector~$\boldsymbol{\ell}$ that defines a categorical distribution~$\stateprob$ that softly assigns each primitive to the static base or ``on''/``off'' object states. \textbf{(right)}~At render time, user specifies list of states that are active 
or inactive 
, and the probability of each gaussian being present in the scene~$\omega(\tilde{\mathbf{p}})$ is computed via Eq.~\eqref{eq:presence}.
}
\Description{An image of an open refrigerator is decomposed into three soft Gaussian-assignment layers representing the static base, the refrigerator's off state, and its on state. State-dependent binary weights of one, zero, and one retain the base and on-state layers while suppressing the off-state layer. The retained layers are summed to render the refrigerator in its open state.}
\label{fig:softmax-placeholder}
\end{figure}

Each 3D Gaussian primitive in our model is defined by its standard geometric and appearance attributes: mean $\mean$, covariance $\cov$, opacity $\opacity$, and view-dependent appearance~$\radiance$. 
To make our representation state-aware, we augment every Gaussian with a learnable categorical state distribution $\stateprob = \operatorname{softmax}(\logits)$ over $1 + 2\numslots$ channels, where $\numslots$ is the total number of object slots.
These channels comprise a single static \emph{base} channel $\logits[0]$, together with a pair of \emph{off} ($\logits[2k{-}1]$) and \emph{on} ($\logits[2k]$) channels for each slot $k \in \{1, \dots, \numslots\}$.
The two channels per slot --- rather than one --- let the softmax assign each Gaussian to a single configuration, so that disjoint subsets of primitives can independently model the geometry and appearance of an object in each of its two states~(see~\Cref{fig:softmax-placeholder}).
This separation is essential because the on/off configurations of an object generally differ not just in appearance, but also in spatial extent --- a closed door and an open door, for instance, occupy different parts of the scene --- which a single shared primitive simply cannot represent.
 
\myparagraph{Test-time rendering}
Assume we have trained our representation, and that we want to render an image corresponding to the slot configuration $\tilde\stateprob \in \{0,1\}^K$.
This is achieved by computing the probability of each gaussian being present in the scene to be rendered:
\begin{equation}
    \visibility(\tilde\stateprob) = \stateprob[0] + \sum_{k=1}^{\numslots} \Big( \tilde\stateprob[k] \cdot \stateprob[2k] + (1 - \tilde\stateprob[k]) \cdot \stateprob[2k-1] \Big)
    \label{eq:presence}
\end{equation}
This formulation ensures that ``base'' Gaussians remain present across all frames, while the presence of the on/off state Gaussians is gated by the probability of the corresponding slot being on.
We then gate the opacity of the gaussian by this probability:
\begin{equation}
    \tilde{\opacity} = \opacity \cdot \visibility(\tilde\stateprob),
    \label{eq:opacity}
\end{equation}
which is then used to compute the final RGB color for a ray $\ray$ via classical volume rendering~\cite{digest}:
\begin{equation}
    \hat{I}(\ray) = \sum_{i \in \mathcal{H}(\ray)} \transmittance_i(\ray) \cdot \tilde{\opacity}_i \cdot \radiance_i(\ray), \quad
    \transmittance_i(\ray) = \prod_{j<i} (1 - \tilde{\opacity}_j),
    \nonumber
\end{equation}
where $\mathcal{H}(r)$ represents the set of Gaussians intersected by the ray in depth order and $\radiance_i(r)$ denotes the view-dependent color.
While $\tilde\stateprob$ will be selected by the user at test-time, the key question in training our representation lies in how to propagate the gradients from image $\hat{I}_t$ at time $t$ to the (per-Gaussian) state distribution~$\stateprob$.

\subsection{Differentiable slot table}
\label{sec:slot-table}
\begin{figure}[t]
    \centering
    \begingroup
    \setlength{\unitlength}{.01\linewidth}
    \begin{picture}(100,18)
        \put(0,0){\makebox(0,0)[bl]{\includegraphics[width=\linewidth]{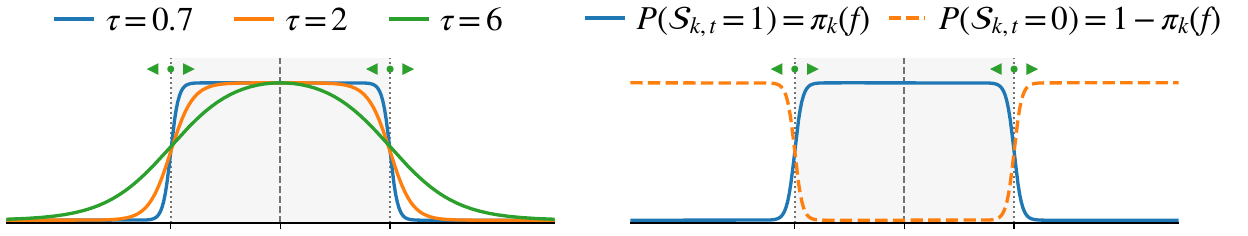}}}
\put(14.47, -2){\makebox(0,0)[c]{$\windowL$}}
        \put(23.79, -2){\makebox(0,0)[c]{$\windowM$}}
        \put(33.10, -2){\makebox(0,0)[c]{$\windowR$}}
\put(64.47, -2){\makebox(0,0)[c]{$\windowL$}}
        \put(73.79, -2){\makebox(0,0)[c]{$\windowM$}}
        \put(83.10, -2){\makebox(0,0)[c]{$\windowR$}}
    \end{picture}
    \endgroup
    \caption{ \textbf{Slot window function -- }
    The slot table $\pi_k(t)$ is parameterized as a normalized product of two sigmoids controlled by learnable endpoints $(\windowL, \windowR)$ and temperature $\temp$.
    \textbf{(left)} Sweeping $\temp$ interpolates between sharp rectangular windows (low $\temp$) and broad, soft transitions (high $\temp$).
    \textbf{(right)} The resulting per-frame Bernoulli PMF over slot $k$'s state, $P(\mathcal{S}_{k,t})$.
    Green markers denote the two learnable endpoints $\windowL, \windowR$, which can shift along the time axis during optimization.
    }
    \Description{Two plots share vertical markers for the left endpoint, midpoint, and right endpoint of a temporal slot. The left plot overlays slot-window curves for temperatures 0.7, 2, and 6: the lowest temperature produces nearly rectangular transitions, while higher temperatures produce progressively broader, smoother shoulders. The right plot shows complementary Bernoulli probabilities, with the active-state probability near one inside the endpoints and the dashed inactive-state probability near one outside. Green markers and arrows identify the learnable endpoints.}
    \label{fig:family_of_functions}
\end{figure}
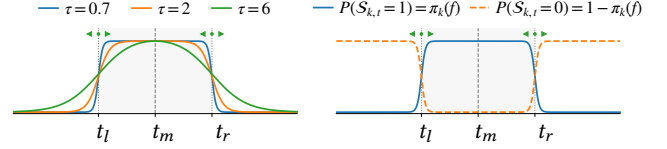

To achieve this, we exploit the key assumption that, within the video, the activation state of a slot is \underline{\smash[b]{compactly supported in time}}.
Let us denotes with the per-slot scalar $\pi_k(t) \in [0, 1]$ the probability of slot $k$ being in its \emph{on} state at time $t$.
Then, $\forall k \neq 0$ we have that $\pi_k(t)$ is zero everywhere outside some bounded, closed region.
Without loss of generality, let us drop the indexing $k$, and focus on a single entry $\pi(t)$.
We can satisfy our compact support assumption by explicitly parameterizing $\pi(t)$ as a \underline{\smash[b]{smooth window function}} defined by its temporal endpoints $(\windowL, \windowR)$\footnote{To ensure $\windowL \leq \windowR$ in practice we optimize auxiliary free variables $u$ and $v$, and then remap them to $\windowL=\min(u,v)$ and $\windowR=\max(u,v)$.}, and a temperature parameter $\temp$ that controls the smoothness of transition; see \Cref{fig:family_of_functions}.
 By defining the midpoint as $\windowM = (\windowL+\windowR)/2$,
and $\sigma_\temp(t) = \sigma(t/\temp)$, the probability of an interaction occurring at time $t$ is computed via the normalized product of two sigmoid functions $\sigma$:
\begin{align}
    \pi(t) = \frac
    {
        \sigma_\temp(t - \windowL) 
        \cdot
        \sigma_\temp(\windowR-t)
    }
    {
        \sigma_\temp(\windowM - \windowL) 
        \cdot 
        \sigma_\temp(\windowR - \windowM)
    }
\end{align}
Therefore, to model the time window of each object interaction, we treat the state of the $k$-th slot at time $t$ as a Bernoulli random variable $\slot_{k,t} \in \{0, 1\}$, where $\slot_{k,t} = 1$ denotes the slot being in its \emph{on} state and $\slot_{k,t} = 0$ its \emph{off} state, whose probability mass function~(PMF) for a specific state $s \in \{0, 1\}$ is given by:
\begin{align}
    P(\mathcal{S}_{k,t} = s) = \pi_k(t)^s (1 - \pi_k(t))^{1-s}
\end{align}

\myparagraph{Train-time rendering}
The quantity $\pi_k(t)$ defined above is our \underline{\smash[b]{slot table}}, which we use to compute the probability of each Gaussian being present in the scene at time $t$:
\begin{align}
    \visibility(t) = \stateprob[0] + \sum_{k=1}^{\numslots} \Big( \pi_k(t) \cdot \stateprob[2k] + (1 - \pi_k(t)) \cdot \stateprob[2k-1] \Big)
\end{align}
from which, like before, we can compute a time-dependent effective opacity for volume rendering the predicted image $\hat{I}_t$:
\begin{align}
    \tilde{\opacity}(t) = \opacity \cdot \visibility(t)
\end{align}
This differentiable rendering framework ensures that gradients propagate seamlessly to the slot activation probabilities, Gaussian state memberships, and geometric parameters, enabling stable and efficient joint optimization of the entire scene representation.

\subsection{Initialization}
\label{sec:init}
In a casual capture setting, neither the total number of slots $\numslots$, nor the precise slot transition timestamps are known a priori.
Requiring the user to annotate them would defeat the very premise of a ``casual'' capture.
To address this, we leverage the reasoning capabilities of Vision-Language Models (VLMs), and in particular GPT $5.6$, to provide an approximate initialization for the slot-table. 
By prompting GPT $5.6$ with a sequence of frames, we extract semantically grounded start and end timestamps for each distinct object interaction. 
These timestamps serve as the initial values for the endpoint parameters $(\windowL, \windowR)$ in our slot-table.
This VLM-informed initialization allows the system to bypass exhaustive temporal searches or complex diagnostics, providing a clean and efficient configuration for final full-resolution training.
To facilitate a robust optimization process, we initialize each interaction slot with a high temperature value $\temp$. 
This high-temperature initialization ensures that the Bernoulli random variable $\slot_{k,t}$ begins with smooth transitions at the boundaries, allowing the gradients to flow more effectively over a wider temporal range during the initial stages of optimization. 

\subsection{Optimization} 
\label{sec:optim}
Leveraging the initial parameters derived from the VLM-based slot discovery phase, we proceed to optimize our final state-aware scene representation. 
We build our state-aware framework upon the 3D Gaussian Unscented Transform (3DGUT) architecture~\cite{3dgut}. 
Following the 3DGUT pipeline, we initialize the 3D Gaussian primitives using COLMAP-derived point clouds and augment each primitive with the learnable state logits $\logits$. 
To provide an unbiased starting point for optimization, these logits are initialized such that the base/on/off state channels are assigned equal membership probabilities.
Throughout this stage, we jointly optimize the geometric attributes of the Gaussians, their state logits, and the temporal parameters of the slot-table $(\windowL, \windowR, \temp)$. 
We strictly adhere to the optimization schedule established by 3DGUT, including its specialized densification and pruning protocols.
To ensure consistent state behavior during optimization, we modify the standard densification process: whenever a Gaussian primitive is cloned or split, its state logits are inherited by the resultant primitives. 
This inheritance mechanism ensures that state-aware attributes remain spatially coherent even as the underlying scene geometry increases in complexity.

\subsection{Training objective}
\label{sec:train_obj}
Beyond the standard $L_1$ photometric and SSIM losses used to preserve perceptual fidelity, we employ two additional regularizers to mitigate floater artifacts, and prevent the emergence of redundant Gaussian primitives by supervising the optimization process via:
\begin{align}
    \losst{} = \losst{rgb} + \lambda_1 \losst{SSIM} + \lambda_2 \losst{sparsity} + \lambda_3 \losst{compactness}
\end{align}

\myparagraph{Sparsity loss} 
To ensure a parsimonious partitioning of the scene, we regularize the Gaussian primitives to favor the static base state, thereby preventing unnecessary scene regions from being categorized as interactive. 
This is achieved by applying a negative log-likelihood penalty on the base state probability for all $N$ Gaussians:
\begin{align}
    \losst{sparsity} = -\frac{1}{N}\sum_n \log \stateprob_{n}[0]
\end{align} 

\myparagraph{Compactness loss} 
To encourage Gaussians within slots to form spatially cohesive clusters, we first compute the opacity-weighted center of mass for primitives assigned to the on state of slot $k$:
\begin{align}
    \overline{\mean}_{k} = \frac{\sum_n \opacity_n \cdot \stateprob_n[2k] \cdot \mean_n}{\sum_n \opacity_n \cdot \stateprob_n[2k] + \epsilon}
\end{align}
and then penalize the weighted variance~\cite{caca} for both the off and on states to encourage local clustering:
\begin{align}
    \losst{compactness} &= \frac{1}{2\numslots} \sum_{j=\{0,1\}} \frac{\sum_n \opacity_n \cdot \stateprob_n[2k-j] \cdot ||\mean_n - \overline{\mean}_{k}||^2}{\sum_n \opacity_n \cdot \stateprob_n[2k-j] + \epsilon}
\end{align}

\begin{table*}[t]
  \caption{
    \textbf{Quantitative comparison on the main benchmark scenes with standard held-out views.}
    We report PSNR (dB) on ten scenes grouped by capture setting.
    Concurrent indoor scenes contain multiple interactions occurring
    simultaneously. Best and second-best results are shown in
    \textbf{bold} and \underline{underlined}, respectively.
  }
  \label{tab:main_psnr_comparison}
  \label{fig:main_qual}
  \centering

  \scriptsize
\setlength{\tabcolsep}{2.5pt}
  \renewcommand{\arraystretch}{1.12}

  \resizebox{\textwidth}{!}{\begin{tabular}{@{}l *{11}{S}@{}}
    \toprule
    Method
    & \multicolumn{2}{c}{Outdoor}
    & \multicolumn{5}{c}{Indoor}
    & \multicolumn{3}{c}{Concurrent Indoor}
    & \\
    \cmidrule(lr){2-3}
    \cmidrule(lr){4-8}
    \cmidrule(lr){9-11}

    & Balcony
    & Car
    & Kitchen 1
    & Flat 1
    & Flat 2
    & Flat 3
    & Chair
    & M Kitchen 1
    & M Kitchen 2
    & M Kitchen 3
    & Average \\
    \midrule

    \multicolumn{12}{c}{\textit{Static method}} \\FullCircle (3DGS)
      & 25.70 & 26.86 & 29.32 & 29.26 & 28.46
      & 28.92 & 29.88 & 30.16 & 30.13 & 30.10
      & 28.88 \\
    FullCircle (3DGUT)
    & 25.77 & 26.77
    & 29.35 & 29.27 & 28.62 & 29.14 & 30.06
    & 30.38 & 30.23 & 30.35
    & 28.99 \\

    \midrule
    \multicolumn{12}{c}{\textit{Dynamic method}} \\DeGauss
    & 25.86 & 27.31
    & 29.69 & 29.36 & 30.11 & 28.91 & 34.10 & 32.72 & 32.29 & 32.68
    &  30.30\\

    4DGS
    & 26.36 & \textbf{29.14}
    & 30.83 & 30.78 & 30.17 & 29.54 & 30.95
    & 31.84 & 32.75 & 33.14
    & 30.55 \\

    \midrule
    \multicolumn{12}{c}{\textit{State-Conditioned method}} \\SwitchSplat (3DGS)
    & \underline{26.89} & 28.92
    & \underline{32.32} & \underline{32.87}
    & \underline{32.58} & \underline{33.33}
    & \underline{35.55}
    & \underline{33.20} & \underline{33.97}
    & \underline{34.44}
    & \underline{32.41} \\

    SwitchSplat (3DGUT)
    & \textbf{26.97} & \underline{29.03}
    & \textbf{32.55} & \textbf{33.17}
    & \textbf{32.78} & \textbf{33.63}
    & \textbf{36.04}
    & \textbf{33.57} & \textbf{34.17}
    & \textbf{34.56}
    & \textbf{32.65} \\

    \bottomrule
  \end{tabular}

  }
\end{table*}
 \section{Experiments}
To validate the effectiveness of our proposed algorithm in the absence of an existing dataset with variable object interactions, we collected a custom dataset using a consumer-grade Insta360 X4 dual-fisheye $360^{\circ}$ camera.  
This hardware choice reflects the typical setup of casual users. 
During capture, we emulated non-expert behaviors, moving through the scene and interacting with objects—transitioning them from resting to active states and back.

Our dataset comprises $12$ scenes organized into two groups.
The \emph{standard set} contains $10$ casually captured scenes ($8$ indoor and $2$ outdoor), each featuring $3$--$9$ object interactions.
The \emph{golden set} contains $2$ additional scenes for which we acquire controlled test sequences with novel camera trajectories and object-state combinations.
We preprocess the raw dual-fisheye data following FullCircle~\cite{fullcircle} to mask the capturer. 
The calibration is performed in COLMAP using the masked fisheye frames, and then the fisheye images are undistorted into standard pinhole images.

For evaluation, we adhere to standard NeRF procedures, holding out every 8th frame as a test set. 
\textit{Crucially}, we exclude test samples that coincide with state transitions (i.e., frames where an object is actively being moved) to reduce noise in our metrics, though these transition frames are retained within the training set to measure whether models can learn robustly from casual capture data. 
For the \underline{\smash[b]{golden set}} sequences, we additionally evaluate \emph{novel camera} frames, which show state configurations observed during training from new camera trajectories, and \emph{novel state-combination} frames, which contain object-state configurations not observed during training.
This set is used to demonstrate our method's ability to generalize beyond the training configurations, and provide true state-aware control.

\subsection{Baselines}
We evaluate our proposed method against several state-of-the-art scene representation baselines. 
For static modeling, we compare against \textbf{FullCircle}~\cite{fullcircle}, utilizing its underlying \textbf{3DGUT}~\cite{3dgut} architecture as our primary reference. 
To assess performance in dynamic contexts, we benchmark our framework against \textbf{4DGS}~\cite{4dgs} and \textbf{DeGauss}~\cite{degauss}. 
For all quantitative and qualitative assessments, we utilize every 8th frame of the captured sequences as our standard evaluation set.

\subsection{Quantitative Results}
In \Cref{fig:main_qual}, we present the quantitative performance of our method relative to both static and dynamic baselines on the undistorted dataset. 
These results validate our framework's ability to maintain high-fidelity visual representations while providing granular, state-controlled manipulation.
More detailed metrics for all $10$ scenes are available in the supplementary material.

\begin{figure*}[t]
    \centering
    \setlength{\tabcolsep}{2pt}
    \renewcommand{\arraystretch}{0.55}

    \newcommand{\qualimg}[1]{\includegraphics[width=0.175\textwidth]{#1}}

    \begin{tabular}{c@{\hspace{3pt}}ccccc}
        &
        \small FullCircle &
        \small DeGauss &
        \small 4DGS &
        \small SwitchSplat &
        \small Ground Truth
        \\[2pt]

        \rotatebox{90}{\small \ \ \ \ \ \ \ \ \  Standard Test}
        &
        \qualimg{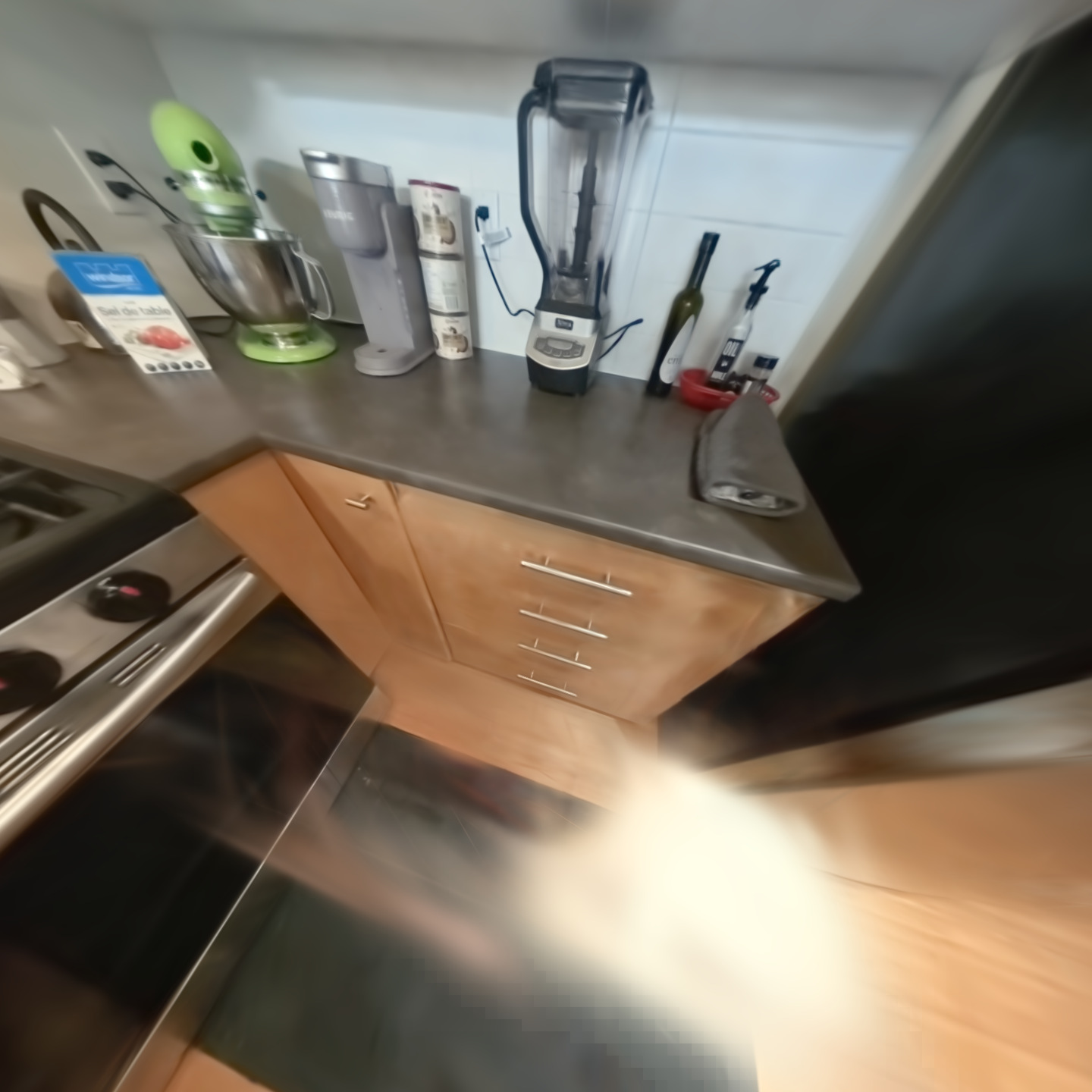} &
        \qualimg{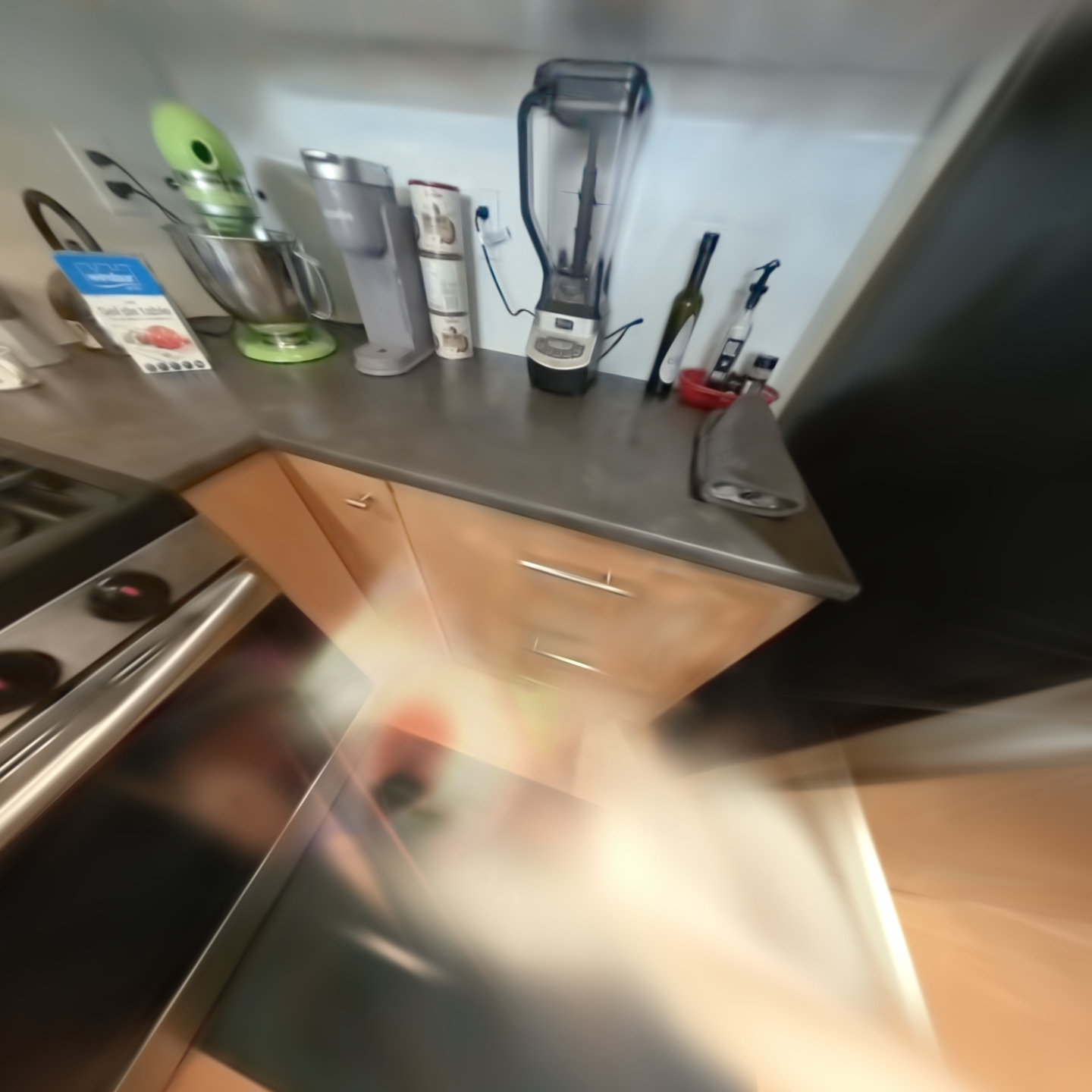} &
        \qualimg{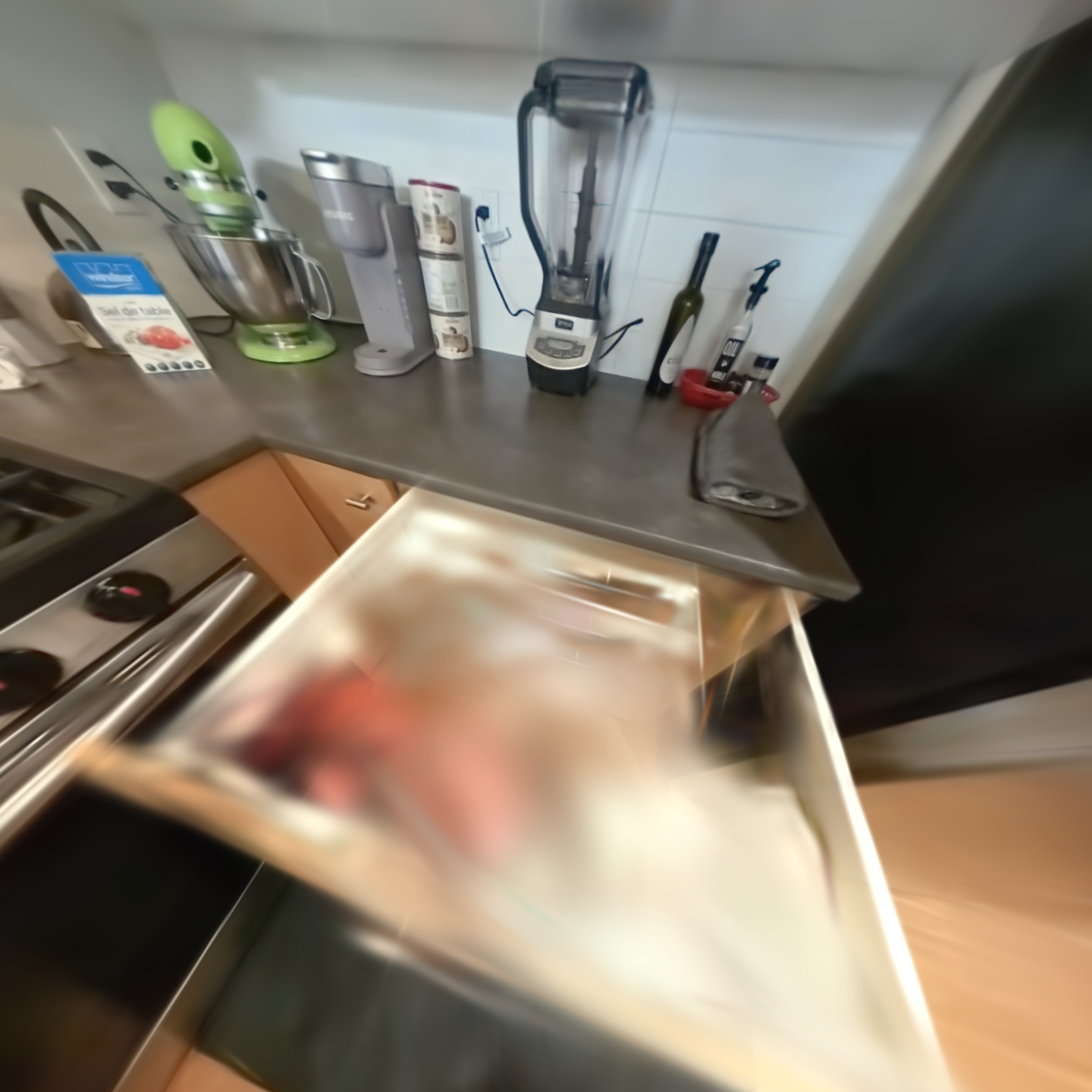} &
        \qualimg{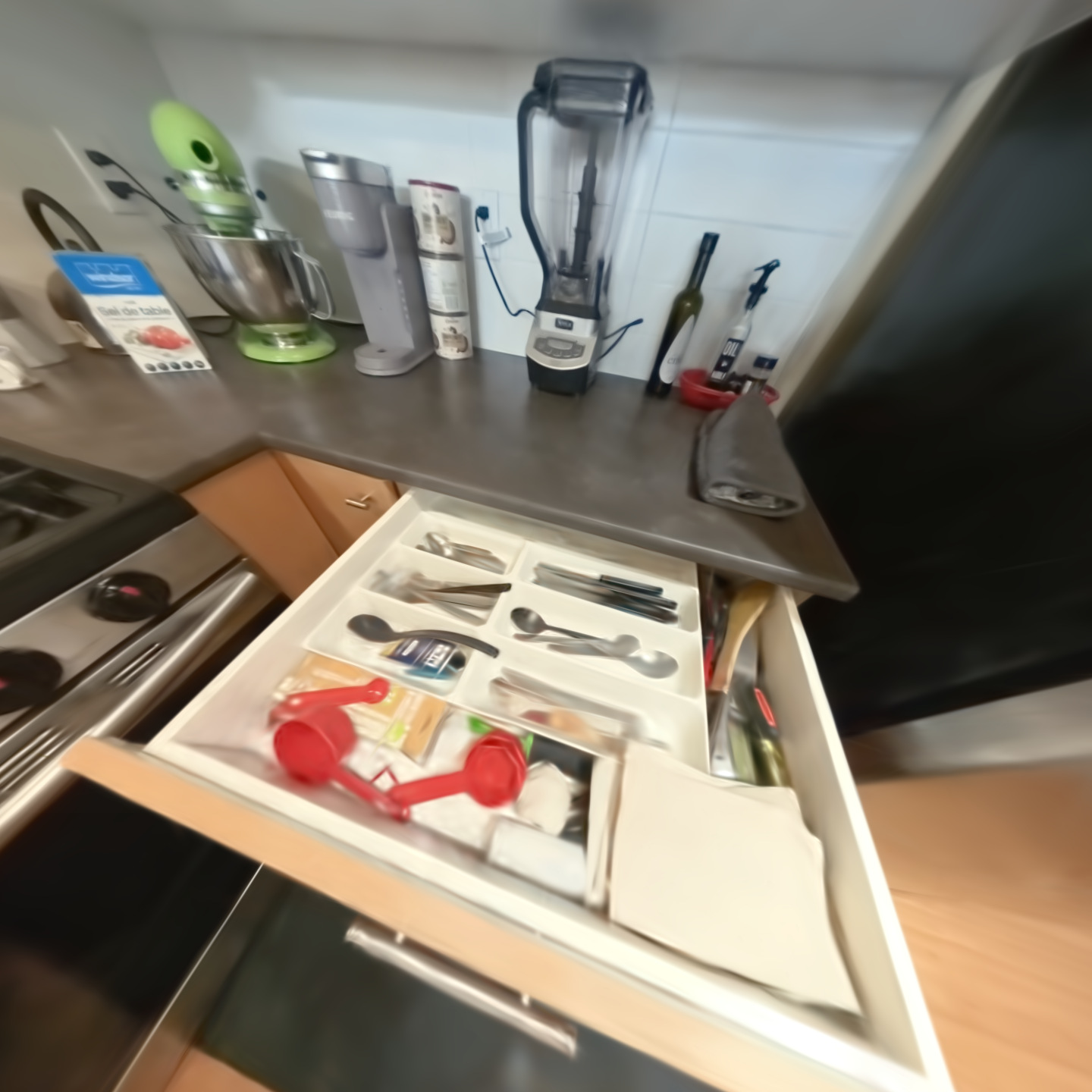} &
        \qualimg{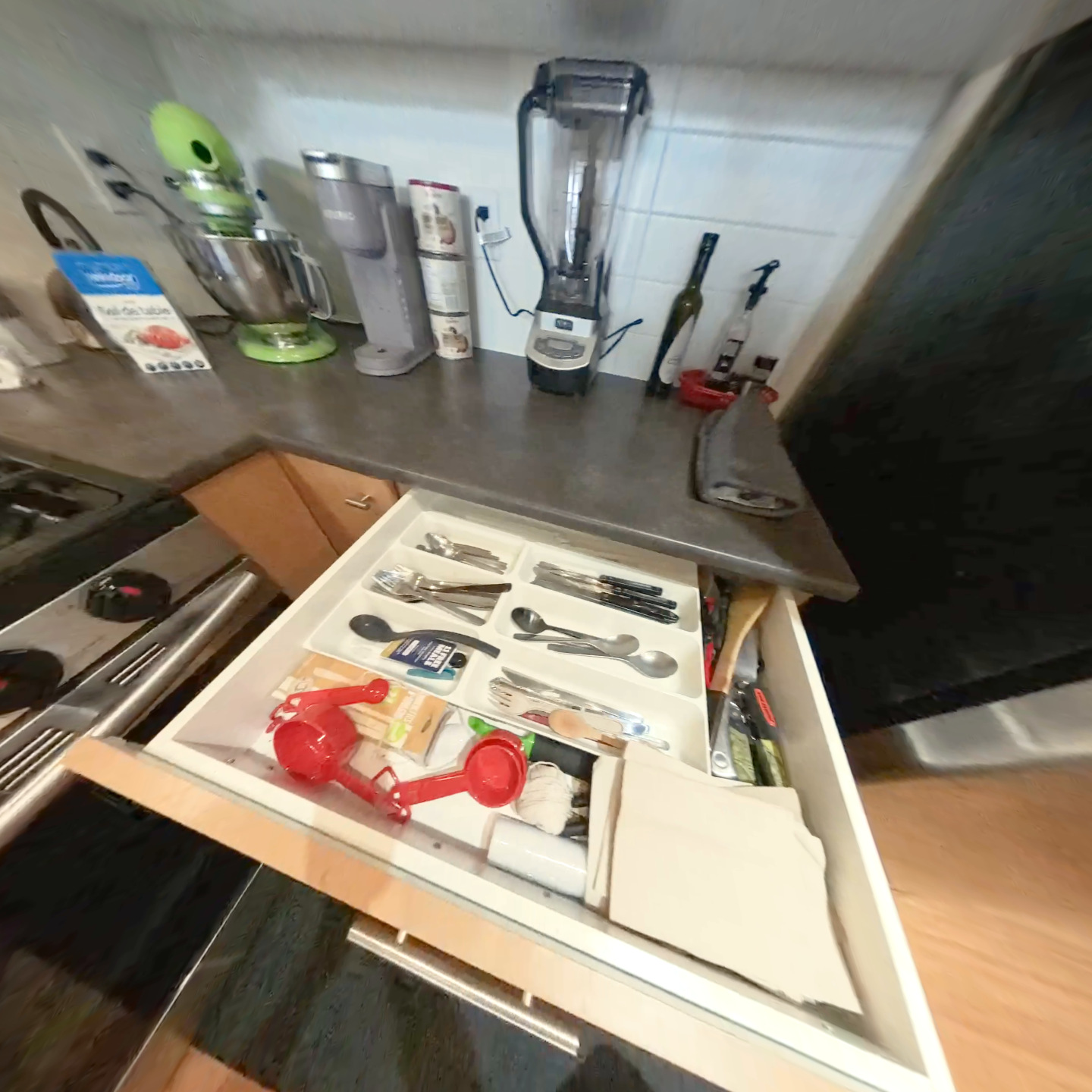}
        \\[5pt]

                \rotatebox{90}{\small \ \ \ \ \ \ \ \ \ \ Novel Camera}
        &
        \qualimg{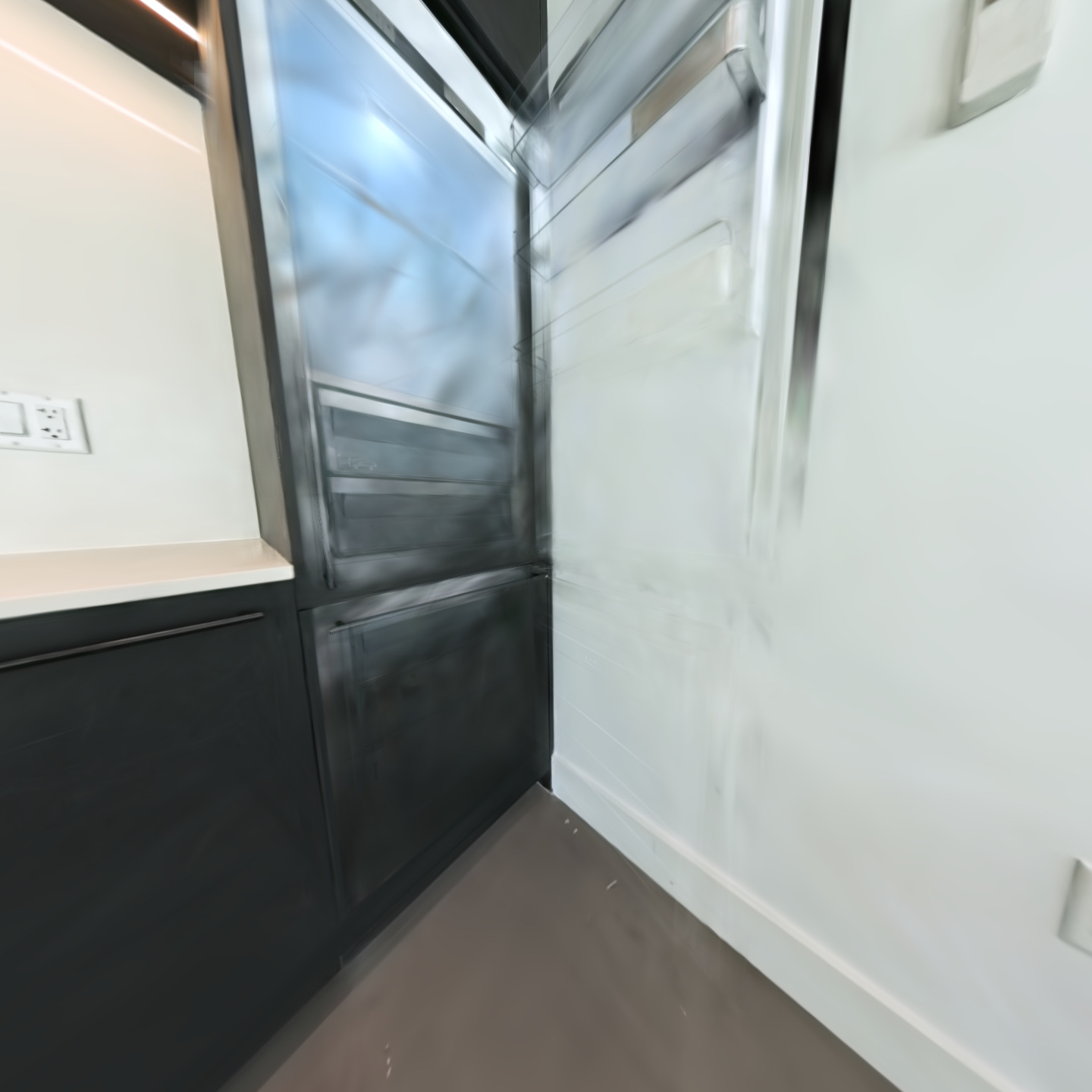} &
        \qualimg{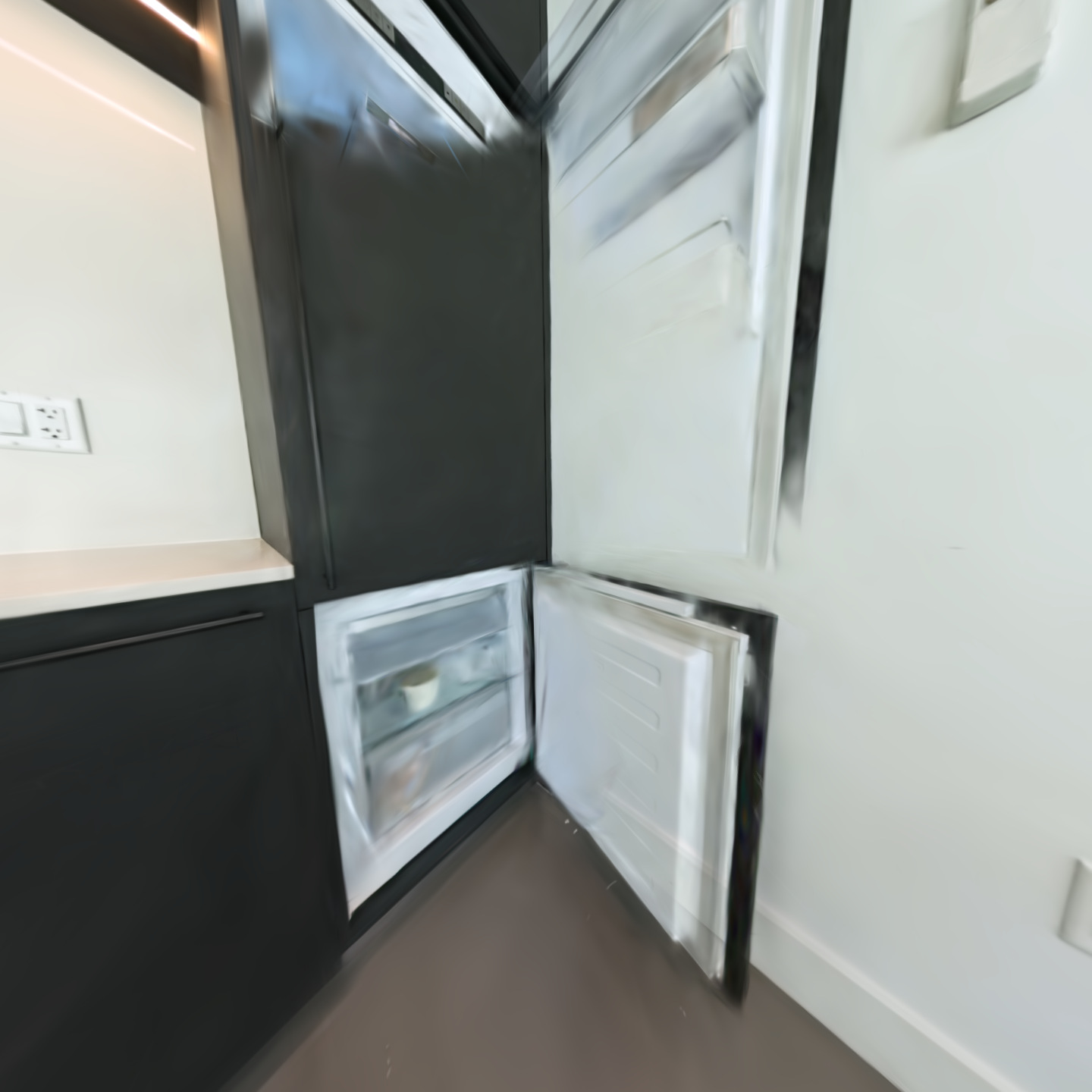} &
        \qualimg{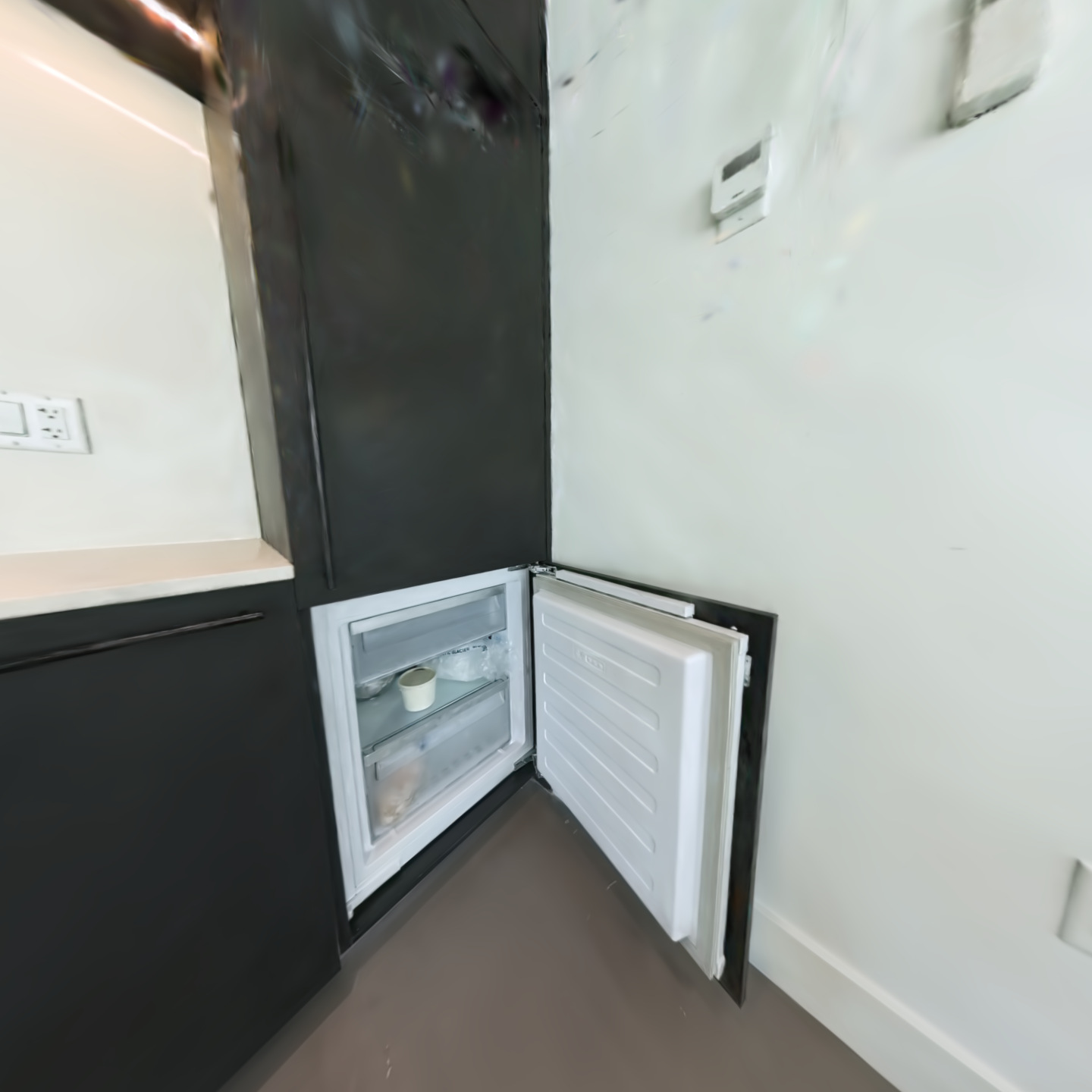} &
        \qualimg{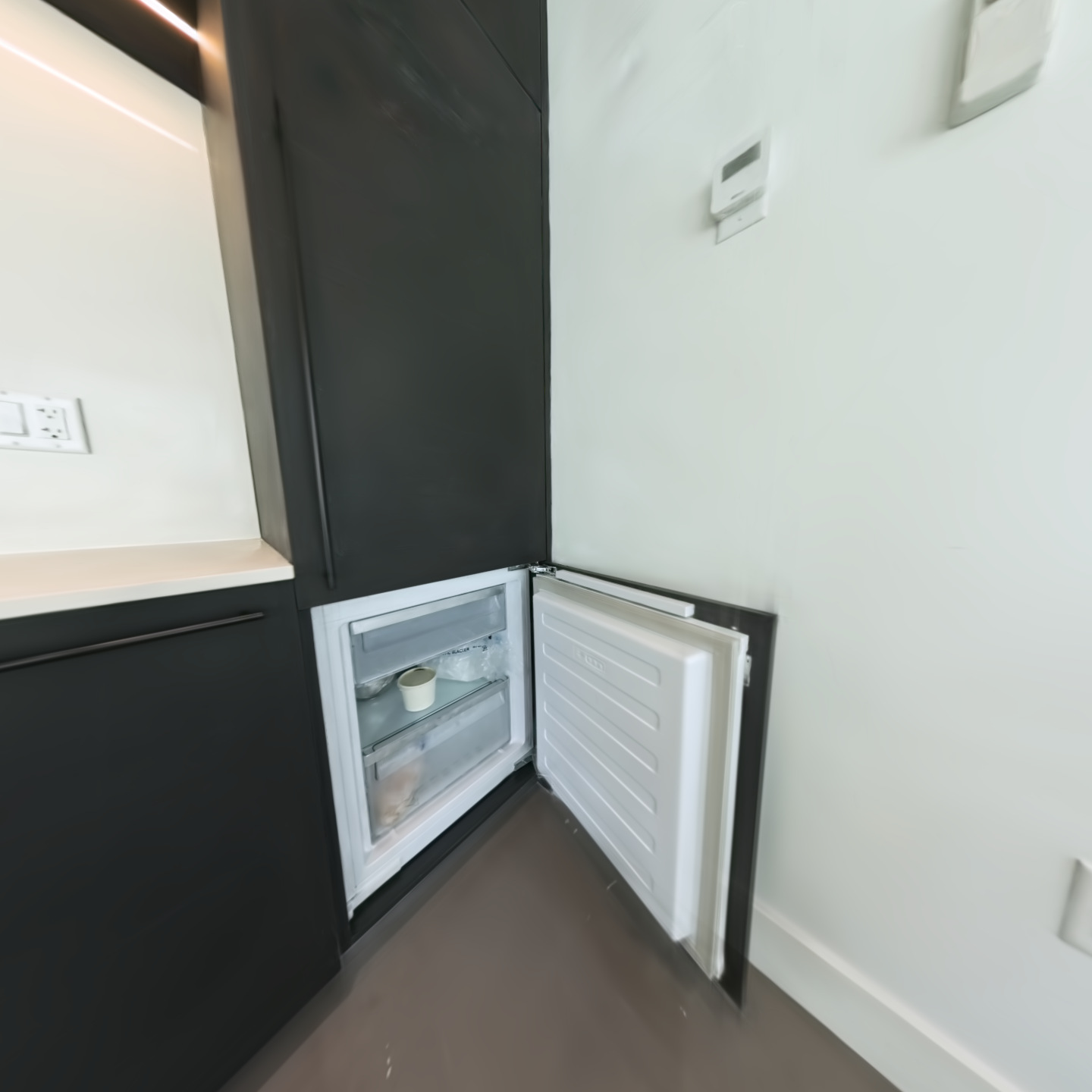} &
        \qualimg{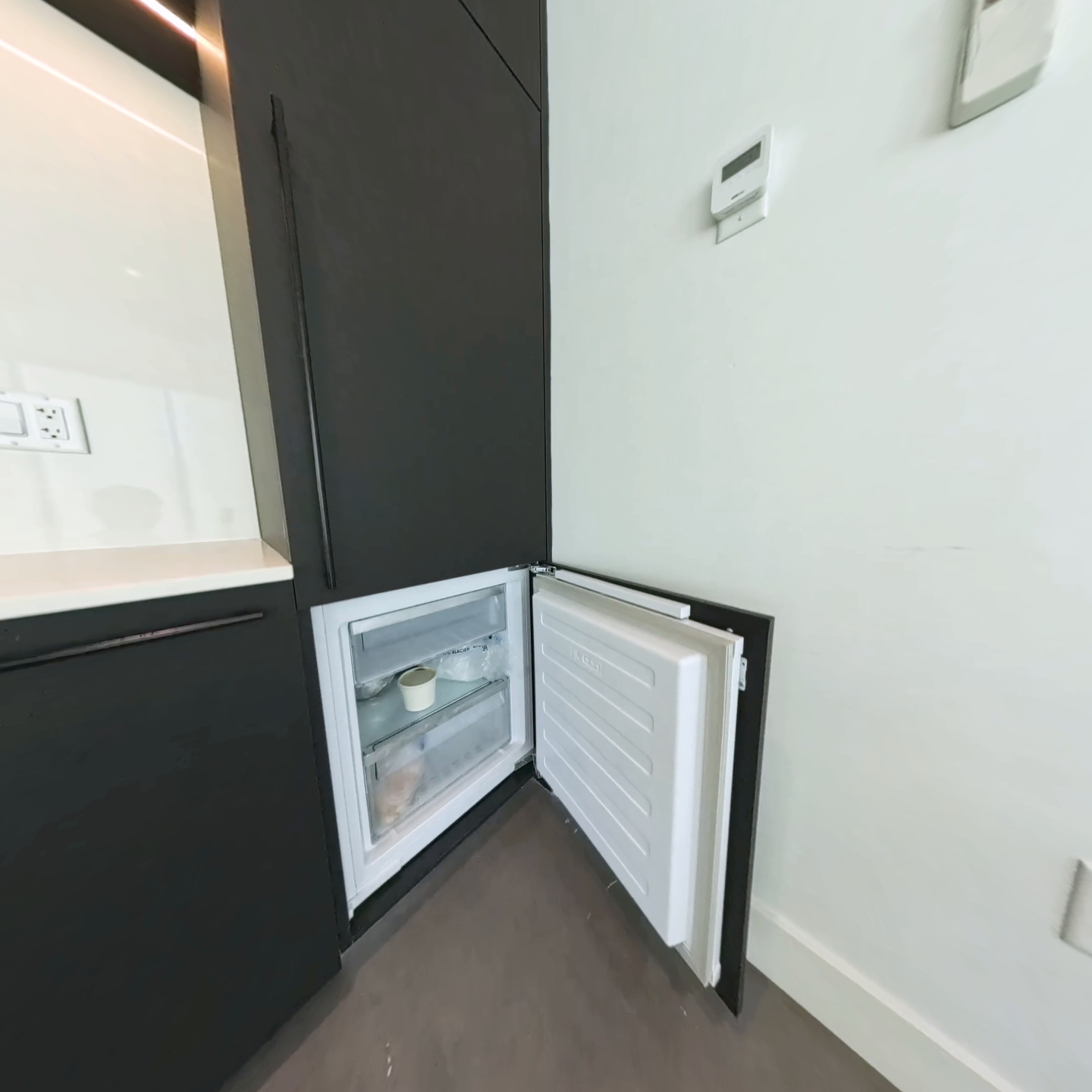}

                \\[5pt]

        \rotatebox{90}{\small \ \ \ \ \ \ \ \  Novel States}
        &
        \qualimg{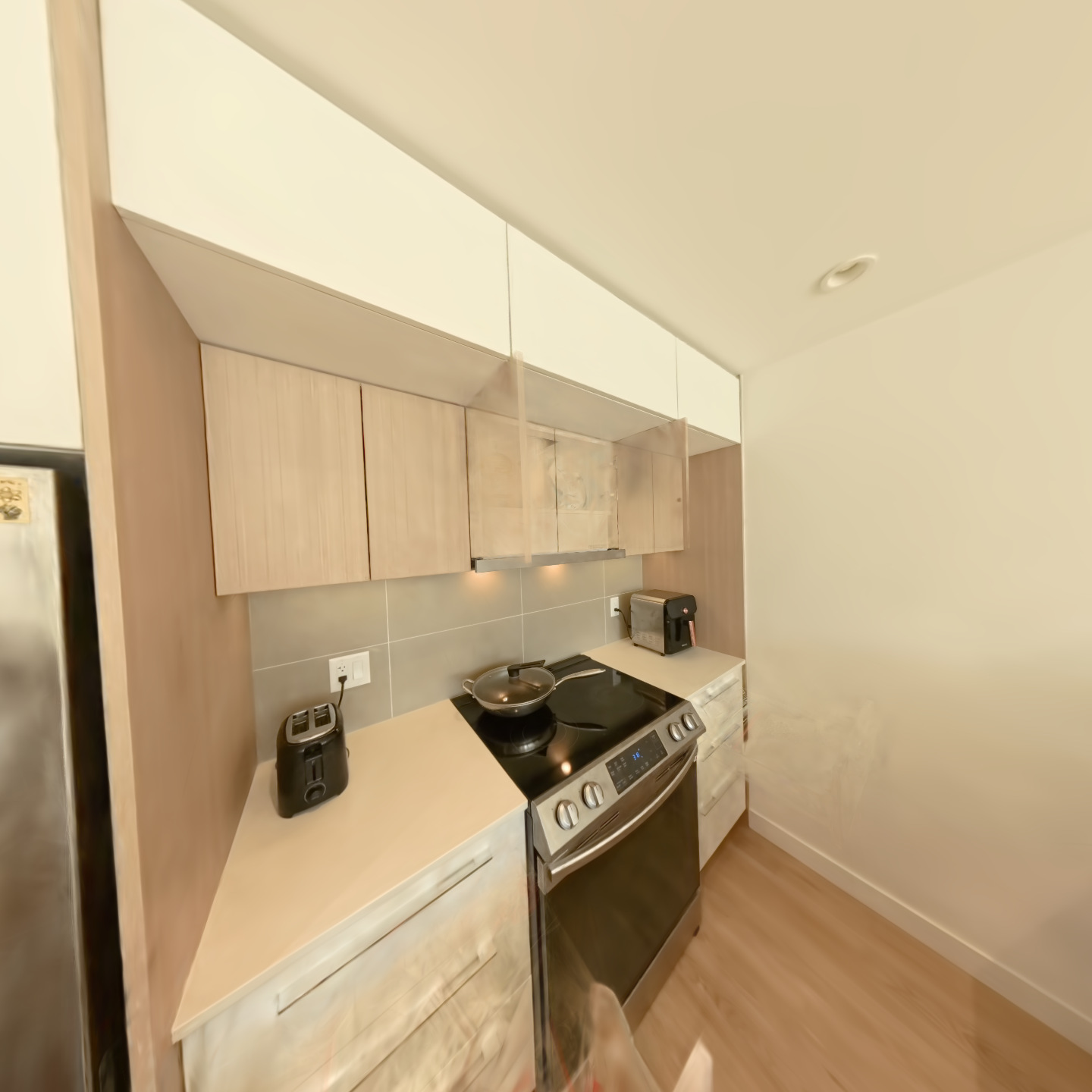} &
        \qualimg{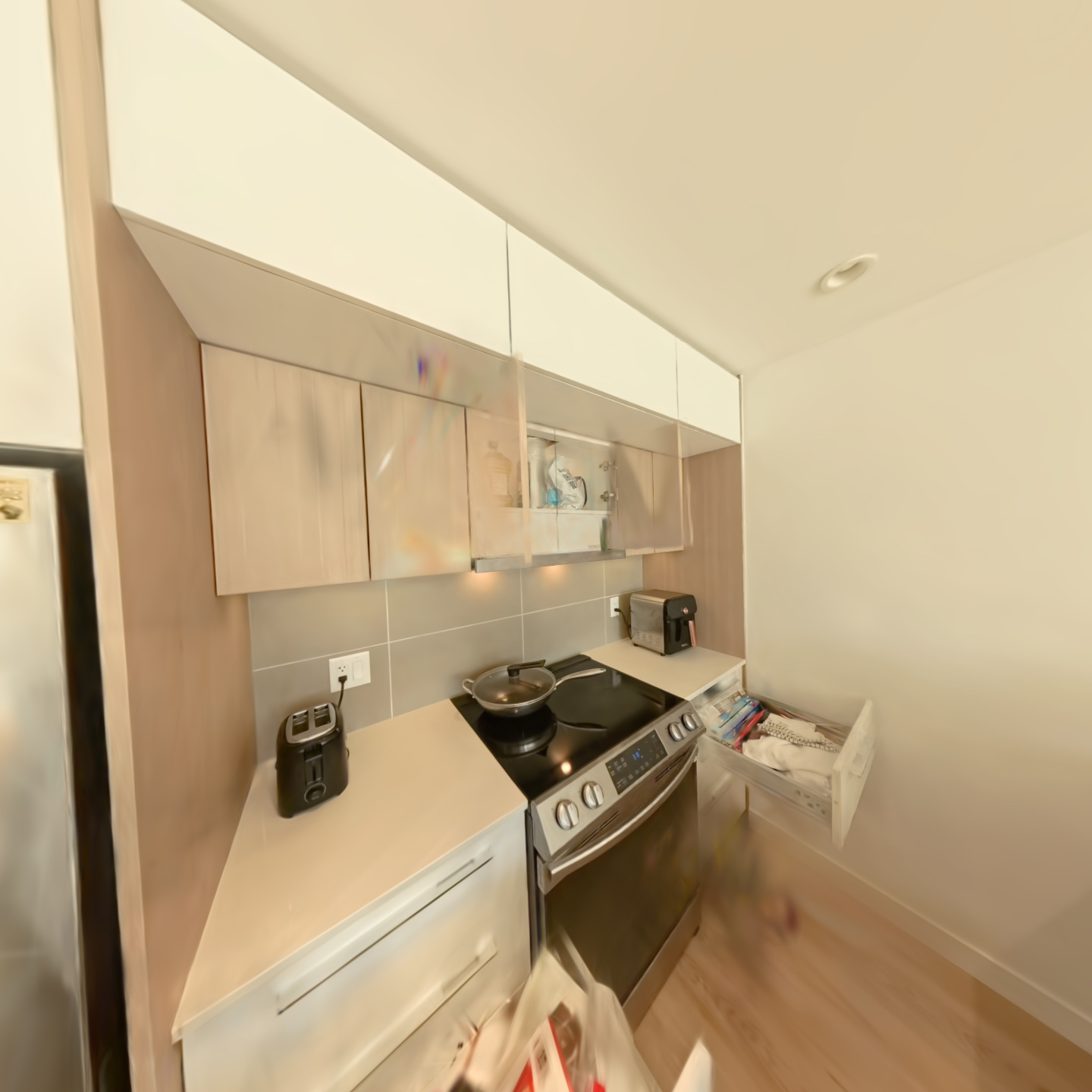} &
        \qualimg{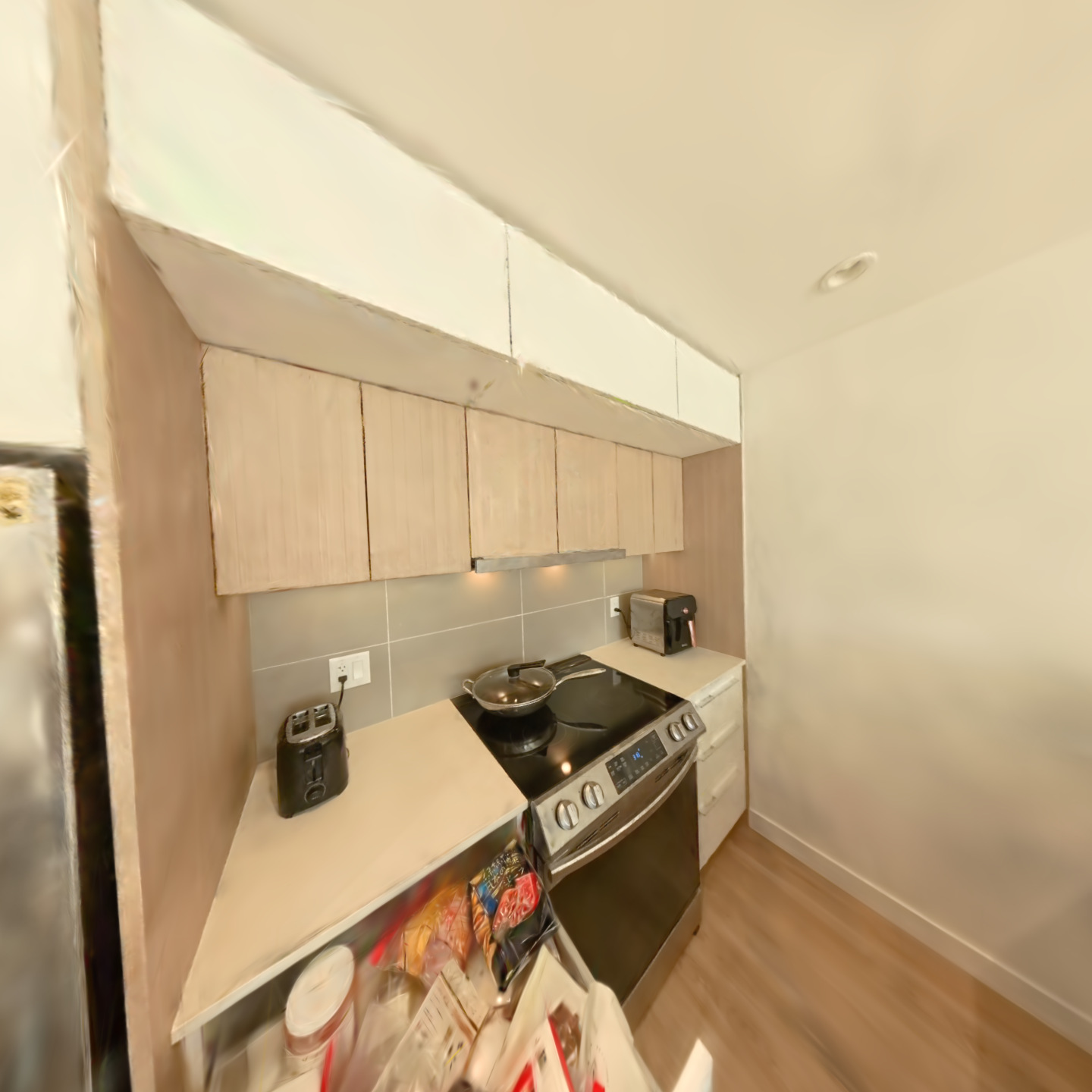} &
        \qualimg{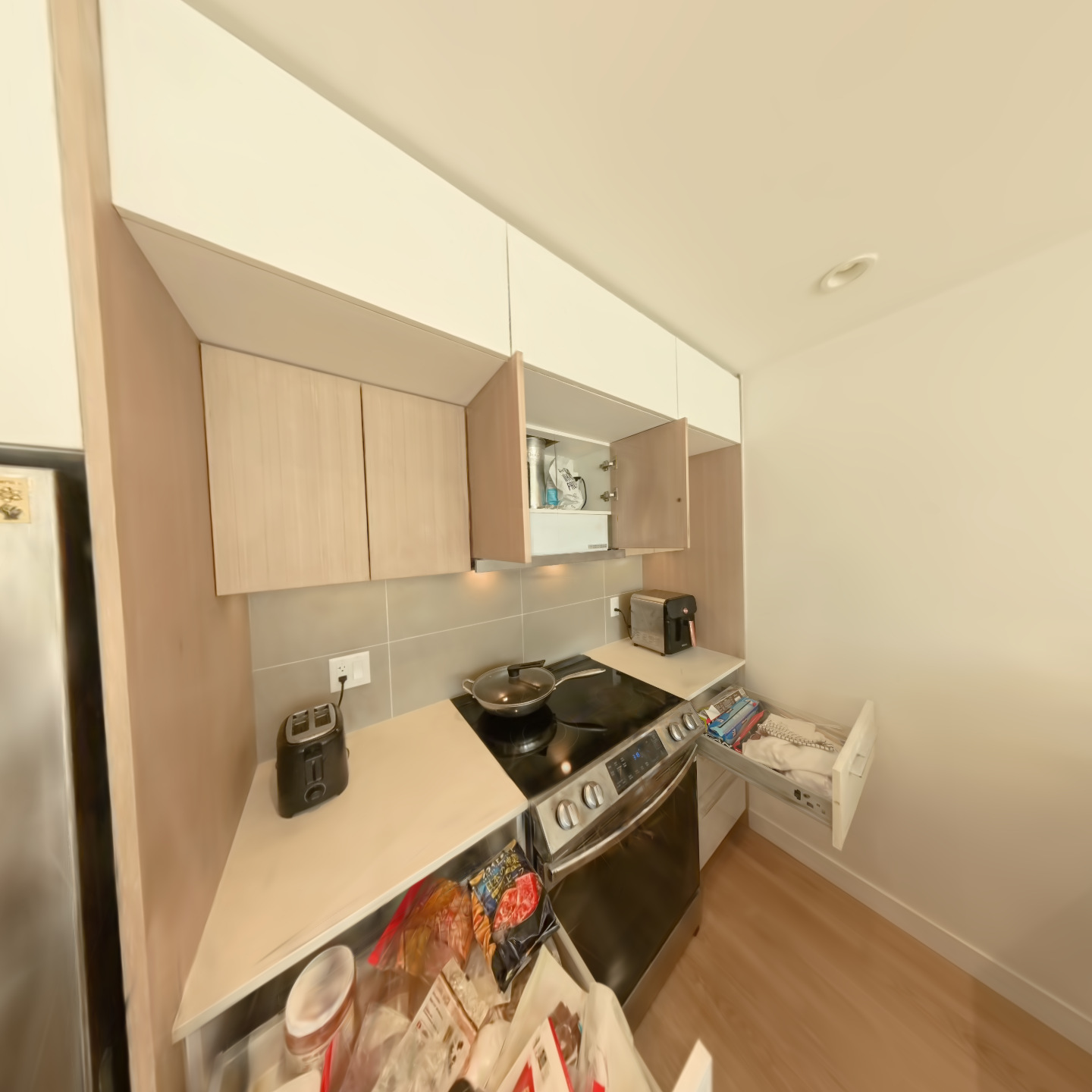} &
        \qualimg{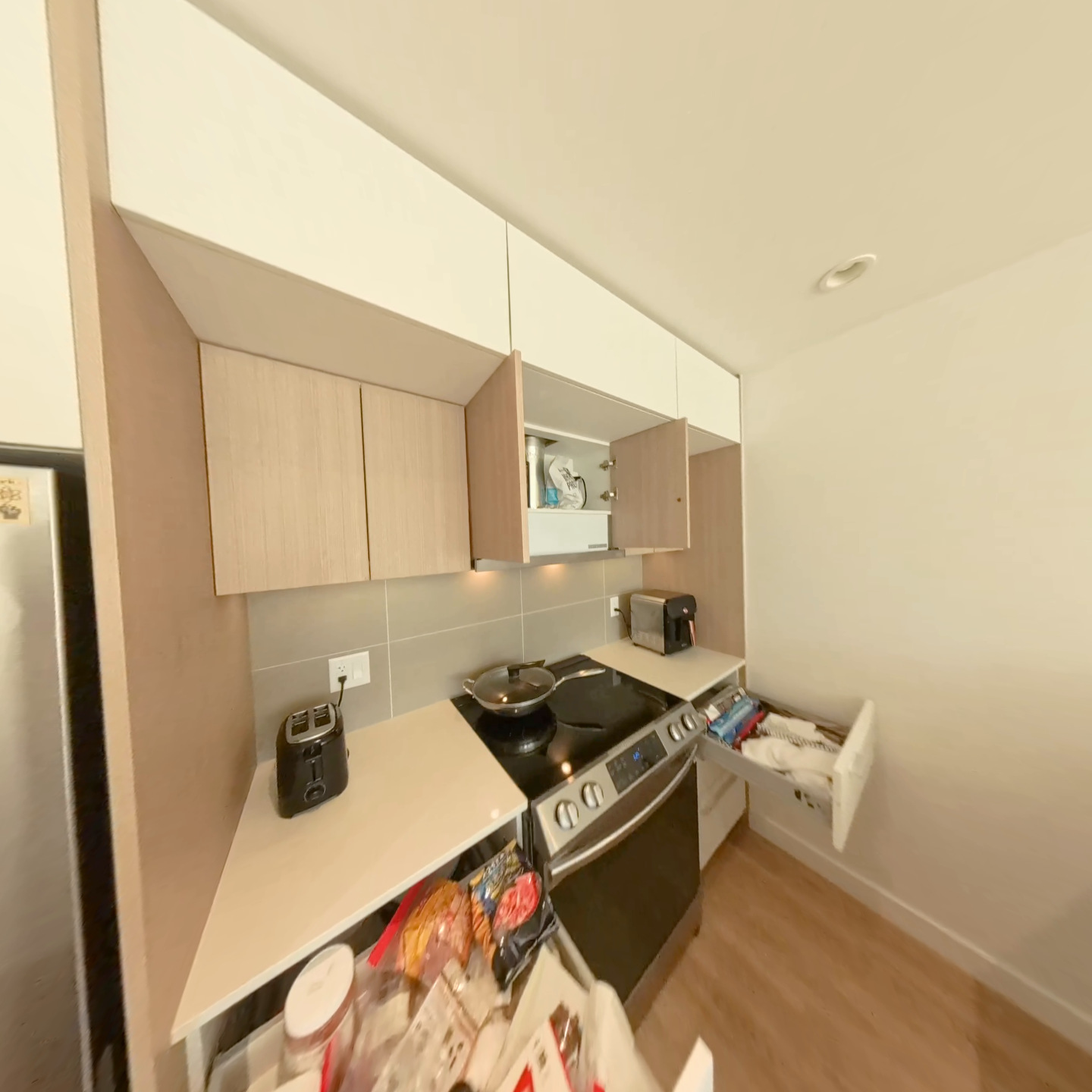}

    \end{tabular}

    \caption{
    \textbf{Qualitative comparison across three evaluation settings}: standard held-out views,
    novel-camera views and novel state combinations. Columns show FullCircle,
    DeGauss, 4DGS, SwitchSplat, and the corresponding ground truth.
    }
    \Description{A three-row, five-column comparison grid. Columns show FullCircle, DeGauss, 4DGS, SwitchSplat, and ground truth. The rows show a standard held-out kitchen view with an open drawer, a novel-camera view of an open refrigerator, and a novel-state kitchen configuration with multiple open compartments. The baseline renderings contain varying amounts of blur, ghosting, or incorrect object-state content, whereas the SwitchSplat renderings more closely preserve the sharp geometry and selected states visible in the ground truth.}
    \label{fig:qualitative_comparison}
\end{figure*}

\subsection{Qualitative Results}
\Cref{fig:qualitative_comparison,fig:comparison1,fig:comparison2} illustrate qualitative comparisons between SwitchSplat and baselines. 
While our method consistently yields high-fidelity renderings, dynamic baselines frequently exhibit pervasive floating artifacts during object interactions. 
For example, in the ``Car'' scene (\Cref{fig:comparison1}), both 4DGS and DeGauss generate significant floaters around the hood region. 
Similarly, in the ``Kitchen'' environment, these baselines produce substantial noise when the refrigerator is in its \emph{on} state. 
Comparable degradations, manifesting as unanchored geometry, are observed during cabinet and wardrobe interactions in the ``Flat'' scenes. 
In contrast, SwitchSplat maintains significantly cleaner reconstructions by effectively disentangling static environment components from discrete state-specific changes.

\begin{table}[t]
  \caption{
    \textbf{Generalization to novel cameras and state combinations.}
  \\
  $^{\dagger}$ Dynamic baselines use a per-image oracle over time, selecting the highest-PSNR rendering from a set of sampled time; these results are upper bounds.}
  \label{tab:golden-camera-ready}
  \label{fig:novel_cam}
  \centering

\small
  \setlength{\tabcolsep}{3.2pt}
  \renewcommand{\arraystretch}{1.10}

\centering
\small
\setlength{\tabcolsep}{2.2pt}
\renewcommand{\arraystretch}{1.15}

\resizebox{\columnwidth}{!}{\begin{tabular}{lccc@{\hspace{6pt}}ccc}
    \toprule
    & \multicolumn{3}{c}{Kitchen 2}
    & \multicolumn{3}{c}{Oslo} \\
    \cmidrule(lr){2-4}
    \cmidrule(lr){5-7}

    Method
    & PSNR $\uparrow$
    & SSIM $\uparrow$
    & LPIPS $\downarrow$
    & PSNR $\uparrow$
    & SSIM $\uparrow$
    & LPIPS $\downarrow$ \\

    \midrule

\multicolumn{7}{c}{\textit{Novel camera frames}} \\
    \addlinespace[1pt]

    DeGauss$^{\dagger}$
    & 32.38 & 0.965 & 0.174
    & 29.40 & 0.960 & 0.207 \\

    4DGS$^{\dagger}$
    & 32.05 & 0.967 & 0.168
    & 29.93 & 0.966 & 0.194 \\

    SwitchSplat (3DGS)
    & 33.90 & 0.974 & 0.155
    & 31.82 & 0.972 & 0.186 \\

    SwitchSplat (3DGUT)
    & \textbf{34.26} & \textbf{0.975} & \textbf{0.154}
    & \textbf{32.55} & \textbf{0.973} & \textbf{0.185} \\

    \addlinespace[3pt]
    \midrule
    \multicolumn{7}{c}{\textit{Novel state-combination frames}} \\
    \addlinespace[1pt]

    DeGauss$^{\dagger}$
    & 30.85 & 0.958 & 0.180
    & 27.51 & 0.940 & 0.214 \\

    4DGS$^{\dagger}$
    & 29.63 & 0.956 & 0.182
    & 26.53 & 0.940 & 0.211 \\

    SwitchSplat (3DGS)
    & 31.29 & 0.962 & 0.164
    & \textbf{29.29} & \textbf{0.959} & 0.186 \\

    SwitchSplat (3DGUT)
    & \textbf{32.58} & \textbf{0.967} & \textbf{0.161}
    & 29.10 & 0.958 & \textbf{0.186} \\

    \bottomrule
\end{tabular}}

\end{table}

\subsection{Golden set comparisons}
Beyond the standard held-out evaluation on the original trajectories, we evaluate SwitchSplat on two additional generalization settings: novel camera frames and novel state-combination frames.
For two representative scenes, we acquire dedicated test trajectories containing previously unseen camera viewpoints and object-state combinations, and compare against 4DGS and DeGauss. Because 4DGS and DeGauss parameterize scene variation by continuous time rather than a discrete state code, we evaluate both dynamic baselines on the novel-camera and novel state-combination splits using a per-image oracle over time. For each target image, we select the rendering with the highest PSNR for both methods; these results therefore constitute upper bounds.
As shown in \Cref{fig:qualitative_comparison}, SwitchSplat consistently outperforms the dynamic baselines on novel camera frames and produces sharper reconstructions with fewer structural artifacts.
We further evaluate novel state-combination frames, where multiple independently controlled interactions are composed into configurations not observed during training. Quantitative results in \Cref{fig:novel_cam} demonstrate that SwitchSplat maintains high reconstruction quality under these unseen configurations, highlighting the compositional generalization enabled by our state-conditioned representation.

\subsection{Ablation study}

We perform a comprehensive ablation study to evaluate the contribution of individual components within our framework. 
These experiments are conducted on the ``Kitchen 2'' scene, which features the golden test set. 
We report PSNR averaged over three evaluation settings: the standard 8th-frame test set, novel camera frames, and novel state-combination frames, with each configuration further averaged over five independent runs.

\myparagraph{Slot representation}
We first evaluate the importance of the slot representation.
When interaction slots are removed, the model reduces to a scene representation without any of our slot-specific components, including slot optimization and the associated regularization terms.
The substantial performance drop in \Cref{tab:ablation} demonstrates that explicitly introducing interaction slots is essential for modeling state-dependent scene changes.

\myparagraph{Slot optimization}
We next assess the importance of optimizing the initialized slot parameters.
For this ablation, slots are initialized normally via VLM outputs but their parameters are kept fixed throughout training.
The observed deterioration in reconstruction quality (\Cref{tab:ablation}) shows that differentiable slot optimization is important for refining the initial estimates according to photometric evidence.

\myparagraph{Compactness regularization}
The compactness loss encourages state-conditioned Gaussians associated with the same interaction to remain spatially localized, preventing them from dispersing across unrelated regions of the scene.
This spatial prior therefore plays an important role in learning coherent state-dependent representations.

\myparagraph{Sparsity regularization}
Removing the sparsity regularization has a modest effect on average reconstruction quality.
The regularizer nevertheless discourages unnecessary assignment of otherwise static Gaussians to interactive states, promoting a more parsimonious state decomposition and reducing spurious state-dependent changes.

\begin{table}[htb]
  \caption{
    \textbf{Ablation Study.}
    We evaluate the impact of various components in our method by systematically excluding them and analyzing the reconstruction quality (PSNR) on the ``Kitchen 2'' scene.
    We report average PSNR across the 3 evaluation splits and 5 independent runs.
  }
  \label{tab:ablation}
  \centering
  \small
  \setlength{\tabcolsep}{4.0pt}
  \renewcommand{\arraystretch}{1.12}

  \begin{tabular}{@{}ccccc@{}}
    \toprule
    Slots &
Slots opt. &
    $\losst{compactness}$ &
    $\losst{sparsity}$ &
    PSNR $\uparrow$ \\
    \midrule

    \xmark & --     & --     & --     & 31.98 \\
    \cmark & \xmark & \cmark & \cmark & 34.25 \\
    \cmark & \cmark & \xmark & \cmark & 32.88 \\
    \cmark & \cmark & \cmark & \xmark & 34.29 \\

    \cmark & \cmark & \cmark & \cmark & \textbf{34.31} \\

    \bottomrule
  \end{tabular}

\end{table}
  \section{Conclusion}
We presented SwitchSplat, a 3D Gaussian Splatting framework that learns state-aware scene representations directly from casual video captures.
By decomposing the scene into a shared static base and a set of independently controllable slots, each modeling the binary ``on''/``off'' states of one interactable object, SwitchSplat decouples scene state from time -- enabling a capability that time-conditioned dynamic methods fundamentally cannot provide: free composition of object configurations at inference, including combinations never observed during capture.
The slot table is initialized via a VLM and jointly refined end-to-end through differentiable rendering, requiring no manual annotation.
Experiments on our new casual-capture dataset show that SwitchSplat outperforms static and dynamic baselines in novel-view synthesis quality, and uniquely generalizes to out-of-distribution camera paths and unseen state combinations.

\myparagraph{Limitations and future work}
The current model assumes binary object states with the constraints of single-operator capture with one free hand.
Extending to continuous or multi-valued states by multiple operators -- would broaden the applicability of the framework.
The slot count and approximate interaction windows are currently seeded by a VLM; a fully unsupervised discovery of both the number and timing of interactions from raw video remains an open problem.
Finally, SwitchSplat focuses on novel-view synthesis; combining our state-aware decomposition with physics-based priors or generative models could enable downstream applications such as scene editing and simulation. \begin{acks}
We thank the anonymous reviewers for their helpful comments and suggestions.
We also thank Zhihan Zhang, Egor Fokin, and Mingrui Zhao for contributing scenes and objects used in our dataset captures.
{We thank Lily Goli for her helpful advice and suggestions.}
This work was supported in part by the Natural Sciences and Engineering Research Council of Canada (NSERC) Discovery Grant [2023-05617], the Canada Foundation for Innovation (CFI) John R. Evans Leaders Fund (JELF), the SFU Visual Computing Research Chair, Google Research, and the Digital Research Alliance of Canada.
\end{acks}

\clearpage
\bibliographystyle{ACM-Reference-Format}
\bibliography{references}

\begin{figure*}[t]
\centering
    \setlength{\tabcolsep}{1pt}\renewcommand{\arraystretch}{1.0}\begin{tabular}{@{}ccccc@{}}
        \footnotesize FullCircle &
        \footnotesize DeGauss &
        \footnotesize 4DGS &
        \footnotesize SwitchSplat &
        \footnotesize Ground Truth \\

        \includegraphics[width=0.18\textwidth]{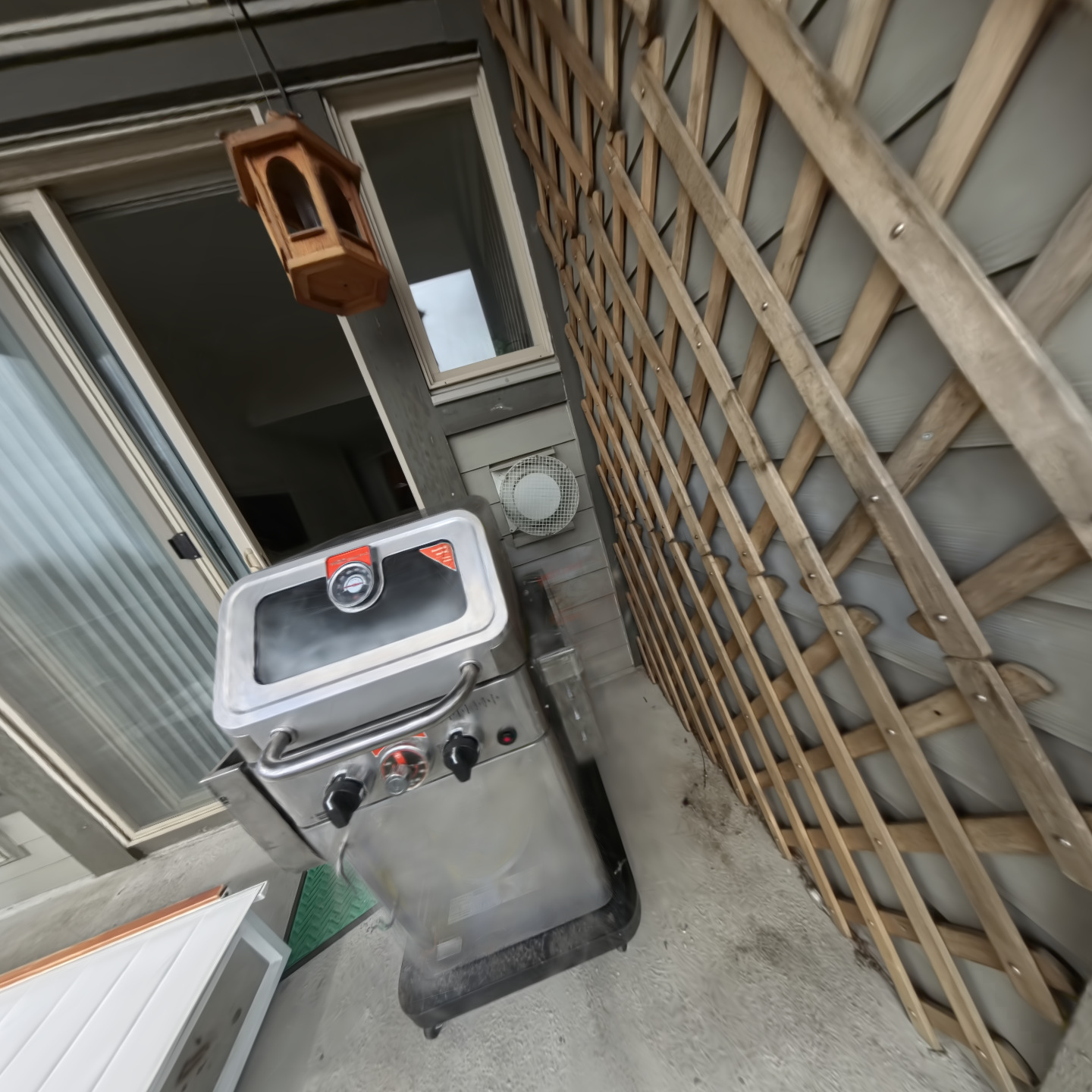} &
        \includegraphics[width=0.18\textwidth]{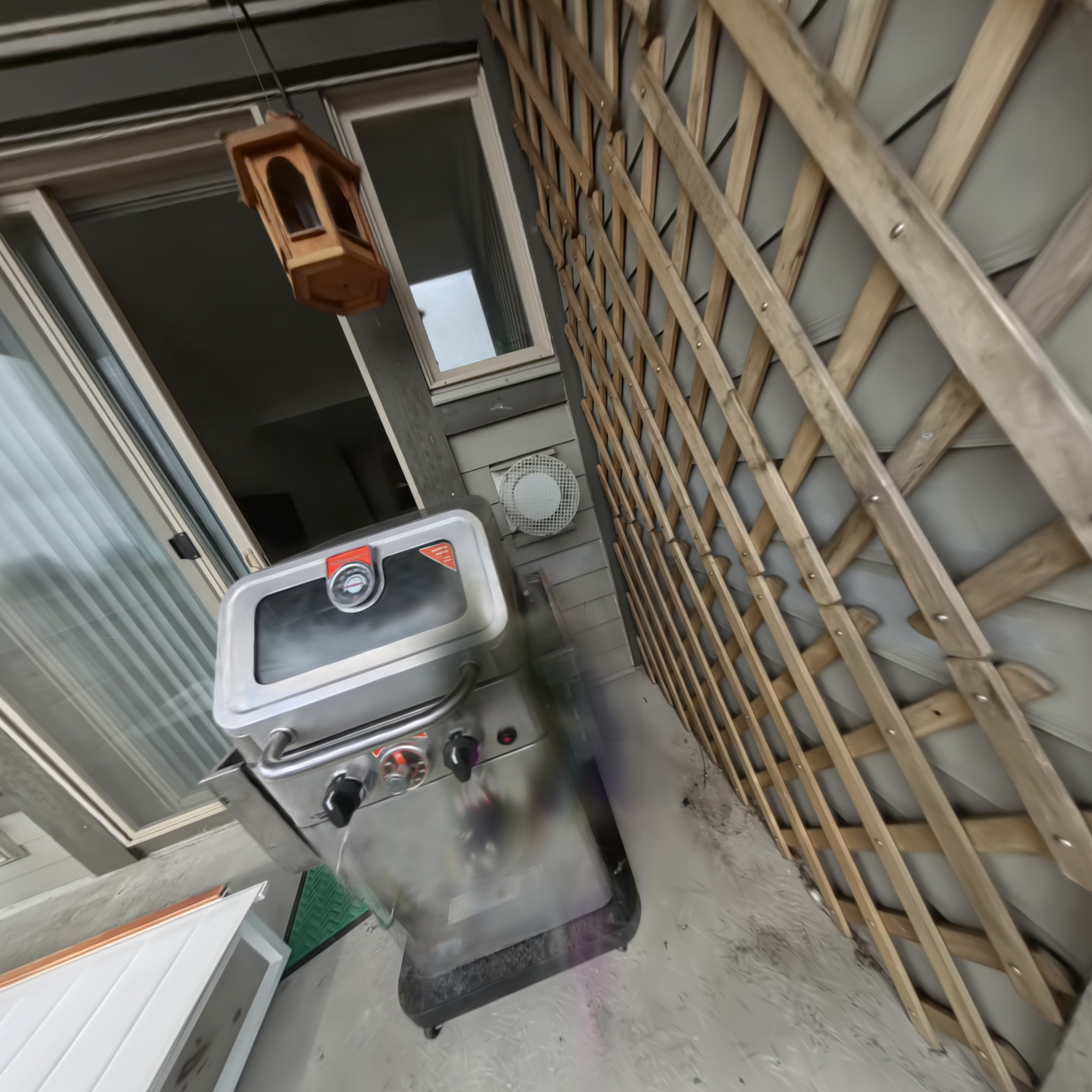} &
        \includegraphics[width=0.18\textwidth]{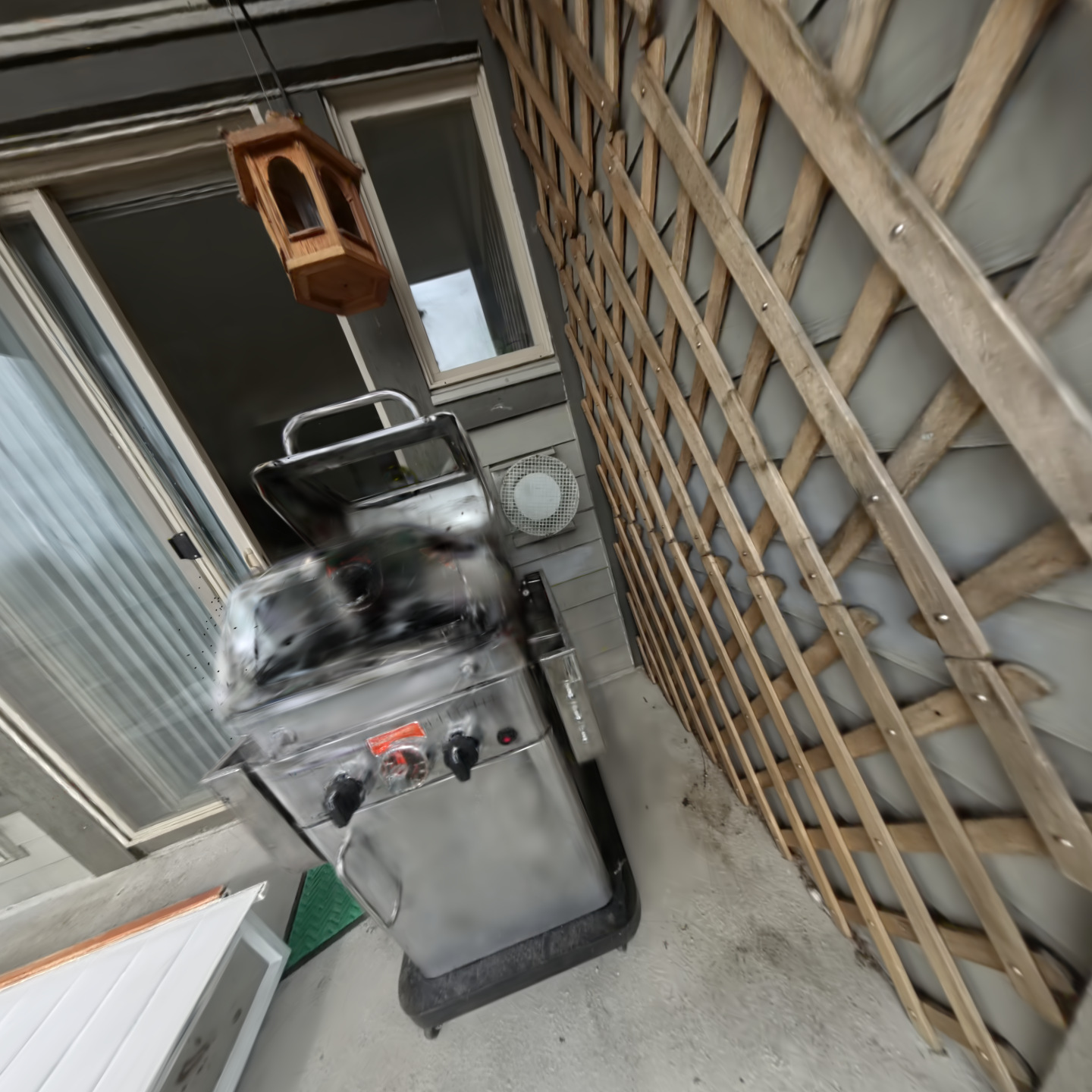} &
        \includegraphics[width=0.18\textwidth]{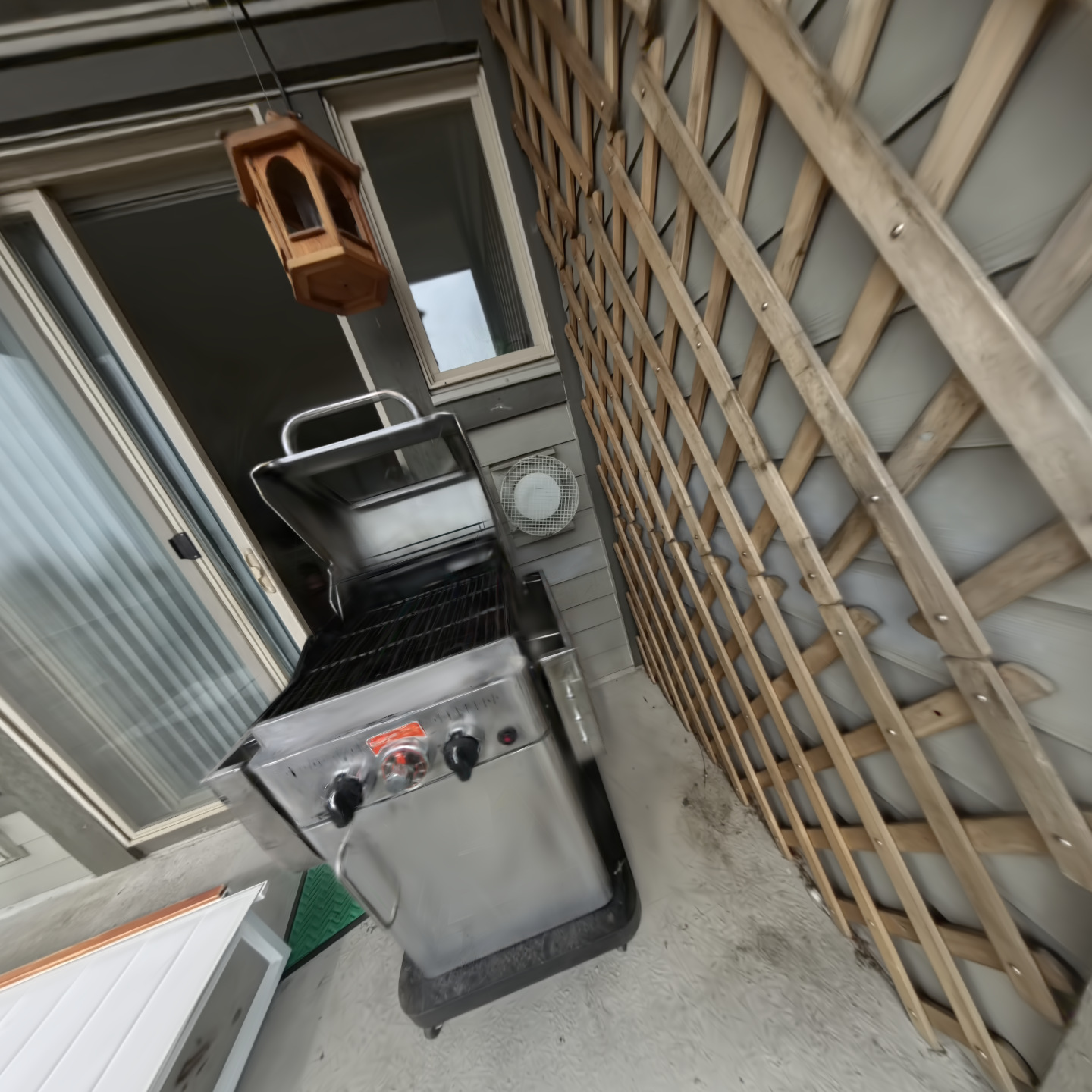} &
        \includegraphics[width=0.18\textwidth]{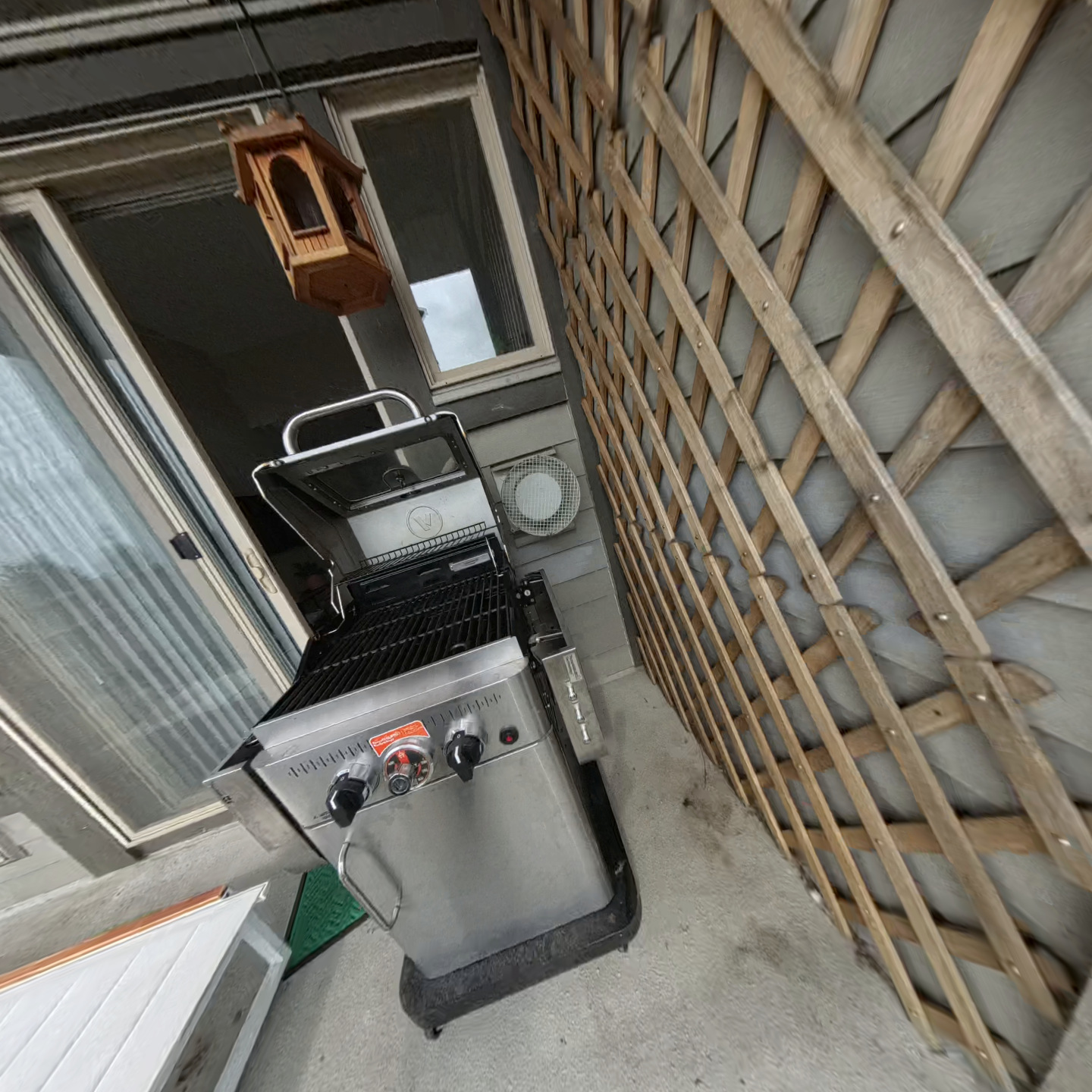} \\

        \includegraphics[width=0.18\textwidth]{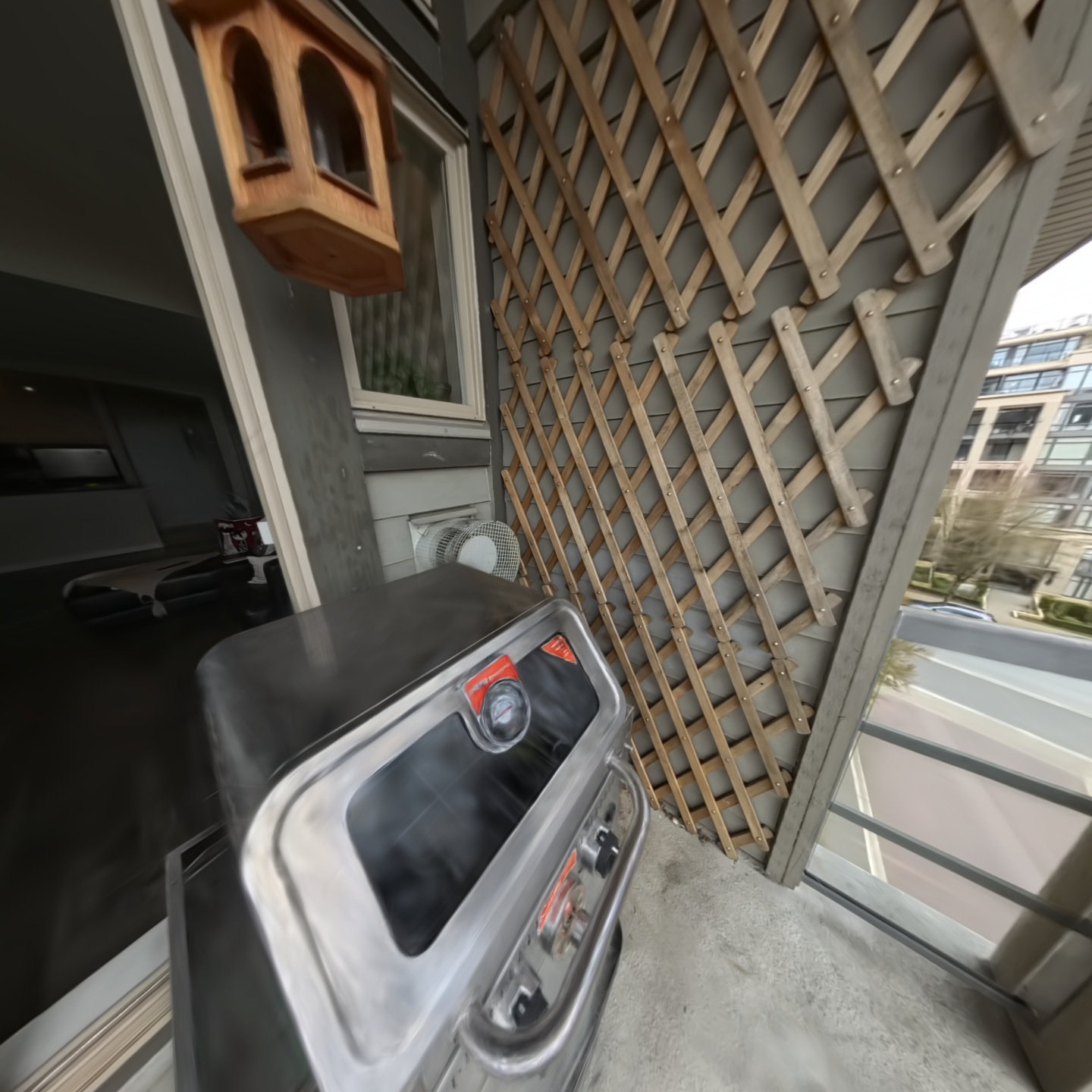} &
        \includegraphics[width=0.18\textwidth]{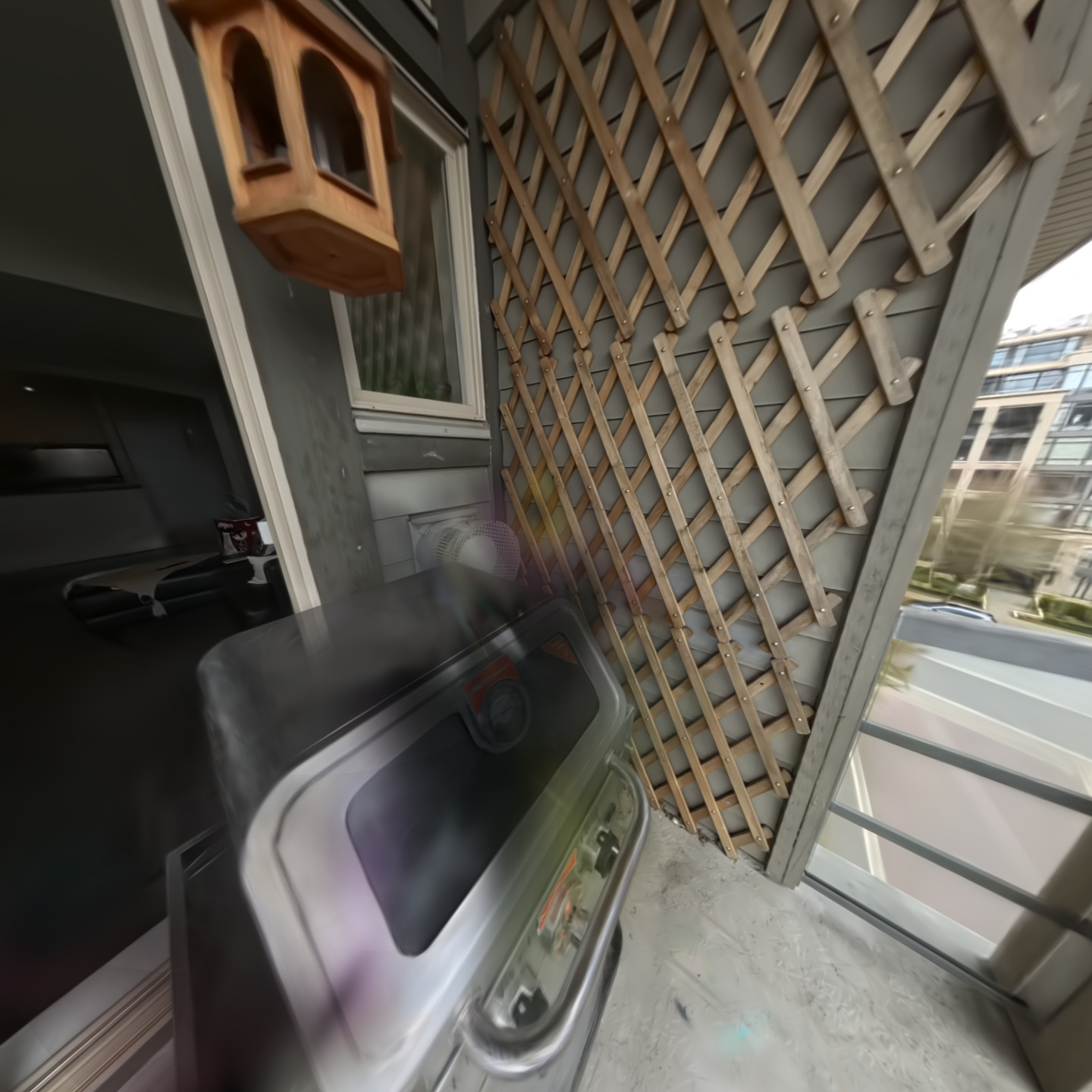} &
        \includegraphics[width=0.18\textwidth]{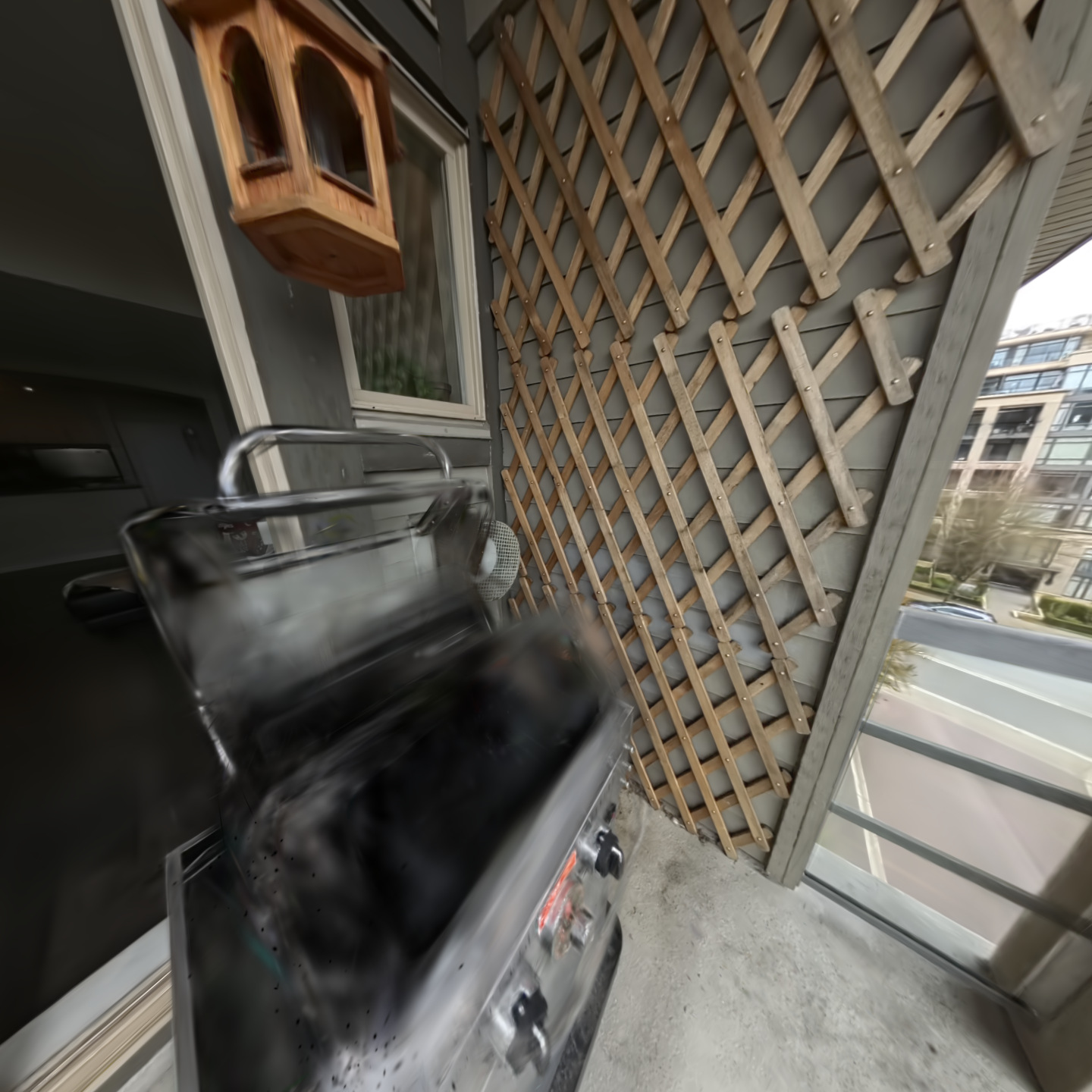} &
        \includegraphics[width=0.18\textwidth]{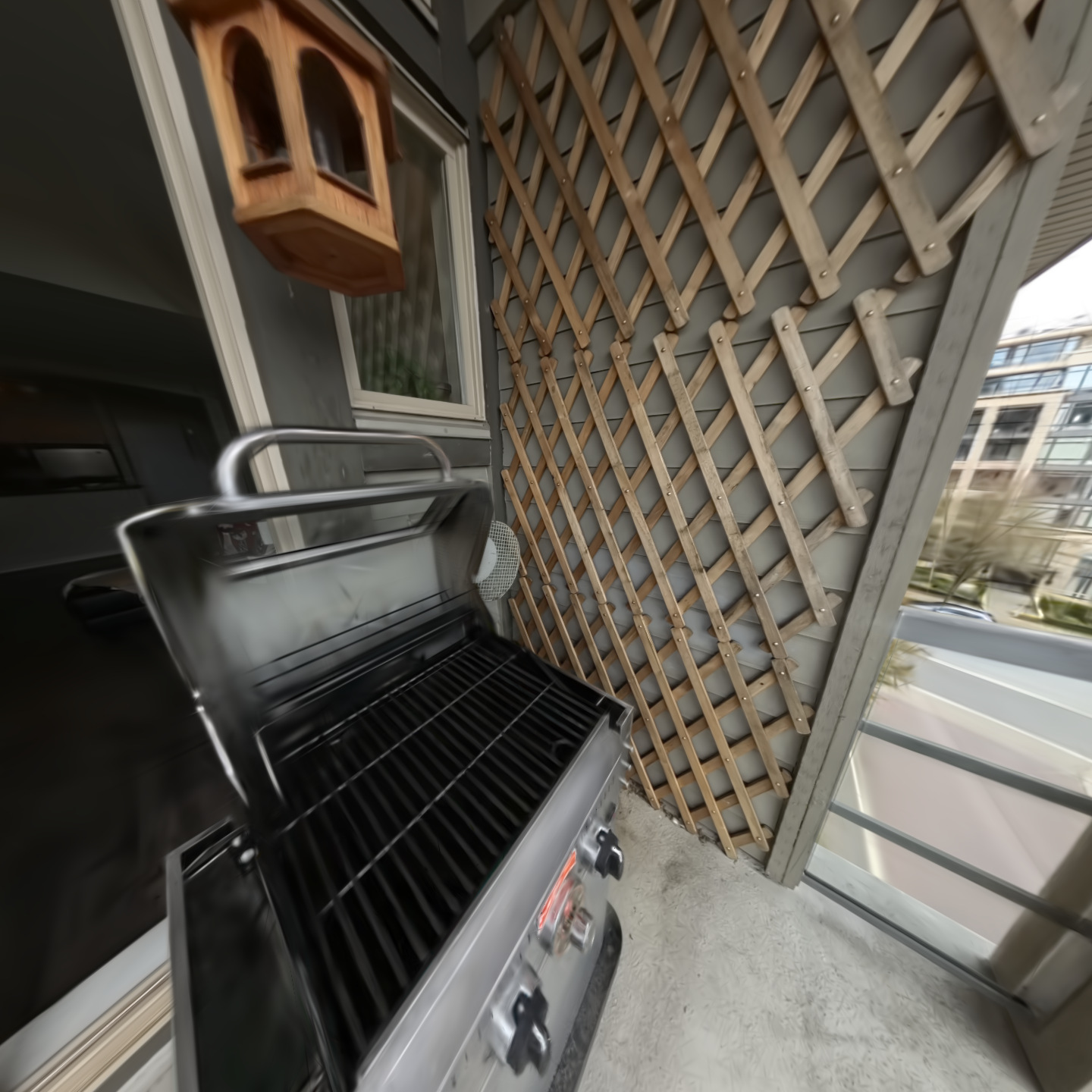} &
        \includegraphics[width=0.18\textwidth]{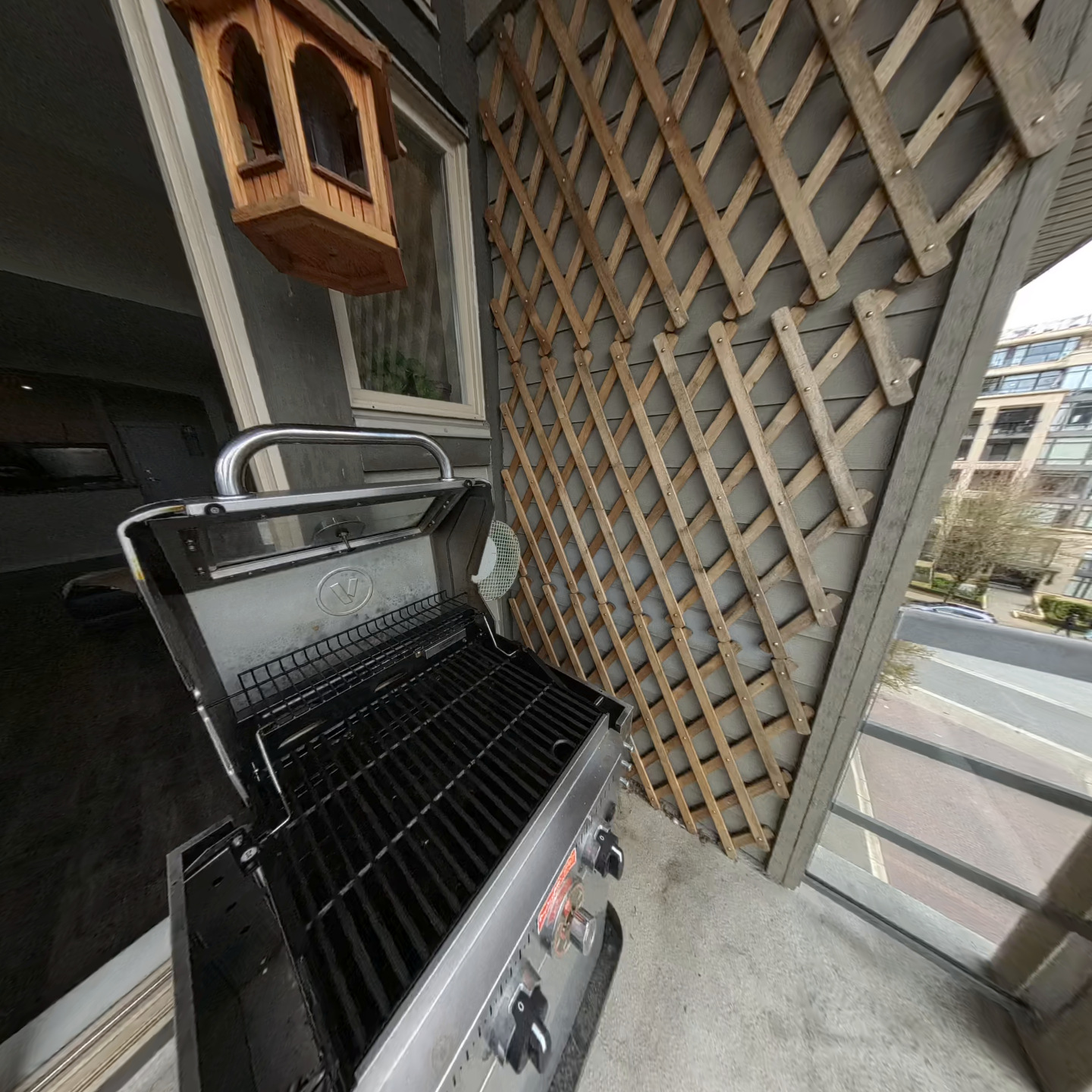} \\

        \includegraphics[width=0.18\textwidth]{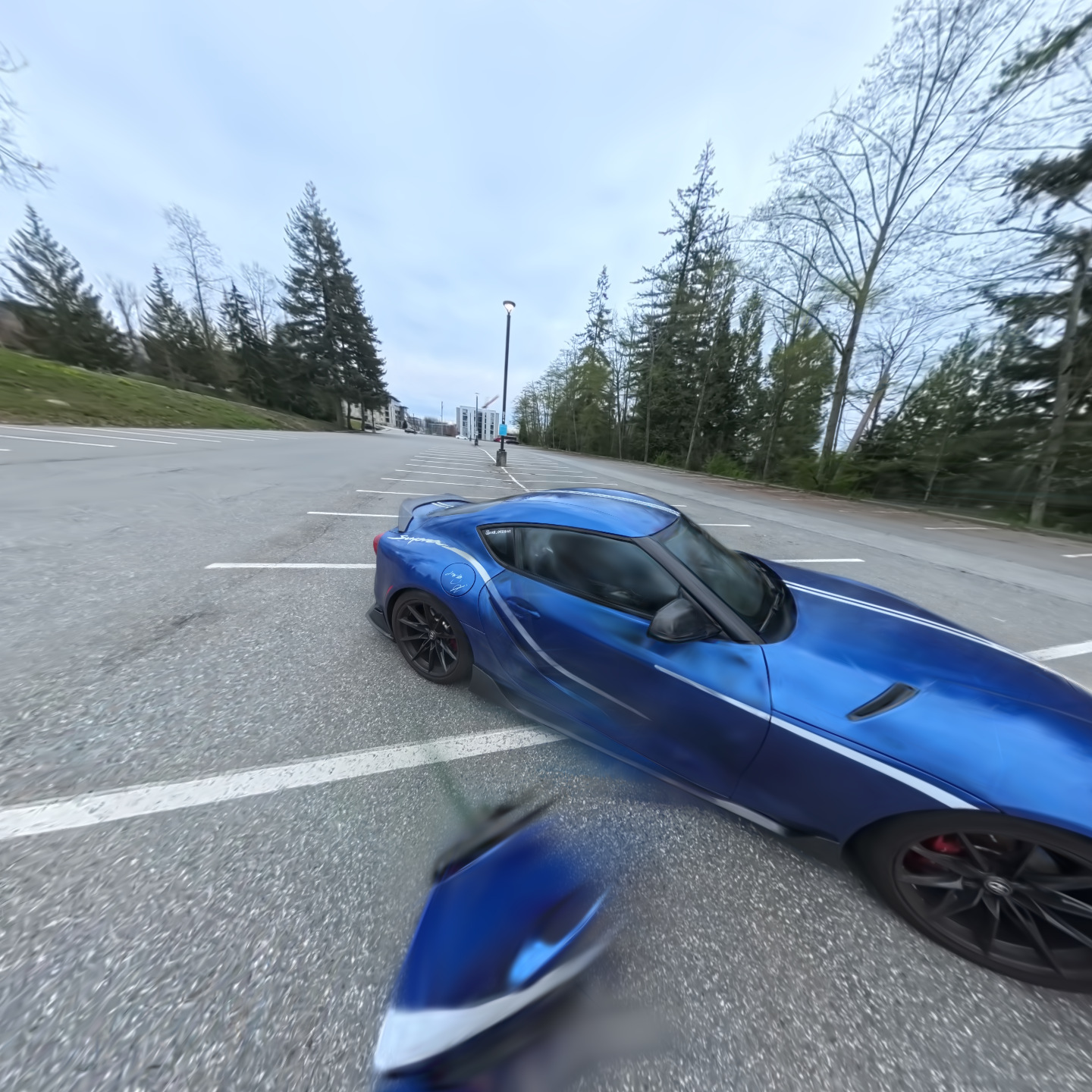} &
        \includegraphics[width=0.18\textwidth]{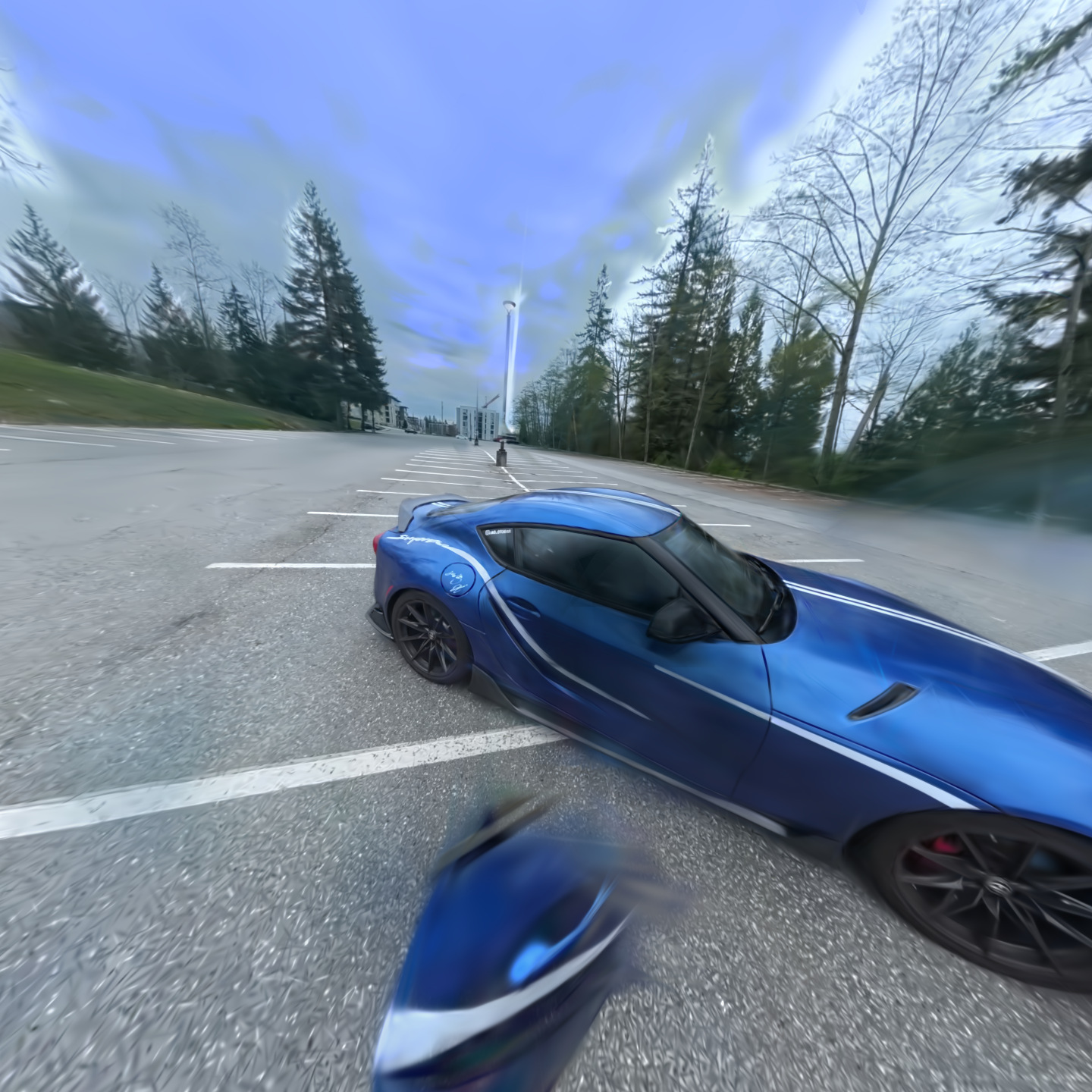} &
        \includegraphics[width=0.18\textwidth]{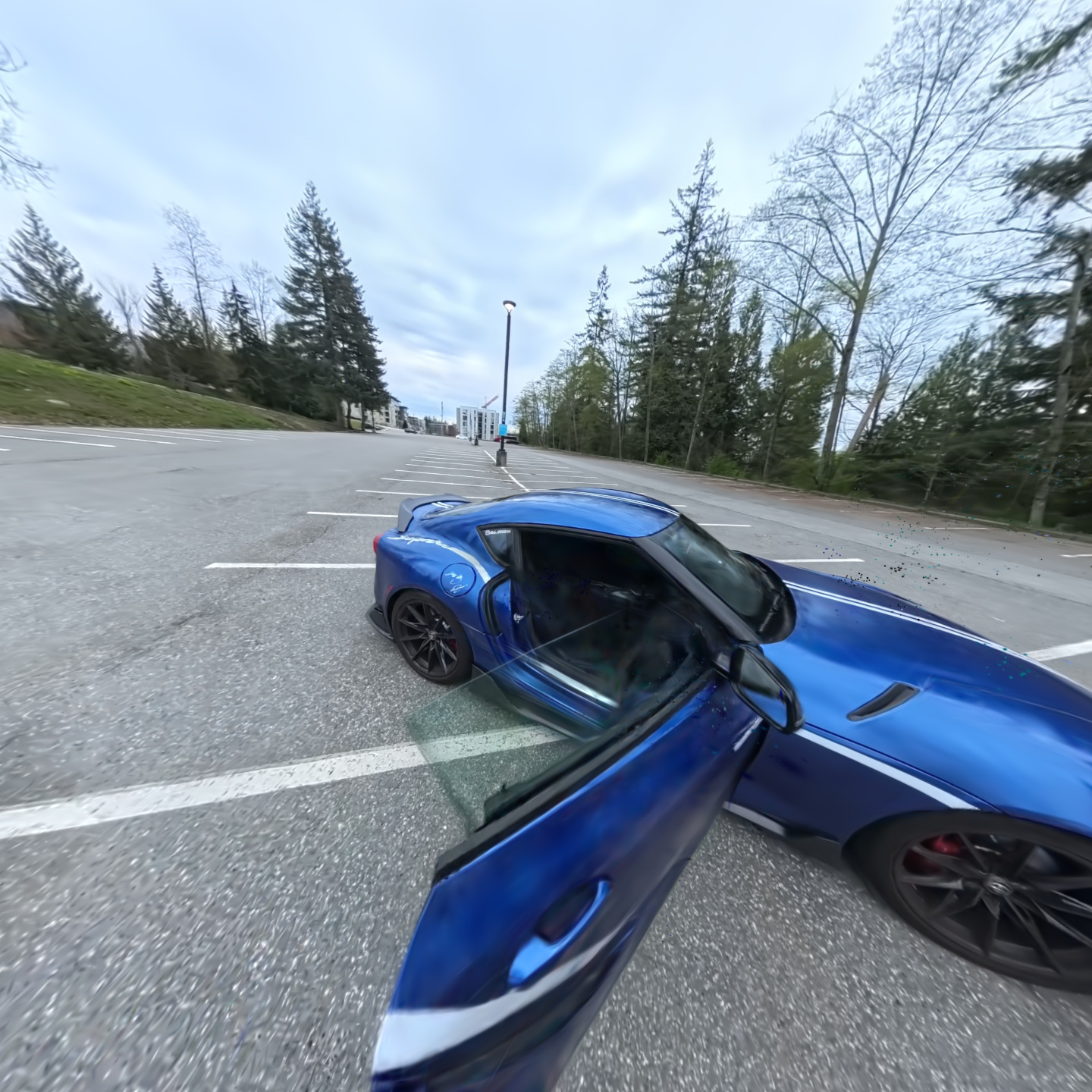} &
        \includegraphics[width=0.18\textwidth]{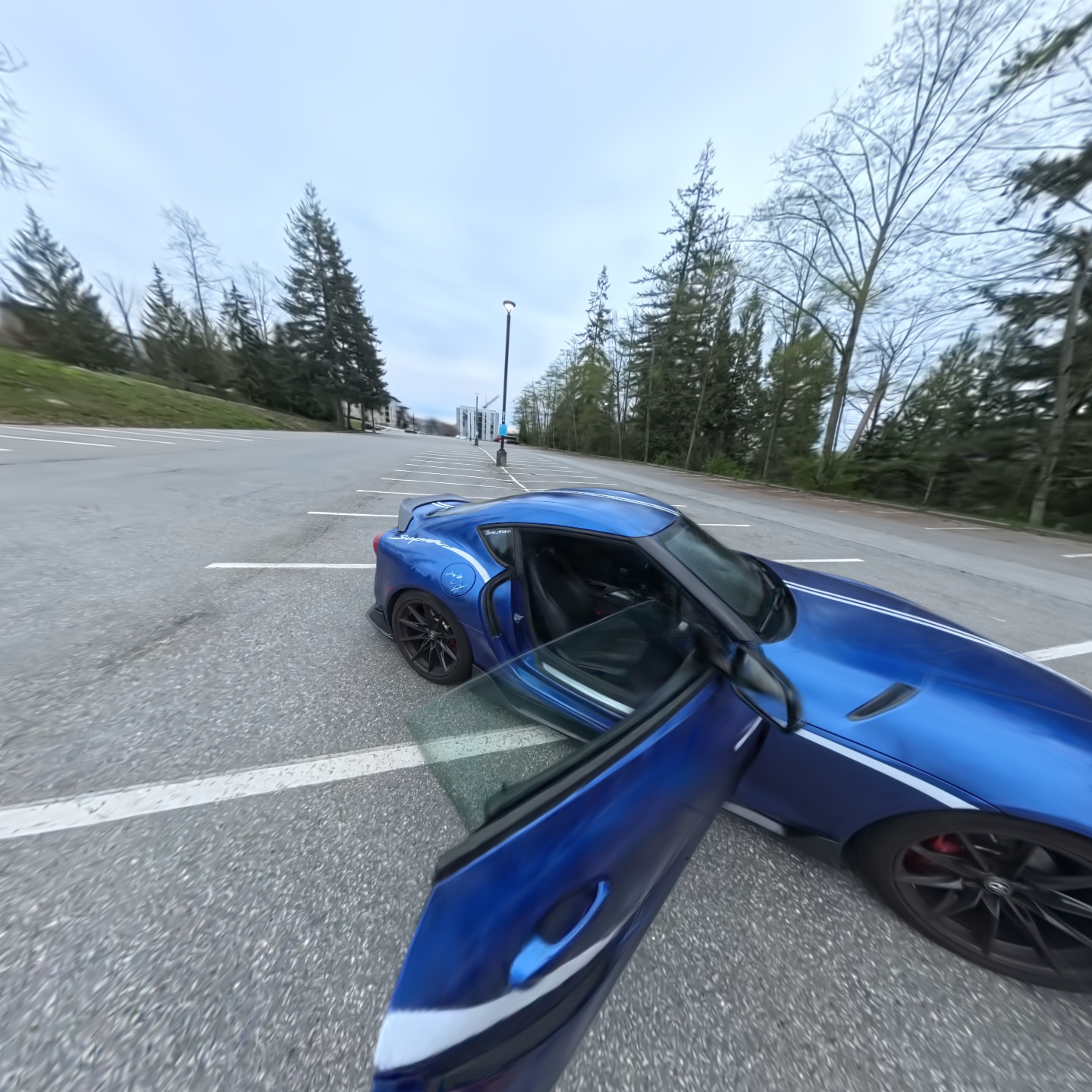} &
        \includegraphics[width=0.18\textwidth]{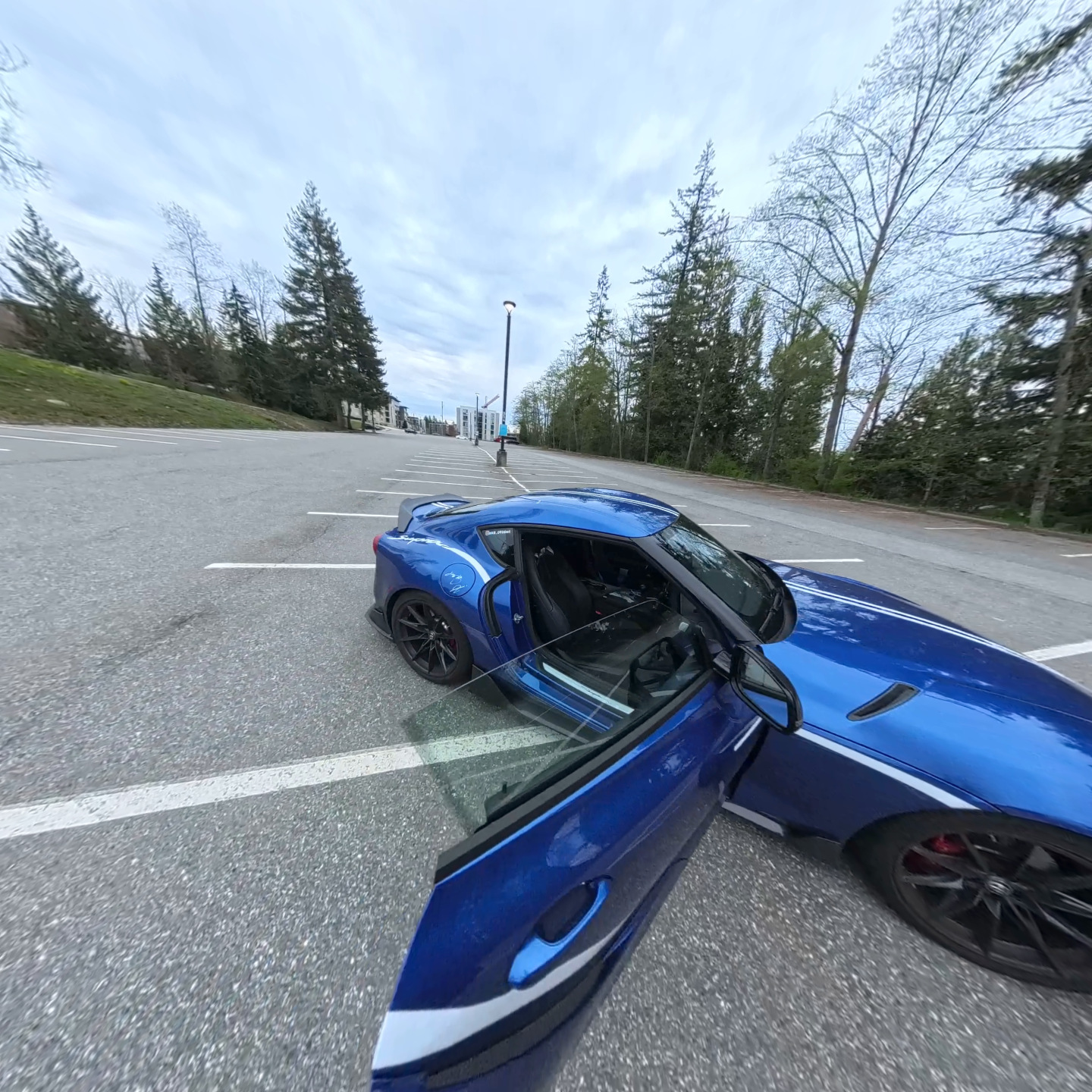} \\

        \includegraphics[width=0.18\textwidth]{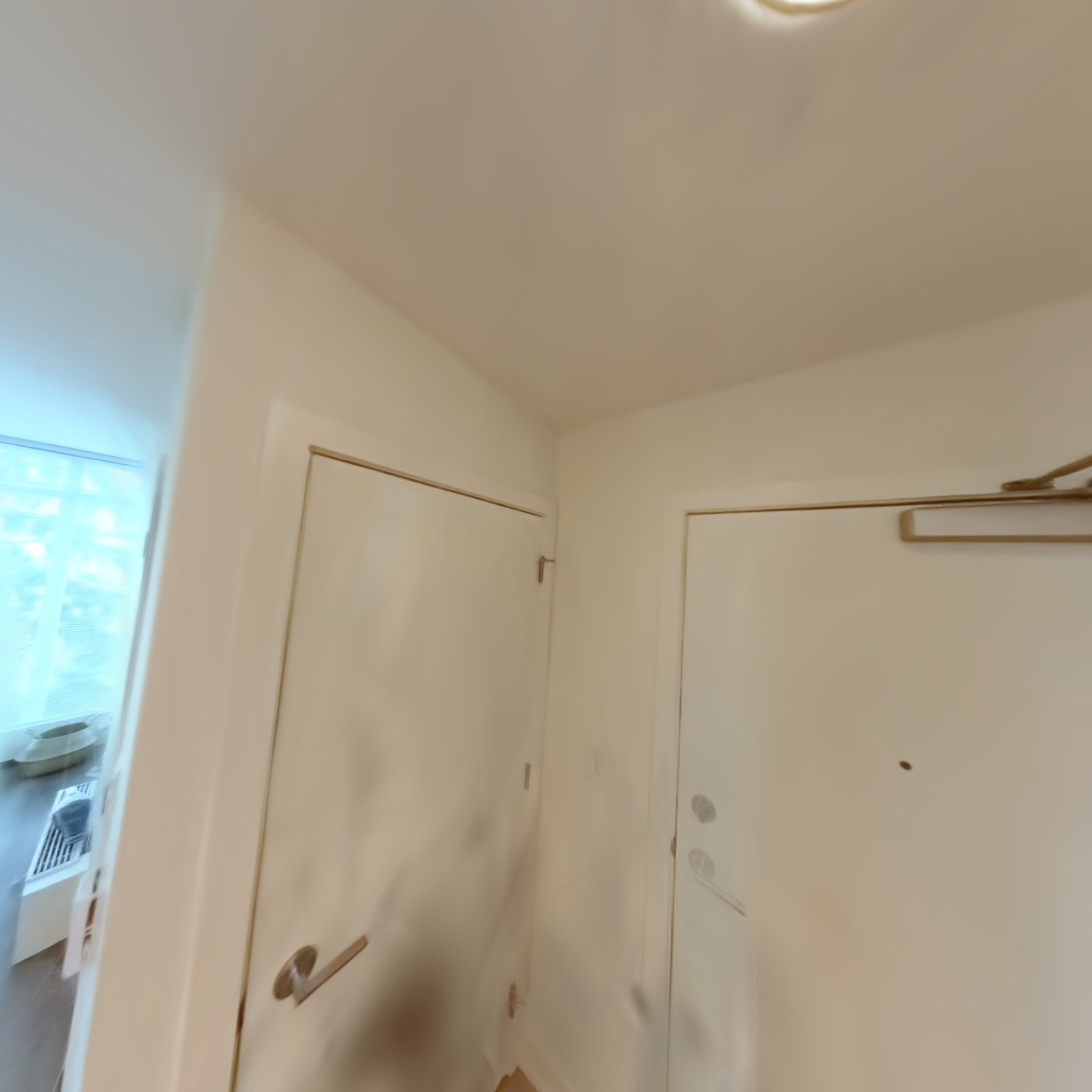} &
        \includegraphics[width=0.18\textwidth]{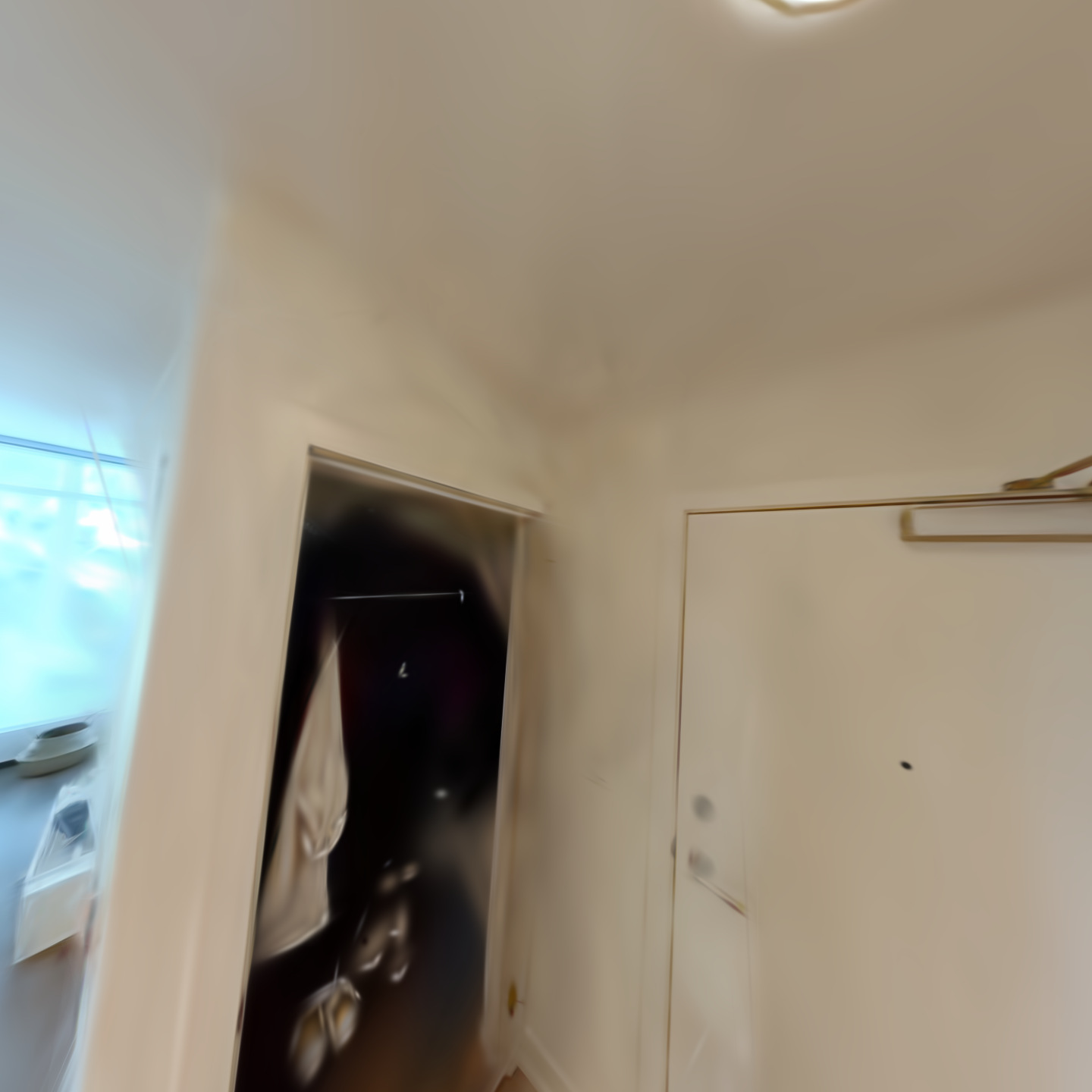} &
        \includegraphics[width=0.18\textwidth]{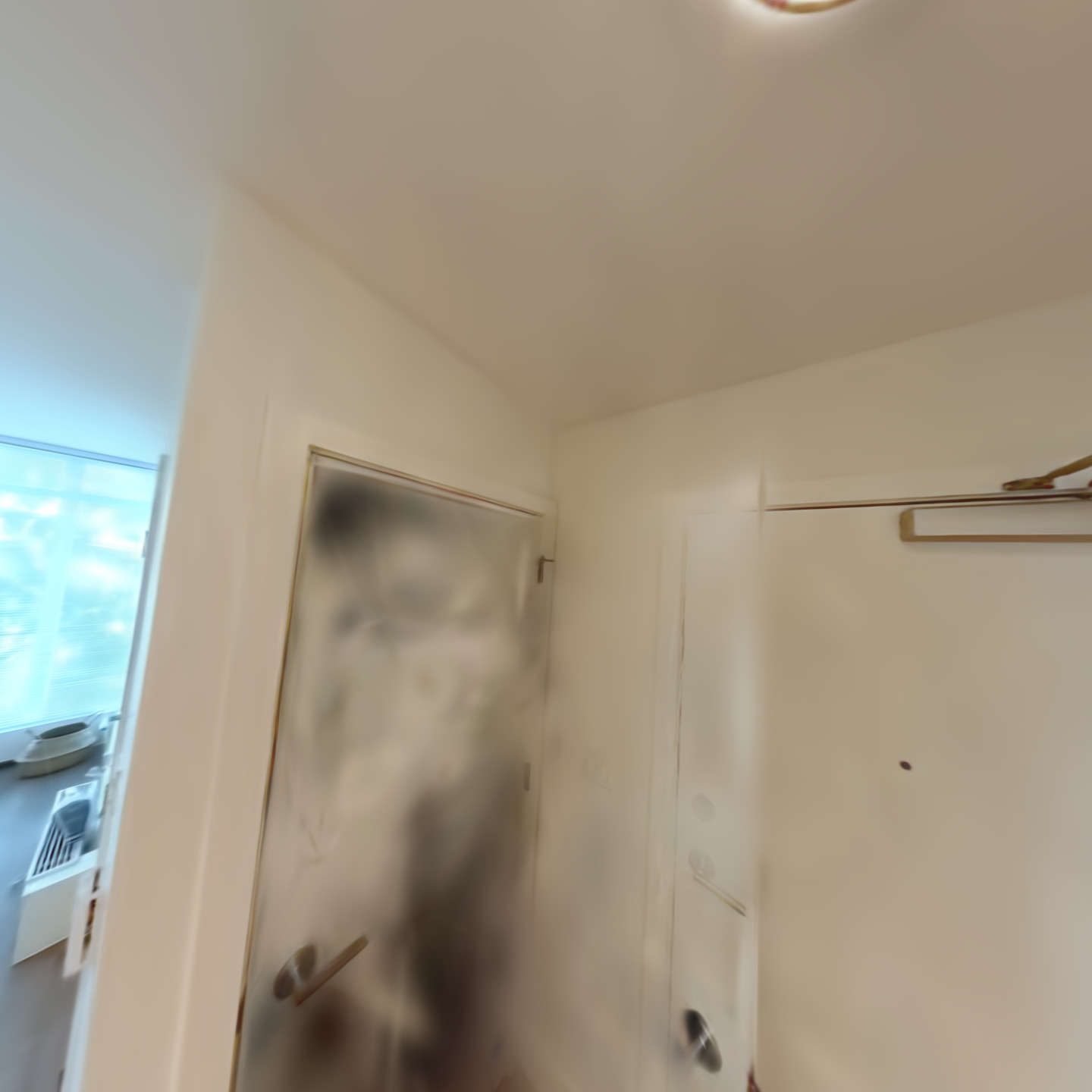} &
        \includegraphics[width=0.18\textwidth]{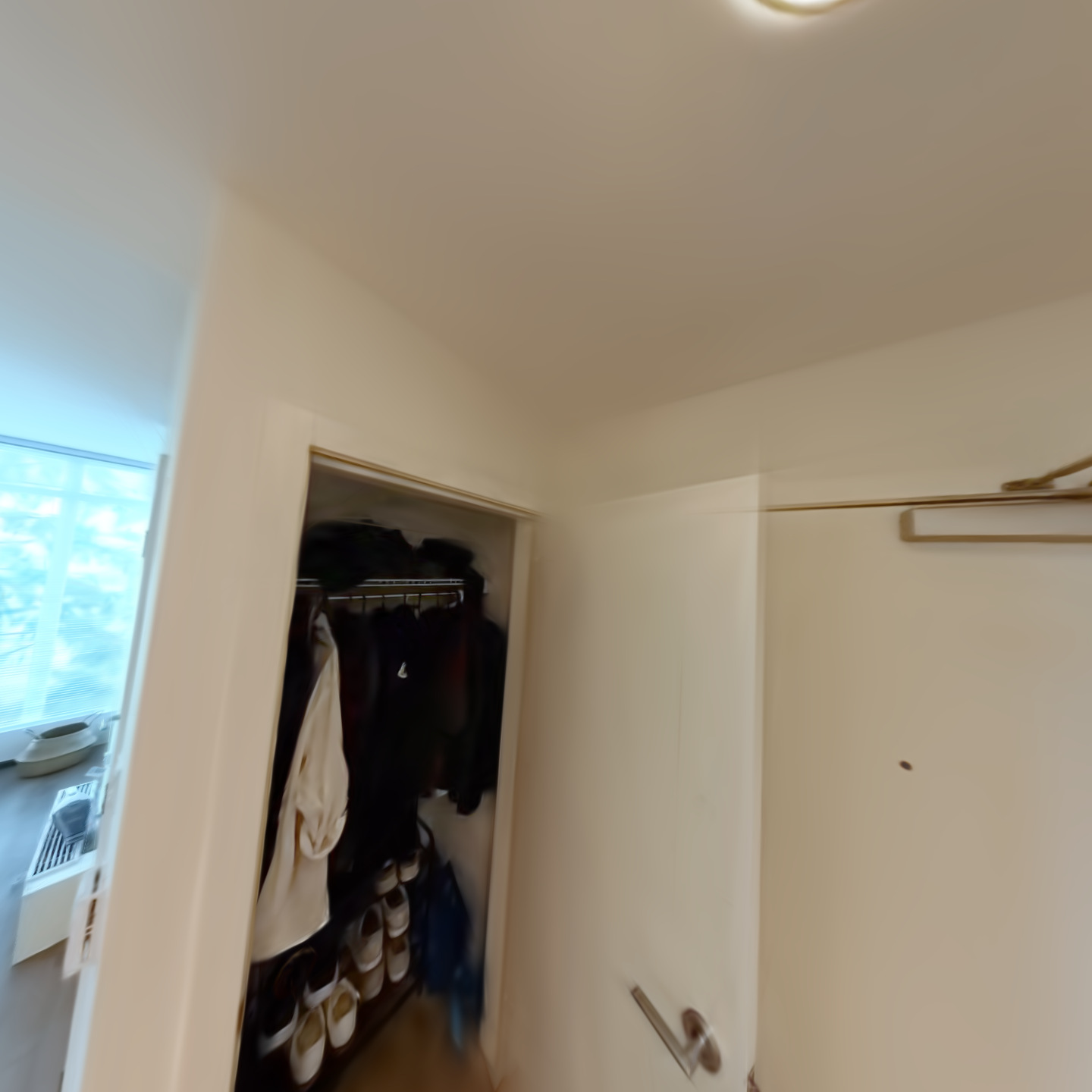} &
        \includegraphics[width=0.18\textwidth]{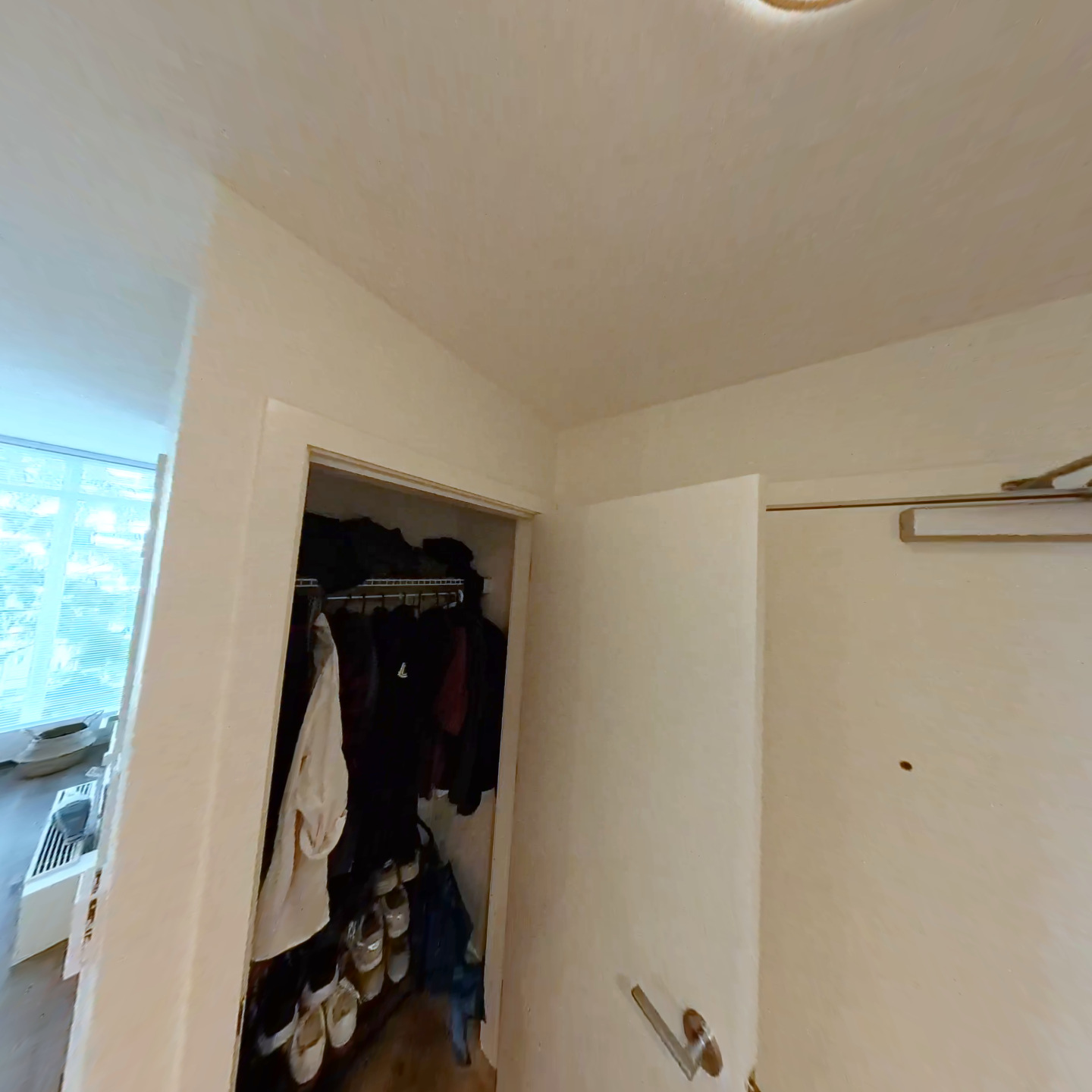} \\

        \includegraphics[width=0.18\textwidth]{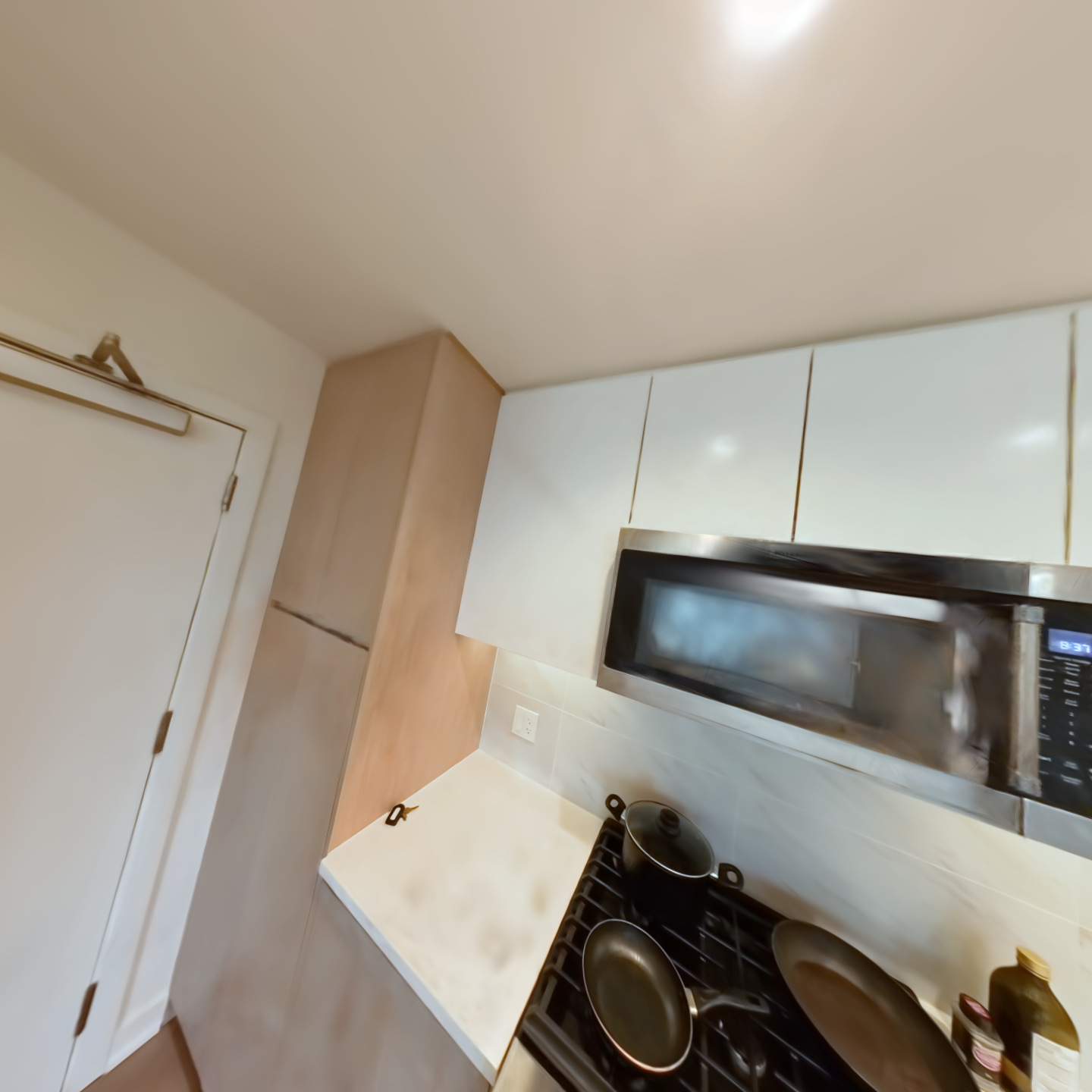} &
        \includegraphics[width=0.18\textwidth]{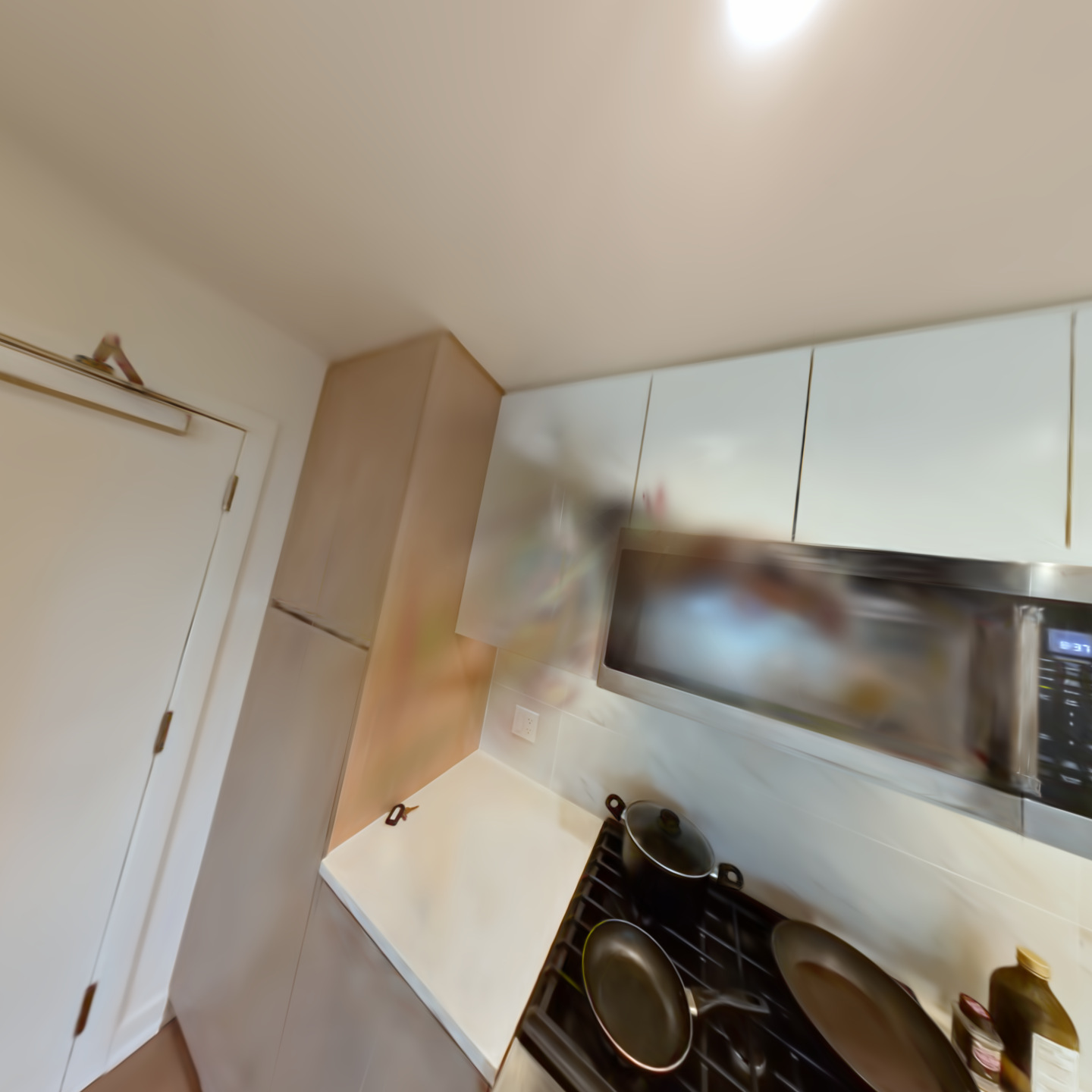} &
        \includegraphics[width=0.18\textwidth]{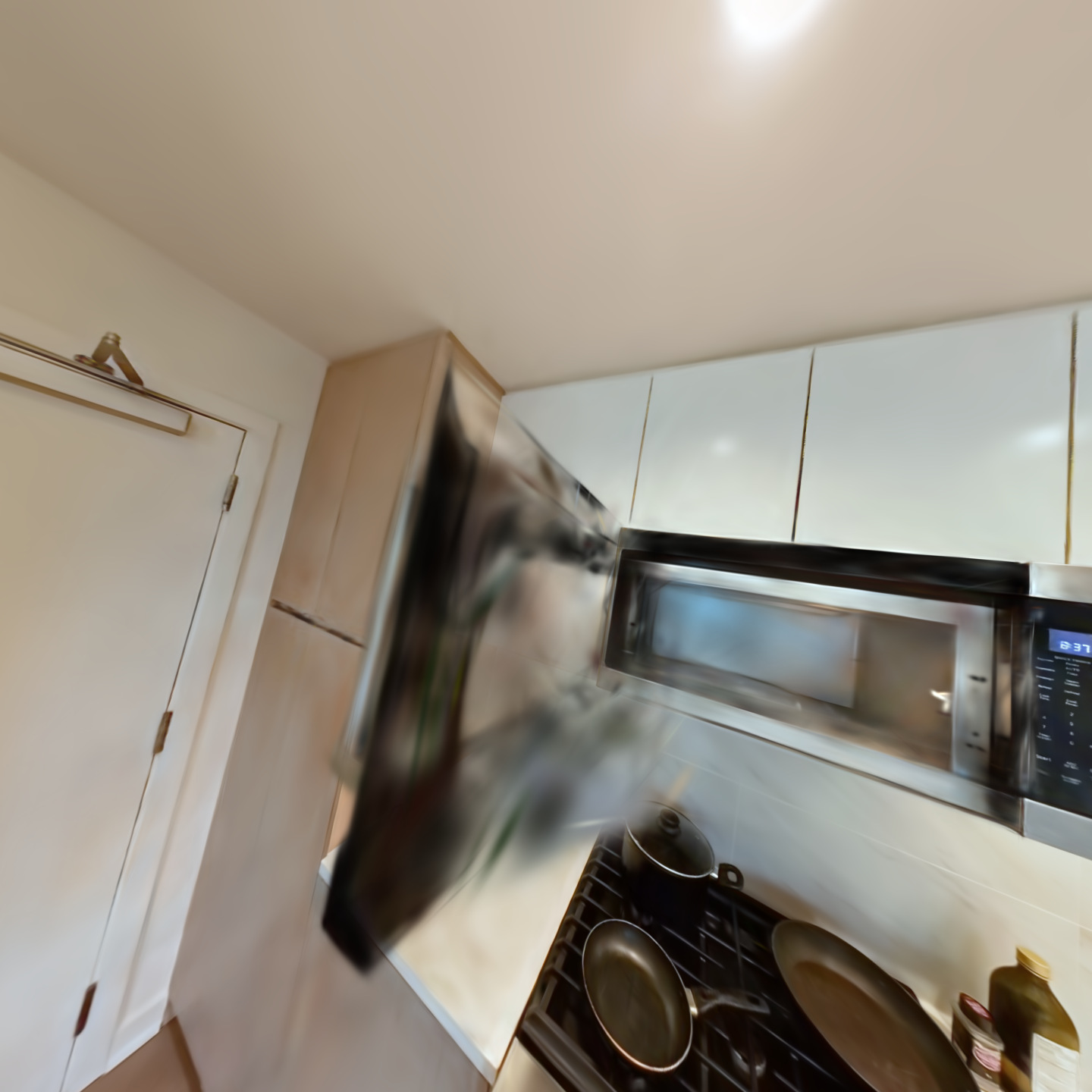} &
        \includegraphics[width=0.18\textwidth]{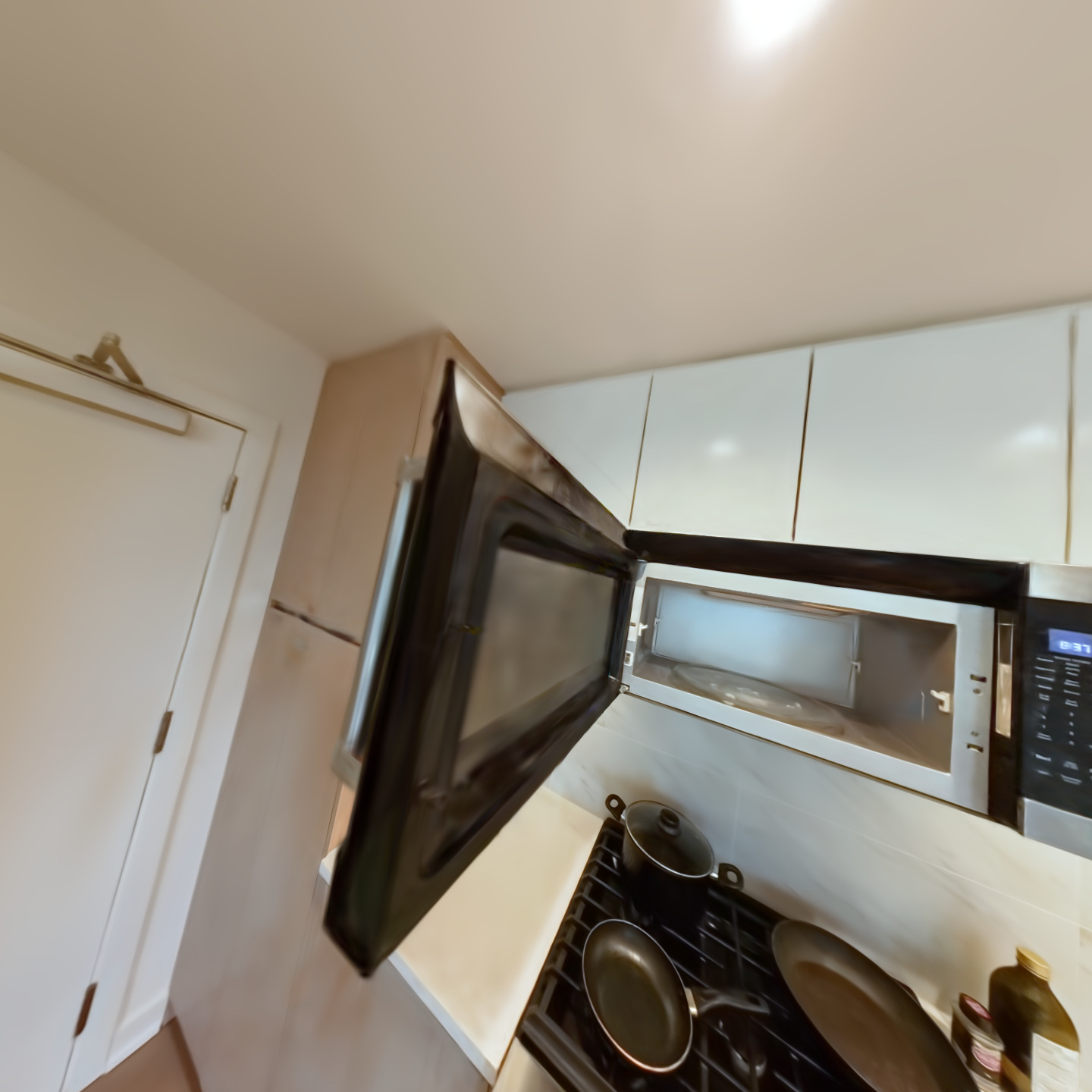} &
        \includegraphics[width=0.18\textwidth]{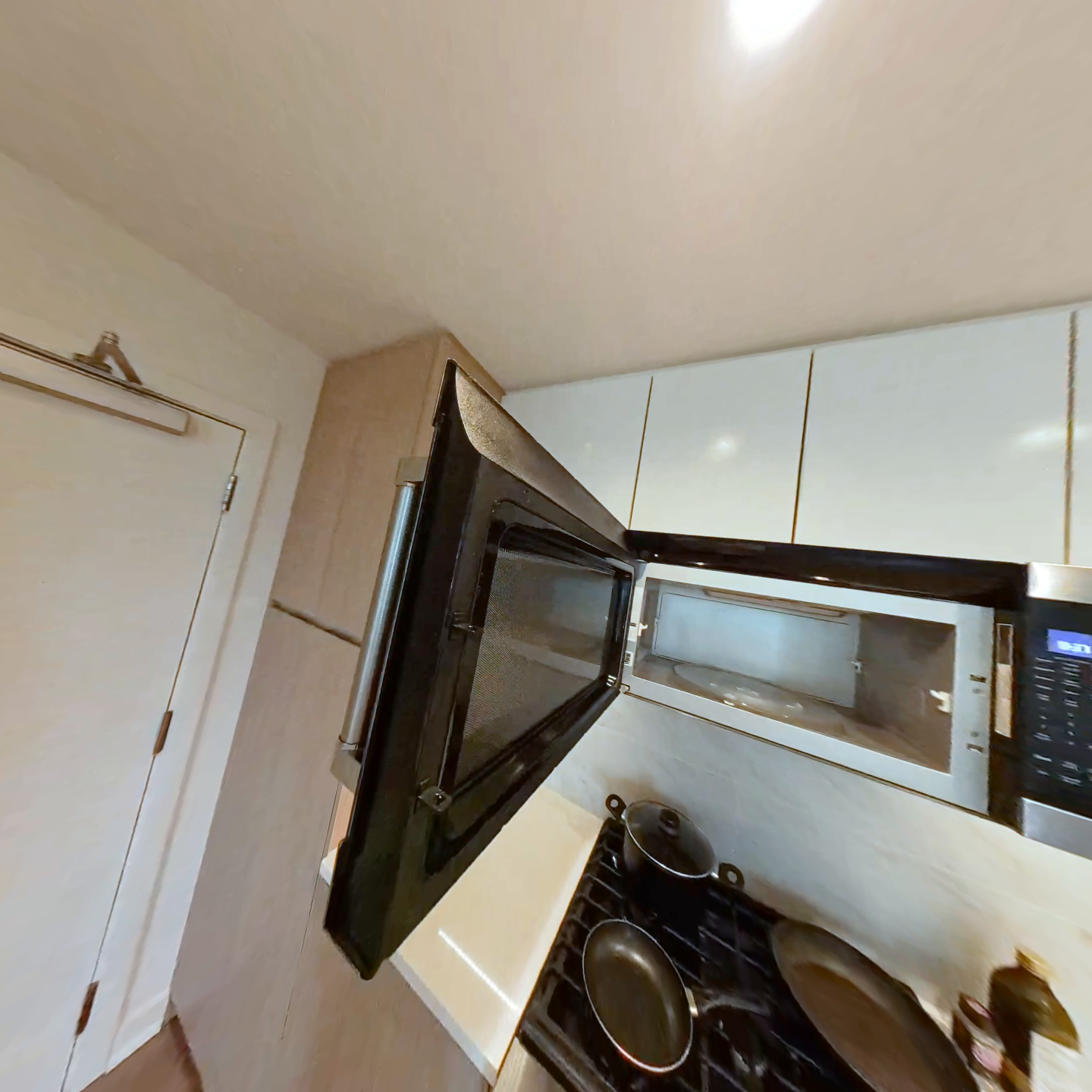} \\

        \includegraphics[width=0.18\textwidth]{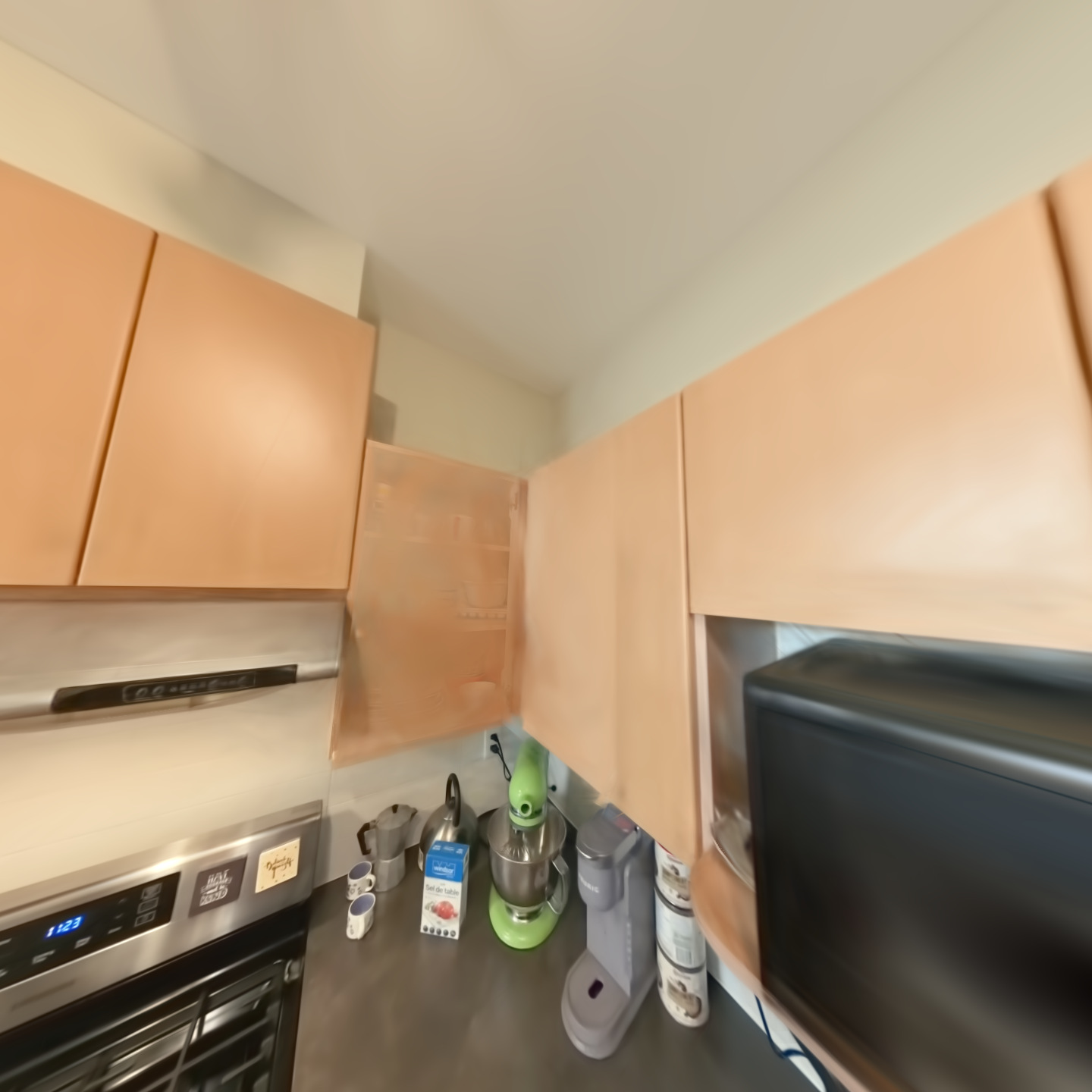} &
        \includegraphics[width=0.18\textwidth]{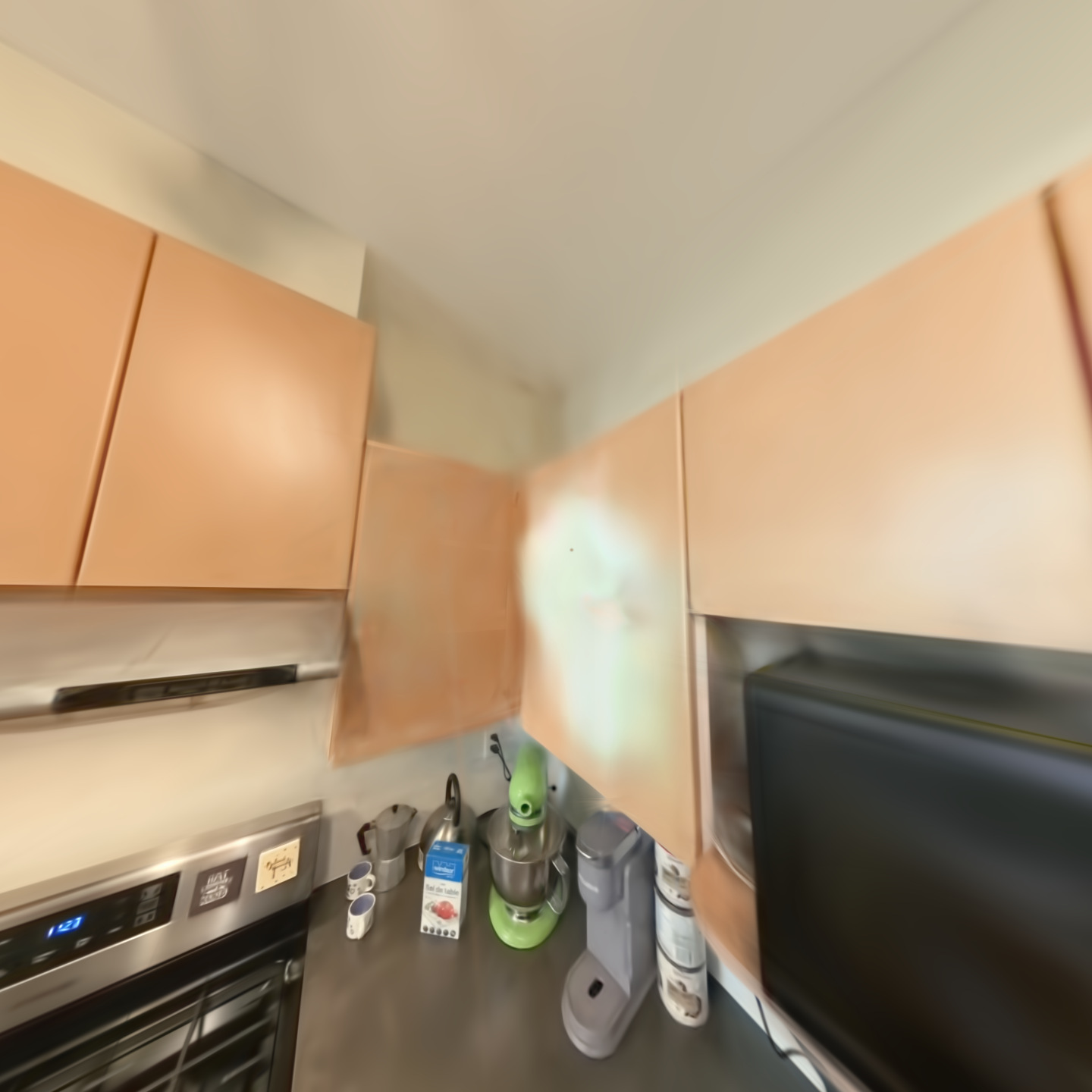} &
        \includegraphics[width=0.18\textwidth]{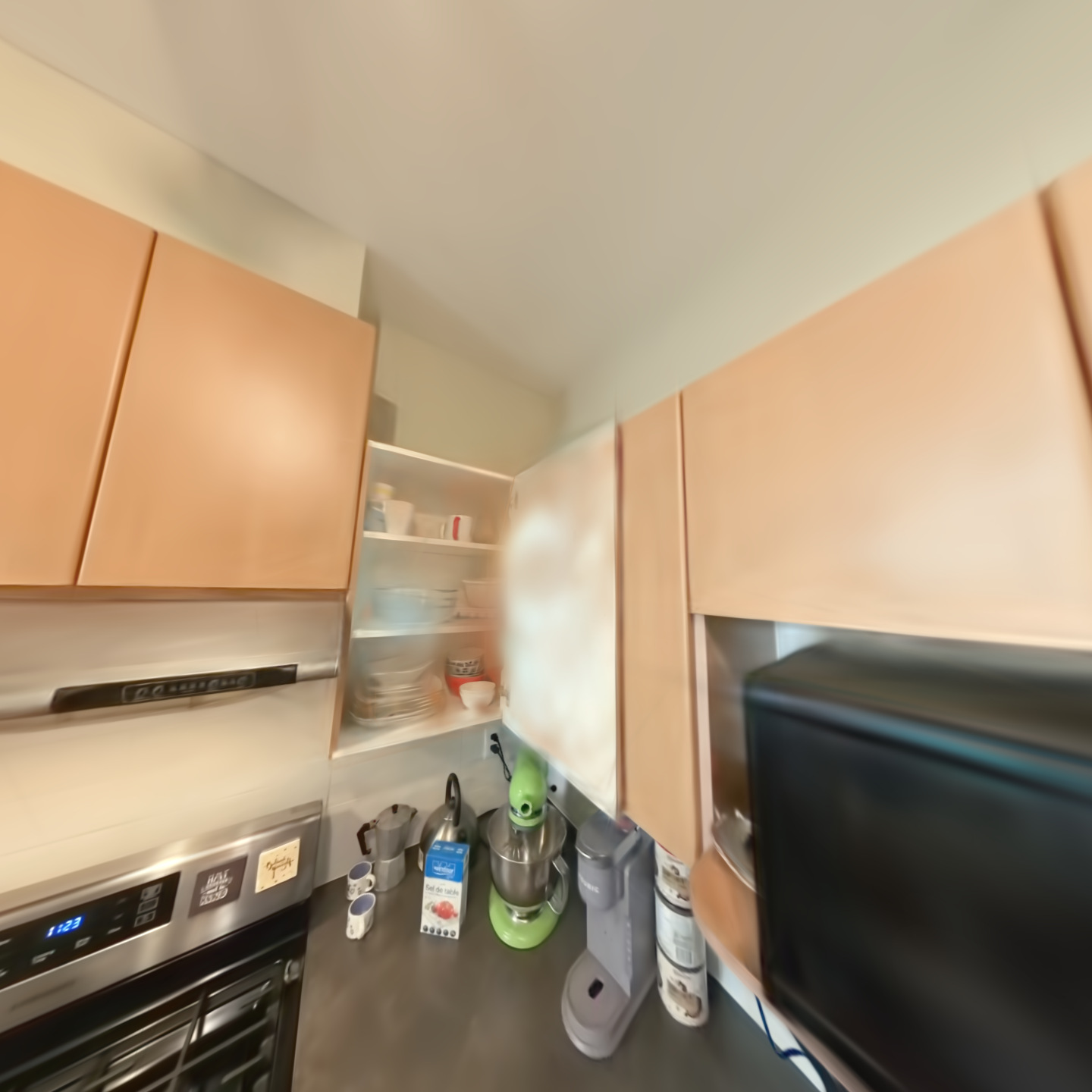} &
        \includegraphics[width=0.18\textwidth]{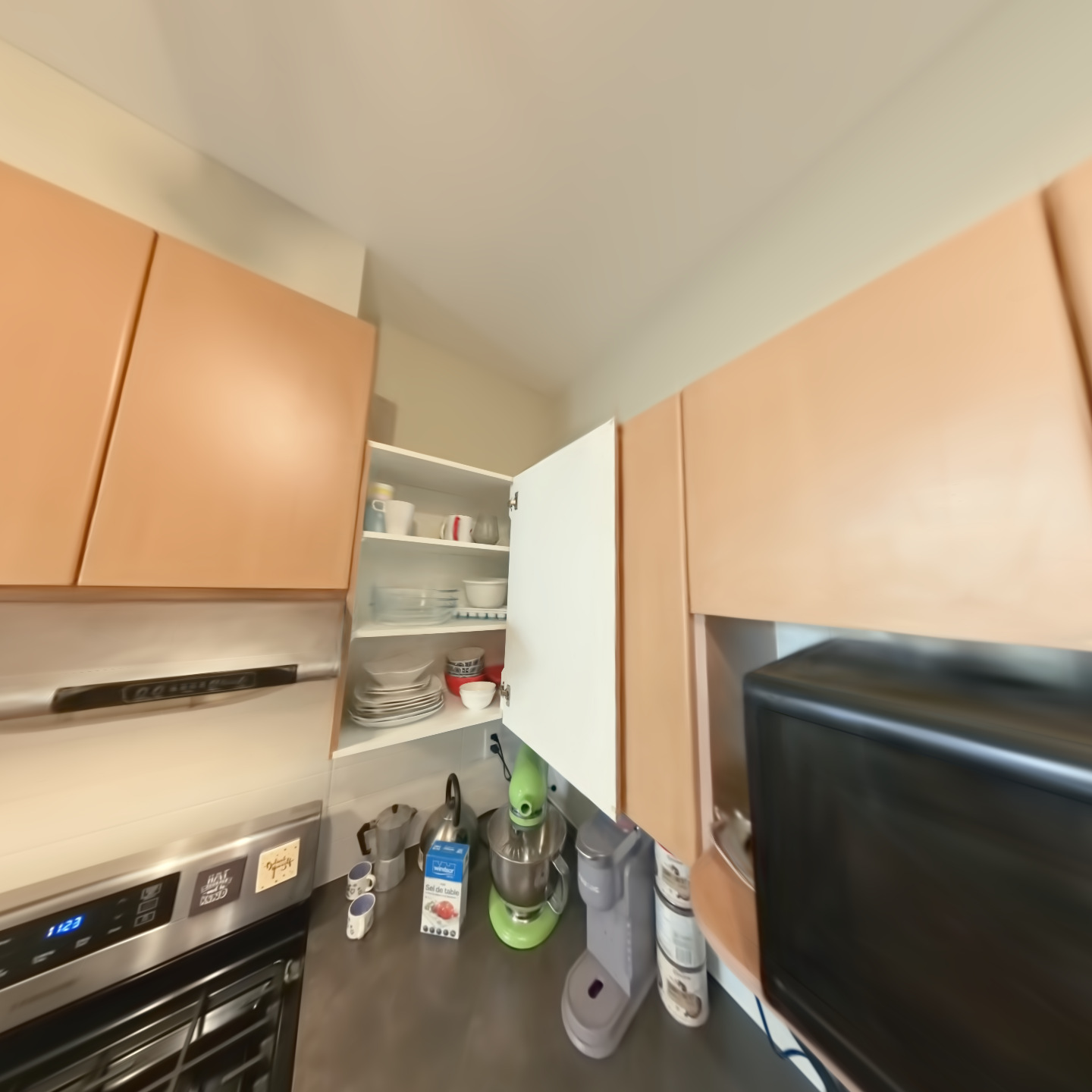} &
        \includegraphics[width=0.18\textwidth]{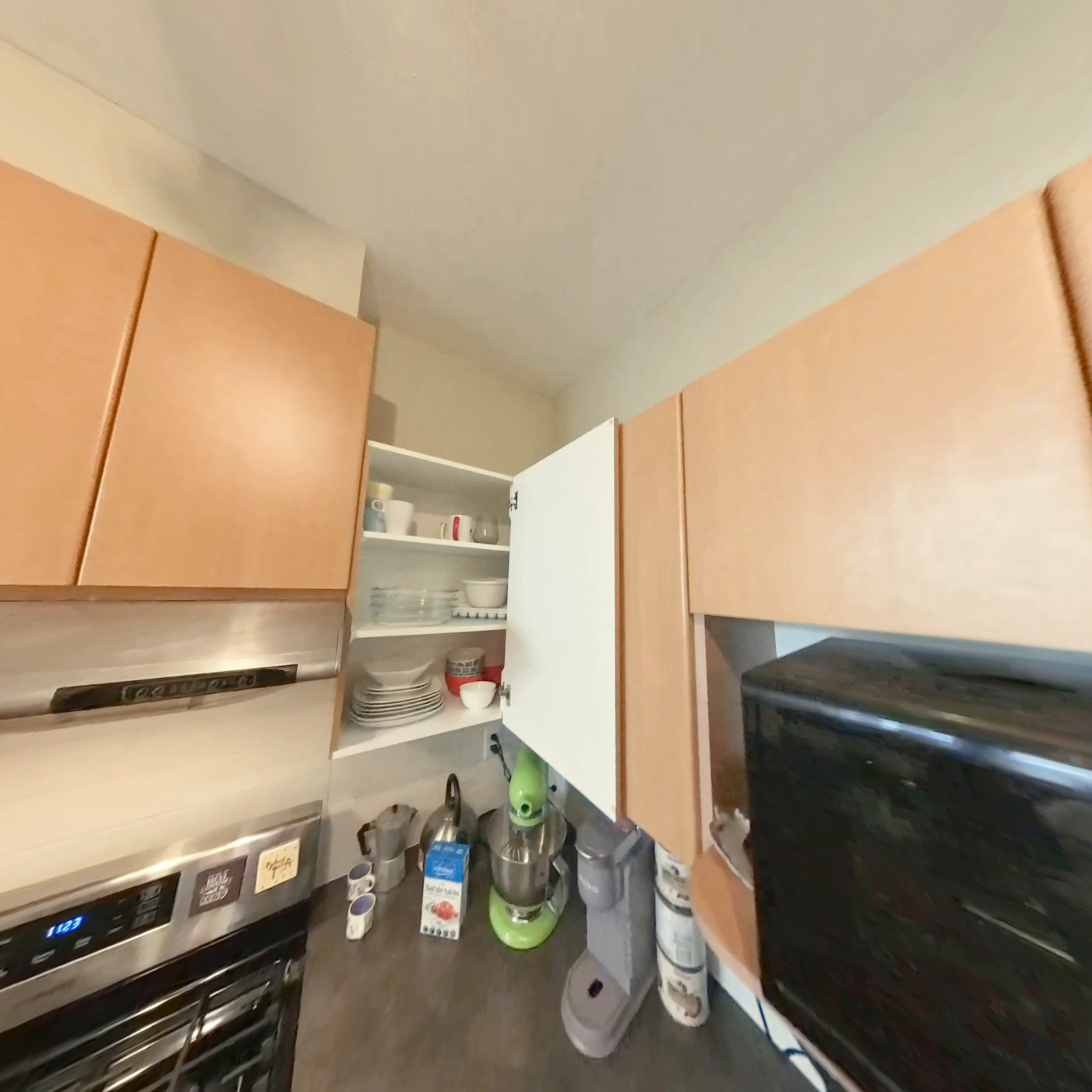} \\
    \end{tabular}
    \caption{\textbf{Qualitative comparison.}
    We compare FullCircle, DeGauss, 4DGS, and SwitchSplat against ground-truth images from standard held-out test views.
    Scenes are shown from top to bottom: Balcony, Balcony, Car, Flat 2, Flat 2, and Flat 3.}
    \Description{A six-row, five-column image grid. Columns show FullCircle, DeGauss, 4DGS, SwitchSplat, and ground truth. From top to bottom, the rows contain two views of an outdoor grill on a balcony, a blue car with an open side door, an open closet, an open microwave, and an open kitchen cabinet. In the interacting object regions, the baseline renderings often contain blur, floating artifacts, or incomplete geometry; the SwitchSplat renderings more closely match the crisp object states and surrounding scene in the ground truth.}
    \label{fig:comparison1}
\end{figure*}

\begin{figure*}[t]
    \centering
    \setlength{\tabcolsep}{1pt}\renewcommand{\arraystretch}{1.0}\begin{tabular}{@{}ccccc@{}}
        \footnotesize FullCircle &
        \footnotesize DeGauss &
        \footnotesize 4DGS &
        \footnotesize SwitchSplat &
        \footnotesize Ground Truth \\

        \includegraphics[width=0.18\textwidth]{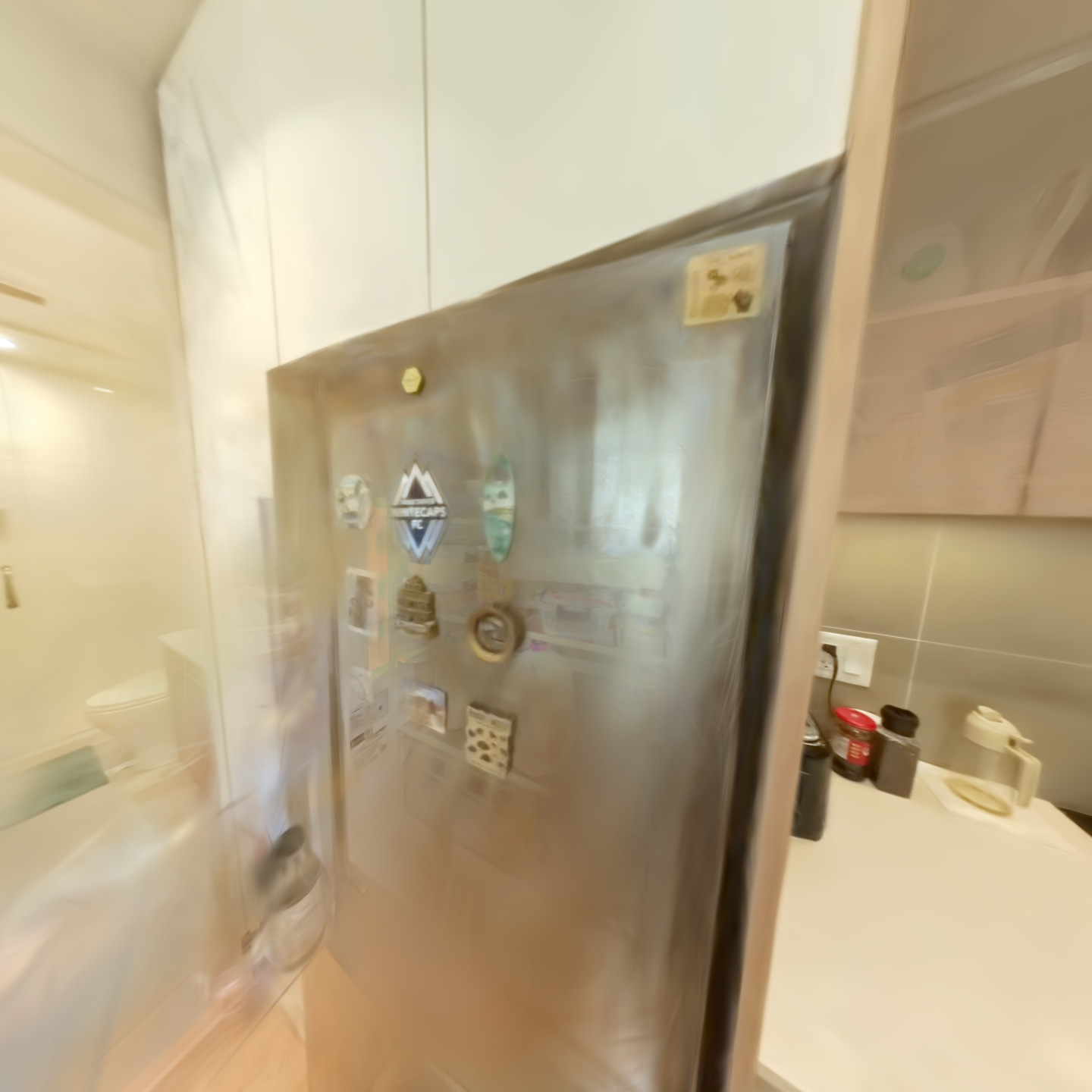} &
        \includegraphics[width=0.18\textwidth]{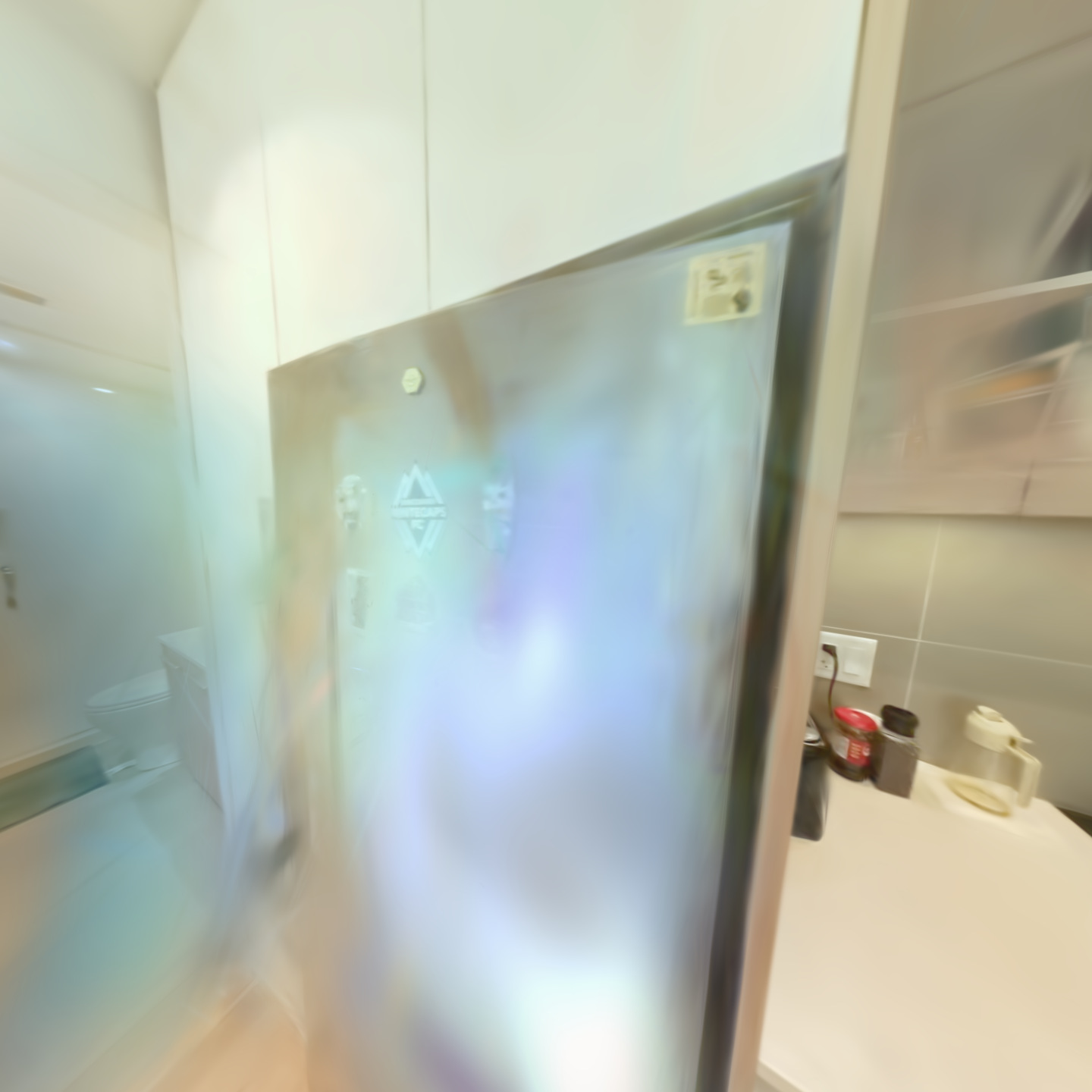} &
        \includegraphics[width=0.18\textwidth]{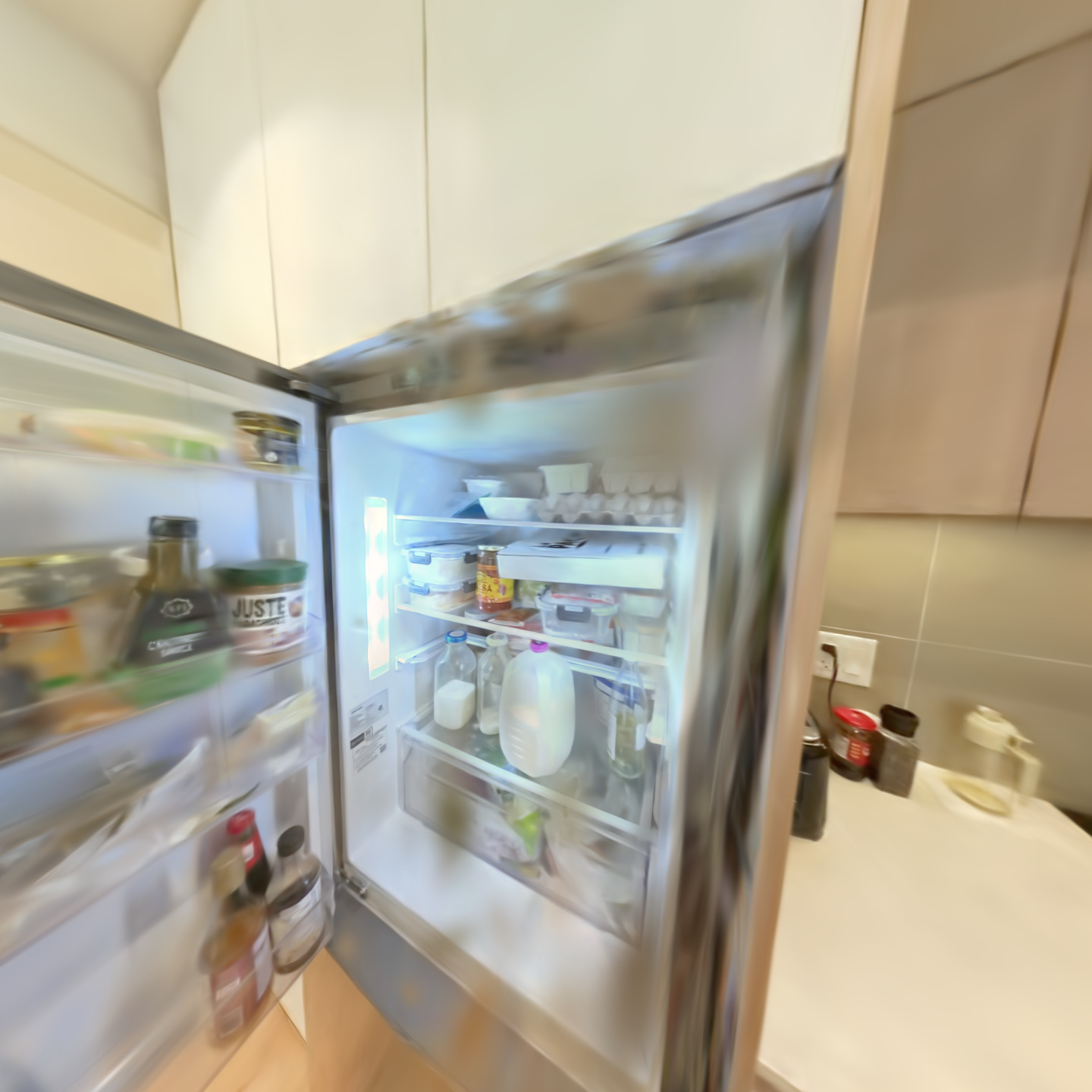} &
        \includegraphics[width=0.18\textwidth]{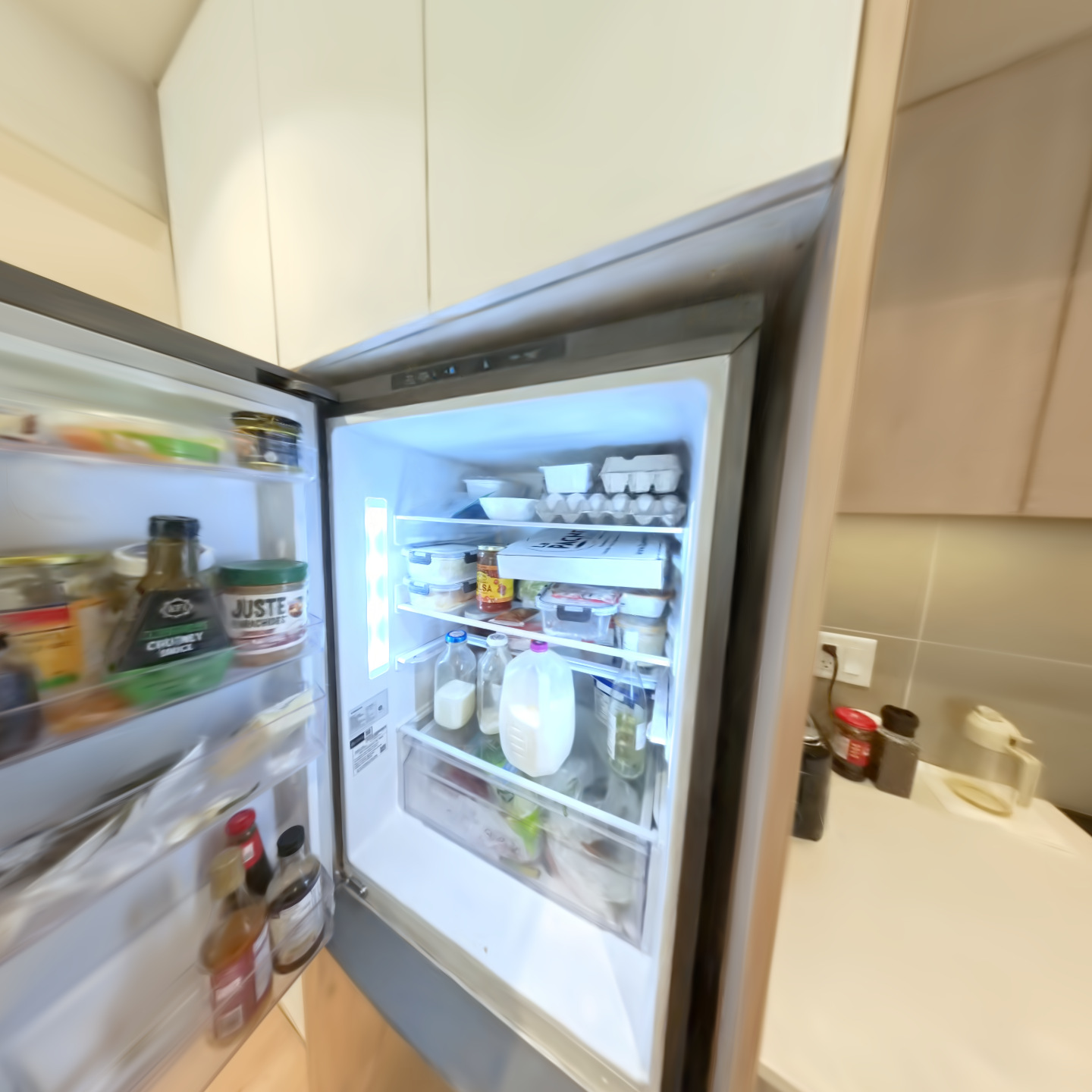} &
        \includegraphics[width=0.18\textwidth]{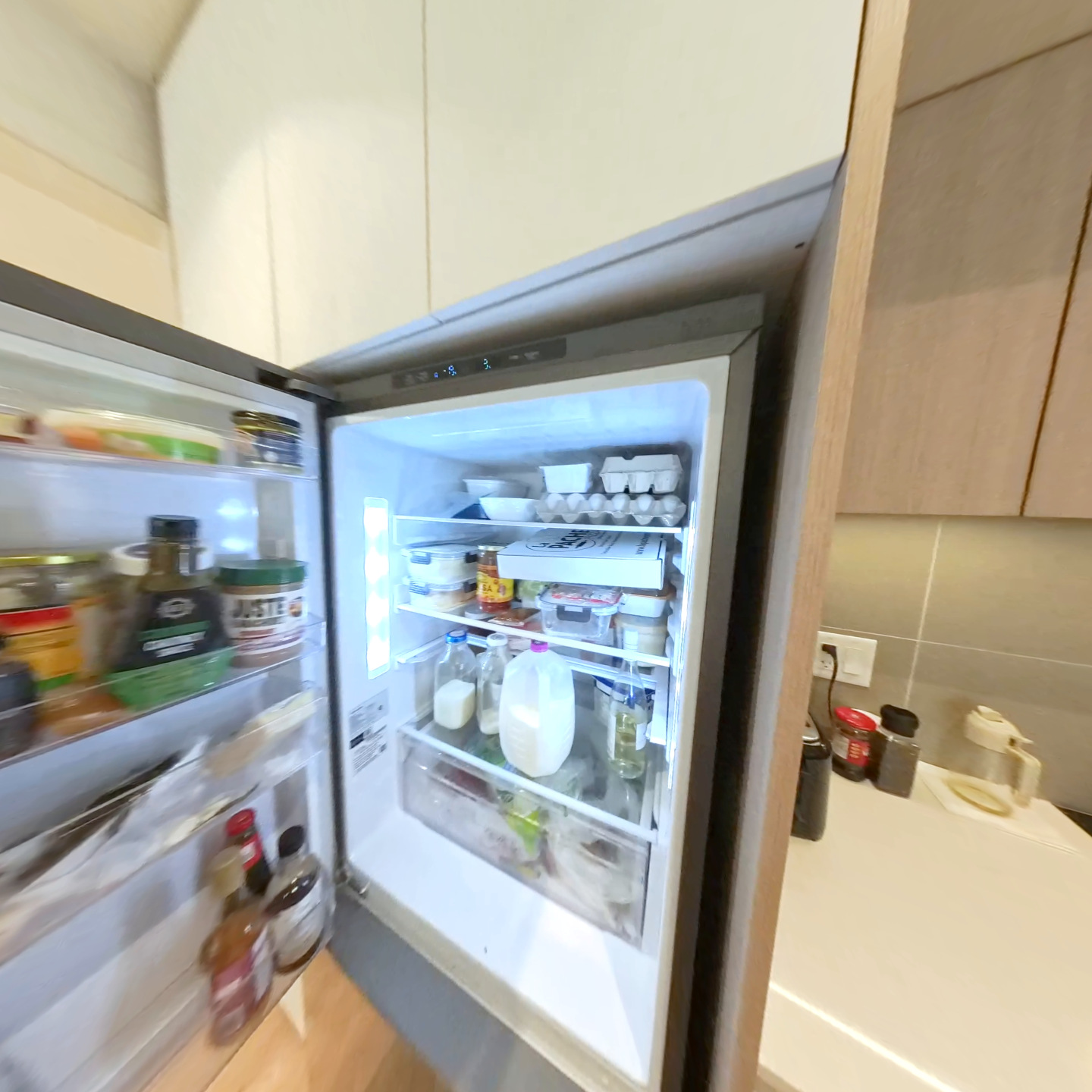} \\

        \includegraphics[width=0.18\textwidth]{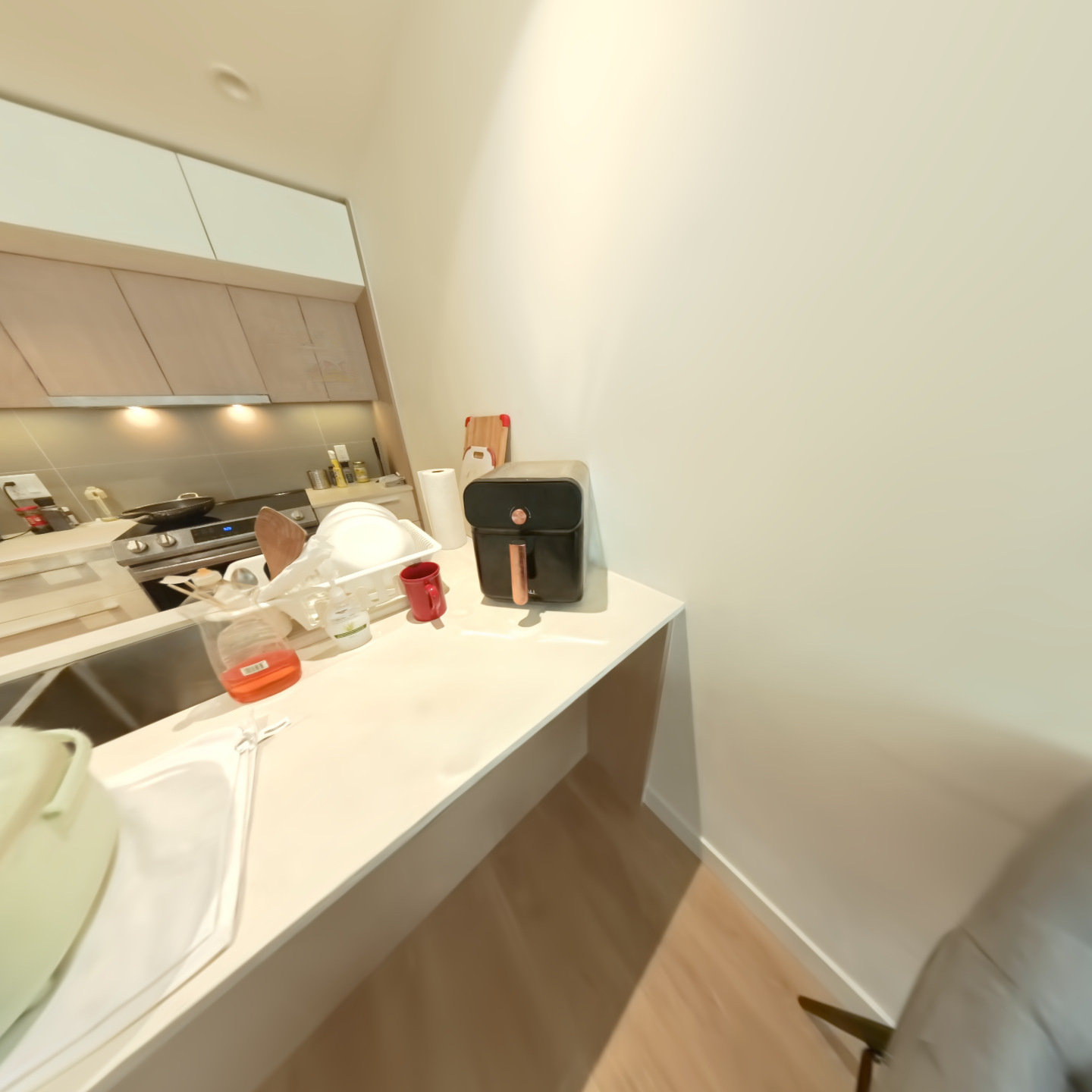} &
        \includegraphics[width=0.18\textwidth]{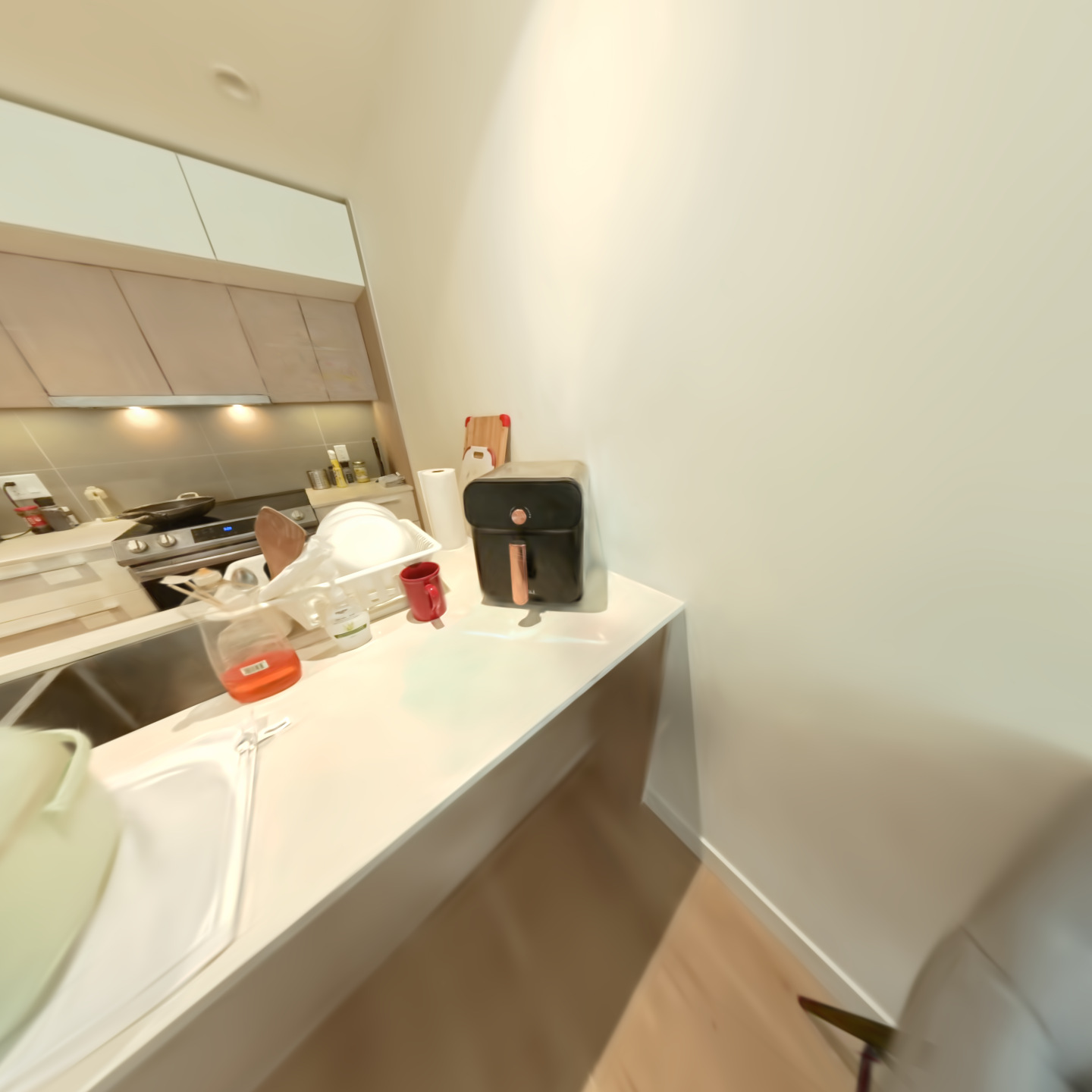} &
        \includegraphics[width=0.18\textwidth]{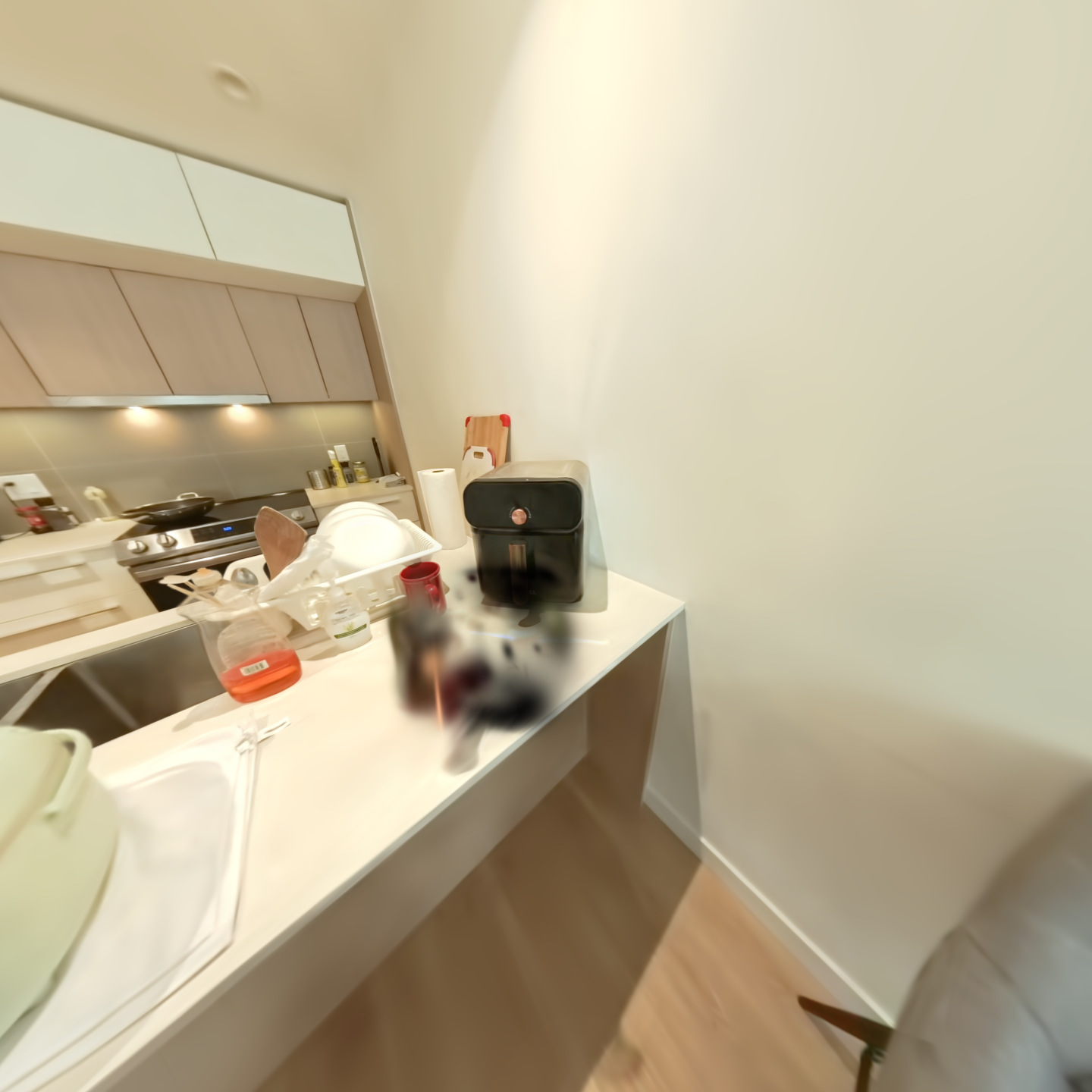} &
        \includegraphics[width=0.18\textwidth]{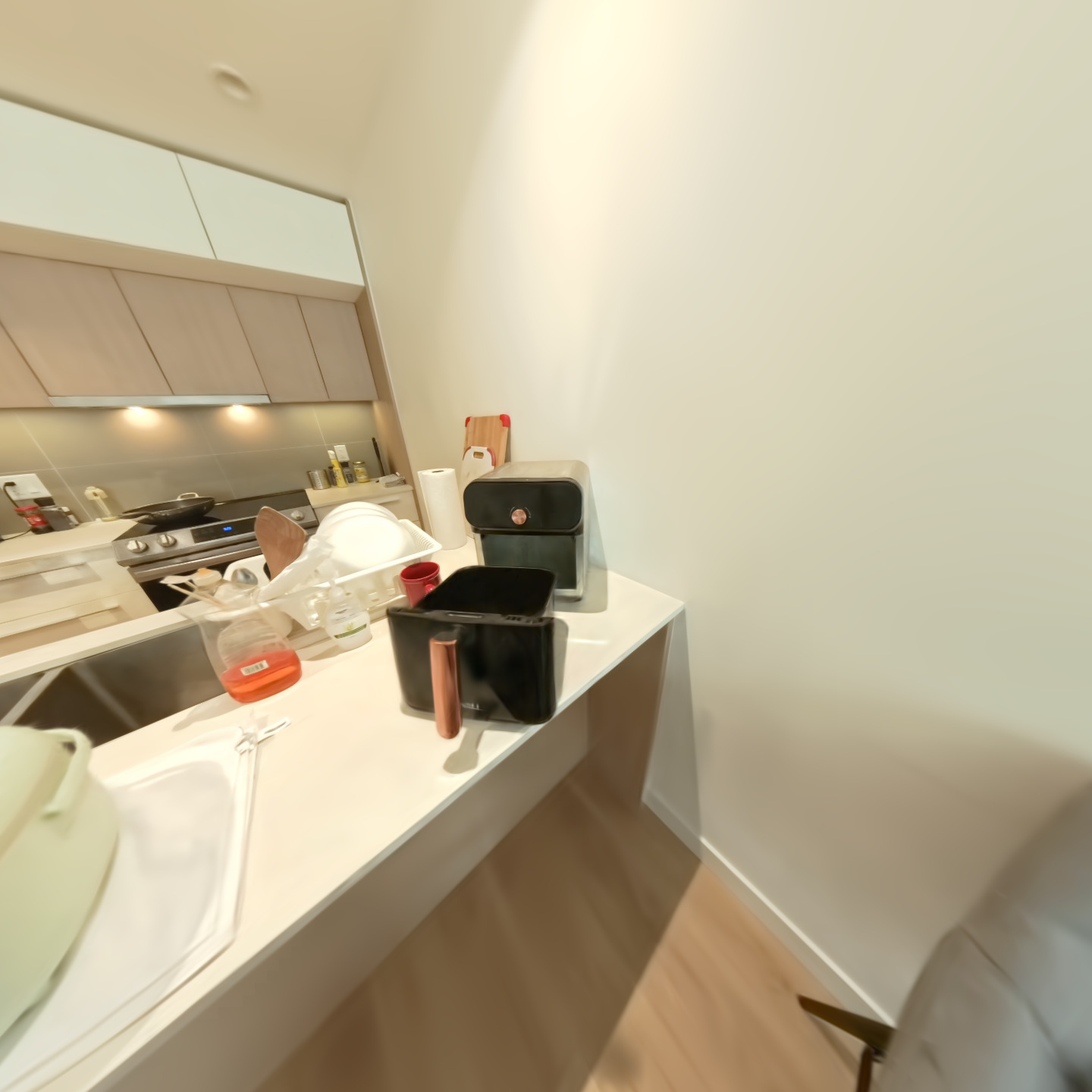} &
        \includegraphics[width=0.18\textwidth]{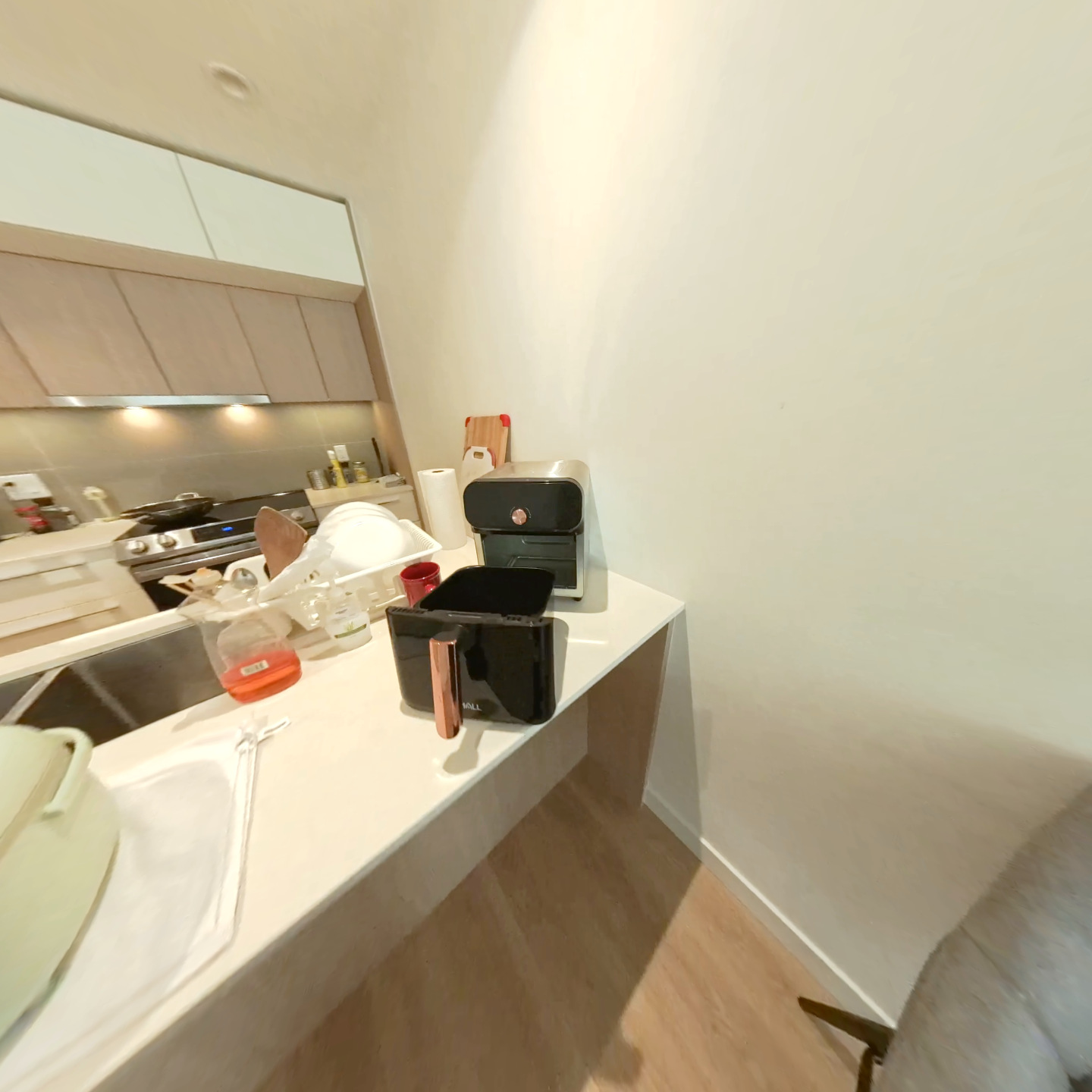} \\

        \includegraphics[width=0.18\textwidth]{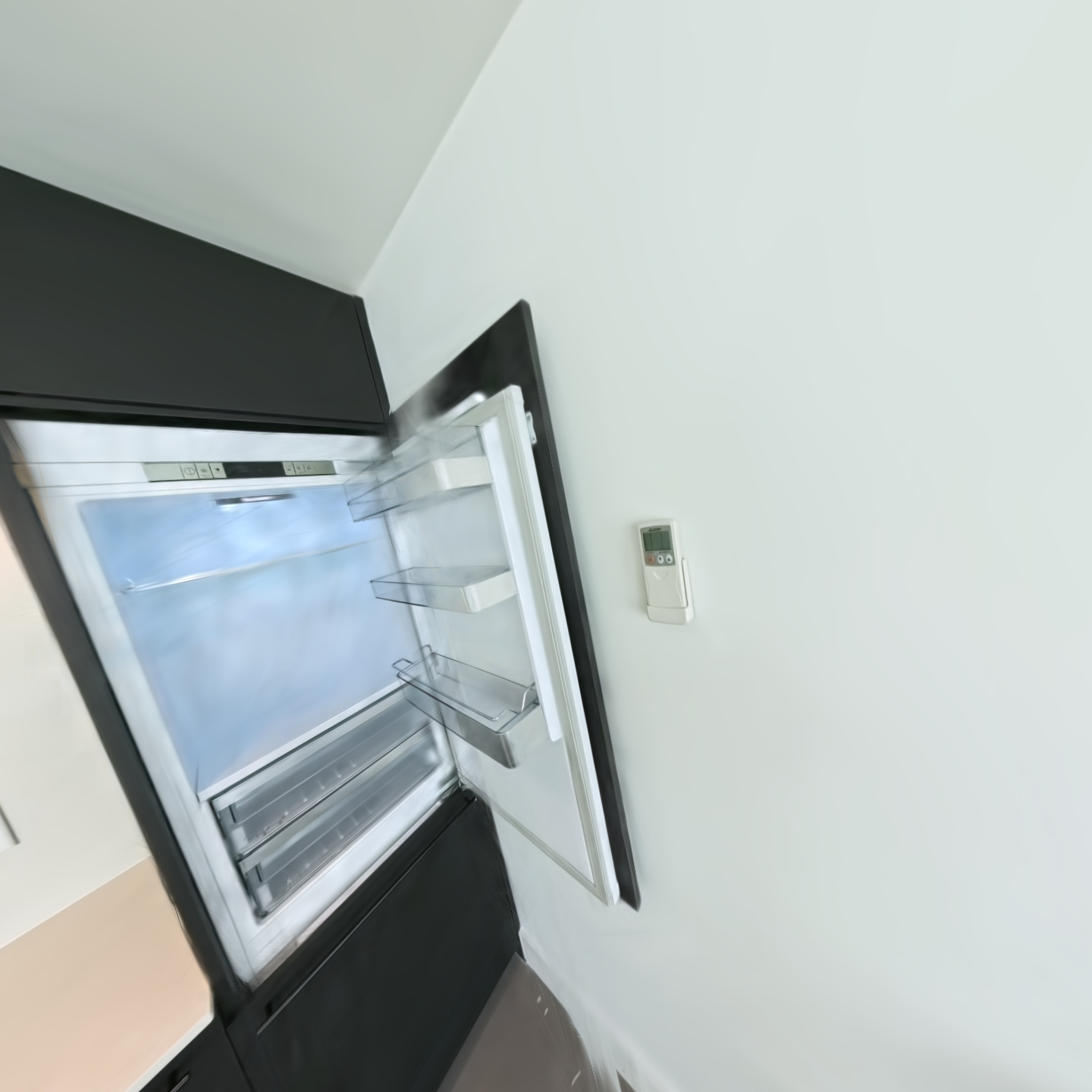} &
        \includegraphics[width=0.18\textwidth]{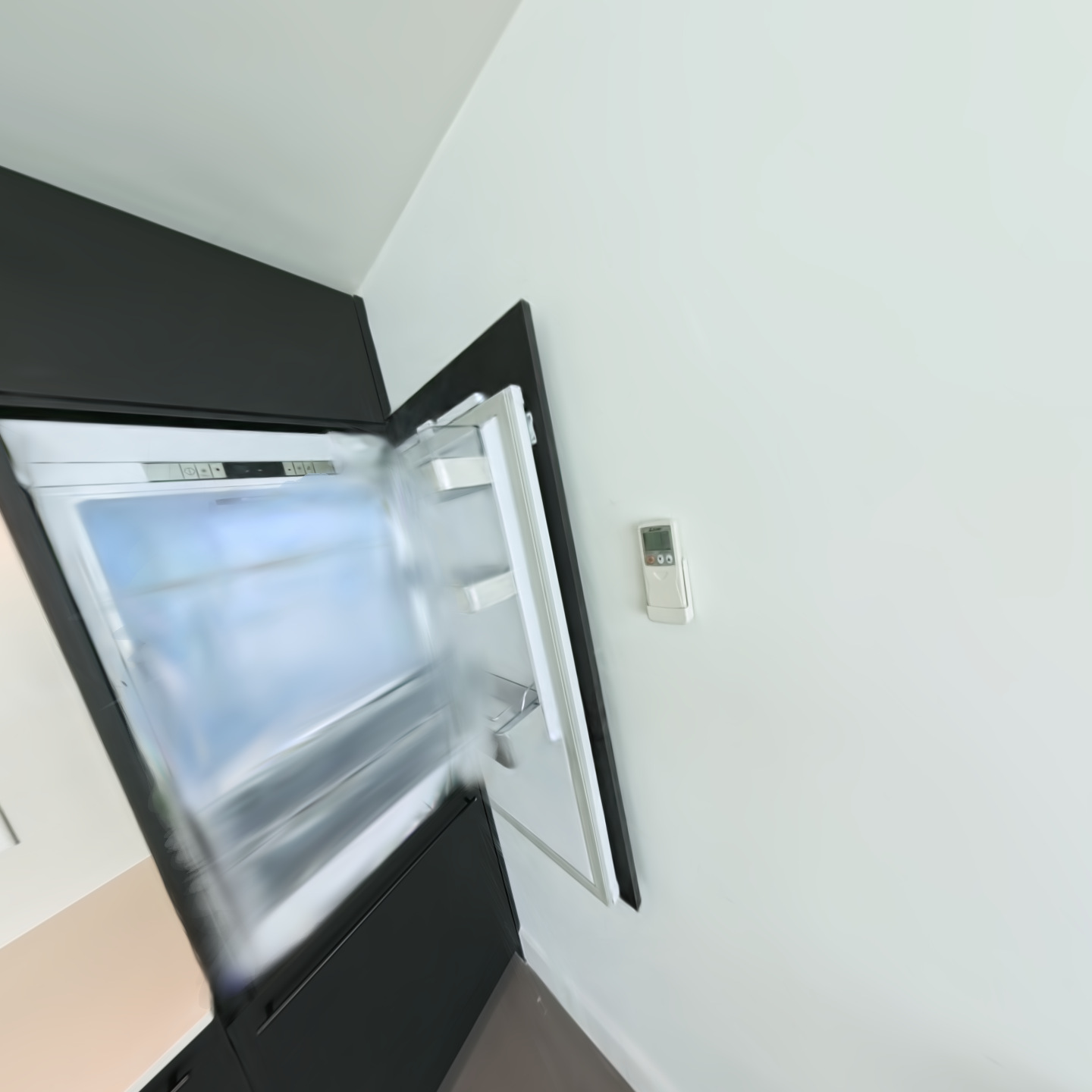} &
        \includegraphics[width=0.18\textwidth]{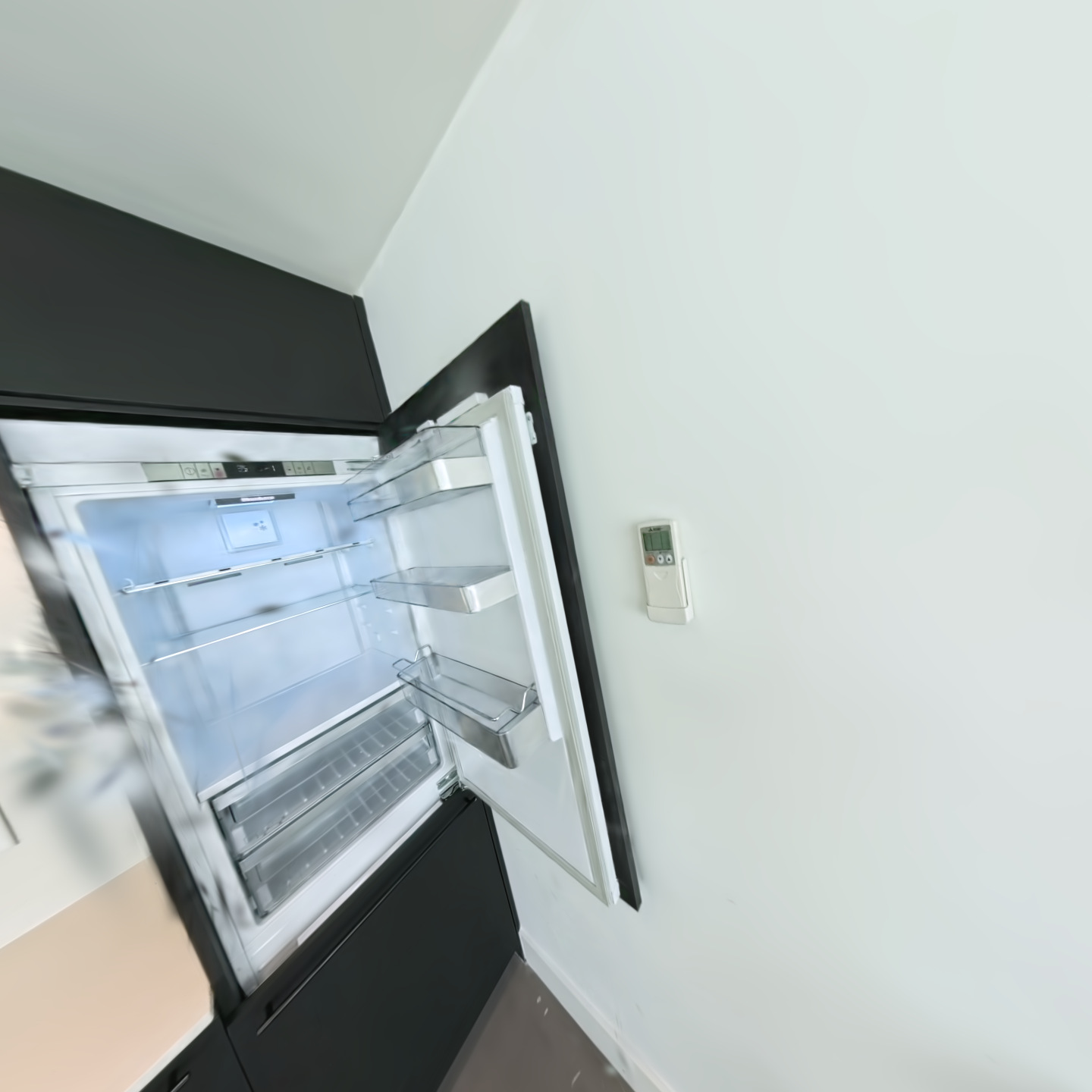} &
        \includegraphics[width=0.18\textwidth]{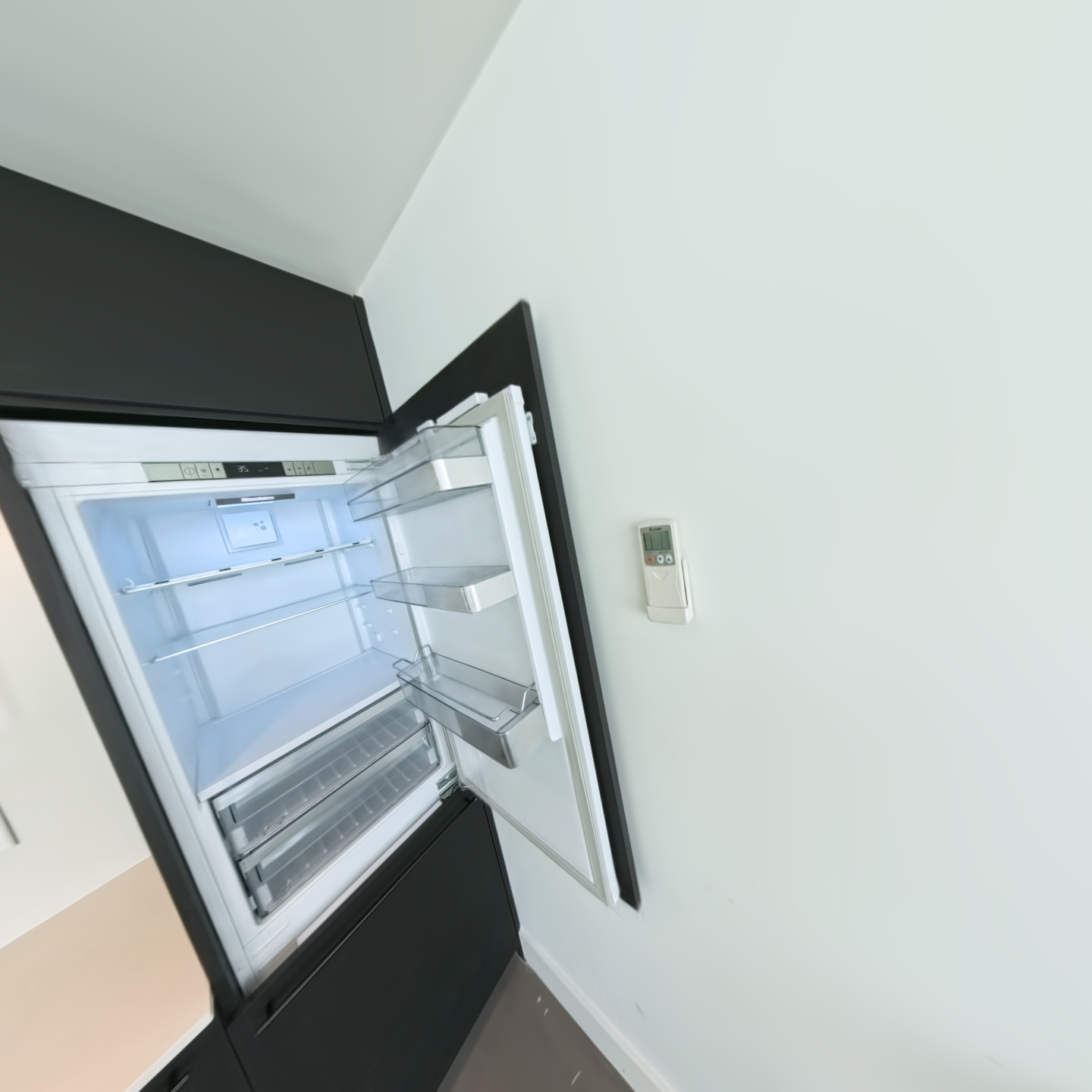} &
        \includegraphics[width=0.18\textwidth]{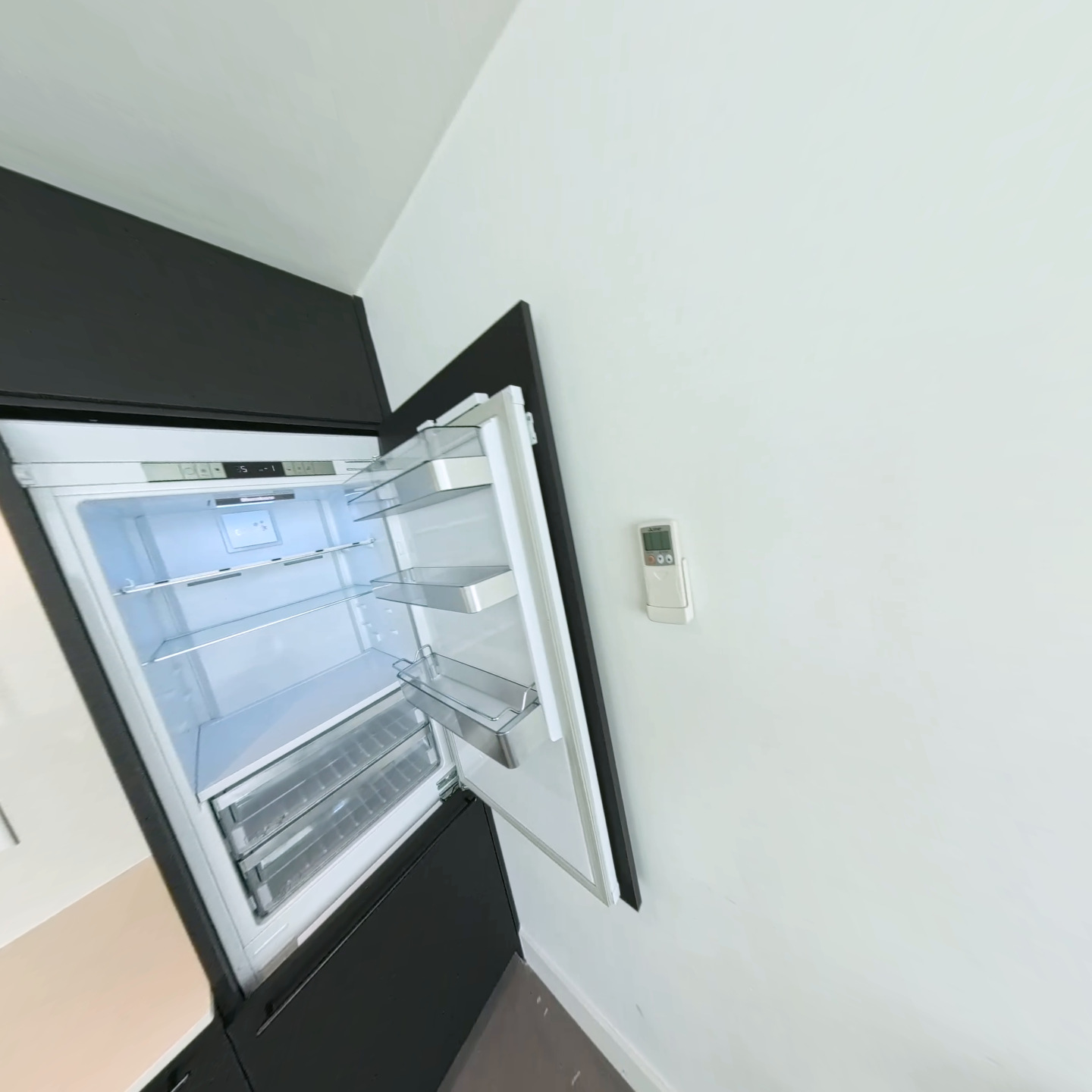} \\

        \includegraphics[width=0.18\textwidth]{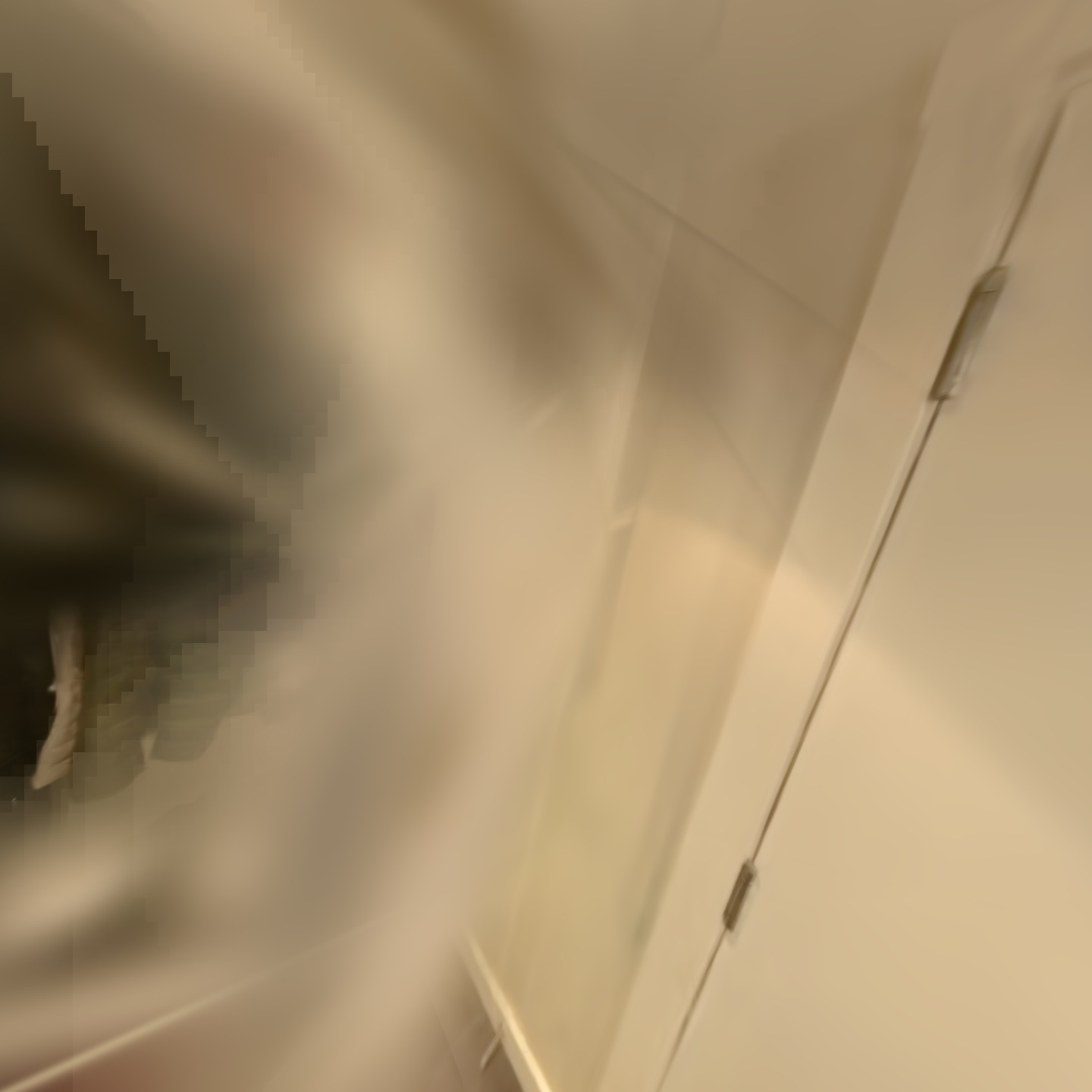} &
        \includegraphics[width=0.18\textwidth]{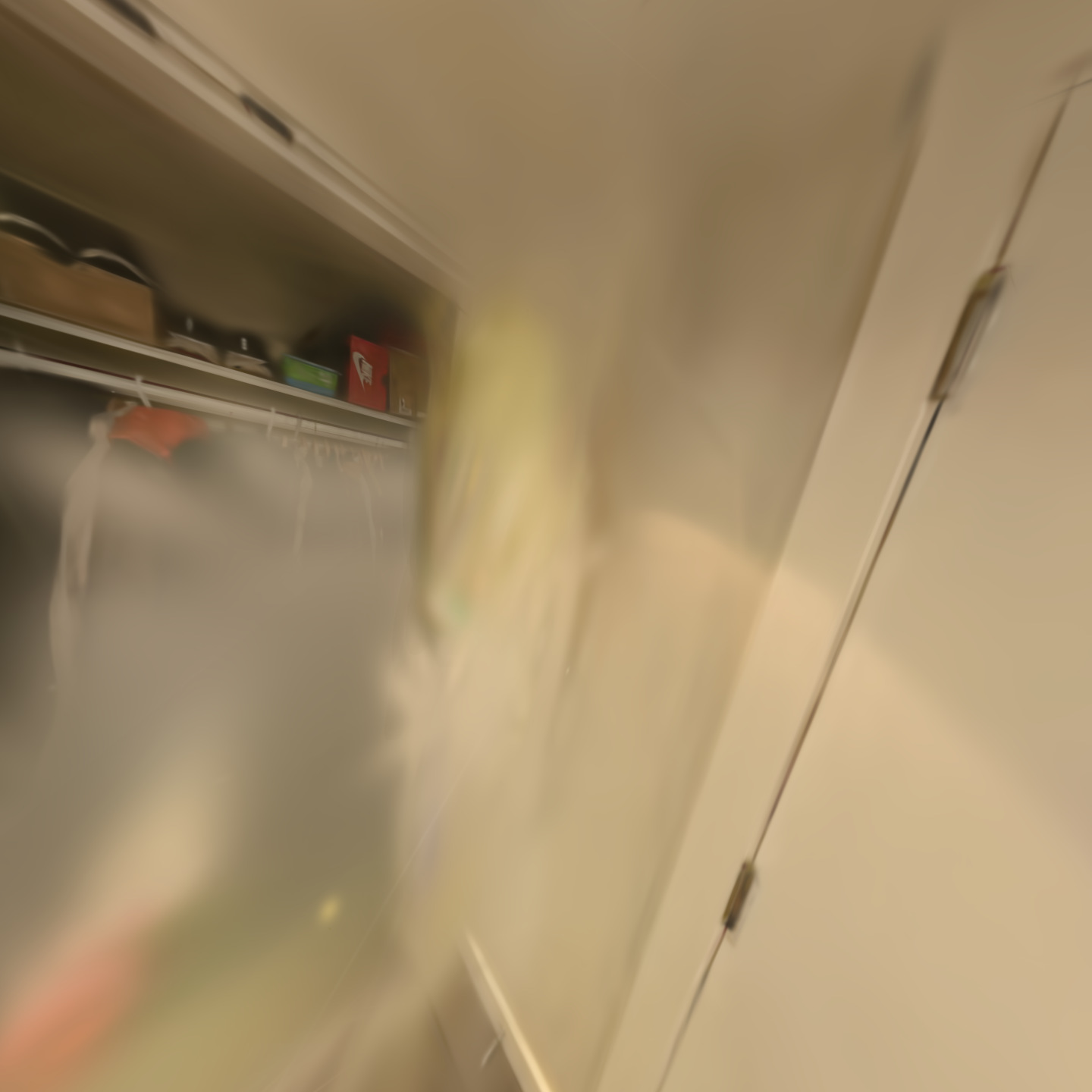} &
        \includegraphics[width=0.18\textwidth]{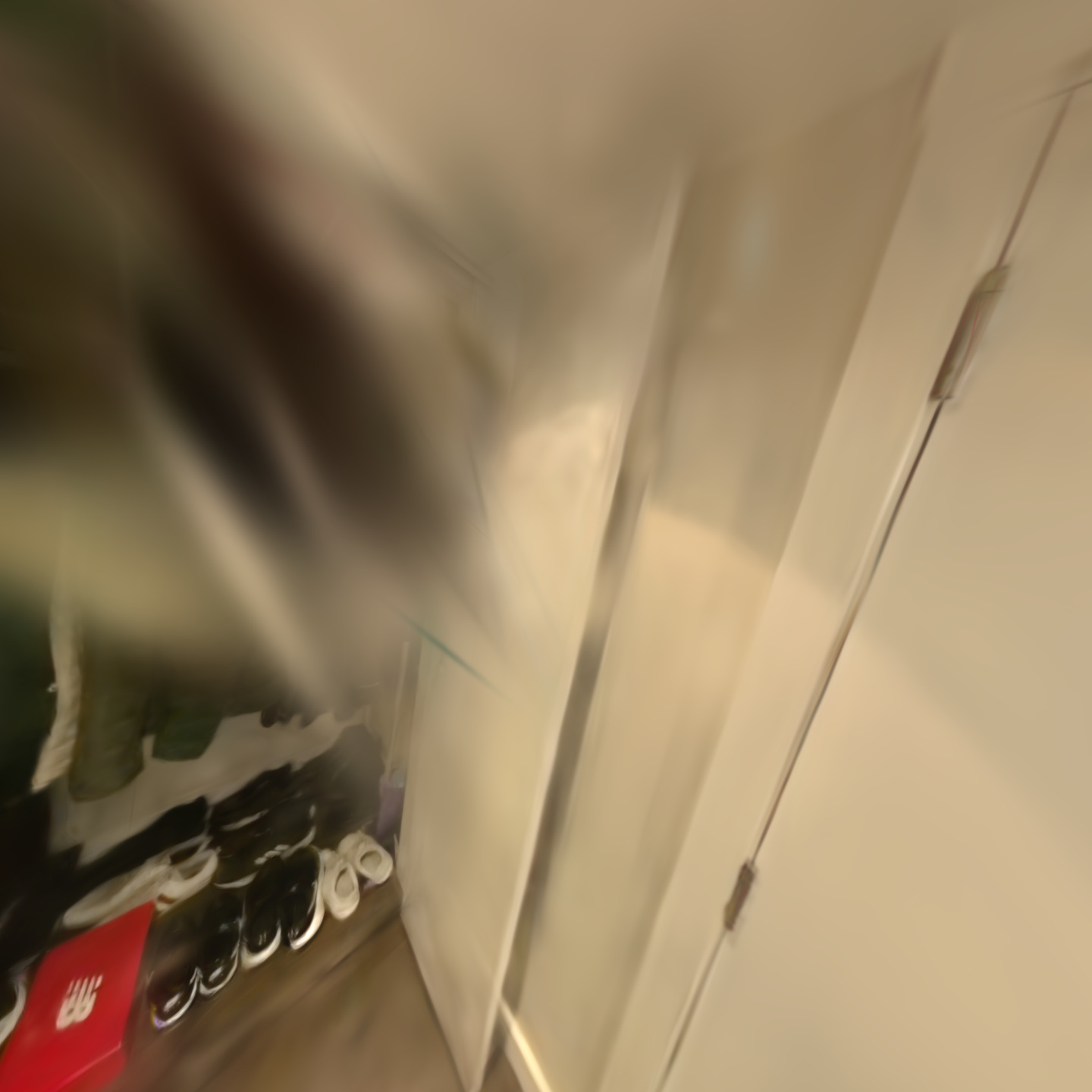} &
        \includegraphics[width=0.18\textwidth]{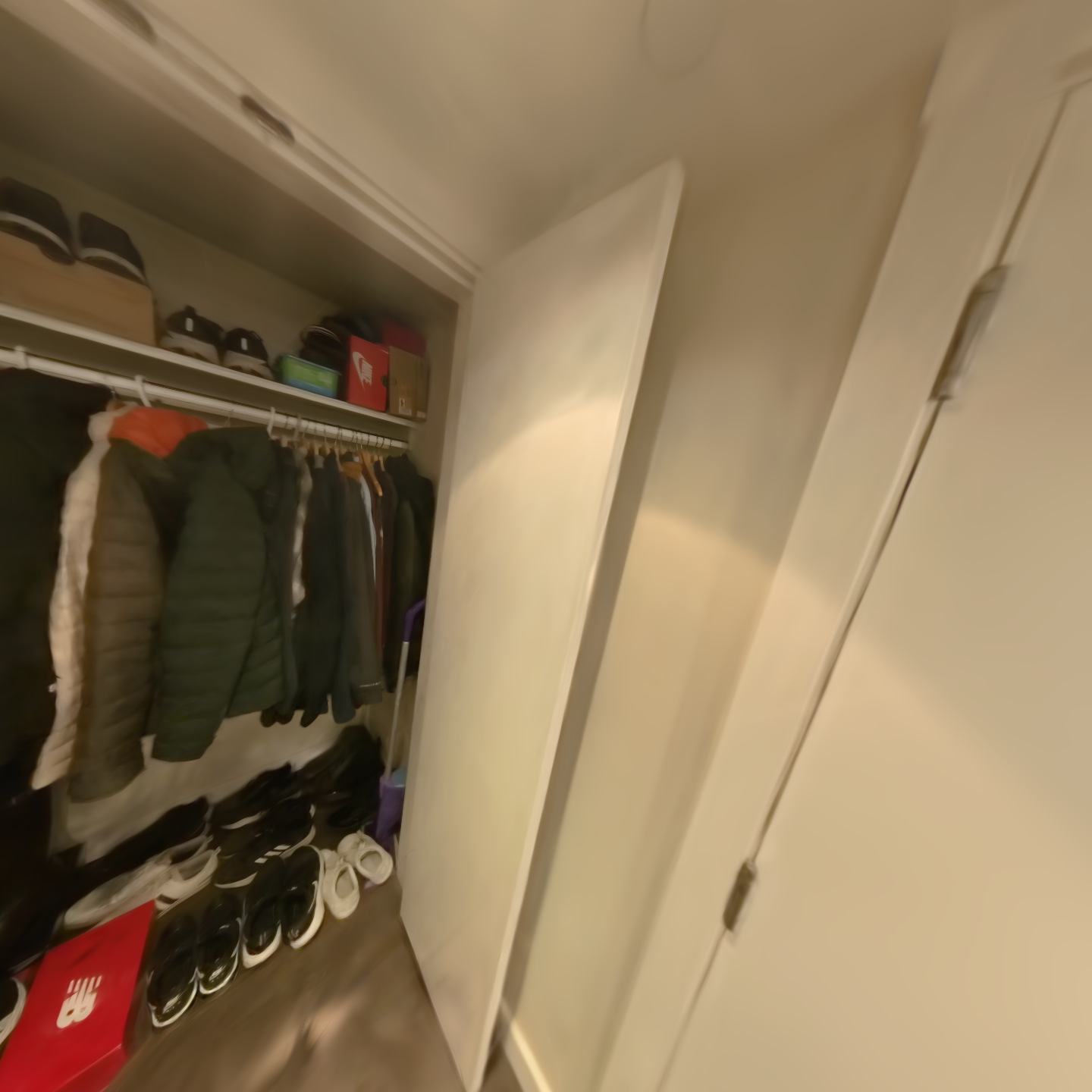} &
        \includegraphics[width=0.18\textwidth]{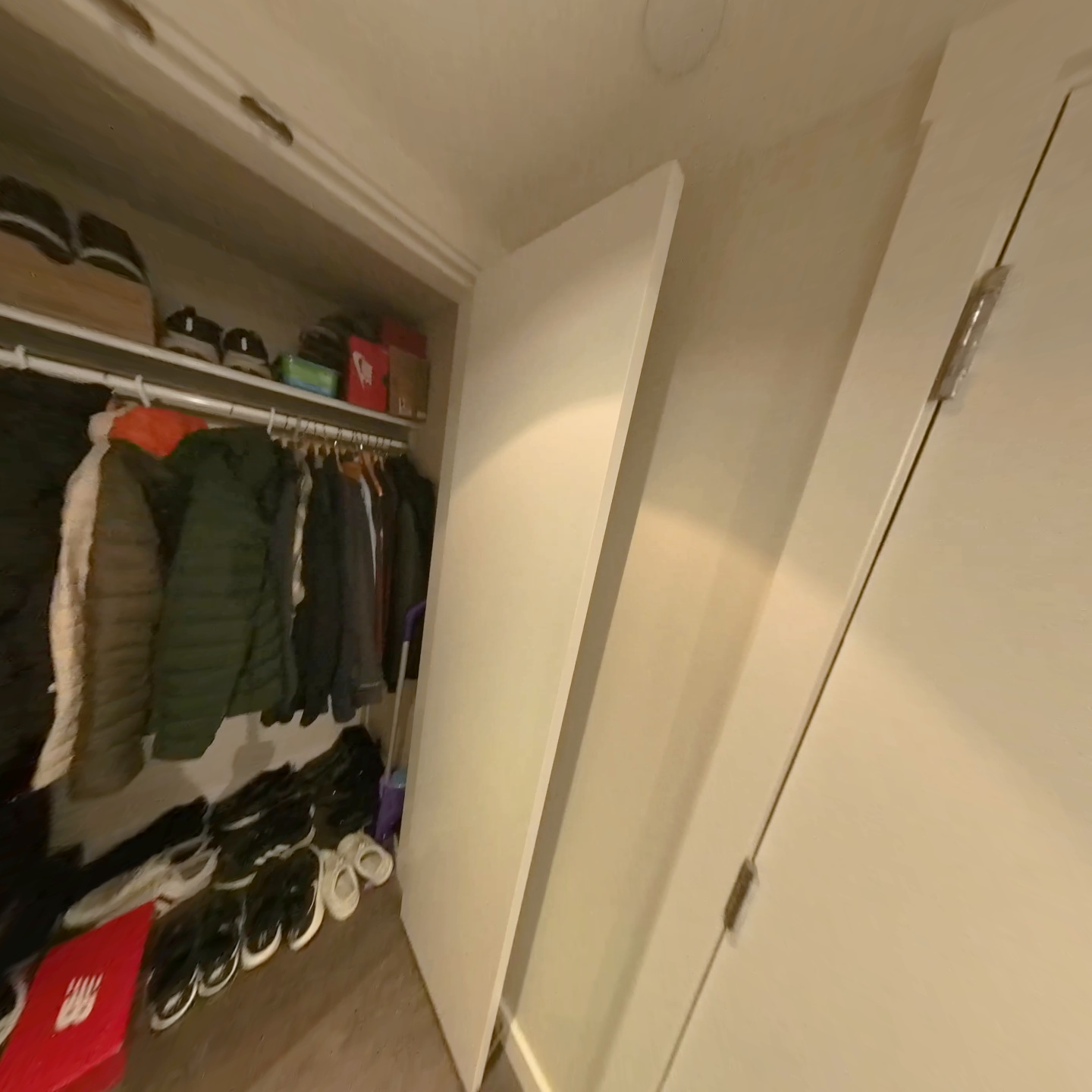} \\

        \includegraphics[width=0.18\textwidth]{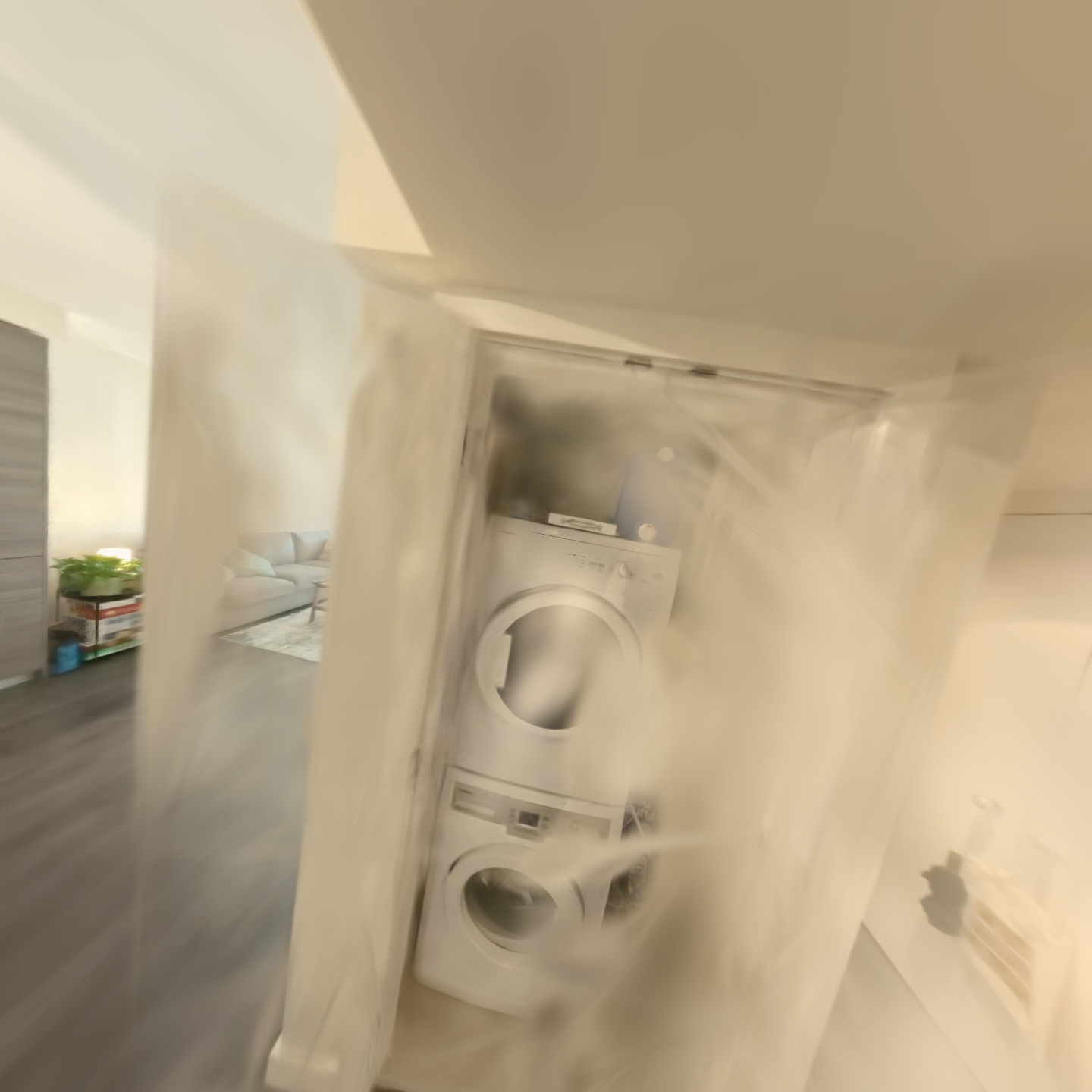} &
        \includegraphics[width=0.18\textwidth]{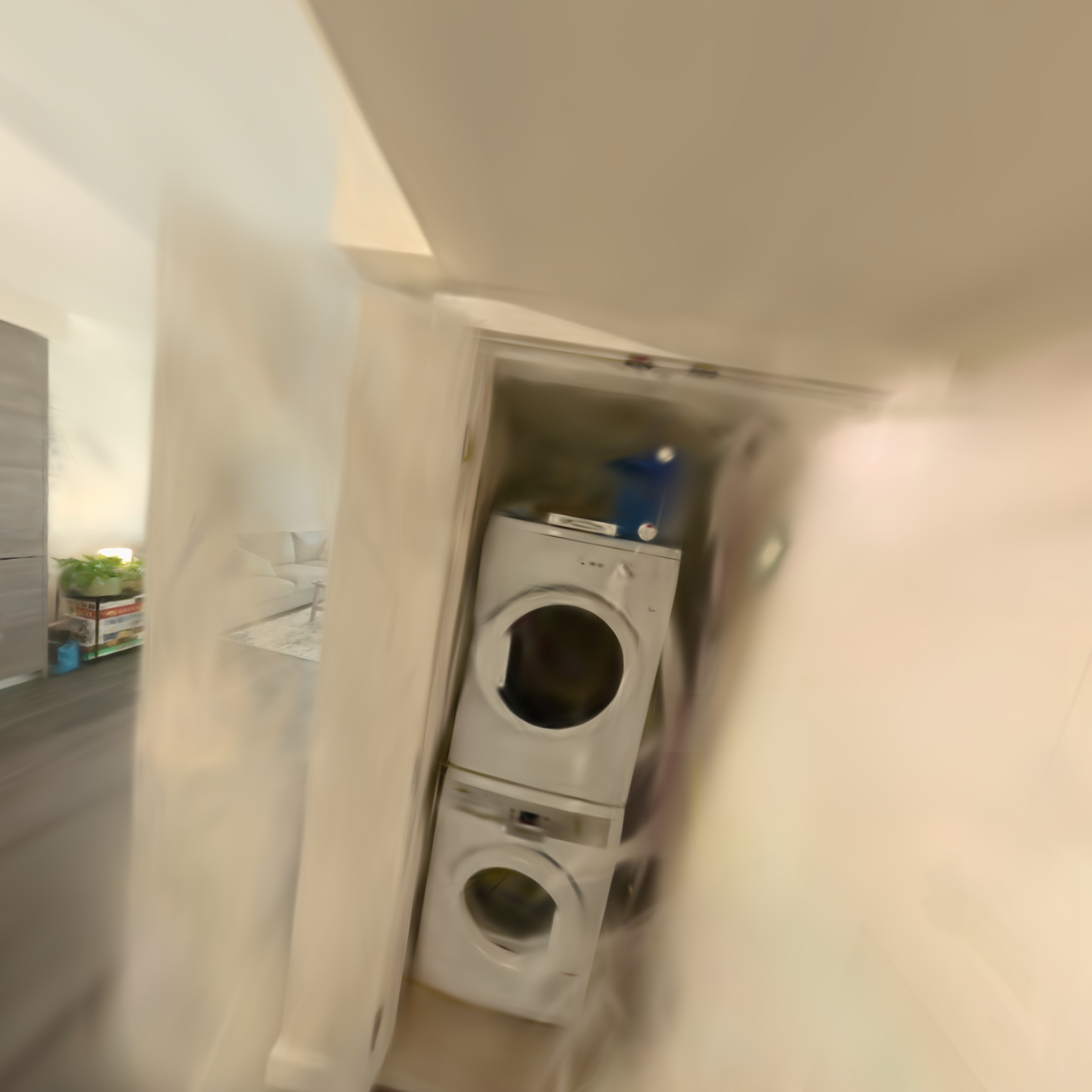} &
        \includegraphics[width=0.18\textwidth]{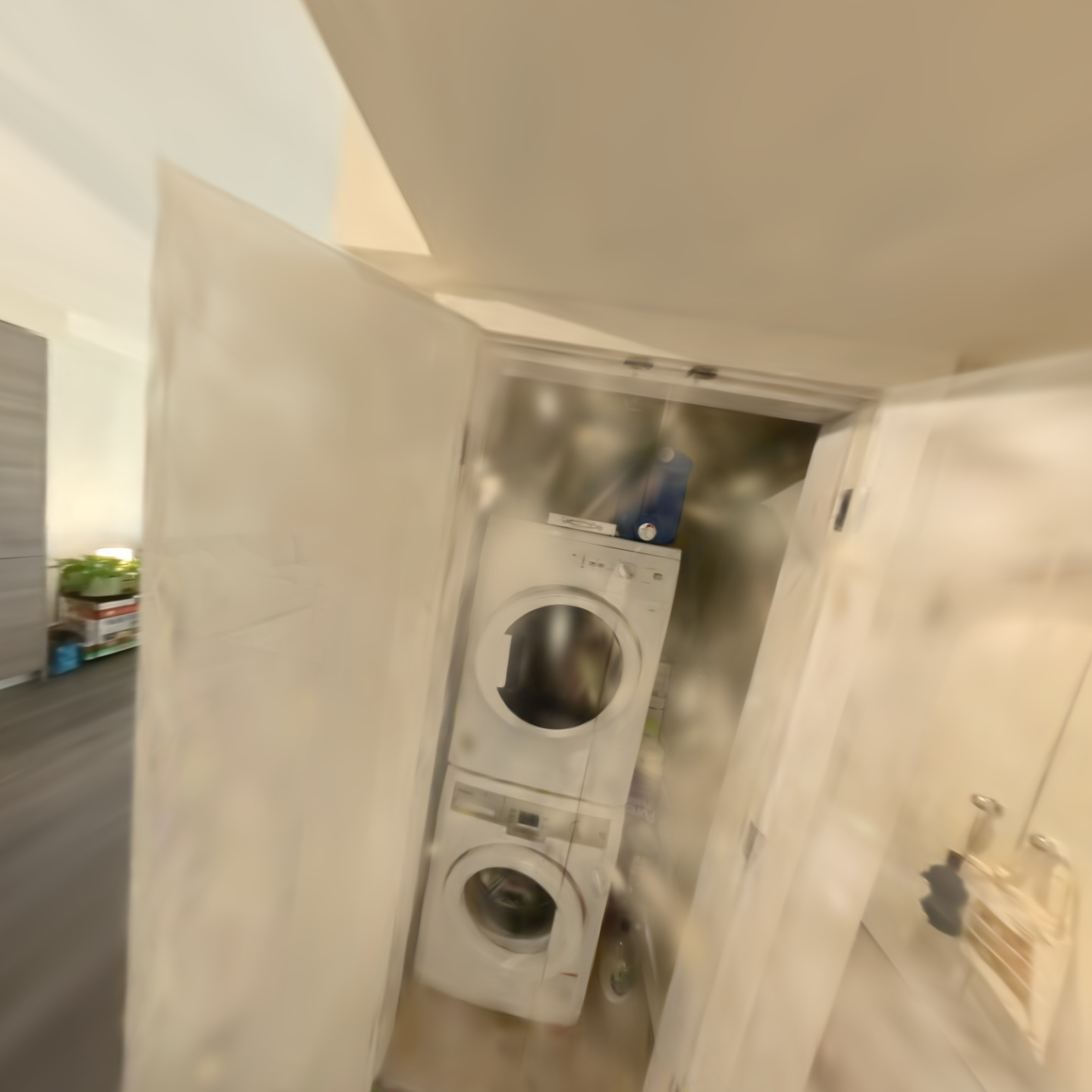} &
        \includegraphics[width=0.18\textwidth]{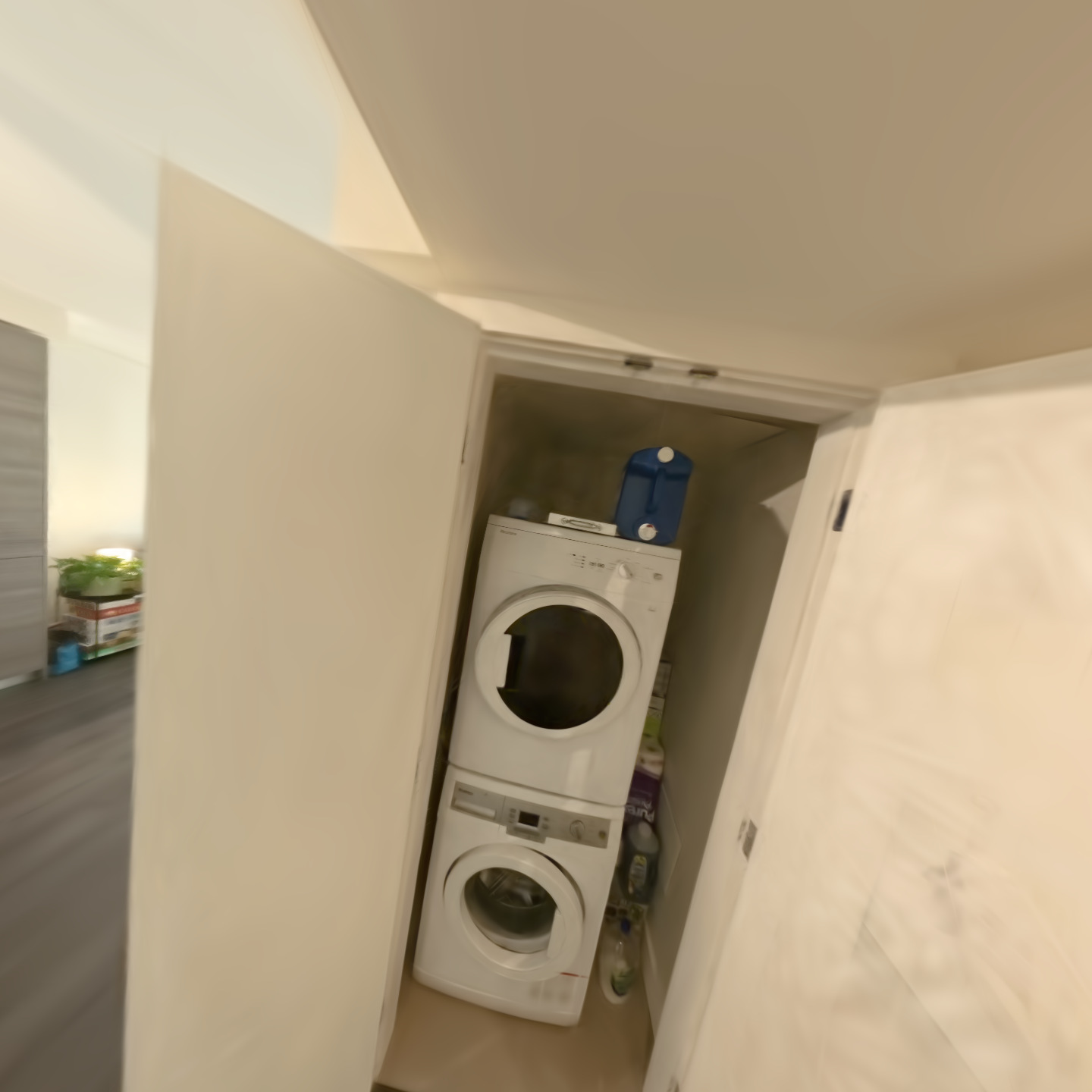} &
        \includegraphics[width=0.18\textwidth]{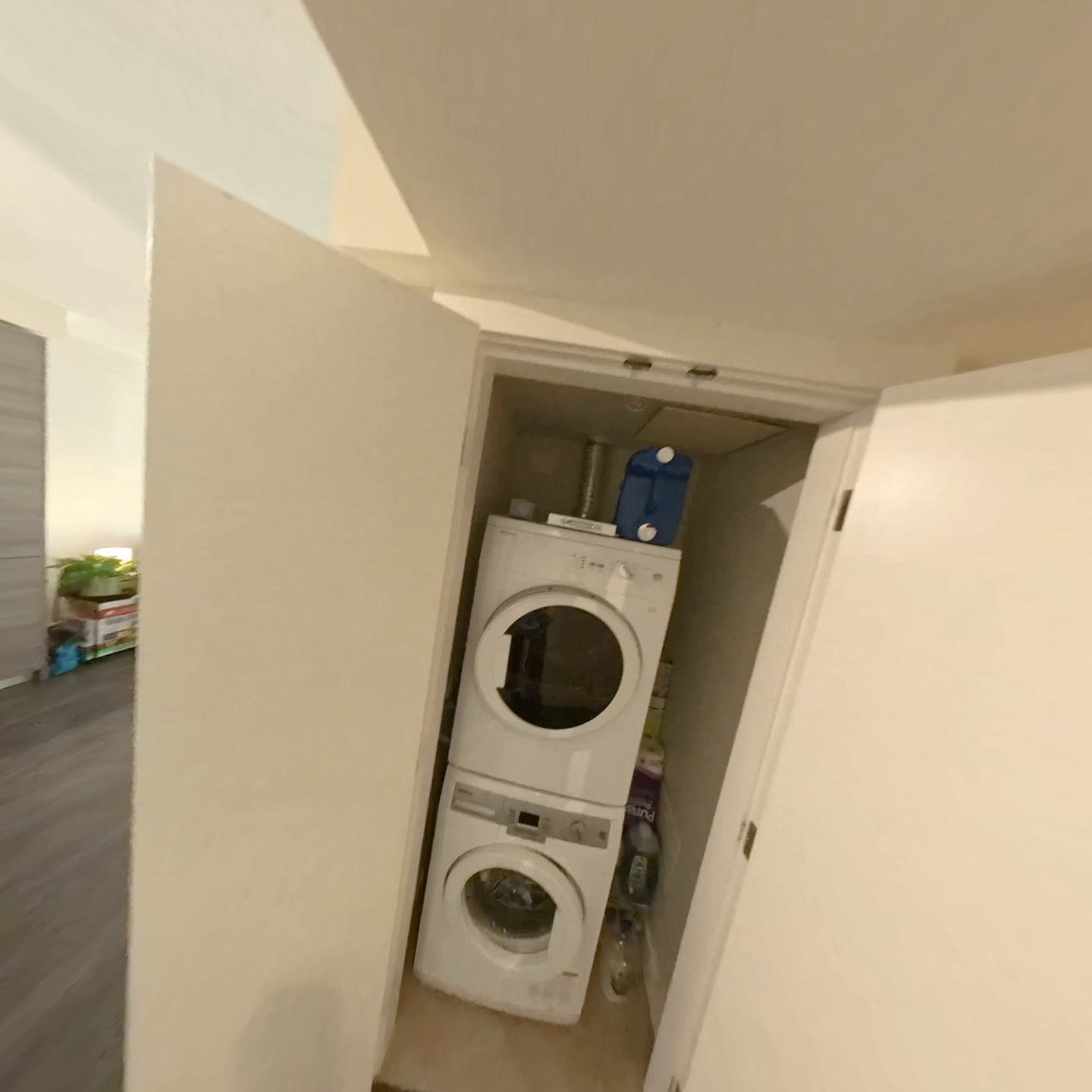} \\

        \includegraphics[width=0.18\textwidth]{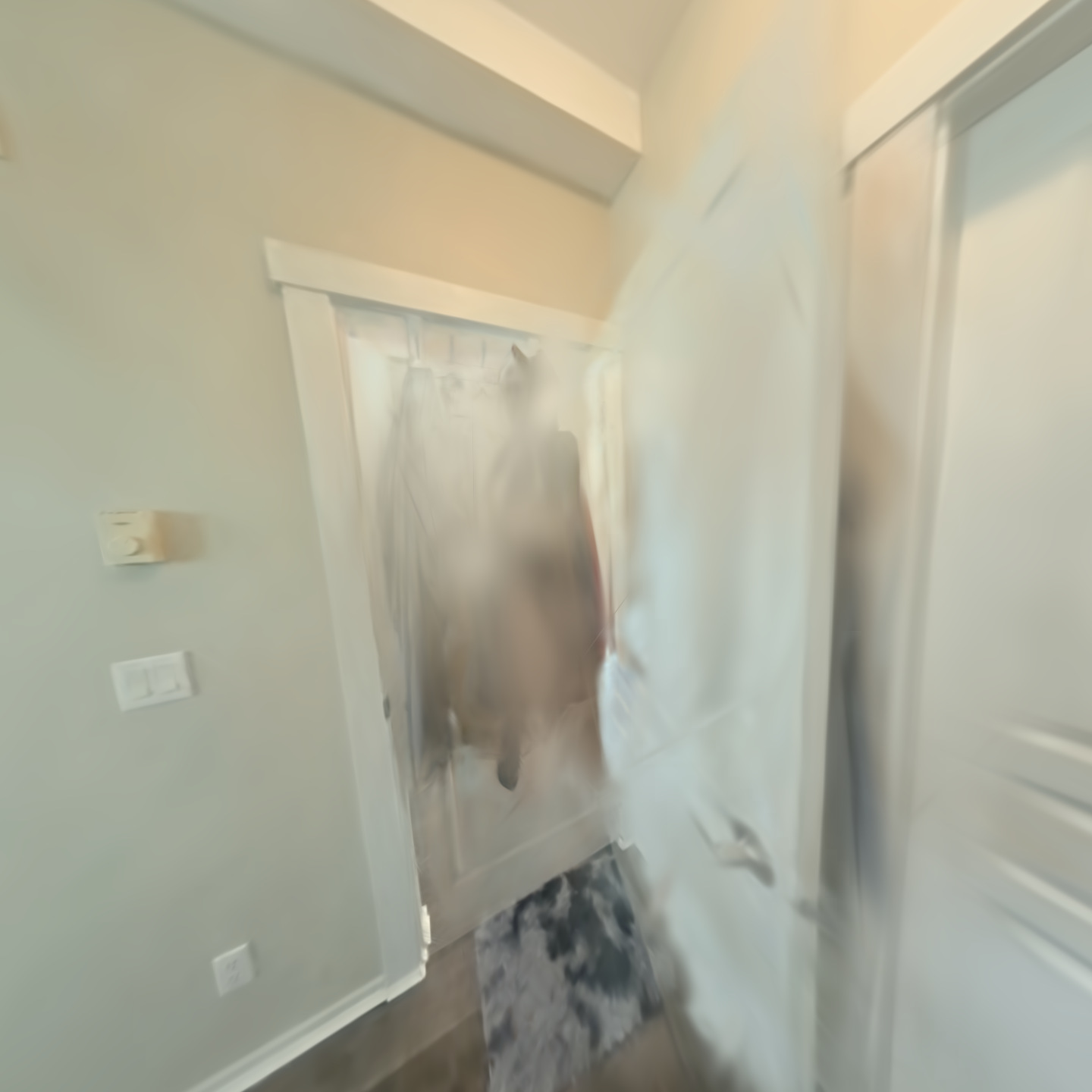} &
        \includegraphics[width=0.18\textwidth]{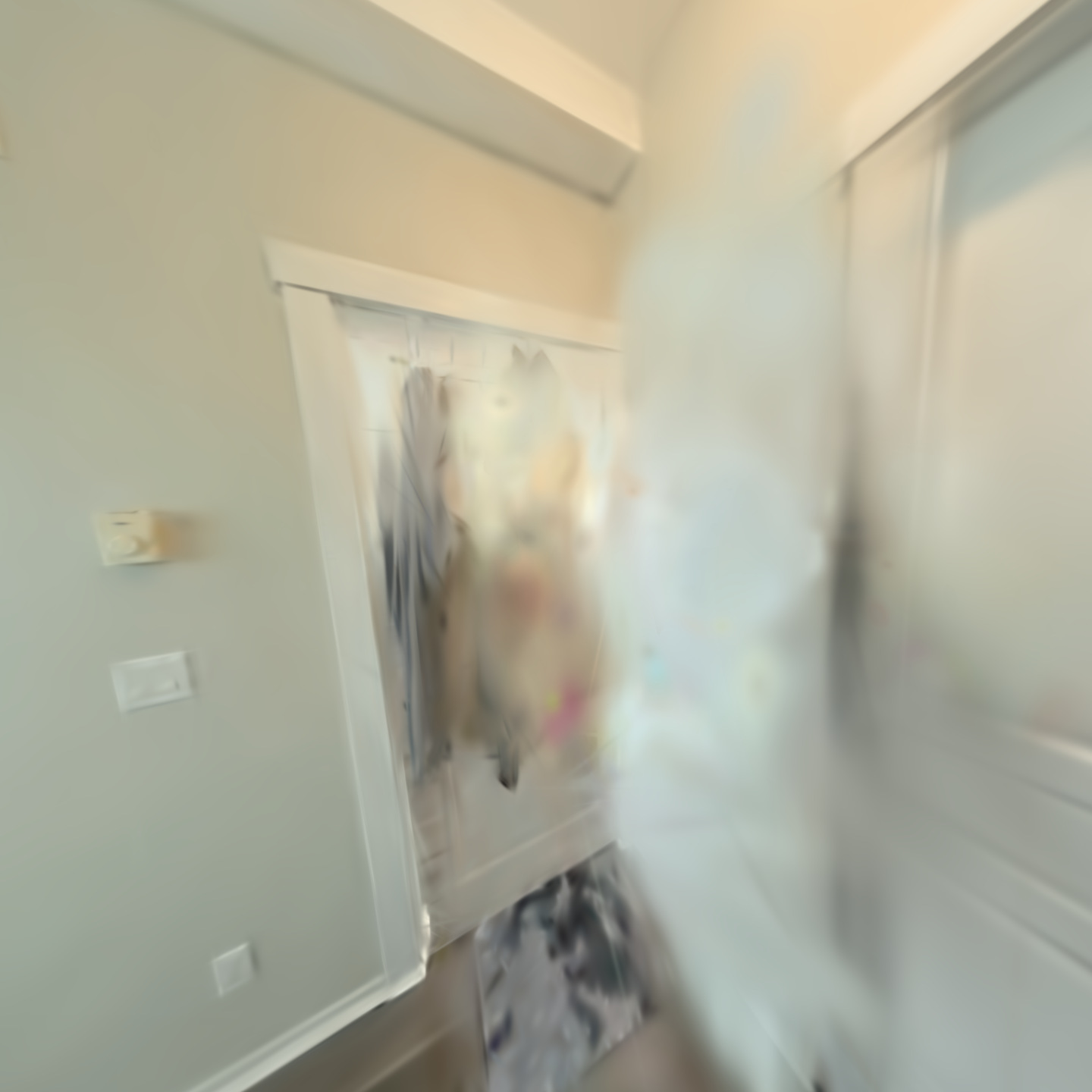} &
        \includegraphics[width=0.18\textwidth]{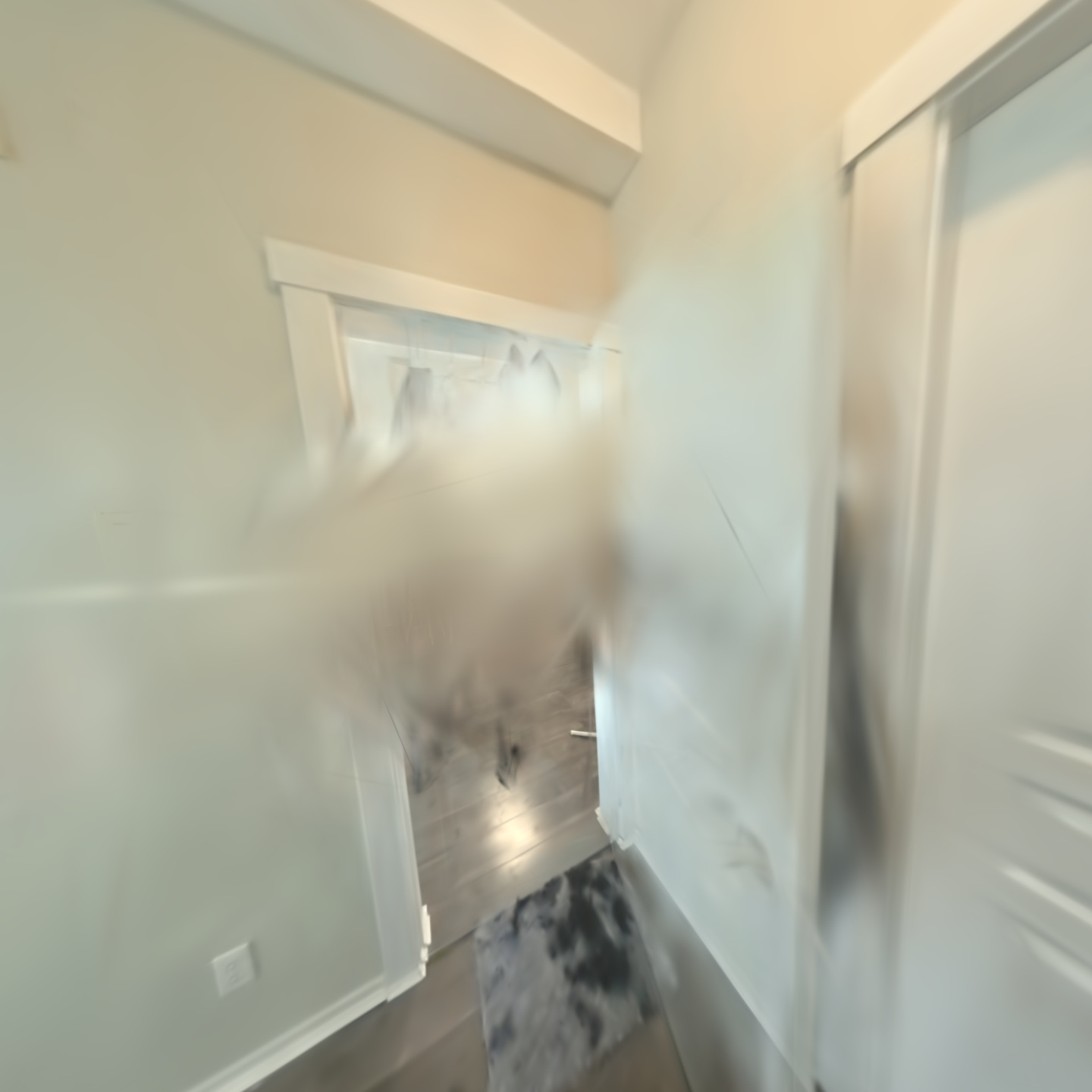} &
        \includegraphics[width=0.18\textwidth]{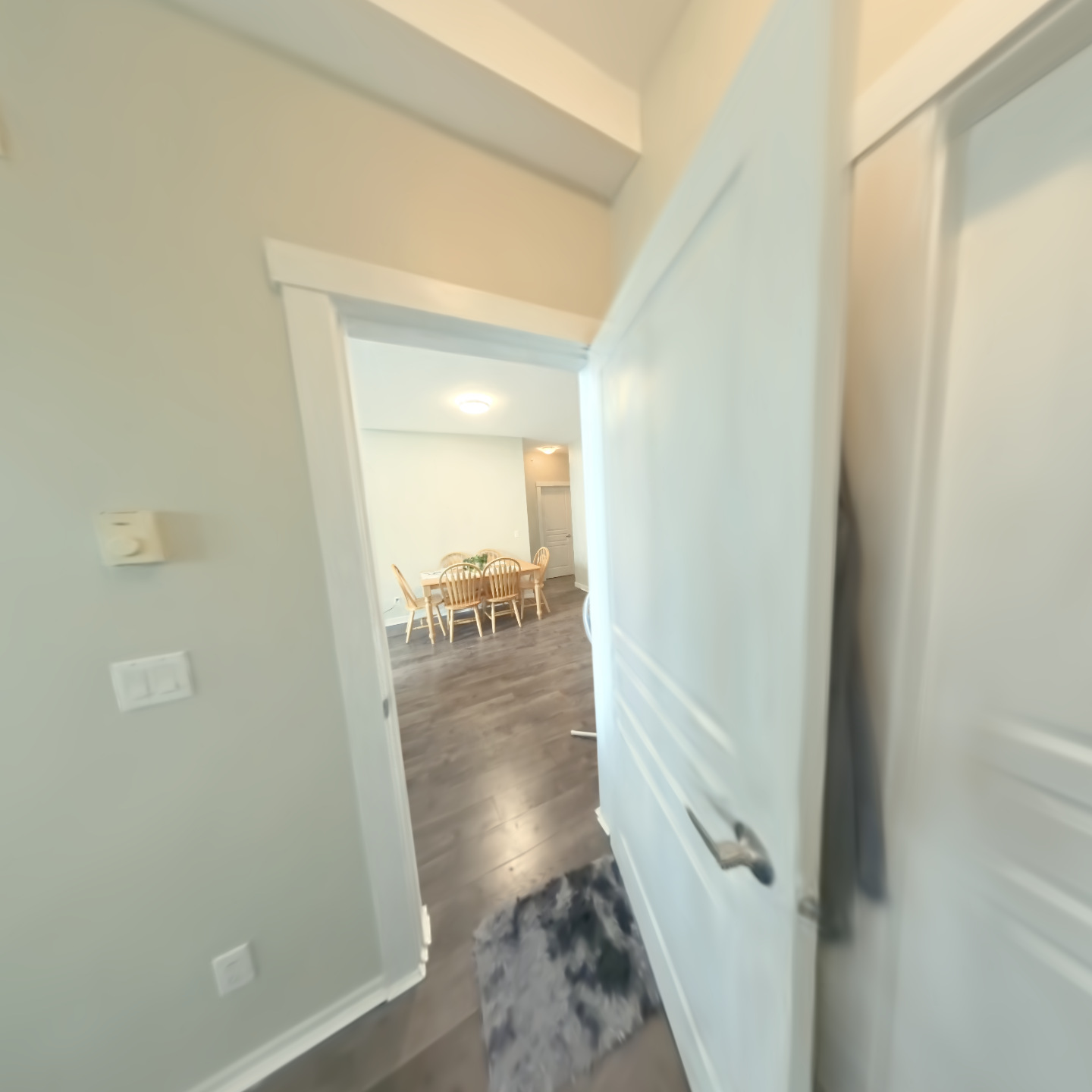} &
        \includegraphics[width=0.18\textwidth]{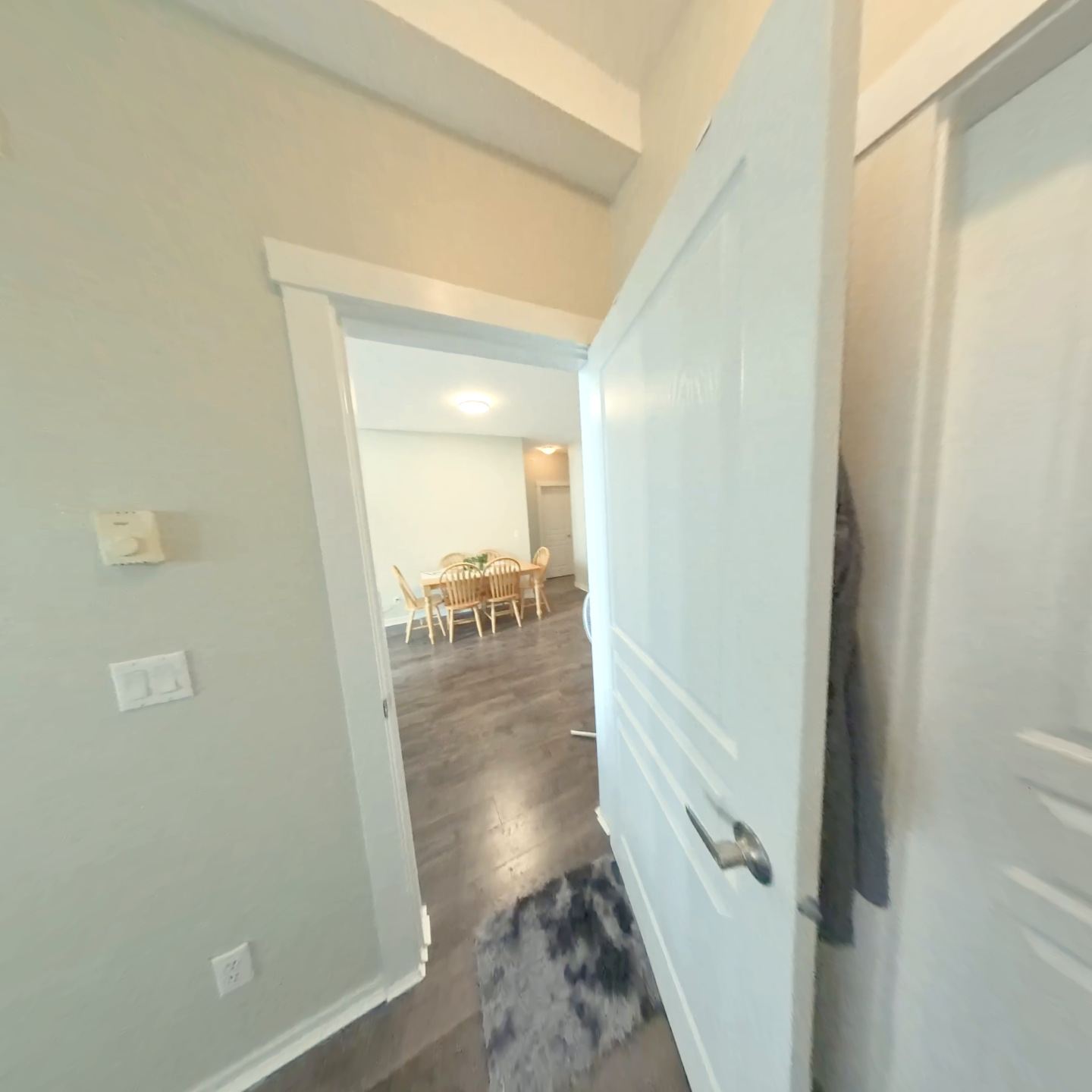} \\
    \end{tabular}

    \caption{\textbf{Qualitative comparison.}
    We compare FullCircle, DeGauss, 4DGS, and SwitchSplat against ground-truth images from standard held-out test views.
    Scenes are shown from top to bottom: Kitchen 1, Kitchen 1, Oslo, Flat 1, Flat 1, and Flat 3.}
    \Description{A six-row, five-column image grid. Columns show FullCircle, DeGauss, 4DGS, SwitchSplat, and ground truth. From top to bottom, the rows show an open refrigerator, a countertop appliance, another open refrigerator, an open coat closet, an open laundry closet, and an open interior door. The baseline renderings frequently blur, duplicate, or omit parts of the interacting objects, while the SwitchSplat renderings more closely reproduce the object configurations and scene structure shown in the ground truth.}
    \label{fig:comparison2}
\end{figure*} \clearpage
\appendix

\section{Supplementary Material}

The supplementary materials are organized as follows.
We first provide additional \textbf{qualitative video results} in the accompanying HTML supplement.
To view them locally, open the provided \texttt{index.html} file in a modern web browser.
Additional interactive results are available on our project page:
\url{https://switchsplat.github.io}.

\subsection{Dataset Capture and Preprocessing}
\begin{figure*}[htb]
    \centering
    \includegraphics[width=0.9\textwidth]
        {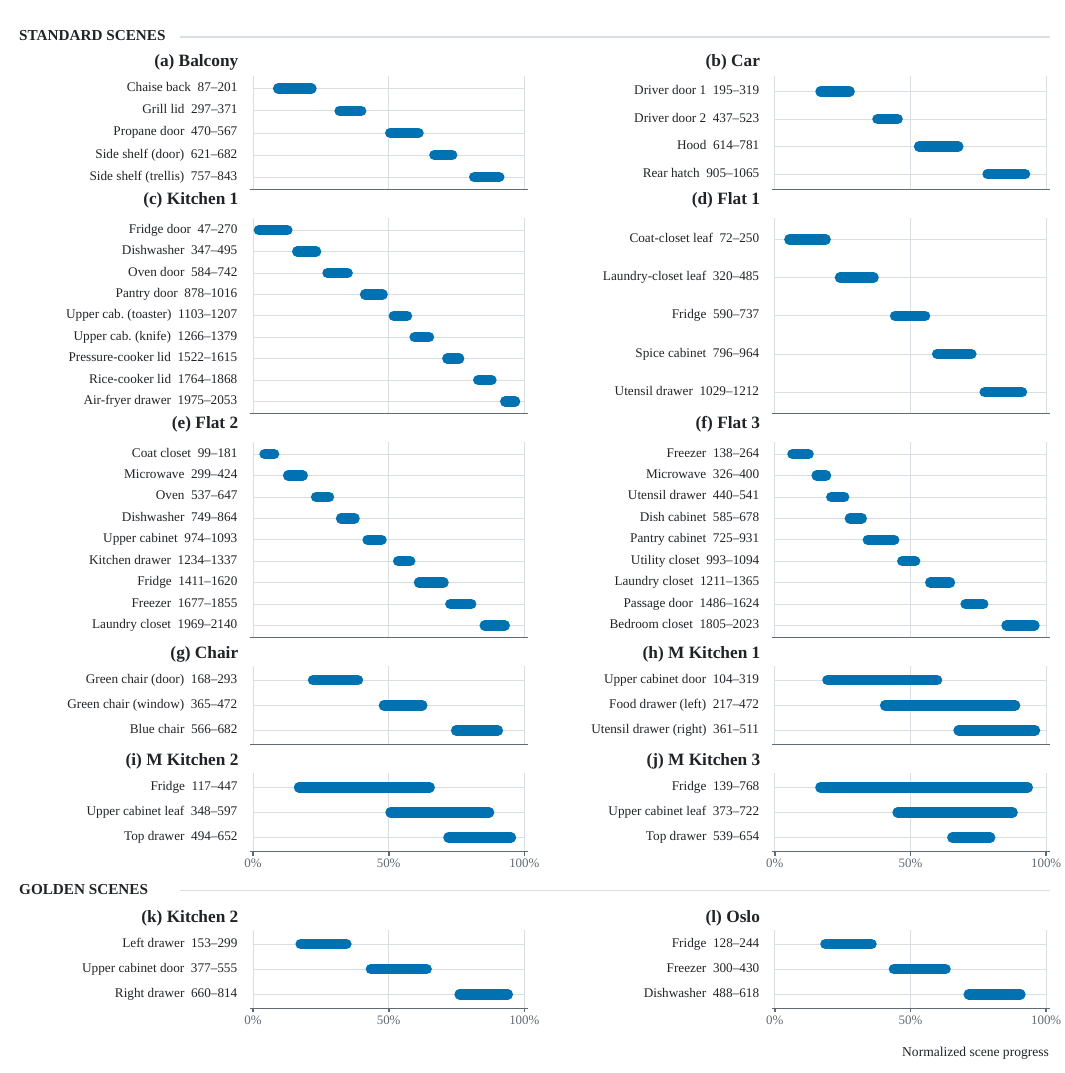}
\caption{SwitchSplat dataset interaction events across 12 scenes featuring indoor, outdoor, rigid and non-rigid scenarios. Scenes contain 3--9 object-interaction episodes with various temporal structure: interactions may occur sequentially or overlap in time. Sementic labels shown are obtained through VLM predictions.}

    \label{fig:dataset_pattern_12}
\end{figure*}

We captured twelve real-world scenes using an Insta360 X4 dual-fisheye 360 camera, comprising ten standard scenes and two golden scenes. Video frames were extracted at 4 FPS. We use the raw fisheye frames for reconstruction to avoid propagating stitching artifacts into the scene representation. Camera-holder masks are automatically obtained using the FullCircle masking pipeline. These masks, together with the fisheye frames, are passed to COLMAP for camera pose estimation and subsequent image undistortion. Input images are downsampled by a factor of two. An overview of the object-interaction intervals and temporal structure of all twelve scenes is provided in \Cref{fig:dataset_pattern_12}.

\subsection{Benchmark Construction and Evaluation Splits}
\label{sec:sup_benchmark_eval}
We evaluate reconstruction and generalization using four evaluation splits.

\noindent\textbf{Base-state test frames.}
This split contains held-out frames in which all objects remain in their initial base state. It evaluates reconstruction of the canonical scene configuration and provides a natural comparison point for static reconstruction methods.

\noindent\textbf{All standard test frames.}
For each scene, we hold out every eight frame from the original capture sequence. Frames occurring during object transitions are excluded, such that evaluation is performed only on stable object configurations. The remaining frames constitute our standard test set and cover the object states observed throughout the original capture.

\begin{figure}[t]
    \centering
    \includegraphics[width=\columnwidth]{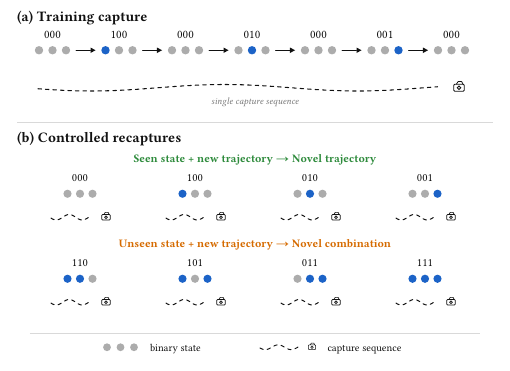}
    \caption{
        \textbf{Golden-set benchmark construction.}
        \textbf{(a)} The training sequence contains one interaction at a time, repeatedly returning to the base state.
        \textbf{(b)} We additionally recapture the scene under controlled object states using new camera trajectories.
        States observed during training evaluate novel camera trajectories, while previously unseen joint states evaluate novel state combinations.
    }
    \label{fig:golden_set_benchmark}
\end{figure}

For the two golden scenes, Kitchen~2 and Oslo, we additionally perform controlled recaptures under prescribed binary object states and new camera trajectories, as illustrated in \Cref{fig:golden_set_benchmark}. These recaptures define two additional evaluation splits.

\noindent\textbf{Novel camera frames.}
We recapture object-state configurations observed during the original training sequence using new camera trajectories. This split evaluates generalization to novel viewpoints while keeping the scene configuration within the set observed during training.

\noindent\textbf{Novel state-combination frames.}
We additionally recapture joint object-state configurations that never occurred during the training sequence, again using new camera trajectories. This split evaluates whether the learned representation can compose individually observed object states into previously unseen scene configurations.

\paragraph{Dynamic-baseline evaluation.}
DeGauss and 4DGS represent scene variation using a continuous timestamp and cannot directly condition on the prescribed state code of a Golden-set test image. For the novel-camera and novel state-combination splits, we therefore evaluate both methods using the same per-image oracle-over-time protocol. We first evaluate 30 uniformly spaced timestamps over $[0,1]$, then evaluate 8 additional timestamps within a $\pm0.03$ neighborhood of the best coarse prediction. The rendering with the highest PSNR across both stages is retained. PSNR, SSIM, and LPIPS are computed from this same selected rendering; SSIM and LPIPS are not independently optimized. Since the ground-truth image is used to select the timestamp, these results should be interpreted as oracle upper bounds.

\subsection{Additional Training Details}
For fair comparison, all methods use corresponding undistorted pinhole images for both training and rendering, with the camera-holder exclusion masks mapped into the respective undistorted image representation.

We initialize our Gaussians from the COLMAP SfM point cloud and use the same densification and pruning schedule as 3DGUT. Gaussian parameters and state logits are optimized with fused Adam, while the slot-window endpoints and temperatures use a separate Adam optimizer; both optimizers are stepped jointly. SwitchSplat, FullCircle, and 4DGS are trained for 30{,}000 iterations with batch size 4. DeGauss uses batch size 4 with a 1{,}000-iteration coarse stage followed by a 30{,}000-iteration fine stage. The learning rates for opacity and state logits are both $5\times10^{-2}$.

The state-sparsity loss has weight $10^{-4}$, starts at iteration 18{,}000, and ramps linearly over 1{,}000 iterations. The compactness loss has weight $5\times10^{-4}$, starts at iteration 21{,}000, and also ramps linearly over 1{,}000 iterations. For each slot, both states are measured relative to the opacity- and assignment-weighted centroid of the active state. After iterations 18{,}000 and 21{,}000, we mean-center the state logits and scale them by $0.3$, reducing assignment confidence as the corresponding regularizer becomes active.

\begin{figure}[h]
  \centering

  \begin{tcblisting}{
    width=\columnwidth,
    listing only,
    listing style=vlmprompt,
    colback=black!1.5,
    colframe=black!22,
    boxrule=0.4pt,
    arc=1.2pt,
    left=5pt,
    right=5pt,
    top=5pt,
    bottom=5pt,
    title={
      \sffamily\bfseries VLM Prompt
      \hfill
      \normalfont\sffamily\fontsize{6.5pt}{7pt}\selectfont
      Exact prompt template
    },
    fonttitle=\footnotesize,
    coltitle=black,
    colbacktitle=black!4,
    boxed title style={
      boxrule=0pt,
      arc=0pt,
    },
    attach boxed title to top left={
      xshift=0pt,
      yshift=0pt
    },
    before skip=0pt,
    after skip=0pt,
  }
You are estimating object interaction times from an ordered sequence of separately attached video frames.

Task:
- The camera observes a user moving through an environment.
- For each interacted object, the user changes the object from off to on, leaves it on for a while, then changes it from on to off.
- Events happen consecutively.
- Each image has its exact frame number rendered on its pixels as t=FRAME.
- Return rough midpoint frame numbers for every visible on transition and off transition.

Rules:
- The top priority is the correct number of interacted objects/events. Rough timing within about 50 frames is still useful.
- Prefer including a low-confidence event over omitting a plausible object-state episode.
- Do not merge two adjacent object episodes just because they look visually similar.
- Detect each distinct localized manipulation episode. Different movable parts or controls of the same larger object are separate events and must not be merged merely because they belong to the same object.
- The sequence may begin with a setup or base-capture traversal in which the camera circles or surveys the scene. Do not count that traversal unless a specific part or control is manipulated and changes state.
- When the manipulated part is obscured, repeated approach-manipulate-withdraw cycles, separated by clear disengagement or repositioning, may establish separate episodes.
- Estimate the midpoint of the user's state-changing action, not only the first frame where the object is perfectly stable or unobstructed.
- Use the rendered t=FRAME labels for transition timing.
- If a transition is uncertain, still give the best approximate frame and lower confidence.
- Keep events in chronological order.
- Do not invent events with no visual support, but missing an event is worse than reporting a plausible low-confidence one.
- Valid frame range is {N_FRAME}

Timestamp convention:
- The attached images are in chronological order.
- They are sampled every 8 frames, with the first and final frames included.
- Each image has its exact frame number rendered on its pixels as t=FRAME.
- The rendered t=FRAME label is authoritative; use the stride only as a consistency check.

Analyze only the attached images. Do not inspect the filesystem, run commands, or use external tools. Return only the JSON object required by the output schema.
  \end{tcblisting}

  \caption{
    \textbf{VLM prompt used for interaction-slot initialization.}
    The prompt template is reproduced verbatim; only line wrapping is
    introduced for typesetting. Scene-dependent quantities, such as the
    valid frame range, are inserted at runtime.
  }
  \label{fig:vlm_prompt}
\end{figure}

\begin{table*}[htb]
  \centering
  \caption{
    \textbf{Full per-scene quantitative results on the standard benchmark.}
    We report PSNR, SSIM, and LPIPS on
    \emph{(a)} base-state test frames outside interaction intervals and
    \emph{(b)} all held-out test frames.
    SwitchSplat with 3DGUT is our canonical configuration; the 3DGS
    variant evaluates effect of the rendering backend.
  }
  \label{tab:supp_full_per_scene}
  \vspace{0.4em}

  \footnotesize
  \setlength{\tabcolsep}{2.5pt}
  \renewcommand{\arraystretch}{1.06}

{\bfseries (a) Base-state test frames\par}
  \vspace{0.25em}

  \begin{tabularx}{0.965\textwidth}{@{}l *{11}{Z}@{}}
    \toprule
    \multirow{2}{*}{Method}
      & \multicolumn{2}{c}{Outdoor}
      & \multicolumn{5}{c}{Indoor}
      & \multicolumn{3}{c}{Concurrent indoor}
      & \multirow{2}{*}{Avg.} \\
    \cmidrule(lr){2-3}
    \cmidrule(lr){4-8}
    \cmidrule(lr){9-11}
      & Balcony
      & Car
      & Kitchen 1
      & Flat 1
      & Flat 2
      & Flat 3
      & Chair
      & \scalebox{0.88}{M Kitchen 1}
      & \scalebox{0.88}{M Kitchen 2}
      & \scalebox{0.88}{M Kitchen 3}
      & \\
    \midrule

    \multicolumn{12}{c}{\textbf{PSNR $\uparrow$}} \\[-1pt]

    FullCircle (3DGS)
      & 26.87 & 27.88 & 31.68 & 29.30 & 30.40
      & 30.72 & 30.57 & 29.76 & 28.32 & 27.30
      & 29.28 \\
    FullCircle (3DGUT)
      & 26.89 & 27.73 & 31.62 & 30.33 & 30.74
      & 31.52 & 31.15 & 30.59 & 28.54 & 28.14
      & 29.72 \\
DeGauss
  & 26.95 & 27.86 & 31.95 & 30.54 & 31.12
  & 30.82 & 34.51 & 33.39 & 31.31 & 31.87
  & 31.03 \\
    4DGS
      & 26.70 & \tabbest{29.19} & 31.37 & 31.58 & 30.66
      & 30.48 & 32.64 & 32.65 & 31.81 & 31.69
      & 30.88 \\

    \addlinespace[1.5pt]
    SwitchSplat (3DGS)
      & 27.19 & 28.89 & 32.82 & 33.80 & 32.89
      & 34.15 & 35.89 & 33.77 & 34.41 & 34.19
      & 32.80 \\
    SwitchSplat (3DGUT)
      & \tabbest{27.28} & 28.95
      & \tabbest{33.01} & \tabbest{34.28}
      & \tabbest{33.09} & \tabbest{34.41}
      & \tabbest{36.52}
      & \tabbest{34.17} & \tabbest{34.80}
      & \tabbest{34.52}
      & \tabbest{33.10} \\

    \midrule
    \multicolumn{12}{c}{\textbf{SSIM $\uparrow$}} \\[-1pt]

    FullCircle (3DGS)
      & 0.856 & 0.893 & 0.967 & 0.951 & 0.961
      & 0.963 & 0.977 & 0.962 & 0.950 & 0.944
      & 0.943 \\
    FullCircle (3DGUT)
      & 0.856 & 0.892 & 0.968 & 0.954 & 0.962
      & 0.965 & 0.978 & 0.965 & 0.950 & 0.949
      & 0.944 \\
DeGauss
  & 0.849 & 0.856 & 0.965 & 0.952 & 0.957
  & 0.959 & 0.980 & 0.971 & 0.962 & 0.965
  & 0.942 \\
    4DGS
      & 0.851 & 0.889 & 0.968 & 0.960 & 0.962
      & 0.962 & 0.981 & 0.975 & 0.970 & 0.970
      & 0.949 \\

    \addlinespace[1.5pt]
    SwitchSplat (3DGS)
      & 0.858 & 0.896
      & \tabbest{0.972} & \tabbest{0.965}
      & \tabbest{0.967} & 0.971
      & \tabbest{0.986}
      & 0.975 & \tabbest{0.975} & 0.975
      & 0.954 \\
    SwitchSplat (3DGUT)
      & \tabbest{0.859} & \tabbest{0.897}
      & \tabbest{0.972} & \tabbest{0.965}
      & \tabbest{0.967} & \tabbest{0.972}
      & \tabbest{0.986}
      & \tabbest{0.976} & \tabbest{0.975}
      & \tabbest{0.976}
      & \tabbest{0.955} \\

    \midrule
    \multicolumn{12}{c}{\textbf{LPIPS $\downarrow$}} \\[-1pt]

    FullCircle (3DGS)
      & 0.279 & \tabbest{0.231}
      & 0.240 & 0.277 & 0.253
      & 0.263 & 0.190 & 0.243 & 0.260 & 0.267
      & 0.250 \\
    FullCircle (3DGUT)
      & \tabbest{0.278} & 0.232
      & 0.241 & 0.275 & 0.253
      & 0.261 & 0.190 & 0.242 & 0.263 & 0.263
      & 0.250 \\
DeGauss
  & 0.306 & 0.276 & 0.251 & 0.288 & 0.266
  & 0.277 & 0.189 & 0.237 & 0.250 & 0.248
  & 0.259 \\
    4DGS
      & 0.284 & 0.232
      & 0.245 & 0.270 & 0.256
      & 0.269 & 0.188
      & \tabbest{0.223} & 0.229 & 0.234
      & 0.243 \\

    \addlinespace[1.5pt]
    SwitchSplat (3DGS)
      & 0.279 & \tabbest{0.231}
      & \tabbest{0.238} & 0.262
      & \tabbest{0.247} & \tabbest{0.253}
      & \tabbest{0.176}
      & 0.230 & \tabbest{0.227}
      & \tabbest{0.229}
      & \tabbest{0.237} \\
    SwitchSplat (3DGUT)
      & 0.279 & \tabbest{0.231}
      & \tabbest{0.238} & \tabbest{0.261}
      & \tabbest{0.247} & \tabbest{0.253}
      & 0.177
      & 0.229 & \tabbest{0.227}
      & \tabbest{0.229}
      & \tabbest{0.237} \\
    \bottomrule
  \end{tabularx}

  \vspace{1em}

{\bfseries (b) All standard test frames\par}
  \vspace{0.25em}

  \begin{tabularx}{0.965\textwidth}{@{}l *{11}{Z}@{}}
    \toprule
    \multirow{2}{*}{Method}
      & \multicolumn{2}{c}{Outdoor}
      & \multicolumn{5}{c}{Indoor}
      & \multicolumn{3}{c}{Concurrent indoor}
      & \multirow{2}{*}{Avg.} \\
    \cmidrule(lr){2-3}
    \cmidrule(lr){4-8}
    \cmidrule(lr){9-11}
      & Balcony
      & Car
      & Kitchen 1
      & Flat 1
      & Flat 2
      & Flat 3
      & Chair
      & \scalebox{0.88}{M Kitchen 1}
      & \scalebox{0.88}{M Kitchen 2}
      & \scalebox{0.88}{M Kitchen 3}
      & \\
    \midrule

    \multicolumn{12}{c}{\textbf{PSNR $\uparrow$}} \\[-1pt]

    FullCircle (3DGS)
      & 25.70 & 26.86 & 29.32 & 29.26 & 28.46
      & 28.92 & 29.88 & 30.16 & 30.13 & 30.10
      & 28.88 \\
    FullCircle (3DGUT)
      & 25.77 & 26.77 & 29.35 & 29.27 & 28.62
      & 29.14 & 30.06 & 30.38 & 30.23 & 30.35
      & 28.99 \\
DeGauss
  & 25.86 & 27.31 & 29.69 & 29.36 & 30.11
  & 28.91 & 34.10 & 32.72 & 32.29 & 32.68
  & 30.30 \\
    4DGS
      & 26.36 & \tabbest{29.14} & 30.83 & 30.78 & 30.17
      & 29.54 & 30.95 & 31.84 & 32.75 & 33.14
      & 30.55 \\

    \addlinespace[1.5pt]
    SwitchSplat (3DGS)
      & 26.89 & 28.92 & 32.32 & 32.87 & 32.58
      & 33.33 & 35.55 & 33.20 & 33.97 & 34.44
      & 32.41 \\
    SwitchSplat (3DGUT)
      & \tabbest{26.97} & 29.03
      & \tabbest{32.55} & \tabbest{33.17}
      & \tabbest{32.78} & \tabbest{33.63}
      & \tabbest{36.04}
      & \tabbest{33.57} & \tabbest{34.17}
      & \tabbest{34.56}
      & \tabbest{32.65} \\

    \midrule
    \multicolumn{12}{c}{\textbf{SSIM $\uparrow$}} \\[-1pt]

    FullCircle (3DGS)
      & 0.841 & 0.881 & 0.954 & 0.947 & 0.950
      & 0.954 & 0.972 & 0.959 & 0.958 & 0.957
      & 0.937 \\
    FullCircle (3DGUT)
      & 0.842 & 0.881 & 0.955 & 0.947 & 0.950
      & 0.954 & 0.972 & 0.959 & 0.959 & 0.959
      & 0.938 \\
DeGauss
  & 0.833 & 0.851 & 0.951 & 0.941 & 0.949
  & 0.949 & 0.978 & 0.964 & 0.964 & 0.965
  & 0.934 \\

    4DGS
  & 0.846 & 0.891 & 0.962 & 0.952 & 0.956
  & 0.955 & 0.973
  & 0.969 & 0.970
  & 0.971
  & 0.944 \\

    \addlinespace[1.5pt]
    SwitchSplat (3DGS)
      & 0.851 & 0.894
      & \tabbest{0.967} & \tabbest{0.959}
      & \tabbest{0.964} & 0.967
      & \tabbest{0.984}
      & 0.969 & 0.971 & 0.973
      & 0.950 \\
    SwitchSplat (3DGUT)
      & \tabbest{0.852} & \tabbest{0.895}
      & \tabbest{0.967} & \tabbest{0.959}
      & \tabbest{0.964} & \tabbest{0.968}
      & \tabbest{0.984}
      & \tabbest{0.970} & \tabbest{0.972}
      & \tabbest{0.974}
      & \tabbest{0.951} \\

    \midrule
    \multicolumn{12}{c}{\textbf{LPIPS $\downarrow$}} \\[-1pt]

    FullCircle (3DGS)
      & 0.293 & 0.247 & 0.263 & 0.290 & 0.270
      & 0.279 & 0.206 & 0.251 & 0.254 & 0.267
      & 0.262 \\
    FullCircle (3DGUT)
      & 0.292 & 0.248 & 0.263 & 0.294 & 0.270
      & 0.280 & 0.207 & 0.251 & 0.255 & 0.266
      & 0.263 \\
DeGauss
  & 0.320 & 0.285 & 0.270 & 0.307 & 0.274
  & 0.292 & 0.199 & 0.246 & 0.247 & 0.258
  & 0.270 \\

    4DGS
  & 0.291 & \tabbest{0.236} & 0.252 & 0.283 & 0.259
  & 0.278 & 0.205
  & \tabbest{0.232} & \tabbest{0.232}
  & \tabbest{0.244}
  & 0.251 \\

    \addlinespace[1.5pt]
    SwitchSplat (3DGS)
      & \tabbest{0.286} & 0.237
      & \tabbest{0.245} & \tabbest{0.273}
      & 0.249 & \tabbest{0.259}
      & \tabbest{0.186}
      & 0.237 & 0.236 & 0.245
      & \tabbest{0.245} \\
    SwitchSplat (3DGUT)
      & \tabbest{0.286} & 0.237
      & \tabbest{0.245} & \tabbest{0.273}
      & \tabbest{0.248} & \tabbest{0.259}
      & \tabbest{0.186}
      & 0.236 & 0.236 & 0.245
      & \tabbest{0.245} \\
    \bottomrule
  \end{tabularx}

  \vspace{1.0em}

\end{table*} 
\subsection{Full Quantitative Evaluation}
For completeness, we report the full per-scene evaluation in terms of PSNR, SSIM, and LPIPS. \Cref{tab:supp_full_per_scene} reports results on the 10 standard benchmark scenes for both the base-state subset and all standard test frames, while \Cref{tab:golden-all-splits} reports results on the two Golden scenes across the base-state, standard-test, novel-camera, and novel state-combination splits. Across these evaluations, SwitchSplat achieves the strongest overall performance, with particularly consistent gains on the more challenging novel-camera and novel state-combination settings. Results with both 3DGS and 3DGUT further show that these improvements are not specific to a particular rendering backend. We provide further qualitative comparisons in \Cref{fig:comparison3} and \Cref{fig:comparison5}.

\subsection{VLM Performance}
\label{sec:vlm_reliability}

We use GPT-5.6 Sol to automatically initialize both the number of interaction
slots and their approximate temporal extents. We uniformly sample the camera
stream every 8 frames, resize each frame to $480\times480$, and render its
frame index directly onto the image. The sampled frames are provided as
separate, chronologically ordered inputs to the VLM, which is prompted to
identify distinct object-state episodes and return approximate on/off
transition timestamps together with a confidence score. We discard predicted
intervals shorter than 20 frames and apply confidence-ranked temporal
non-maximum suppression with an IoU threshold of $0.8$.

\begin{figure}[t]
    \centering
    \includegraphics[width=\columnwidth]
    {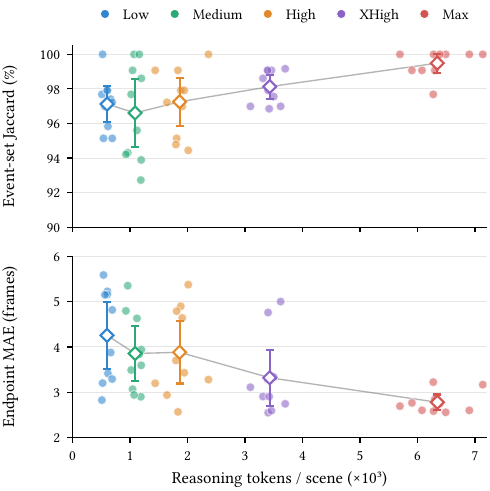}
\caption{\textbf{VLM performance across reasoning-effort levels.}
We evaluate 10 independent predictions at each reasoning-effort level.
Top: event-set Jaccard measures whether the VLM recovers the correct set of interaction slots; bottom: endpoint MAE measures temporal localization error.
Dots denote individual runs and diamonds denote the mean at each effort level.
Increasing reasoning effort generally improves both event recovery and endpoint localization.
At maximum reasoning effort, 6 of 10 runs recover the event set perfectly ($100\%$ Jaccard), with predicted transition endpoints localized to within only a few frames.}
    \label{fig:vlm_prediction_reliability}
\end{figure}

As shown in
Fig.~\ref{fig:vlm_prediction_reliability}, prediction quality generally improves
with increased reasoning effort. At the maximum reasoning setting, the VLM
recovers the complete event set perfectly in 6 out of 10 independent runs,
i.e., achieving $100\%$ event-set Jaccard. Moreover, when the event set is
recovered correctly, the predicted transition times are highly
accurate, with a mean absolute endpoint error within approximately two
frames. 

\begin{figure}[t]
    \centering
    \includegraphics[width=\columnwidth]
        {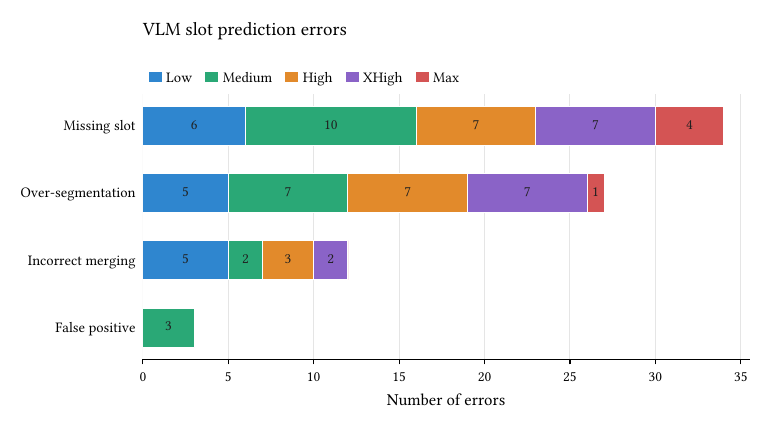}
\caption{\textbf{VLM slot-prediction failure modes.}
We aggregate approximately 3,000 predictions across five reasoning-effort levels, with 10 independent runs per level.
Each bar reports the total number of occurrences of a failure mode, with colored segments indicating contributions from different reasoning-effort levels.
Most errors arise from missing slots or over-segmentation, while incorrect merging is less common and false positives are rare.
At higher reasoning effort, only a small number of failures remain, indicating that the VLM provides a highly reliable initialization in practice.}
    \label{fig:vlm_failure_modes}
    \vspace{-2em}
\end{figure}

We further characterize the remaining prediction errors in
Fig.~\ref{fig:vlm_failure_modes}. Across approximately 3,000 predictions from
five reasoning-effort levels, failures are sparse and are dominated by two
modes: \emph{missing slots}, where an interaction episode is not detected,
and \emph{over-segmentation}, where a single episode is split into multiple
predictions. \emph{Incorrect merging}, where two distinct interaction episodes
are combined into a single predicted slot, occurs less frequently, while
false-positive interactions are rare. At maximum reasoning effort, only a small
number of such errors remain. Together, these results indicate that modern VLMs
can provide a remarkably accurate and repeatable initialization for the
discrete interaction structure required by SwitchSplat.

\paragraph{Prompt and output format.}
We use the same prompt template for all scenes. The complete prompt is
reproduced verbatim in Fig.~\ref{fig:vlm_prompt}. Scene-dependent quantities, such as the valid
frame range, are inserted at runtime.
The VLM is constrained to return structured JSON containing the predicted
object, on/off transition frames, confidence score, uncertainty description,
and supporting evidence for each detected event. 

\subsection{Ablations on Slot Initialization}

We examine how errors from slot initialization affect the subsequent optimization with two factors: the number of interaction slots and the temporal endpoints of each slot. 

\paragraph{Number of slots.}
\Cref{tab:slot_count_ablation} varies the number of slots on Kitchen~2, for which the correct value is $K=3$. Removing slots causes a consistent degradation in reconstruction quality. By contrast, overestimating the slot count has little effects since extra slots can remain inactive during optimization.

\begin{table}[htb]
    \centering
    \caption{
        \textbf{Sensitivity to VLM-predicted slot count on the Kitchen 2 scene.}
        The VLM predicts three interaction slots ($K{=}3$).
        Performance degrades when slots are removed, while adding up to three
        extra slots has little effect.
    }
    \label{tab:slot_count_ablation}
    \small
    \setlength{\tabcolsep}{5pt}
    \resizebox{\columnwidth}{!}{\begin{tabular}{lccccccc}
        \toprule
        & \multicolumn{3}{c}{Missing slots}
        & \multicolumn{1}{c}{Exact slots}
        & \multicolumn{3}{c}{Extra slots} \\
        \cmidrule(lr){2-4}
        \cmidrule(lr){5-5}
        \cmidrule(lr){6-8}
        Number of slots ($K$)
            & 0 & 1 & 2
            & \cellcolor{gray!10}3
            & 4 & 5 & 6 \\
        \midrule
        PSNR $\uparrow$
            & 33.31 & 34.05 & 35.11
            & \cellcolor{gray!10}\textbf{35.98}
            & 35.92 & \textbf{35.98} & 35.96 \\
        SSIM $\uparrow$
            & 0.97 & 0.97 & \textbf{0.98}
            & \cellcolor{gray!10}\textbf{0.98}
            & \textbf{0.98} & \textbf{0.98} & \textbf{0.98} \\
        LPIPS $\downarrow$
            & 0.19 & 0.18 & 0.18
            & \cellcolor{gray!10}\textbf{0.17}
            & \textbf{0.17} & \textbf{0.17} & \textbf{0.17} \\
        \bottomrule
    \end{tabular}}
\end{table}

\begin{figure}[htb]
    \centering
    \includegraphics[width=\columnwidth]{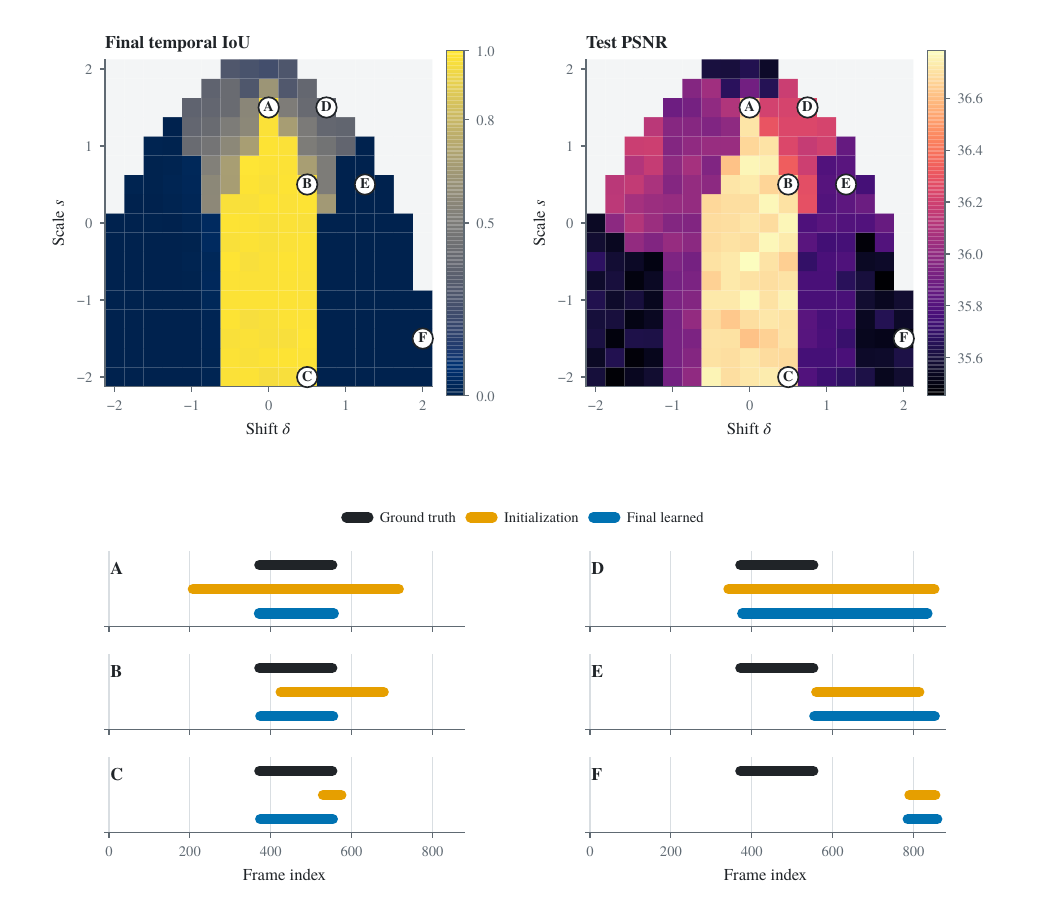}
\caption{
\textbf{Sensitivity of slot initialization on Kitchen 2.}
The heatmaps report final temporal IoU and test PSNR across endpoint shifts and scales.
The bottom rows compare the ground truth, initialization, and final learned intervals for six marked settings.
Cases A--C recover from substantial perturbations, while D--F show failures under more extreme perturbations.
}
    \label{fig:slot_init_sensitivity}
\end{figure}

\paragraph{Slot endpoints.}
We examine how accurately each interaction must be localized in time. \Cref{fig:slot_init_sensitivity} perturbs the initialization of the middle slot in Kitchen~2 by independently varying its center and duration. Let $c$ denote the interval center and $w$ its duration. We parameterize slot center displacement by
\[
\delta =
\frac{c_{\mathrm{init}}-c_{\mathrm{GT}}}{w_{\mathrm{GT}}},
\]
and duration scales by
\[
s =
\log_2
\frac{w_{\mathrm{init}}}{w_{\mathrm{GT}}}.
\]
All runs use $8\times$ image downsampling.

The resulting heatmaps reveal a broad region in which substantially different initial intervals converge to nearly the same learned solution. Within this region, both the final temporal IoU and test PSNR remain high despite large errors in the initial center or duration; cases A--C illustrate several such trajectories. Only for sufficiently extreme perturbations does optimization converge to an incorrect temporal interval, as illustrated by cases D--F.

\subsection{Effect of Dynamic Moving Backgrounds} 
Our outdoor scenes contain substantial temporal variation unrelated to the discrete object-state changes targeted by SwitchSplat, including wind-driven vegetation, changing sky and clouds, and distant street activity. Such variation can disproportionately affect full-frame image metrics and partially obscure differences in the reconstruction quality of the interactive scene content. We therefore additionally evaluate all methods after excluding these moving-background regions, while otherwise retaining the same evaluation protocol. As shown in \Cref{tab:moving_background}, under the original evaluation, SwitchSplat (3DGUT) achieves $29.03$\,dB compared with $29.14$\,dB for 4DGS, despite producing visibly cleaner reconstructions. After excluding the moving background, SwitchSplat improves to $32.87$\,dB, compared with $31.87$\,dB for 4DGS, while also achieving the best SSIM and LPIPS. A similar trend is observed on Balcony, where SwitchSplat remains the strongest method. These results suggest that the apparent discrepancy between quantitative and qualitative performance on the outdoor scenes is largely attributable to unrelated background dynamics.

\begin{table}[htb]
\centering
\small
\setlength{\tabcolsep}{2.2pt}
\renewcommand{\arraystretch}{1.15}

\resizebox{\columnwidth}{!}{\begin{tabular}{lccc@{\hspace{6pt}}ccc}
\toprule
& \multicolumn{3}{c}{Kitchen 2}
& \multicolumn{3}{c}{Oslo} \\
\cmidrule(lr){2-4}
\cmidrule(lr){5-7}

Method
& PSNR $\uparrow$
& SSIM $\uparrow$
& LPIPS $\downarrow$
& PSNR $\uparrow$
& SSIM $\uparrow$
& LPIPS $\downarrow$ \\

\midrule
\multicolumn{7}{c}{\textit{Base-state test frames}} \\
\addlinespace[1pt]

FullCircle (3DGS)
& 33.82 & 0.972 & 0.175
& 33.20 & 0.976 & 0.190 \\

FullCircle (3DGUT)
& 33.88 & 0.972 & 0.175
& 32.84 & 0.975 & 0.189 \\

DeGauss
& 34.37 & 0.971 & 0.184
& 34.97 & 0.980 & 0.184 \\

4DGS
& 33.79 & 0.975 & 0.171
& 34.68 & 0.981 & 0.178 \\

SwitchSplat (3DGS)
& 35.95 & 0.979 & \textbf{0.167}
& 36.52 & 0.984 & 0.178 \\

SwitchSplat (3DGUT)
& \textbf{36.17} & \textbf{0.979} & 0.167
& \textbf{36.73} & \textbf{0.984} & \textbf{0.177} \\

\addlinespace[3pt]
\midrule
\multicolumn{7}{c}{\textit{All standard test frames}} \\
\addlinespace[1pt]

FullCircle (3DGS)
& 33.21 & 0.969 & 0.185
& 31.12 & 0.967 & 0.205 \\

FullCircle (3DGUT)
& 33.31 & 0.970 & 0.186
& 31.30 & 0.968 & 0.204 \\

DeGauss
& 34.21 & 0.970 & 0.187
& 32.87 & 0.969 & 0.198 \\

4DGS
& 34.08 & 0.975 & 0.174
& 33.65 & 0.977 & \textbf{0.183} \\

SwitchSplat (3DGS)
& 35.84 & 0.978 & \textbf{0.173}
& 34.97 & 0.979 & 0.184 \\

SwitchSplat (3DGUT)
& \textbf{35.98} & \textbf{0.978} & 0.173
& \textbf{35.18} & \textbf{0.980} & 0.184 \\

\addlinespace[3pt]
\midrule
\multicolumn{7}{c}{\textit{Novel camera frames}} \\
\addlinespace[1pt]

FullCircle (3DGS)
& 32.25 & 0.967 & 0.166
& 28.33 & 0.956 & 0.211 \\

FullCircle (3DGUT)
& 32.55 & 0.967 & 0.166
& 28.73 & 0.957 & 0.210 \\

DeGauss$^{\dagger}$
& 32.38 & 0.965 & 0.174
& 29.40 & 0.960 & 0.207 \\

4DGS$^{\dagger}$
& 32.05 & 0.967 & 0.168
& 29.93 & 0.966 & 0.194 \\

SwitchSplat (3DGS)
& 33.90 & 0.974 & 0.155
& 31.82 & 0.972 & 0.186 \\

SwitchSplat (3DGUT)
& \textbf{34.26} & \textbf{0.975} & \textbf{0.154}
& \textbf{32.55} & \textbf{0.973} & \textbf{0.185} \\

\addlinespace[3pt]
\midrule
\multicolumn{7}{c}{\textit{Novel state-combination frames}} \\
\addlinespace[1pt]

FullCircle (3DGS)
& 29.91 & 0.955 & 0.178
& 26.14 & 0.932 & 0.216 \\

FullCircle (3DGUT)
& 30.10 & 0.955 & 0.177
& 26.44 & 0.934 & 0.215 \\

DeGauss$^{\dagger}$
& 30.85 & 0.958 & 0.180
& 27.51 & 0.940 & 0.214 \\

4DGS$^{\dagger}$
& 29.63 & 0.956 & 0.182
& 26.53 & 0.940 & 0.211 \\

SwitchSplat (3DGS)
& 31.29 & 0.962 & 0.164
& \textbf{29.29} & \textbf{0.959} & 0.186 \\

SwitchSplat (3DGUT)
& \textbf{32.58} & \textbf{0.967} & \textbf{0.161}
& 29.10 & 0.958 & \textbf{0.186} \\

\bottomrule
\end{tabular}}

\caption{
\textbf{Full quantitative evaluation on the Golden scenes.}
We report PSNR, SSIM, and LPIPS on Kitchen~2 and Oslo across base-state, standard-test, novel-camera, and novel state-combination frames.
For the two OOD splits, DeGauss$^{\dagger}$ and 4DGS$^{\dagger}$ are evaluated using the oracle-over-time upper bound described in \Cref{sec:sup_benchmark_eval}, while SwitchSplat is evaluated directly from the requested state code.
Best results are shown in \textbf{bold}.
}
\label{tab:golden-all-splits}
\end{table} 

\begin{table}[htb]
    \centering
\caption{ \textbf{Evaluation on outdoor scenes excluding moving backgrounds.} We compare the original full-frame evaluation against an evaluation that excludes temporally varying background regions, including foliage, sky, clouds, and street activity. Results are reported on the full test set.} \label{tab:moving_background}
    \label{tab:moving_background_diagnostic}

    \small
    \setlength{\tabcolsep}{2.2pt}
    \renewcommand{\arraystretch}{1.15}

    \resizebox{\columnwidth}{!}{\begin{tabular}{lccc@{\hspace{6pt}}ccc}
        \toprule
        & \multicolumn{3}{c}{Balcony}
        & \multicolumn{3}{c}{Car} \\
        \cmidrule(lr){2-4}
        \cmidrule(lr){5-7}

        Method
        & PSNR $\uparrow$
        & SSIM $\uparrow$
        & LPIPS $\downarrow$
        & PSNR $\uparrow$
        & SSIM $\uparrow$
        & LPIPS $\downarrow$ \\

        \midrule
        \multicolumn{7}{c}{\textit{Full evaluation}} \\
        \addlinespace[1pt]

DeGauss
& 25.86 & 0.833 & 0.320
& 27.31 & 0.851 & 0.285 \\

4DGS
& 26.36 & 0.846 & 0.291
& \textbf{29.14} & 0.891 & \textbf{0.236} \\

SwitchSplat (3DGS)
& 26.89 & 0.851 & \textbf{0.286}
& 28.92 & 0.894 & 0.237 \\

SwitchSplat (3DGUT)
& \textbf{26.97}
& \textbf{0.852}
& \textbf{0.286}
& 29.03
& \textbf{0.895}
& 0.237 \\

        \addlinespace[3pt]
        \midrule
        \multicolumn{7}{c}{\textit{Moving background excluded}} \\
        \addlinespace[1pt]

        DeGauss
        & 26.34 & 0.843 & 0.303
        & 30.58 & 0.903 & 0.148 \\

        4DGS
        & 26.83 & 0.852 & 0.279
        & 31.87 & 0.924 & 0.119 \\

SwitchSplat (3DGS)
& 27.44 & 0.860 & 0.274
& 32.84 & 0.932 & 0.111 \\

        SwitchSplat (3DGUT)
        & \textbf{27.50}
        & \textbf{0.860}
        & \textbf{0.274}
        & \textbf{32.87}
        & \textbf{0.932}
        & \textbf{0.111} \\

        \bottomrule
    \end{tabular}}
\end{table}

\subsection{Comparison with Articulation baselines}
\begin{table}[htb]
\centering
\small
\caption{\textbf{Quantitative comparison on the PARIS~\cite{paris} dataset.}
We report PSNR (dB) across all ten articulated objects.
For ArticulatedGS~\cite{articulatedgs}, we average the published
start- and end-state PSNR for each object; SplArt~\cite{splart}
reports the corresponding per-object results directly. For 3DGUT~\cite{3dgut} we report the average of models trained on separate states.
Best and second-best results are shown in \textbf{bold} and
\underline{underline}, respectively.}
\label{tab:paris}
\setlength{\tabcolsep}{4.0pt}
\renewcommand{\arraystretch}{1.05}
\begin{tabular}{lcccc}
\toprule
Scene
& \shortstack{Articulated\\GS}
& SplArt
& \shortstack{Independent\\3DGUT}
& \shortstack{SwitchSplat\\(3DGUT)} \\
\midrule
USB
    & 41.68 & 45.70 & \underline{49.18} & \textbf{49.72} \\
Blade
    & 41.34 & 46.15 & \textbf{50.85} & \underline{50.06} \\
Folding chair
    & 42.26 & 43.98 & \underline{50.22} & \textbf{50.42} \\
Fridge
    & 37.26 & 41.40 & \underline{43.10} & \textbf{44.18} \\
Laptop
    & 36.23 & 40.90 & \underline{45.55} & \textbf{45.84} \\
Oven
    & 36.66 & 41.59 & \underline{43.03} & \textbf{44.62} \\
Scissors
    & 41.06 & 46.15 & \textbf{49.81} & \underline{49.73} \\
Stapler
    & 43.10 & 44.81 & \textbf{50.59} & \underline{49.71} \\
Storage
    & 38.37 & 42.62 & \underline{43.28} & \textbf{44.16} \\
Washer
    & 40.31 & 43.75 & \underline{46.51} & \textbf{47.78} \\
\midrule
\textbf{Average}
    & 39.82 & 43.70 & \underline{47.21} & \textbf{47.62} \\
\bottomrule
\end{tabular}
\end{table} SwitchSplat targets general object-state changes in full scenes, whereas ArticulatedGS~\cite{articulatedgs} and SplArt~\cite{splart} are designed specifically for part-level articulation of isolated objects. Their articulation-specific constraints prevent them from training on our multi-object dataset. However, SwitchSplat can be directly applied to PARIS~\cite{paris}, the benchmark used by both ArticulatedGS and SplArt. As shown in \Cref{tab:paris}, SwitchSplat achieves $47.62$\,dB, substantially outperforming ArticulatedGS ($39.82$\,dB) and SplArt ($43.70$\,dB). To rule out an evaluation discrepancy, we trained independent 3DGUT models on each articulation state separately, obtaining $47.21$\,dB; this sanity check confirms that the high PSNR achieved by SwitchSplat is indeed attainable. We attribute the gap to the additional constraints imposed by explicit articulation modeling, which create a harder optimization landscape at the expense of photometric reconstruction quality.

\subsection{Training on Native Fisheye Frames}
\label{sec:sup_pinhole_fisheye}
\begin{table}[htb]
\centering
\small
\setlength{\tabcolsep}{2.2pt}
\renewcommand{\arraystretch}{1.15}

{\resizebox{\columnwidth}{!}{\begin{tabular}{lccc@{\hspace{6pt}}ccc}
\toprule
& \multicolumn{3}{c}{Tested on pinhole}
& \multicolumn{3}{c}{Tested on fisheye} \\
\cmidrule(lr){2-4}
\cmidrule(lr){5-7}

Trained on
& PSNR $\uparrow$
& SSIM $\uparrow$
& LPIPS $\downarrow$
& PSNR $\uparrow$
& SSIM $\uparrow$
& LPIPS $\downarrow$ \\

\midrule
\multicolumn{7}{c}{\textit{Base-state test frames}} \\
\addlinespace[1pt]

Pinhole
& \textbf{36.17} & \textbf{0.979} & \textbf{0.167}
& 34.88 & 0.979 & 0.090 \\

Fisheye
& 35.66 & 0.978 & 0.167
& \textbf{37.09} & \textbf{0.983} & \textbf{0.084} \\

\addlinespace[3pt]
\midrule
\multicolumn{7}{c}{\textit{All standard test frames}} \\
\addlinespace[1pt]

Pinhole
& \textbf{35.98} & \textbf{0.978} & \textbf{0.173}
& 34.12 & 0.976 & 0.094 \\

Fisheye
& 35.49 & 0.977 & 0.174
& \textbf{36.59} & \textbf{0.982} & \textbf{0.088} \\

\addlinespace[3pt]
\midrule
\multicolumn{7}{c}{\textit{Novel camera frames}} \\
\addlinespace[1pt]

Pinhole
& 34.26 & 0.975 & 0.154
& 34.63 & 0.979 & 0.080 \\

Fisheye
& \textbf{34.63} & \textbf{0.976} & \textbf{0.152}
& \textbf{35.74} & \textbf{0.982} & \textbf{0.076} \\

\addlinespace[3pt]
\midrule
\multicolumn{7}{c}{\textit{Novel state-combination frames}} \\
\addlinespace[1pt]

Pinhole
& 32.58 & 0.967 & 0.161
& 33.56 & 0.976 & 0.083 \\

Fisheye
& \textbf{33.12} & \textbf{0.972} & \textbf{0.156}
& \textbf{34.69} & \textbf{0.980} & \textbf{0.079} \\

\bottomrule
\end{tabular}}}

\caption{{\textbf{Pinhole vs.\ fisheye training on Kitchen~2.}
Rows give the training camera, and columns give the test camera.
Compare numbers within a column only, because the two columns score different images.
Best result per column in \textbf{bold}.
On fisheye views, both models sort Gaussians by distance from the camera.
Each row is a single training run.
}}
\label{tab:pinhole_fisheye}
\end{table}
 \leavevmode{This section and \Cref{sec:sup_mask_finetune} describe two extensions of the pipeline presented in the main paper.
The main paper trains all methods on undistorted pinhole images.
This choice serves only a fair comparison, since existing baselines do not support our fisheye setup.
SwitchSplat itself can train directly on the raw fisheye frames, which cover a wider field of view.

To evaluate whether this is beneficial, we train one model on the pinhole frames of Kitchen~2 and one on its fisheye frames.
We then evaluate both models on both camera types (\Cref{tab:pinhole_fisheye}).
The pinhole frames are the COLMAP-undistorted fisheye frames used in the main paper, so the two datasets share camera poses, splits, and object timings.

On the base-state and standard test frames, each model performs best on its own camera type, which produces a diagonal pattern in \Cref{tab:pinhole_fisheye}.
These frames are close to the training views, so each model can fit its own image type closely.
The two gaps differ in size, however.
The fisheye-trained model is only about 0.5\,dB behind on pinhole views, whereas the pinhole-trained model is more than 2\,dB behind on fisheye views.

On novel cameras and novel state combinations, the settings that matter most for our task, the fisheye-trained model performs best in both columns.
Part of its advantage on fisheye views comes from the outer rim of the lens, which a pinhole camera never observes.
\Cref{fig:pinhole_fisheye_example} shows a qualitative example.
We therefore recommend training on fisheye frames whenever they are available.
}
\begin{figure}[t]
    \centering
    \newcommand{\pfrow}[1]{\rotatebox{90}{\makebox[0.47\columnwidth][c]{\footnotesize #1}}}
    {\setlength{\tabcolsep}{1pt}\renewcommand{\arraystretch}{1.0}\begin{tabular}{@{}ccc@{}}
        &
        \footnotesize Tested on pinhole &
        \footnotesize Tested on fisheye \\

        \pfrow{Ground truth} &
        \includegraphics[width=0.47\columnwidth]{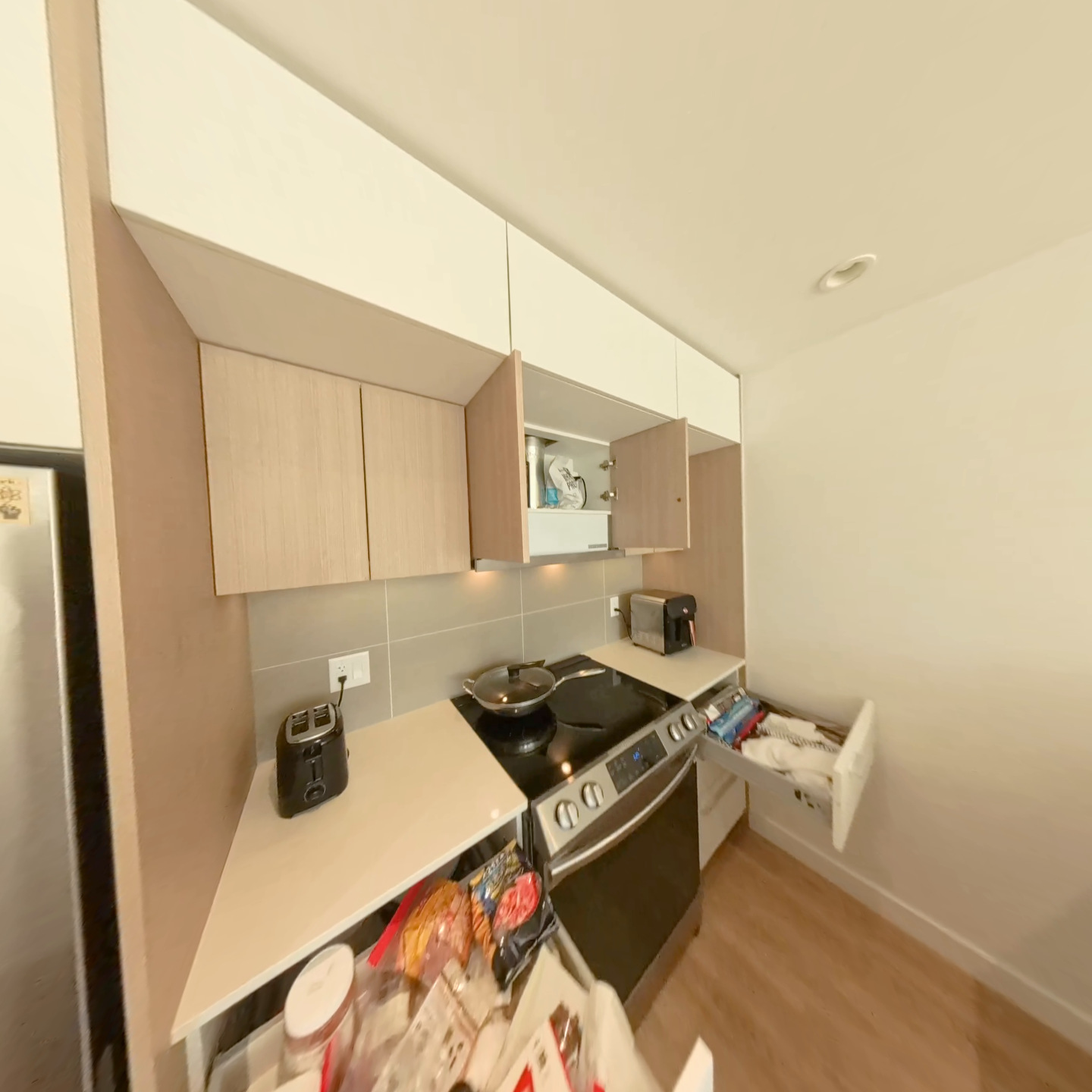} &
        \includegraphics[width=0.47\columnwidth]{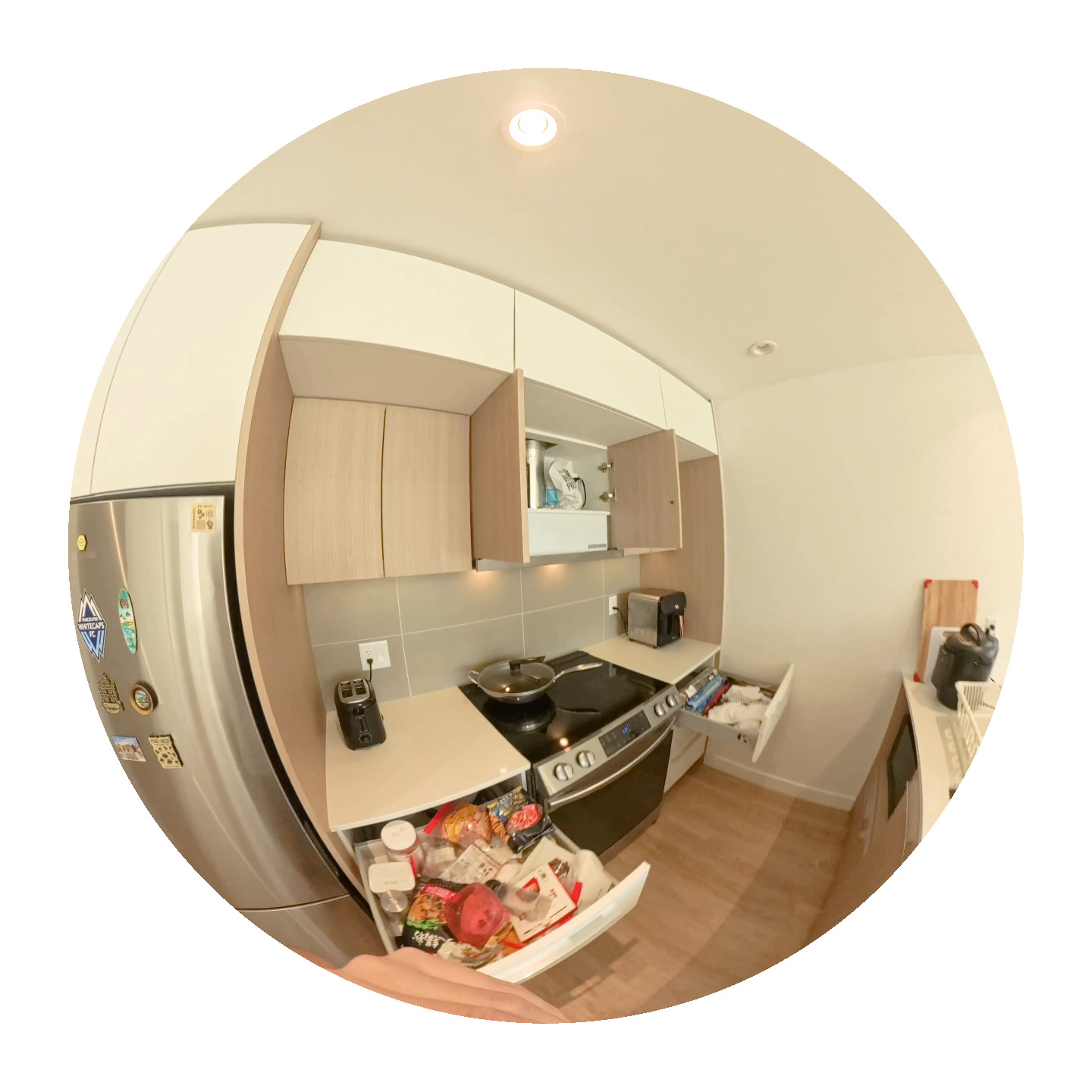} \\

        \pfrow{Trained on pinhole} &
        \includegraphics[width=0.47\columnwidth]{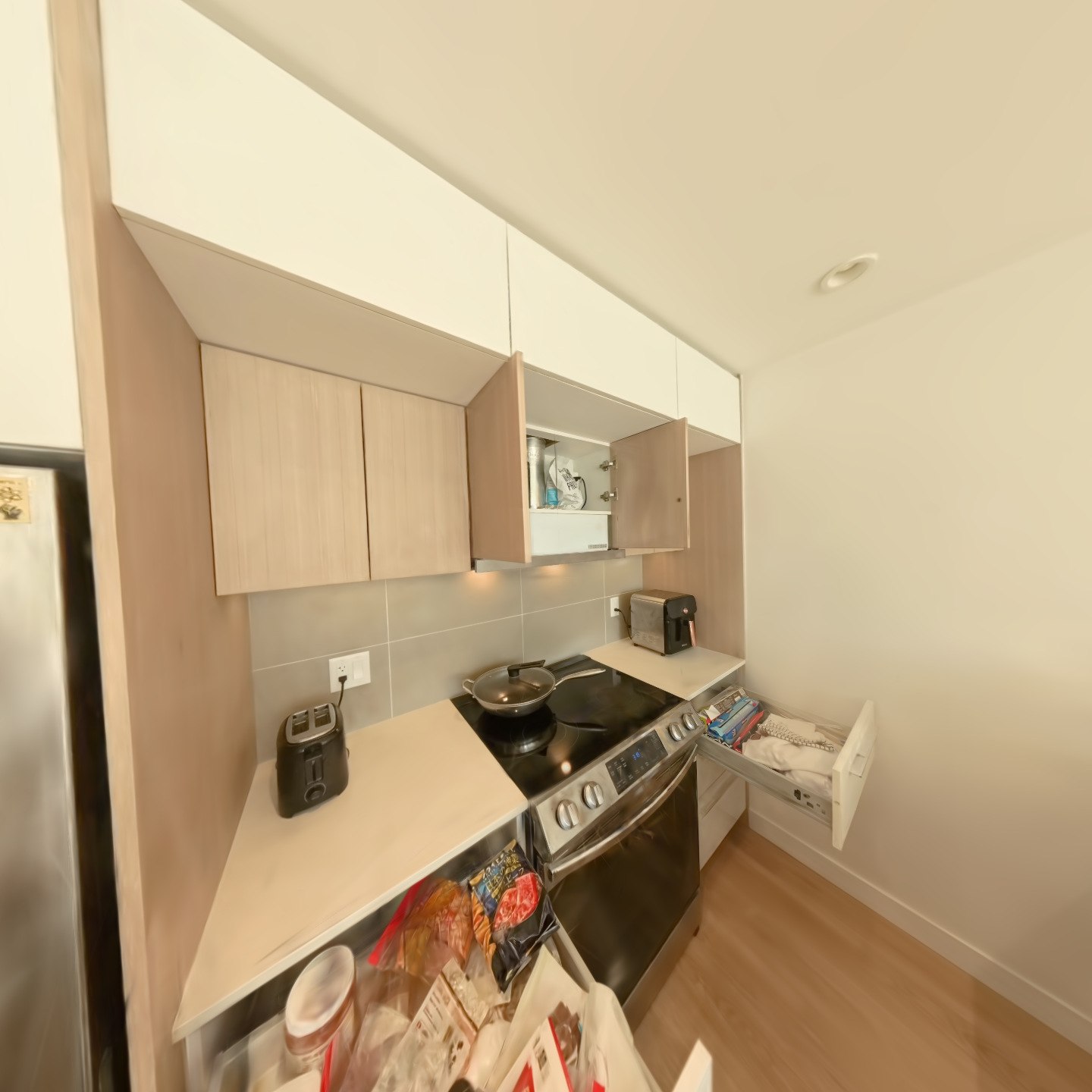} &
        \includegraphics[width=0.47\columnwidth]{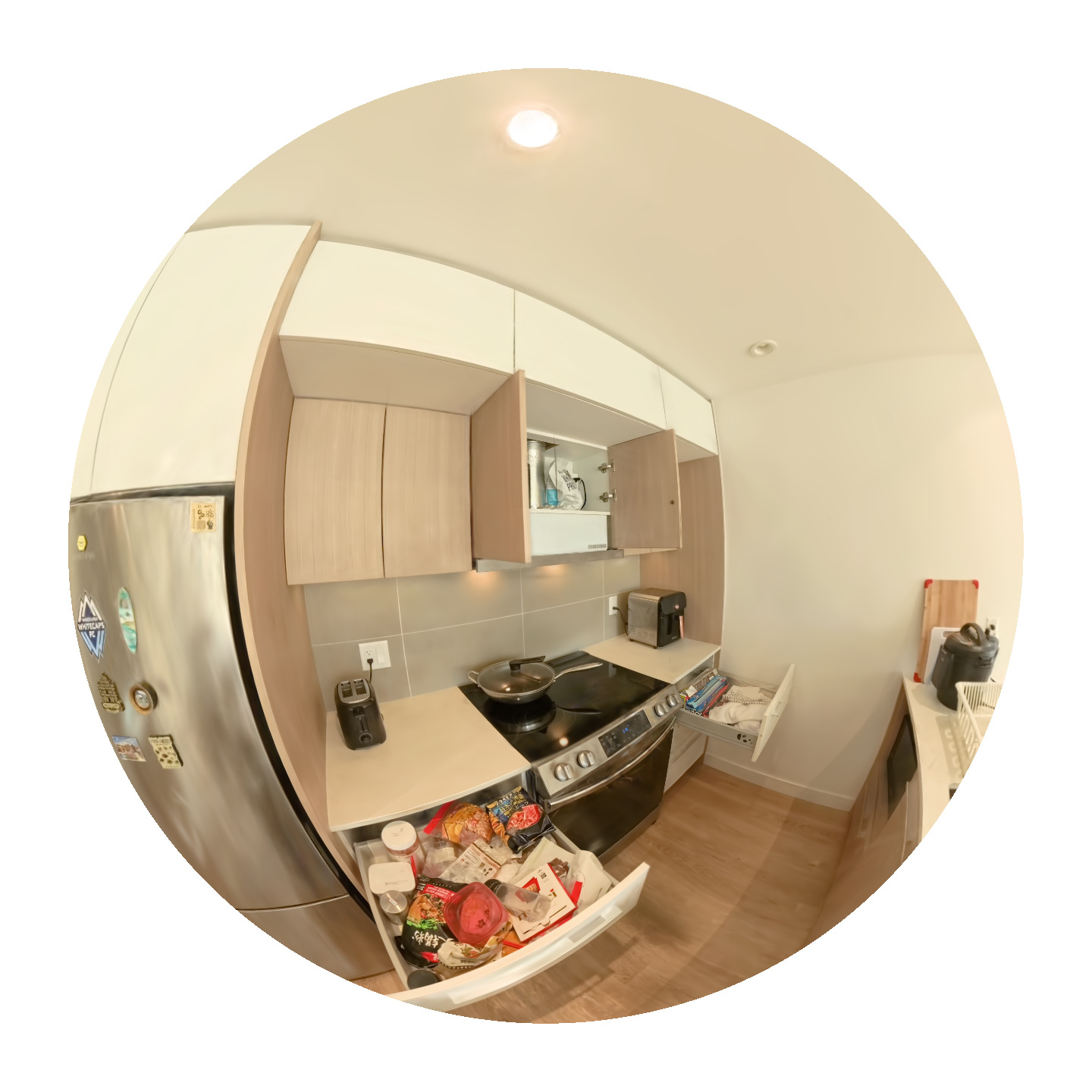} \\

        \pfrow{Trained on fisheye} &
        \includegraphics[width=0.47\columnwidth]{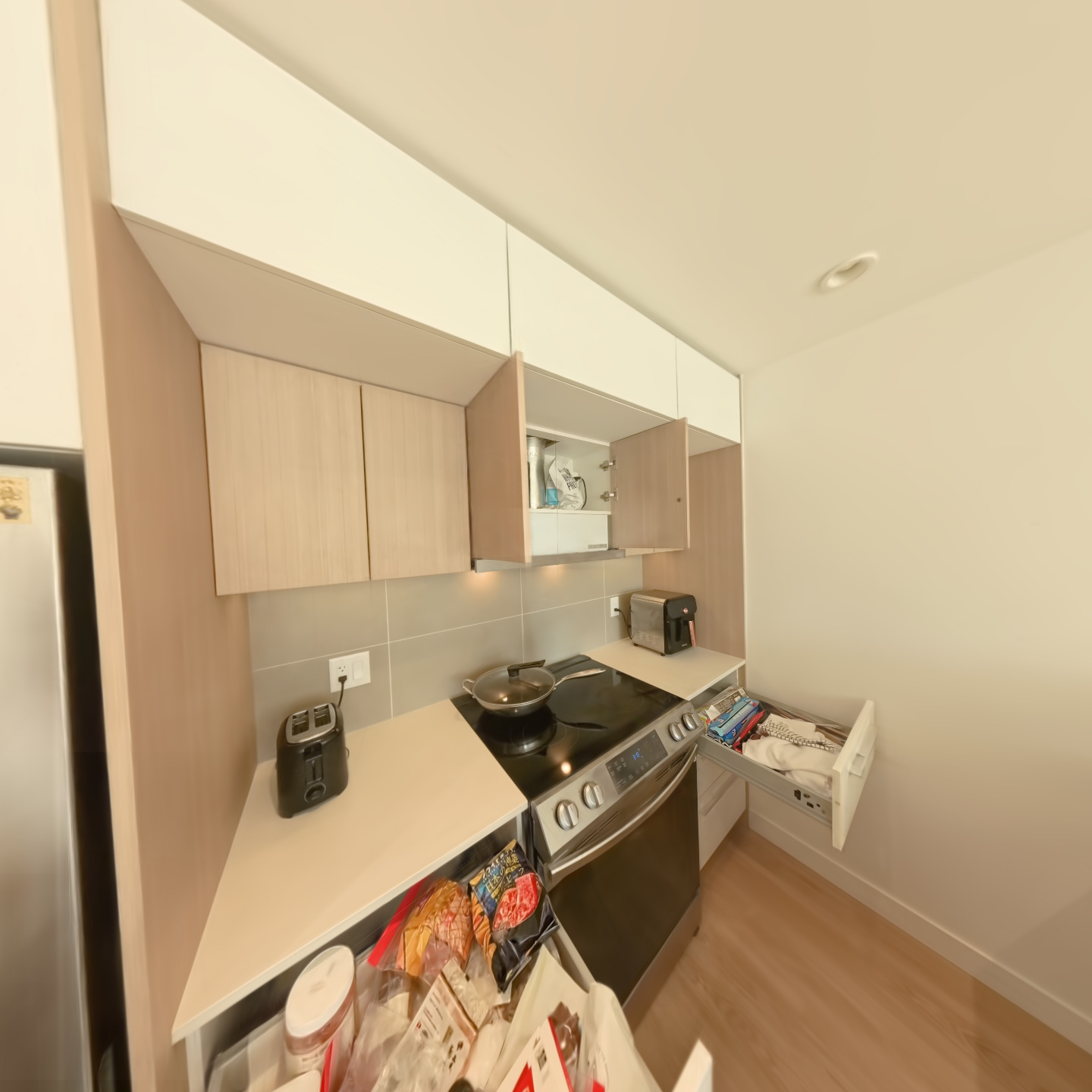} &
        \includegraphics[width=0.47\columnwidth]{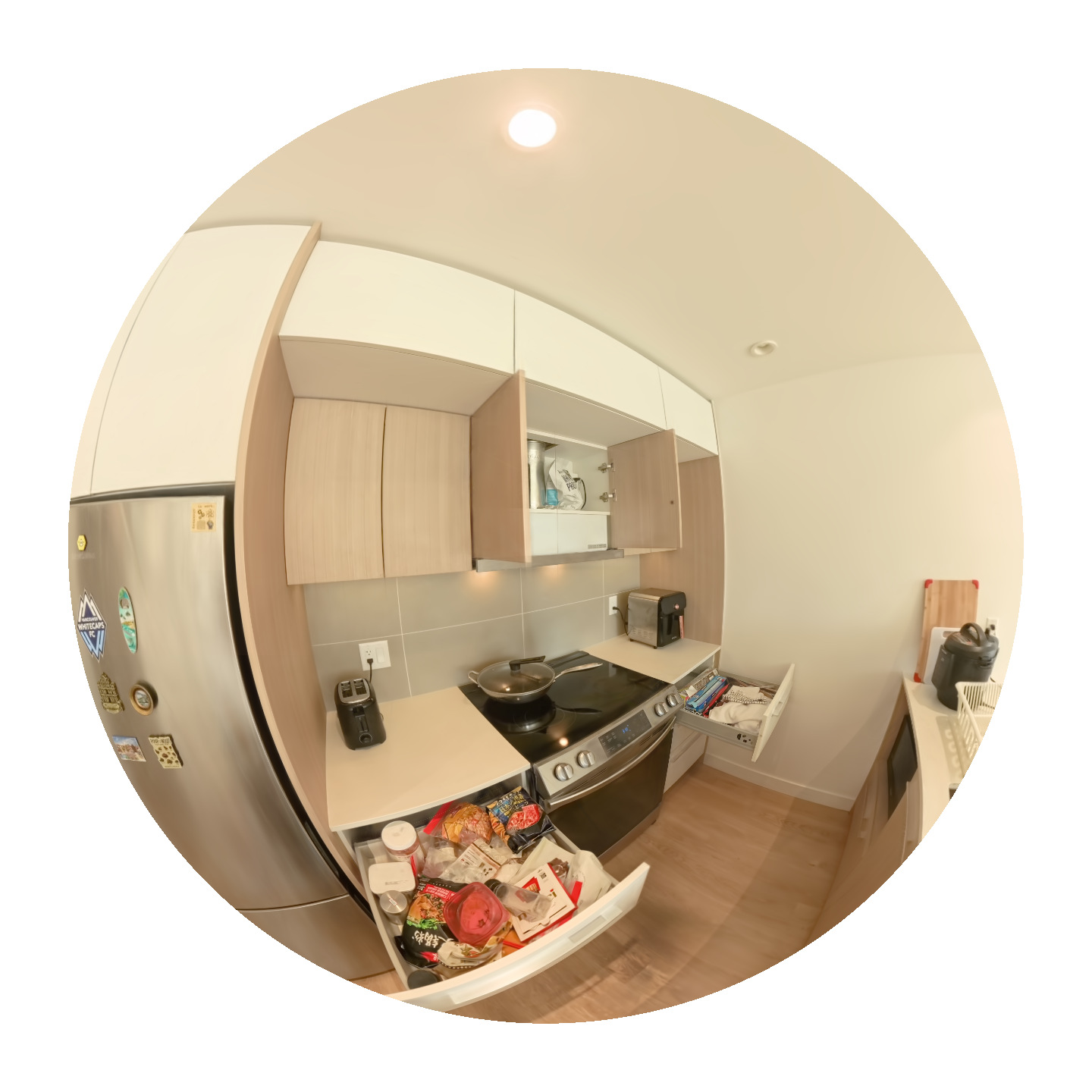} \\
    \end{tabular}}
    \caption{{\textbf{Pinhole vs.\ fisheye training, qualitative example.}
        A novel state-combination view of Kitchen~2, with all three objects switched at once.
        The top row shows the ground truth, the other rows give the training camera, and the columns give the test camera.
        The pinhole-trained model blurs the fridge door and leaves floaters in front of the oven and on the floor, most visibly on the fisheye view.
        The fisheye-trained model is cleaner overall and closer to the ground truth.
    }}
    \Description{A three-row, two-column image grid of a kitchen. The left column shows pinhole images and the right column shows circular fisheye images on a white background. The top row is the ground truth, the middle row is the pinhole-trained model, and the bottom row is the fisheye-trained model.}
    \label{fig:pinhole_fisheye_example}
\end{figure}
 
\subsection{Automatic Mask-Guided Slot Refinement}
\label{sec:sup_mask_finetune}
\begin{table}[htb]
\centering
\small
\setlength{\tabcolsep}{2.2pt}
\renewcommand{\arraystretch}{1.15}

{\resizebox{\columnwidth}{!}{\begin{tabular}{lccc@{\hspace{6pt}}ccc}
\toprule
& \multicolumn{3}{c}{Without mask fine-tuning}
& \multicolumn{3}{c}{With mask fine-tuning} \\
\cmidrule(lr){2-4}
\cmidrule(lr){5-7}

Split
& PSNR $\uparrow$
& SSIM $\uparrow$
& LPIPS $\downarrow$
& PSNR $\uparrow$
& SSIM $\uparrow$
& LPIPS $\downarrow$ \\

\midrule
Base state
& \textbf{37.09} & \textbf{0.983} & \textbf{0.084}
& 36.92 & 0.983 & 0.084 \\

Standard test
& \textbf{36.59} & \textbf{0.982} & \textbf{0.088}
& 36.34 & 0.981 & 0.089 \\

Novel camera
& \textbf{35.74} & 0.982 & 0.076
& 35.73 & \textbf{0.982} & \textbf{0.076} \\

Novel combination
& 34.69 & 0.980 & 0.079
& \textbf{35.23} & \textbf{0.981} & \textbf{0.078} \\

\bottomrule
\end{tabular}}}

\caption{{\textbf{Mask fine-tuning improves novel state combinations.}
Kitchen~2, fisheye test views.
The model without mask fine-tuning is the fisheye-trained model of \Cref{tab:pinhole_fisheye}.
Best result per split in \textbf{bold}.
Each column group is a single training run.
}}
\label{tab:mask_finetune}
\end{table}
 \leavevmode{During an interaction, the video also records changes away from the object itself.
Some changes are unrelated to the object, such as the reflection of the person who holds the camera on the steel fridge in \Cref{fig:mask_pipeline}.
Others are caused by the object but appear elsewhere, because a change of its state can alter the lighting of the surrounding scene, especially when the object is close to a light source.
The Gaussians of the corresponding slot can learn to explain both kinds of change.
When objects are switched later, these Gaussians appear as haze and floaters away from the object.
The effect is strongest when many objects are switched at once (\Cref{fig:maskft_before_after}).

To address this problem, we compute an accurate mask for each object.
We then briefly fine-tune the model, so that each slot keeps its Gaussians inside the object mask and discards those outside it.

\noindent\textbf{Object masks.}
The mask pipeline is fully automatic and requires no human annotation.
For each training frame during an interaction, we render the same view with the object switched back to its base state.
We then compare this render with the photo.
Thresholding this difference yields pseudo-labels: pixels with a large difference are labeled as object, and pixels with almost no difference as background.
Following RoMo~\cite{romo}, we train a small MLP on these pseudo-labels, with DINOv3~\cite{dinov3} features and pixel positions as input.
The MLP then predicts a mask for the entire frame.

Finally, SAM~2.1~\cite{sam} refines the masks with video propagation.
It propagates the MLP masks of the odd frames to the even frames, and the masks of the even frames to the odd frames.
Each final mask thus draws on neighboring frames rather than on a single prediction, which reduces errors that are specific to one frame.
Each mask covers the whole region that changes with the object, including the space that the object reveals when it opens.
\Cref{fig:mask_pipeline} illustrates these stages for one frame.

\noindent\textbf{Mask fine-tuning.}
We fine-tune the fisheye model of \Cref{sec:sup_pinhole_fisheye} for 3{,}000 iterations with one additional loss.
This loss renders each object state in isolation and penalizes its opacity outside the object mask.

\noindent\textbf{Results.}
\Cref{tab:mask_finetune} shows the expected trade-off.
On the base-state and standard test frames, the scores decrease slightly, by less than 0.3\,dB.
The removed Gaussians also explained lighting effects in these held-out frames.
On novel state combinations, where objects are switched independently, PSNR improves by 0.54\,dB.
Novel-camera scores remain unchanged.
\Cref{fig:maskft_before_after} shows a novel view without and with mask fine-tuning.
}
\begin{figure}[t]
    \centering
    \newcommand{\mpimg}[1]{\includegraphics[width=0.325\columnwidth]{figs/mask_pipeline/#1.png}}
    {\setlength{\tabcolsep}{1pt}\renewcommand{\arraystretch}{1.0}\begin{tabular}{@{}ccc@{}}
        \mpimg{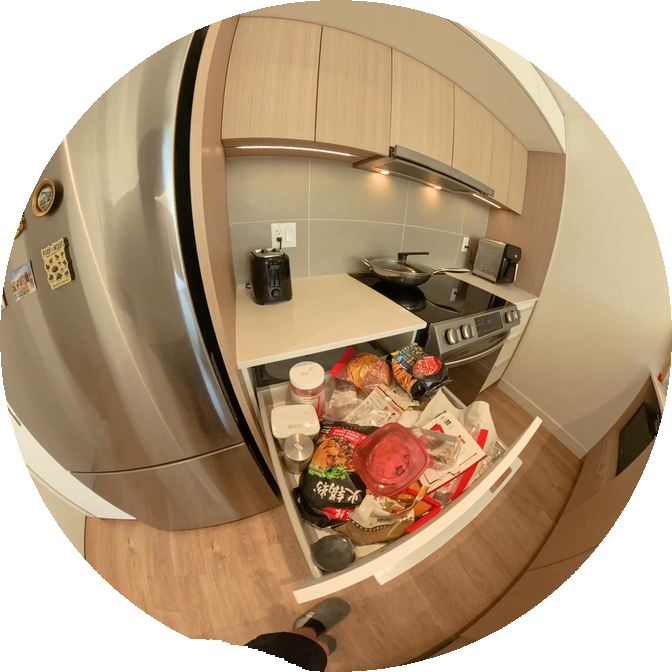} & \mpimg{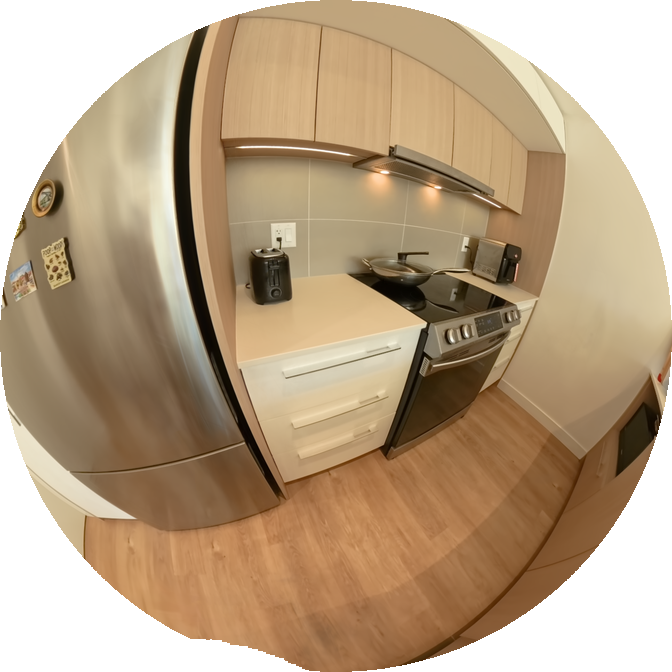} & \mpimg{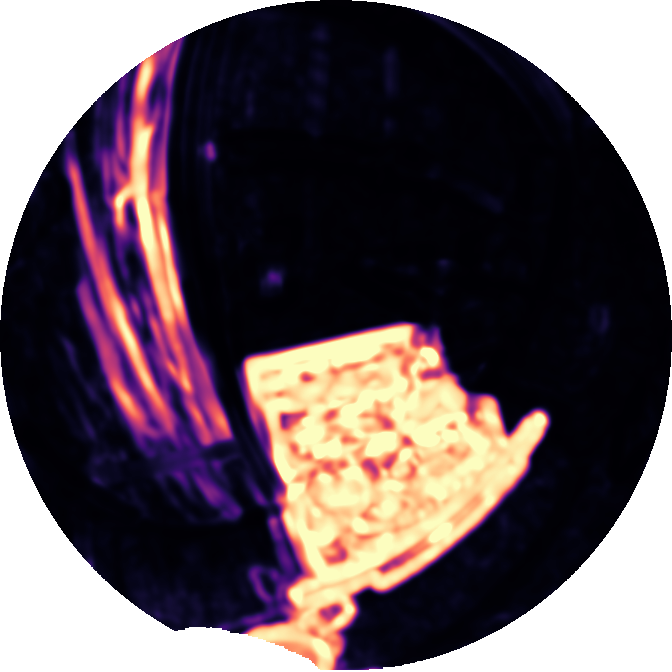} \\
        \footnotesize (a) Photo &
        \footnotesize (b) Render &
        \footnotesize (c) $1-\mathrm{SSIM}$ \\[3pt]
        \mpimg{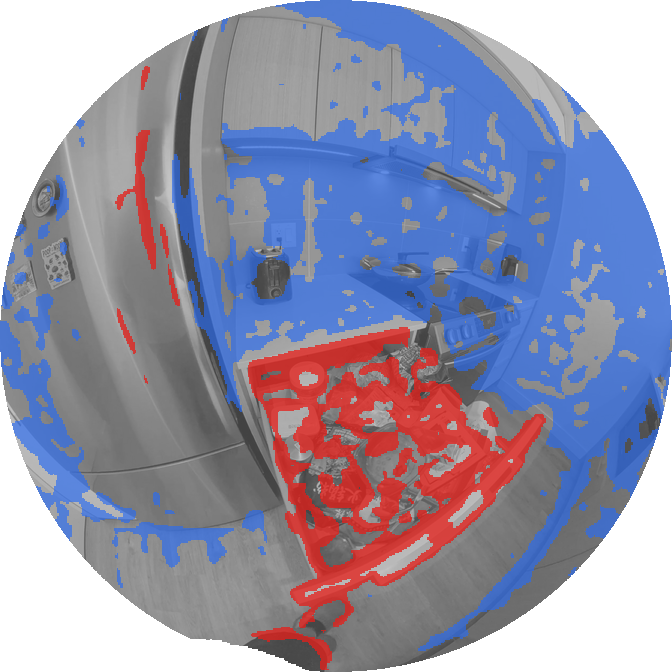} & \mpimg{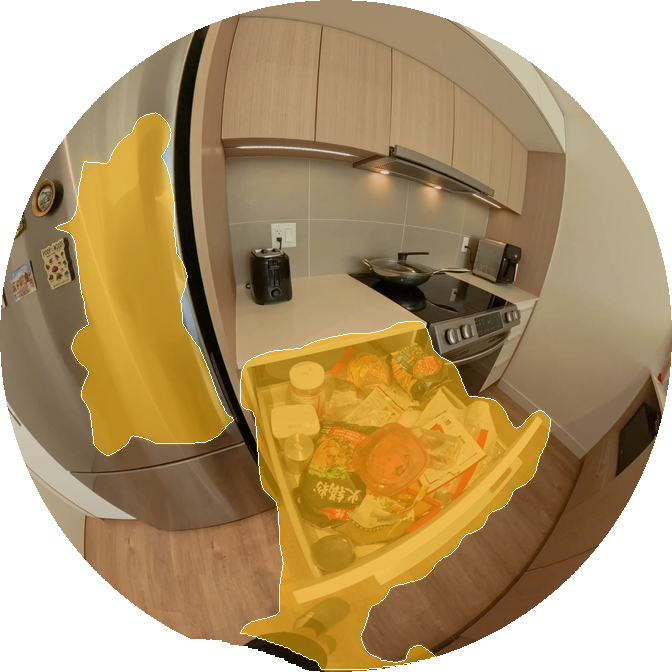} & \mpimg{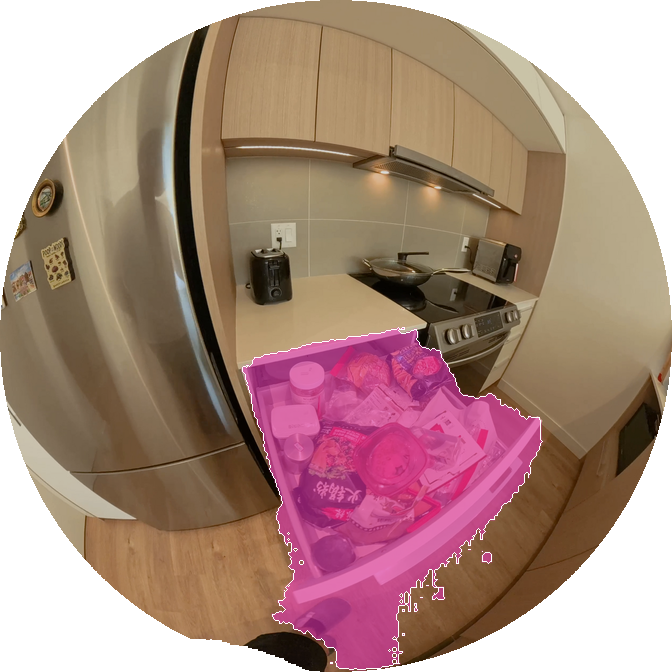} \\
        \footnotesize (d) Pseudo-labels &
        \footnotesize (e) MLP mask &
        \footnotesize (f) Final mask \\
    \end{tabular}}
    \caption{{\textbf{Automatic object mask.}
        The open drawer of Kitchen~2 in one training frame.
        (a)~Photo.
        (b)~Render with the drawer switched back to its closed state.
        (c)~$1-\mathrm{SSIM}$ between (a) and (b).
        (d)~Pseudo-labels from thresholding (c): object (red) and background (blue).
        (e)~Mask predicted by the MLP.
        (f)~Final mask after SAM~2.1 propagation.
        Reflections on the steel fridge, including one of the person who holds the camera, also appear as change and produce wrong object pseudo-labels.
        The MLP inherits these errors and spills onto the fridge and the floor, whereas the propagated mask stays on the drawer.
    }}
    \Description{A two-row, three-column grid of circular fisheye images of a kitchen with an open freezer drawer, on a white background. The top row shows the photo, a render with the drawer closed, and a heat map of image differences that is bright on the drawer and on reflections of the steel fridge. The bottom row shows red and blue pseudo-label pixels, a yellow MLP mask that covers the drawer but also part of the fridge door and the floor, and a magenta final mask that covers only the drawer region.}
    \label{fig:mask_pipeline}
\end{figure}
 \begin{figure}[b]
    \centering
    {\setlength{\tabcolsep}{0pt}\renewcommand{\arraystretch}{1.0}\begin{tabular}{@{}c@{}}
        \footnotesize Without mask fine-tuning \\
        \includegraphics[width=\columnwidth]{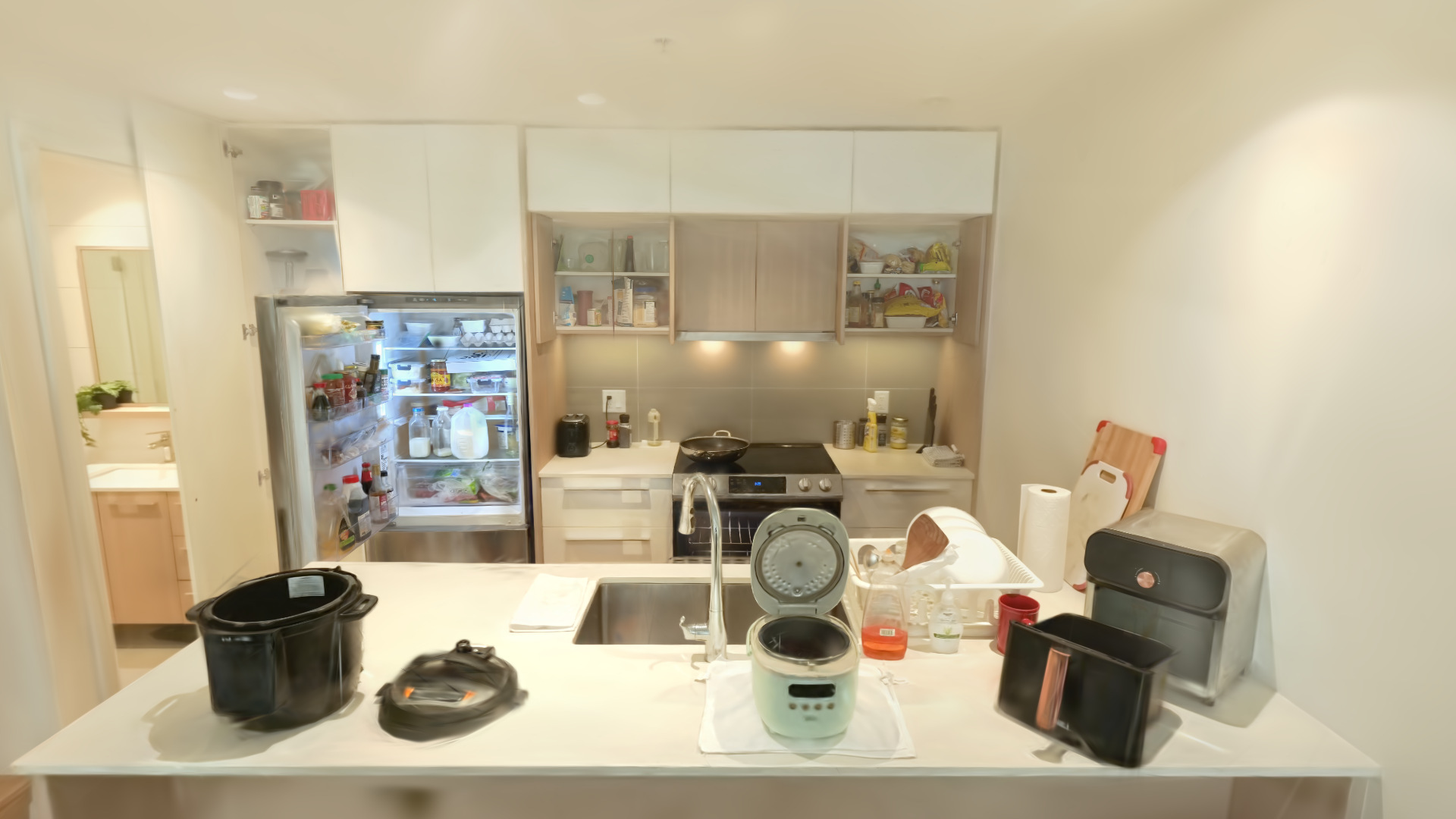} \\[3pt]
        \footnotesize With mask fine-tuning \\
        \includegraphics[width=\columnwidth]{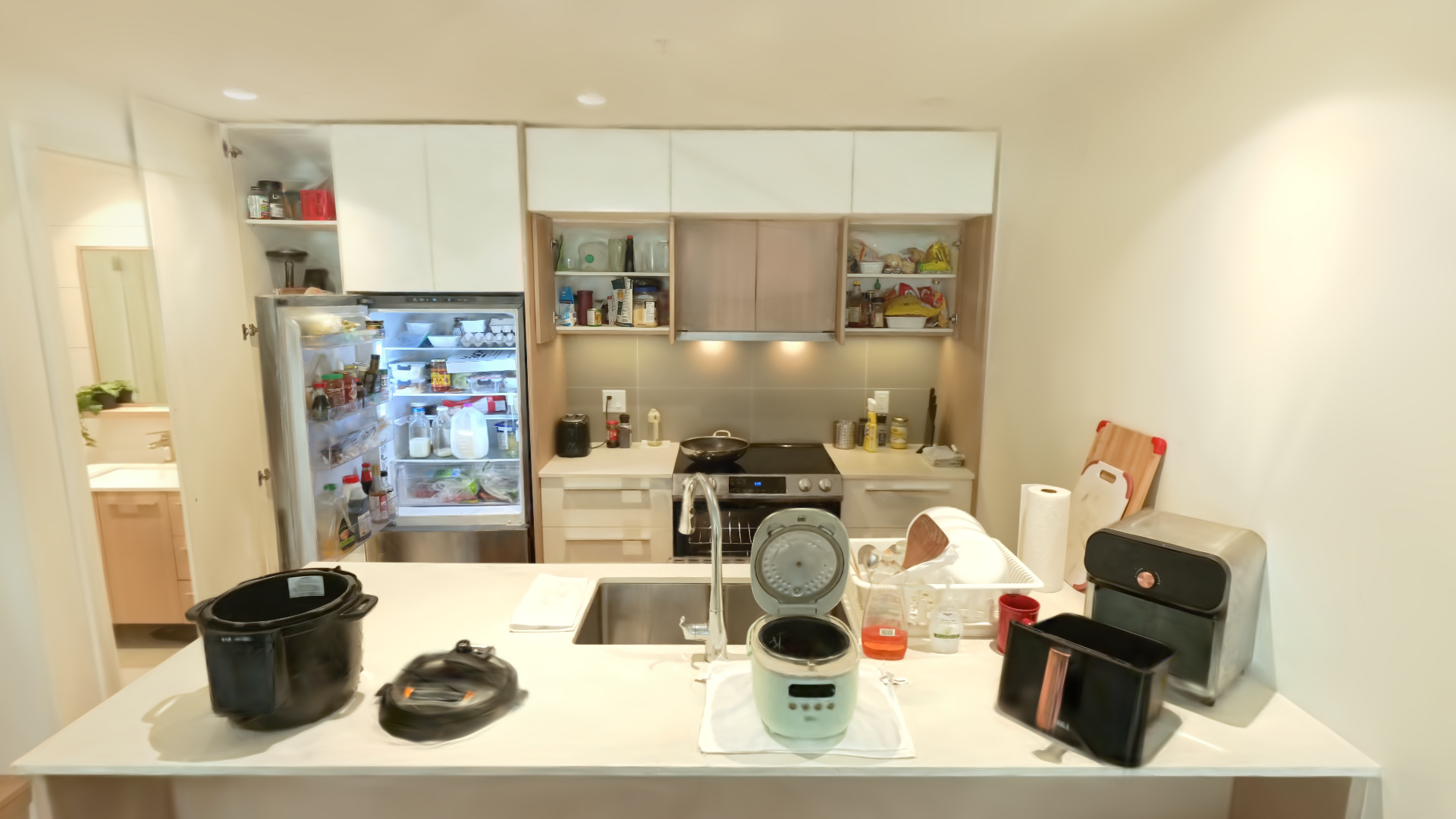} \\
    \end{tabular}}
    \caption{{\textbf{Mask fine-tuning removes haze and floaters.}
        Kitchen~1 with all nine objects switched at once, seen from a new viewpoint with no ground truth.
        (Top)~Without mask fine-tuning, haze and floaters appear in the scene.
        (Bottom)~With mask fine-tuning, the scene is cleaner.
    }}
    \Description{Two renders of the same kitchen view, one above the other, without and with mask fine-tuning. A fridge, cabinets, and several appliances are open.}
    \label{fig:maskft_before_after}
\end{figure}

\begin{figure*}[p]
    \centering
    \setlength{\tabcolsep}{1pt}\renewcommand{\arraystretch}{1.0}

    \begin{tabular}{@{}ccccc@{}}
        \footnotesize FullCircle &
        \footnotesize DeGauss &
        \footnotesize 4DGS &
        \footnotesize SwitchSplat &
        \footnotesize Ground Truth \\[2pt]

        \includegraphics[width=0.18\textwidth]{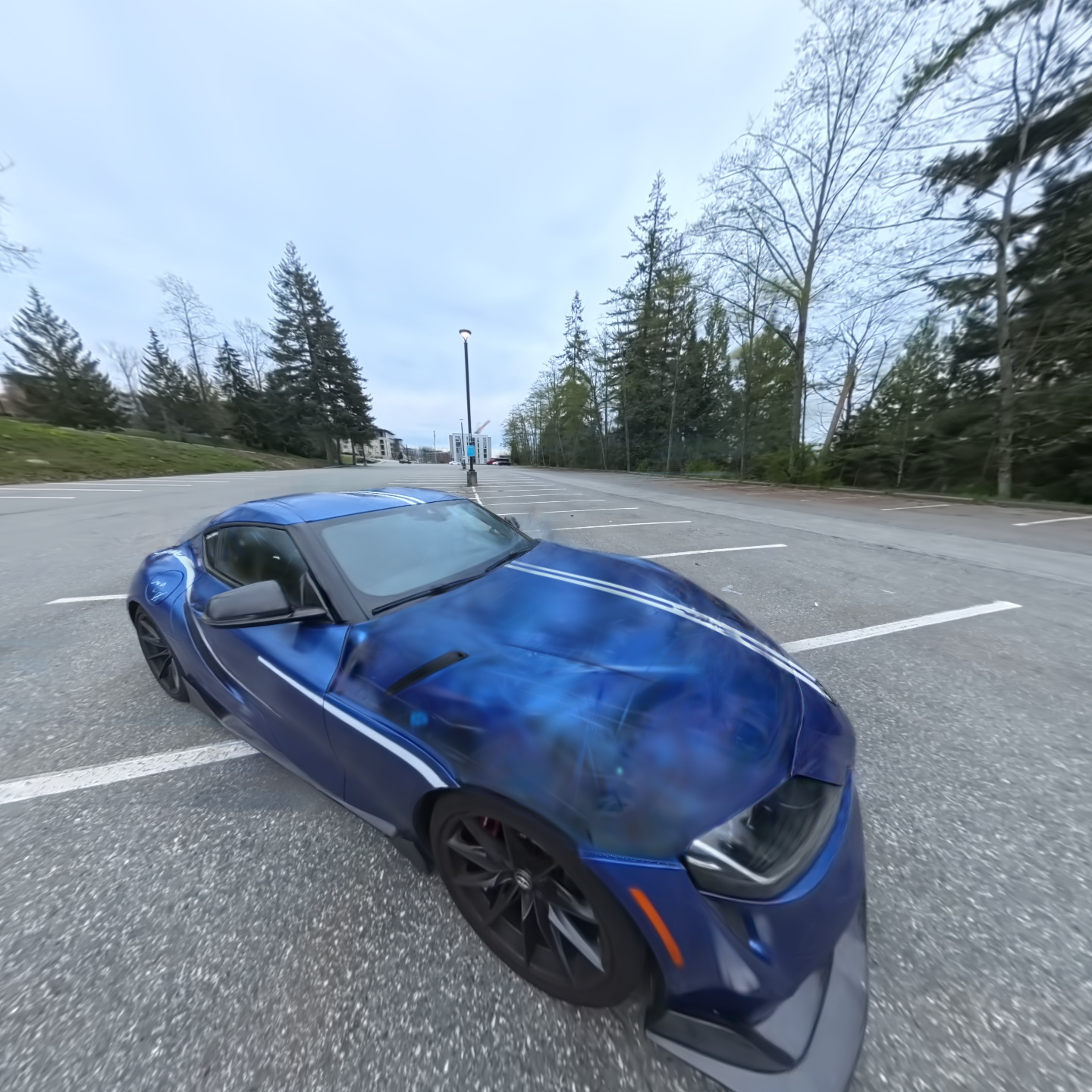} &
        \includegraphics[width=0.18\textwidth]{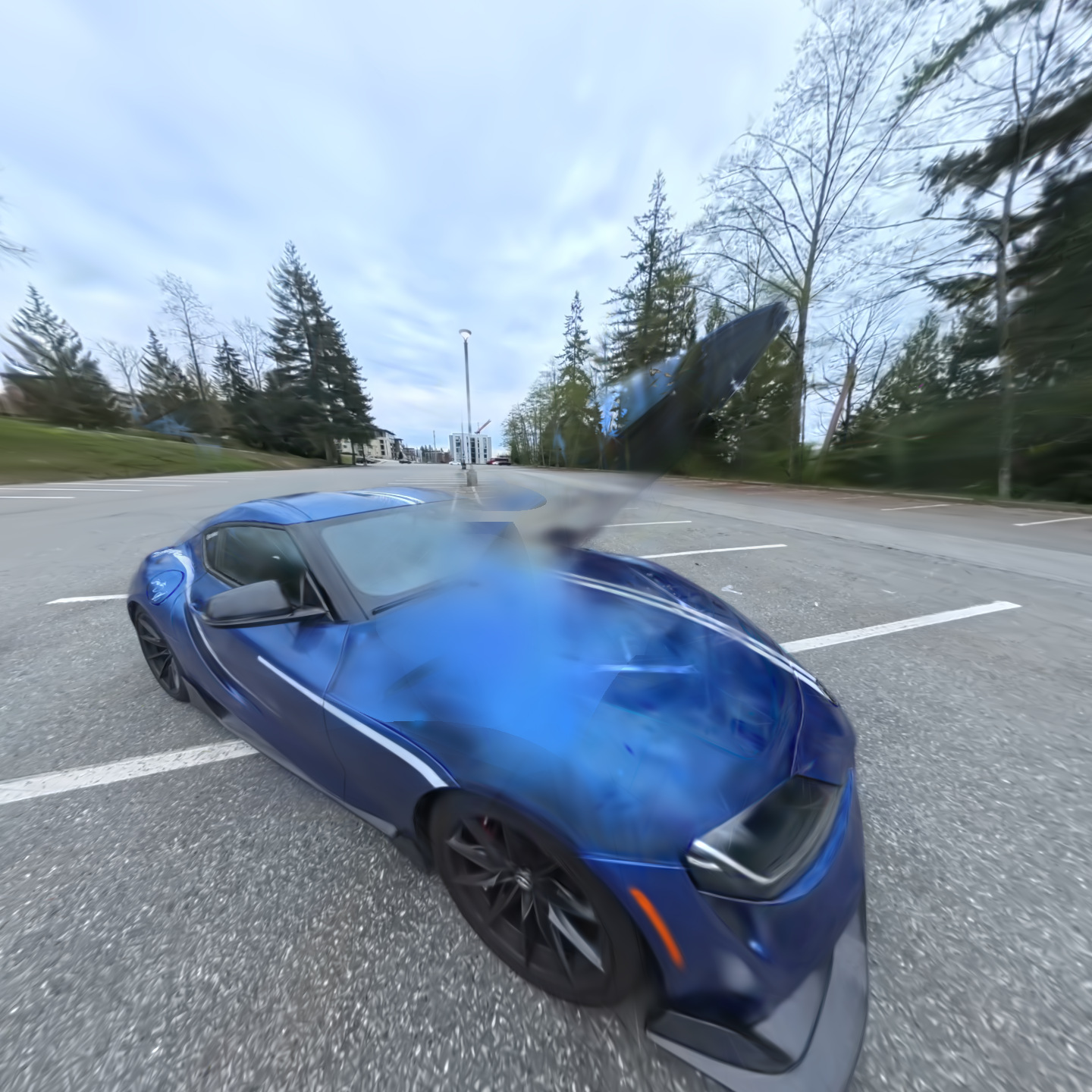} &
        \includegraphics[width=0.18\textwidth]{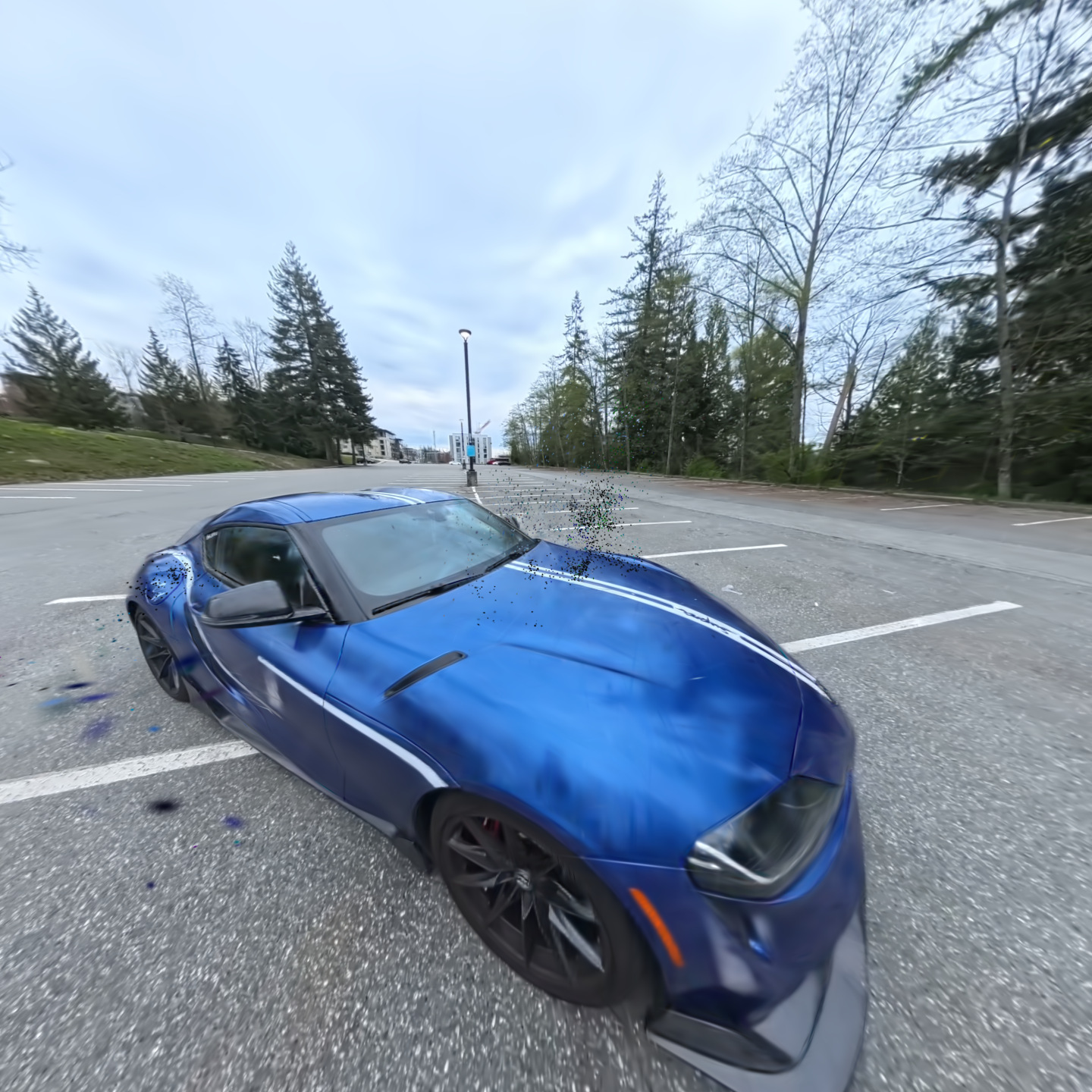} &
        \includegraphics[width=0.18\textwidth]{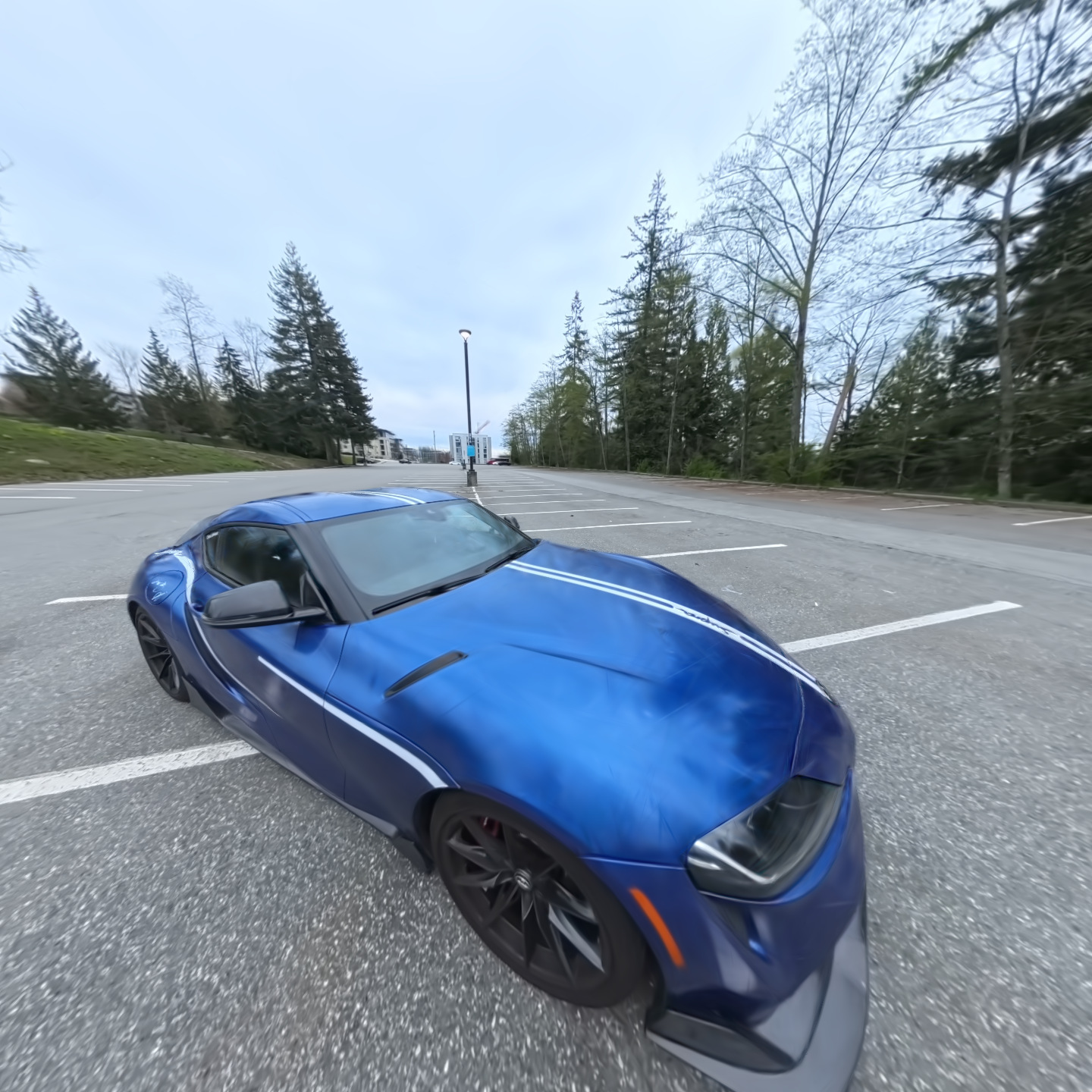} &
        \includegraphics[width=0.18\textwidth]{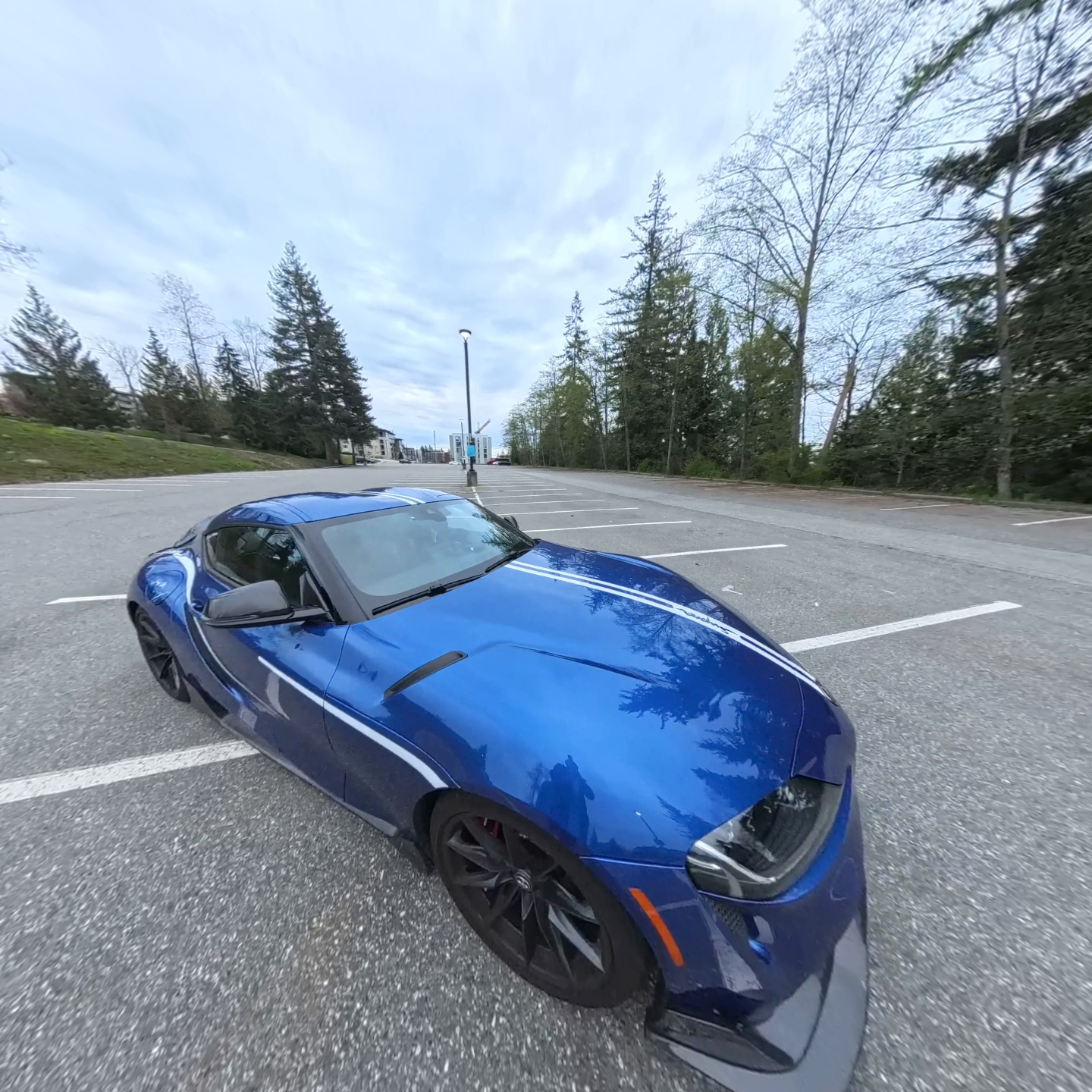} \\[1pt]

        \includegraphics[width=0.18\textwidth]{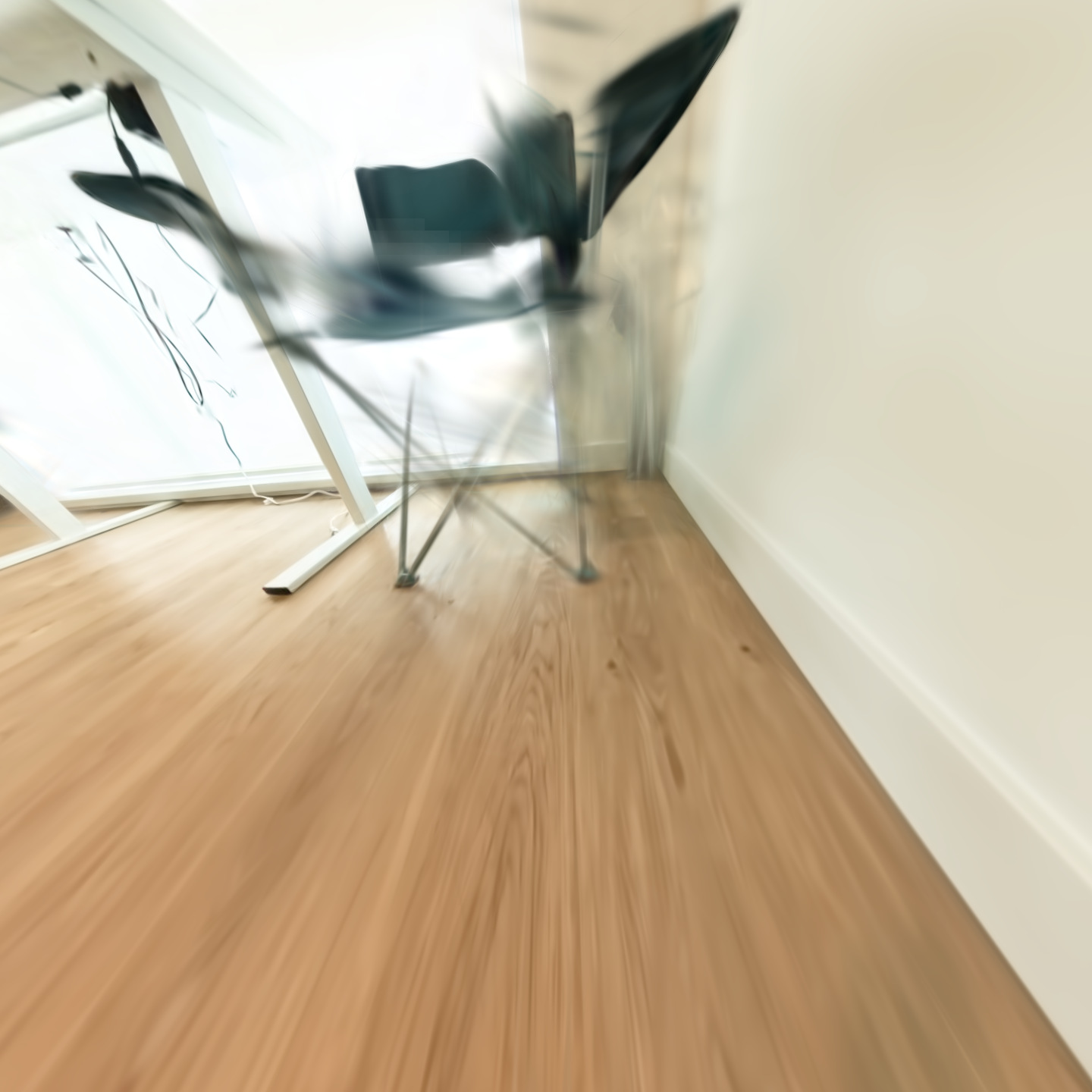} &
        \includegraphics[width=0.18\textwidth]{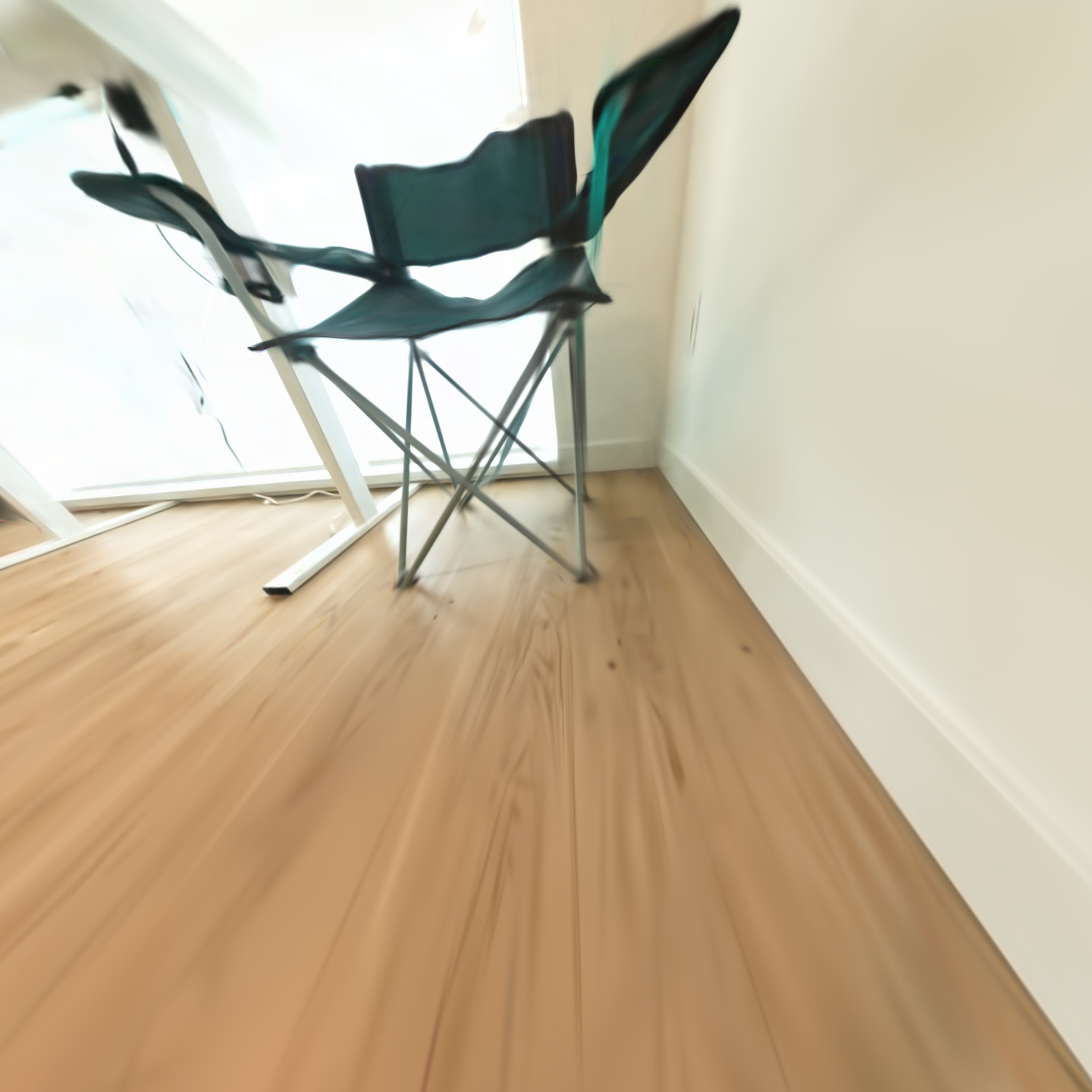} &
        \includegraphics[width=0.18\textwidth]{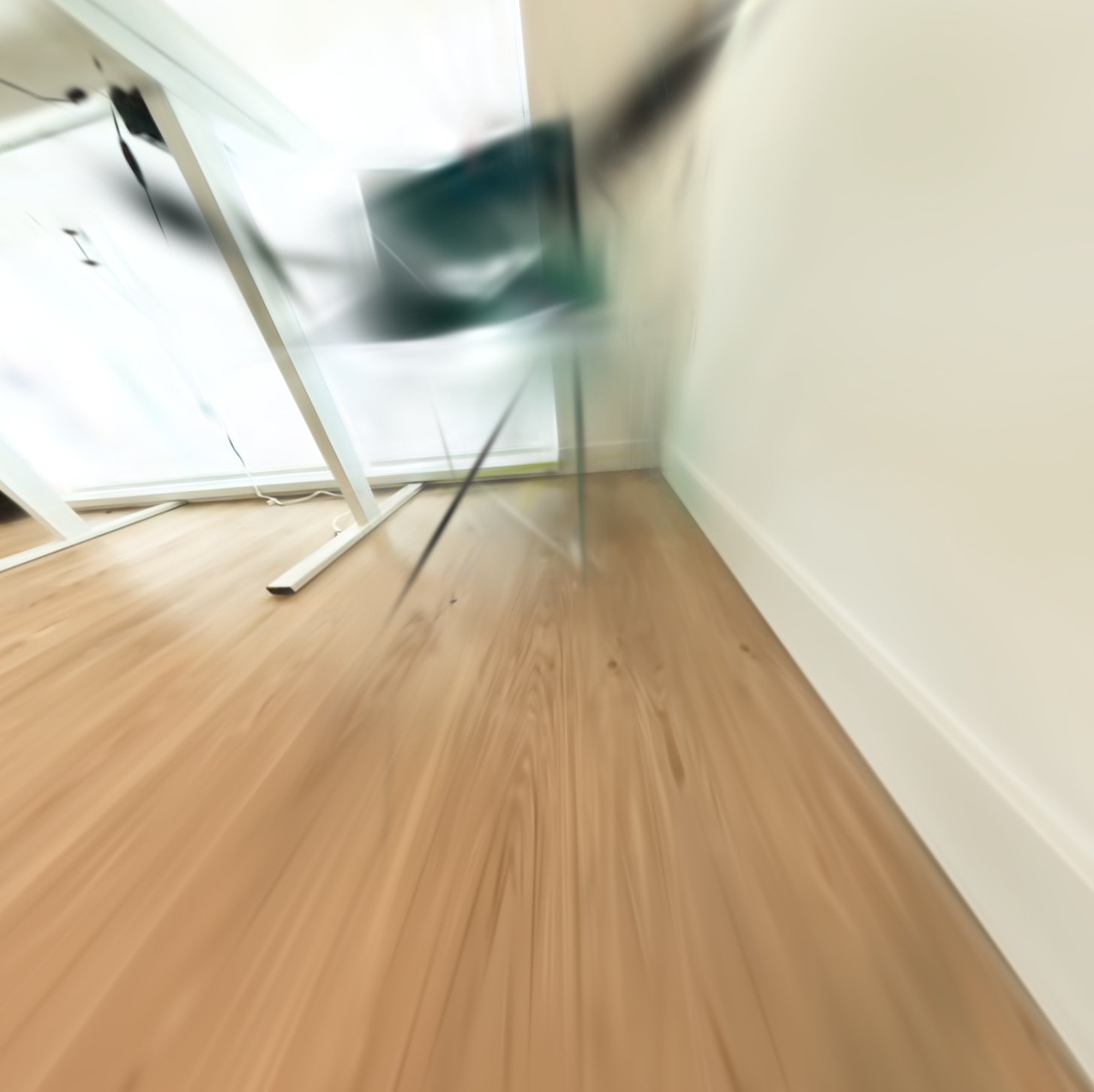} &
        \includegraphics[width=0.18\textwidth]{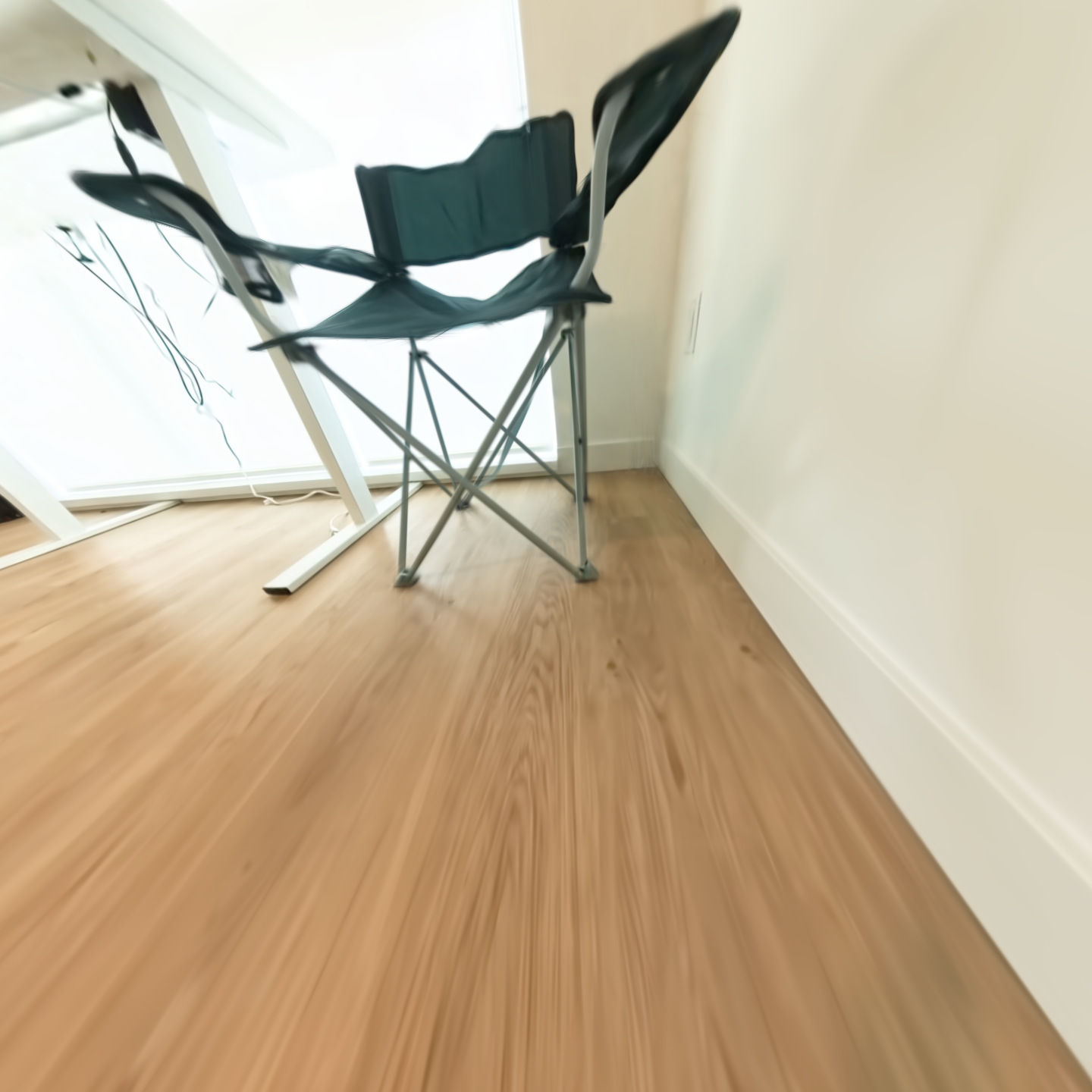} &
        \includegraphics[width=0.18\textwidth]{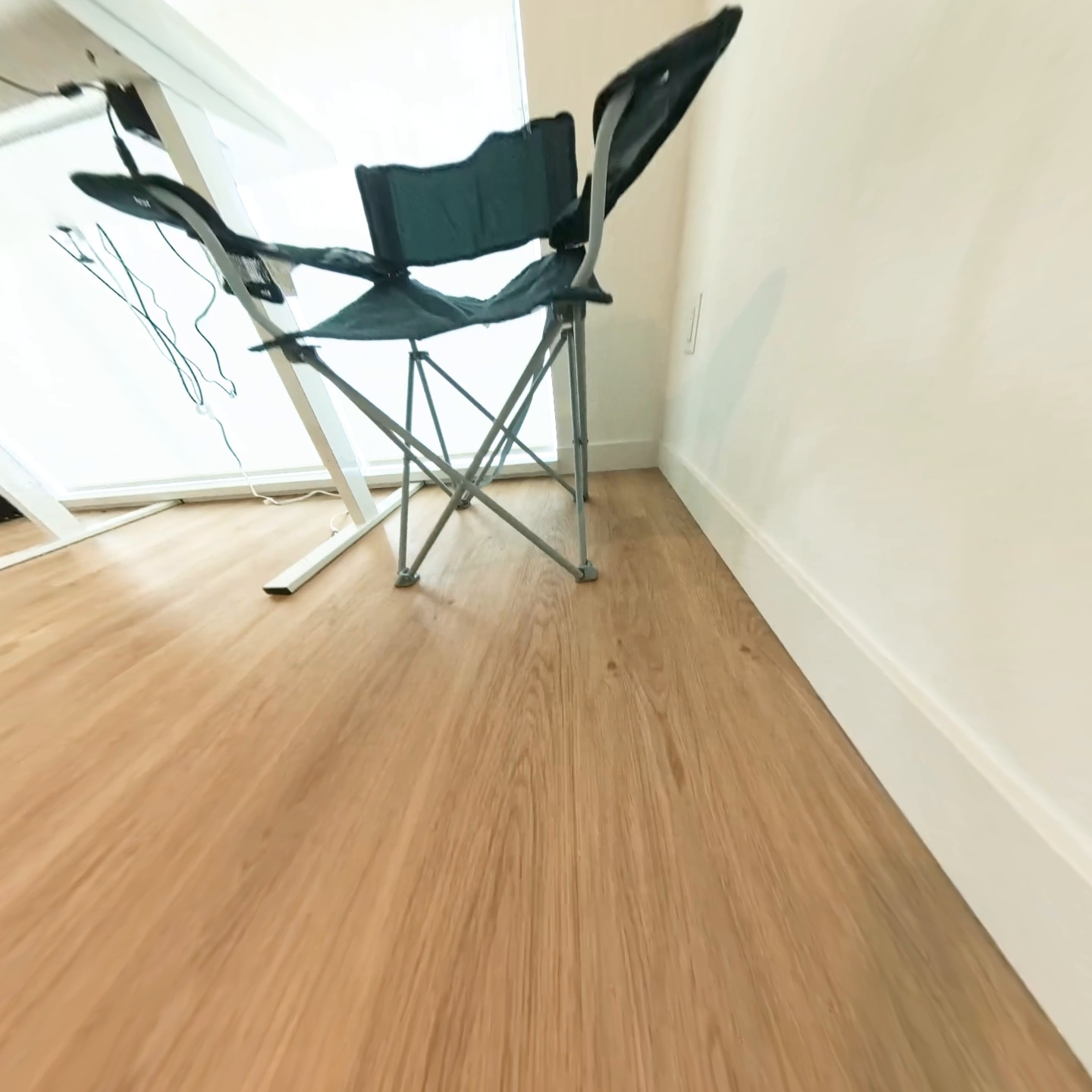} \\[1pt]

        \includegraphics[width=0.18\textwidth]{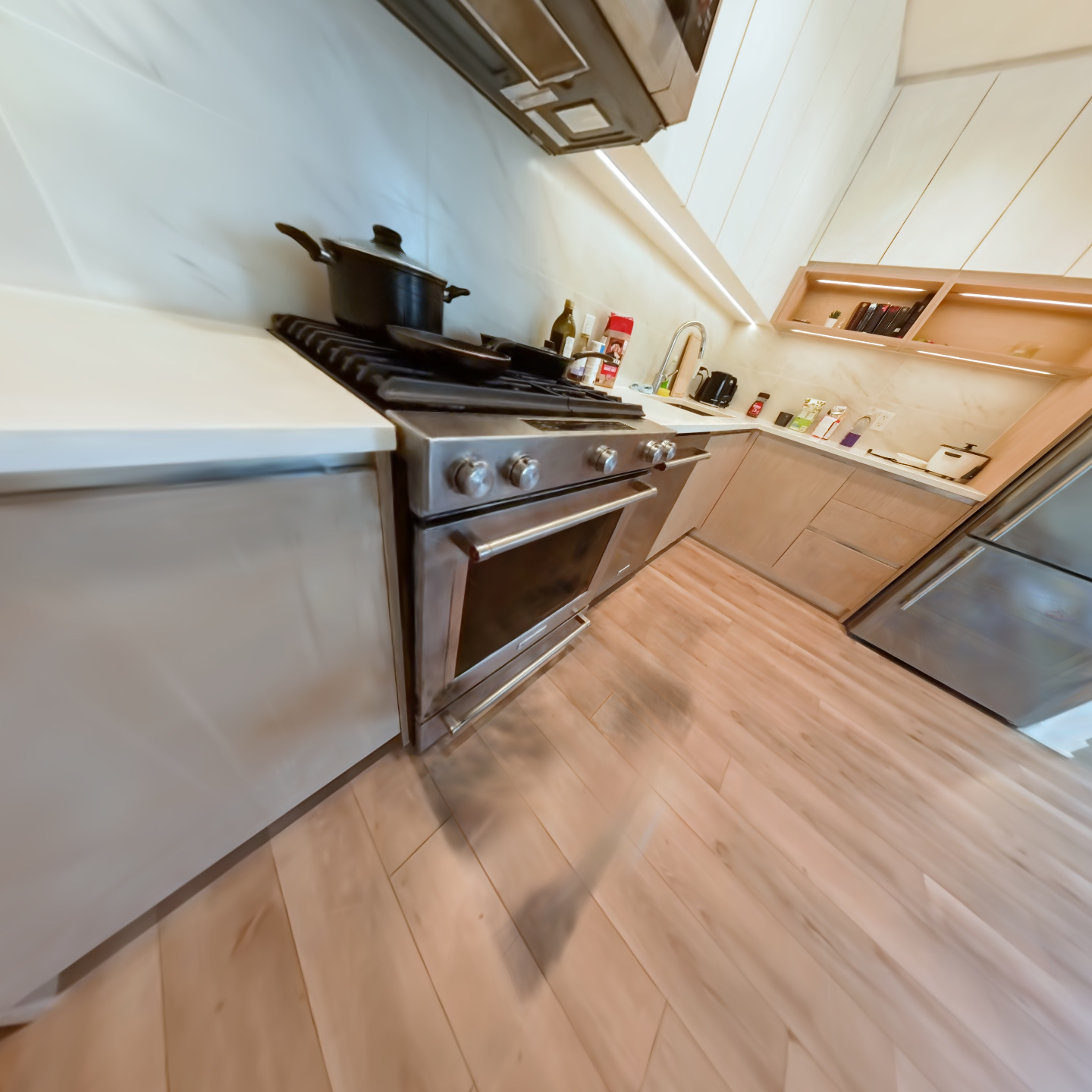} &
        \includegraphics[width=0.18\textwidth]{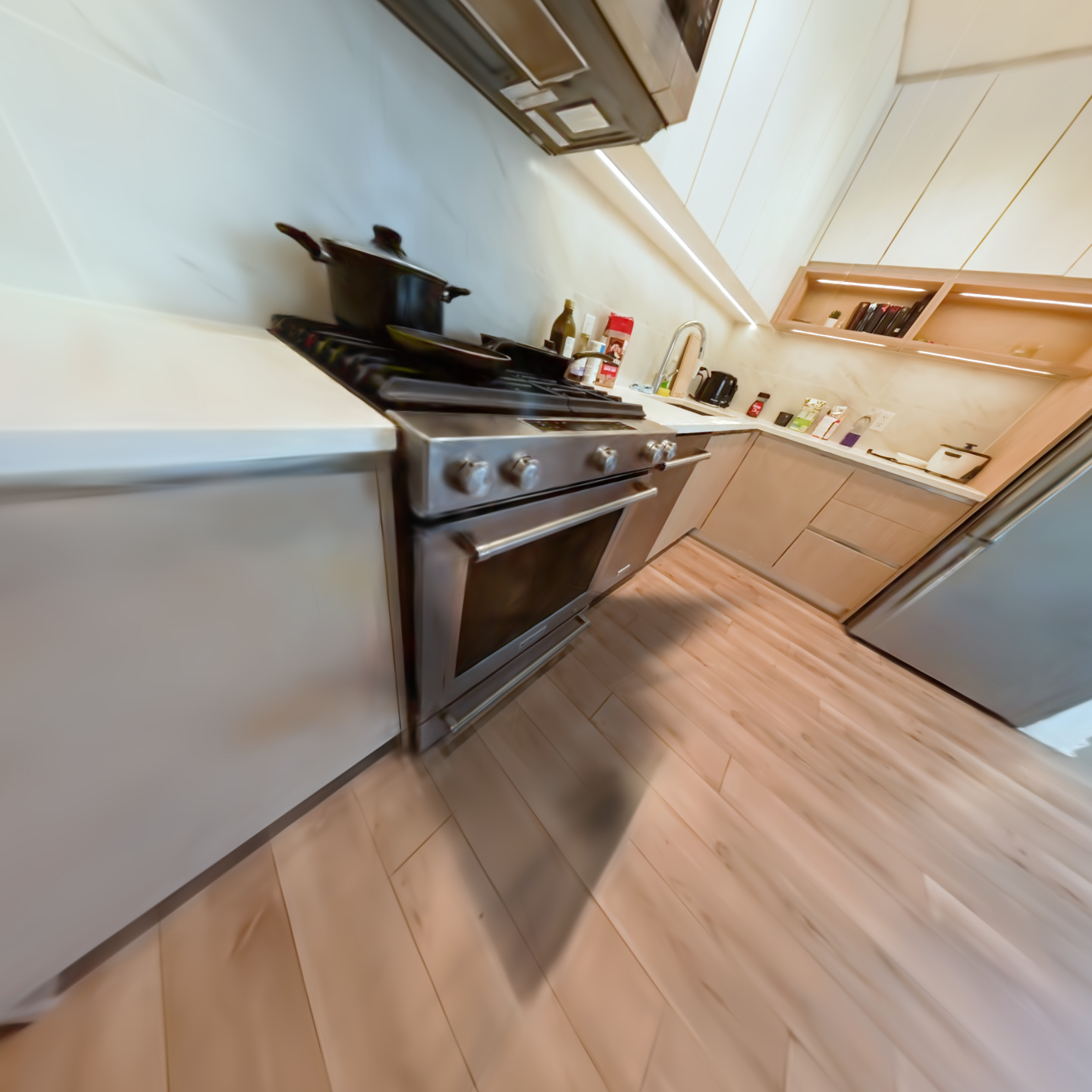} &
        \includegraphics[width=0.18\textwidth]{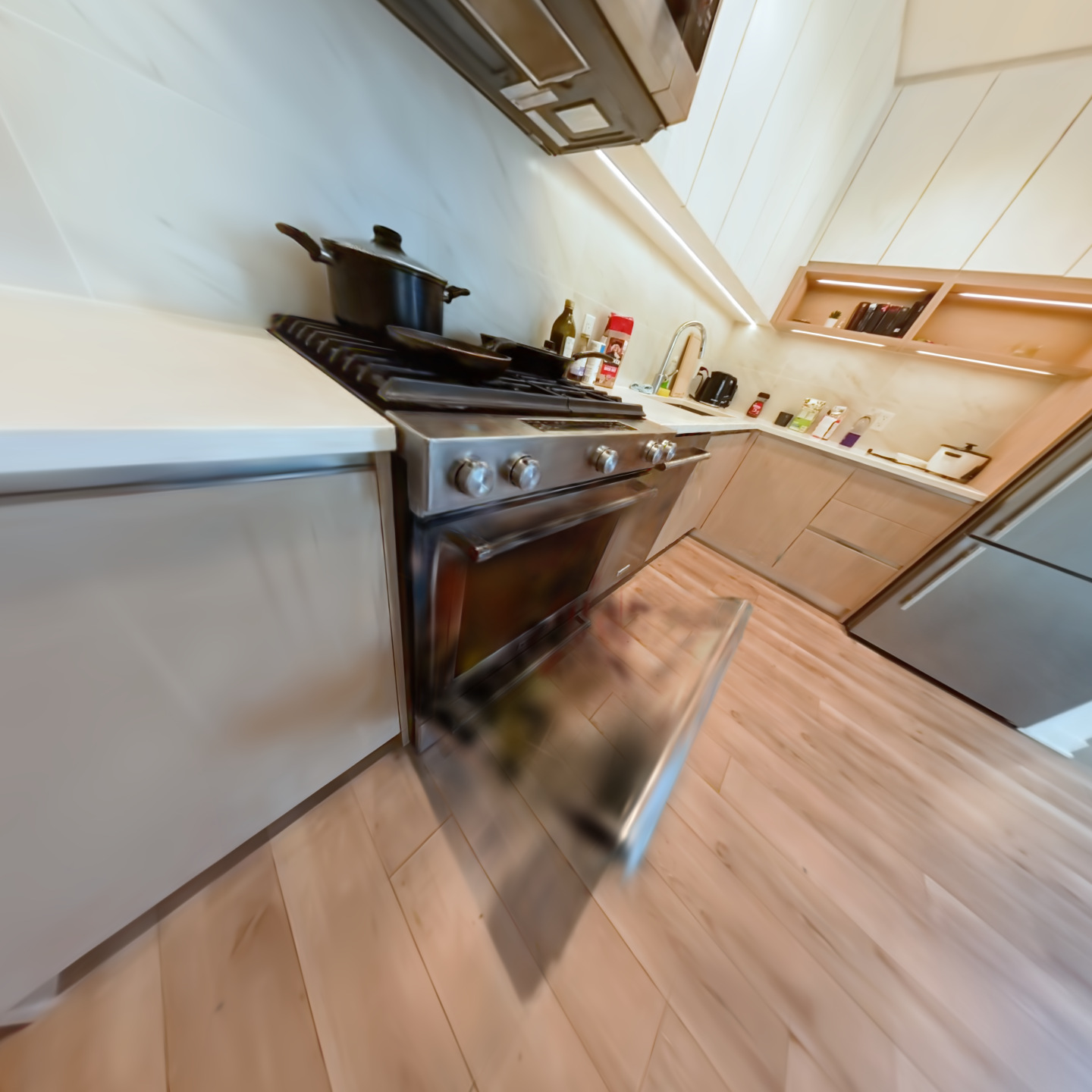} &
        \includegraphics[width=0.18\textwidth]{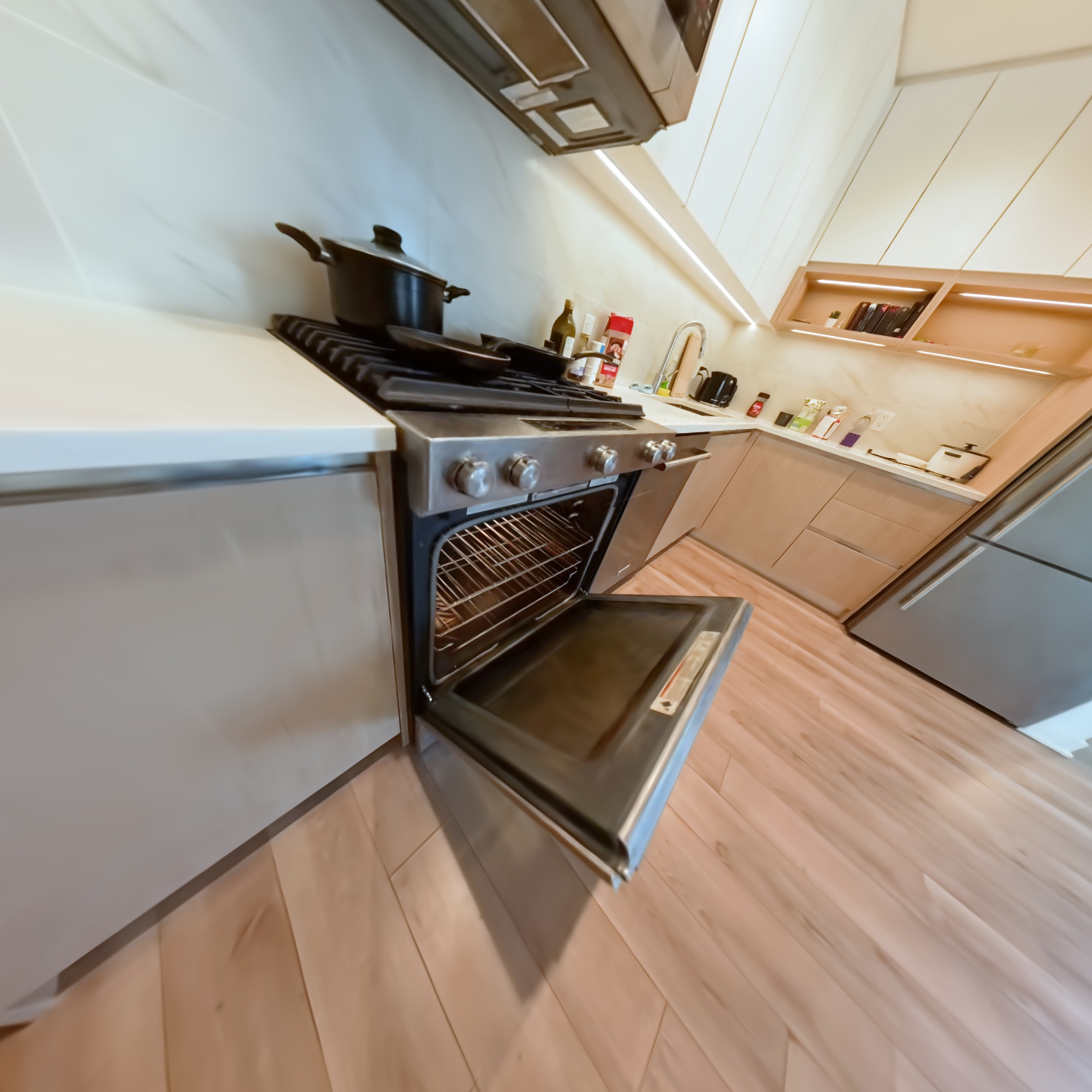} &
        \includegraphics[width=0.18\textwidth]{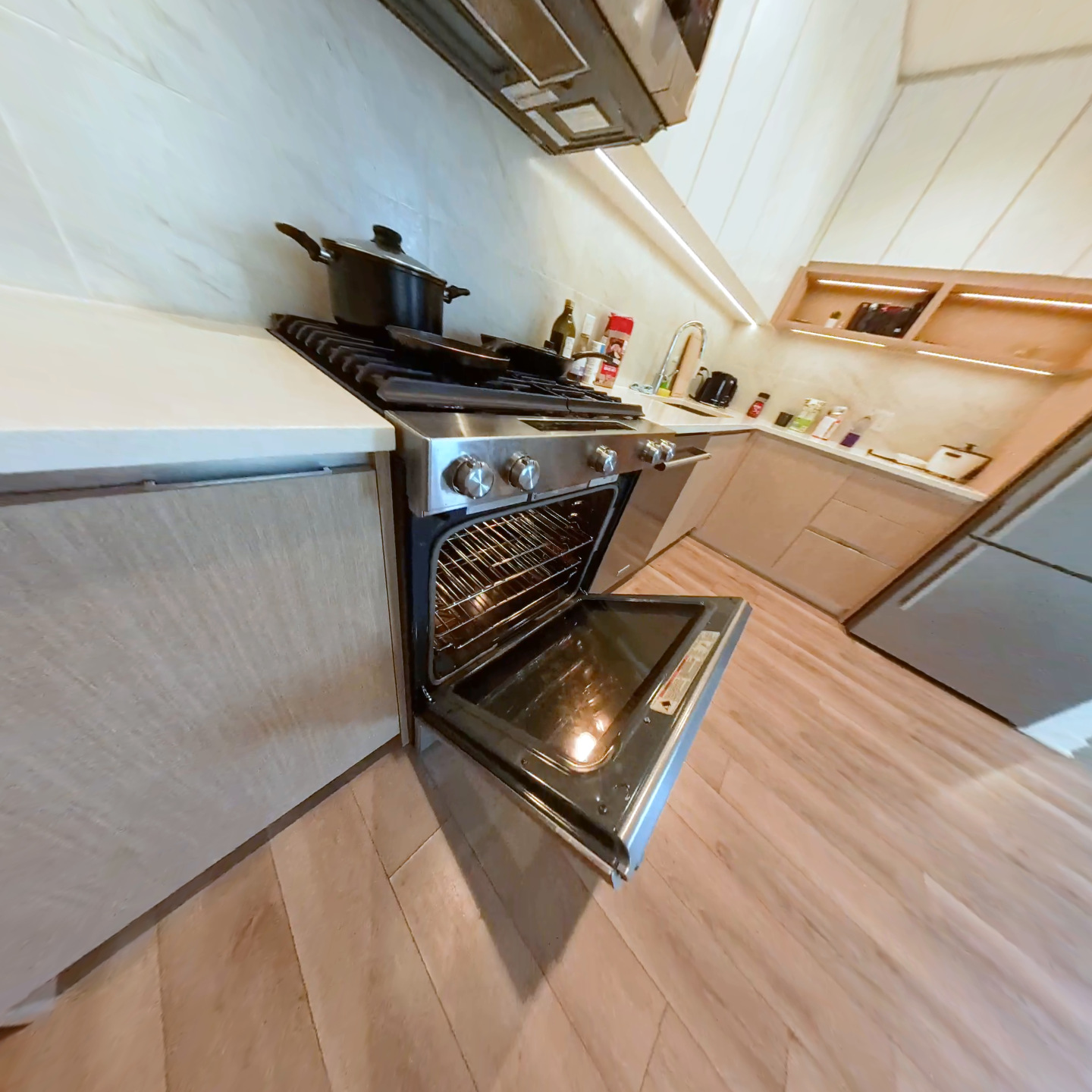} \\[1pt]

        \includegraphics[width=0.18\textwidth]{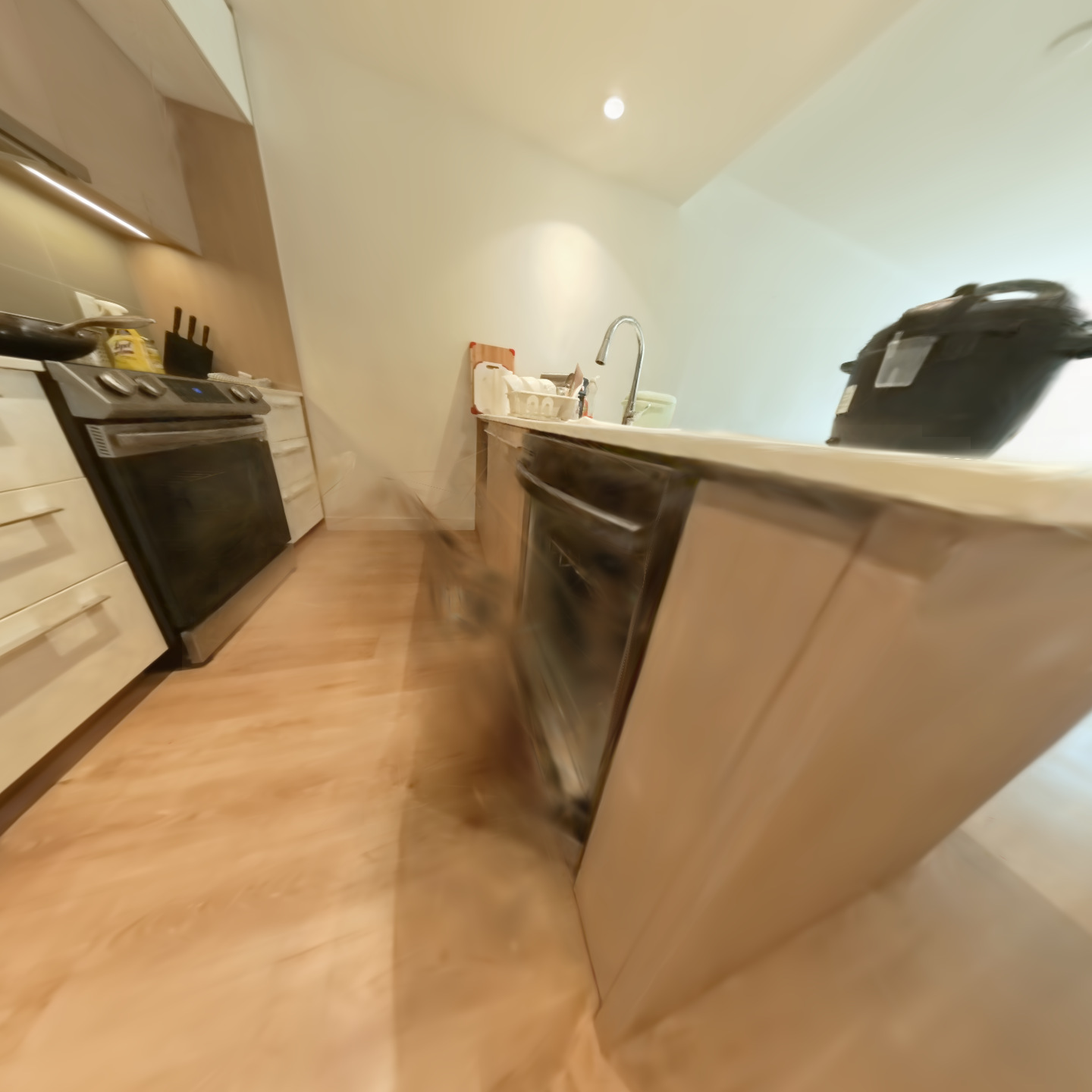} &
        \includegraphics[width=0.18\textwidth]{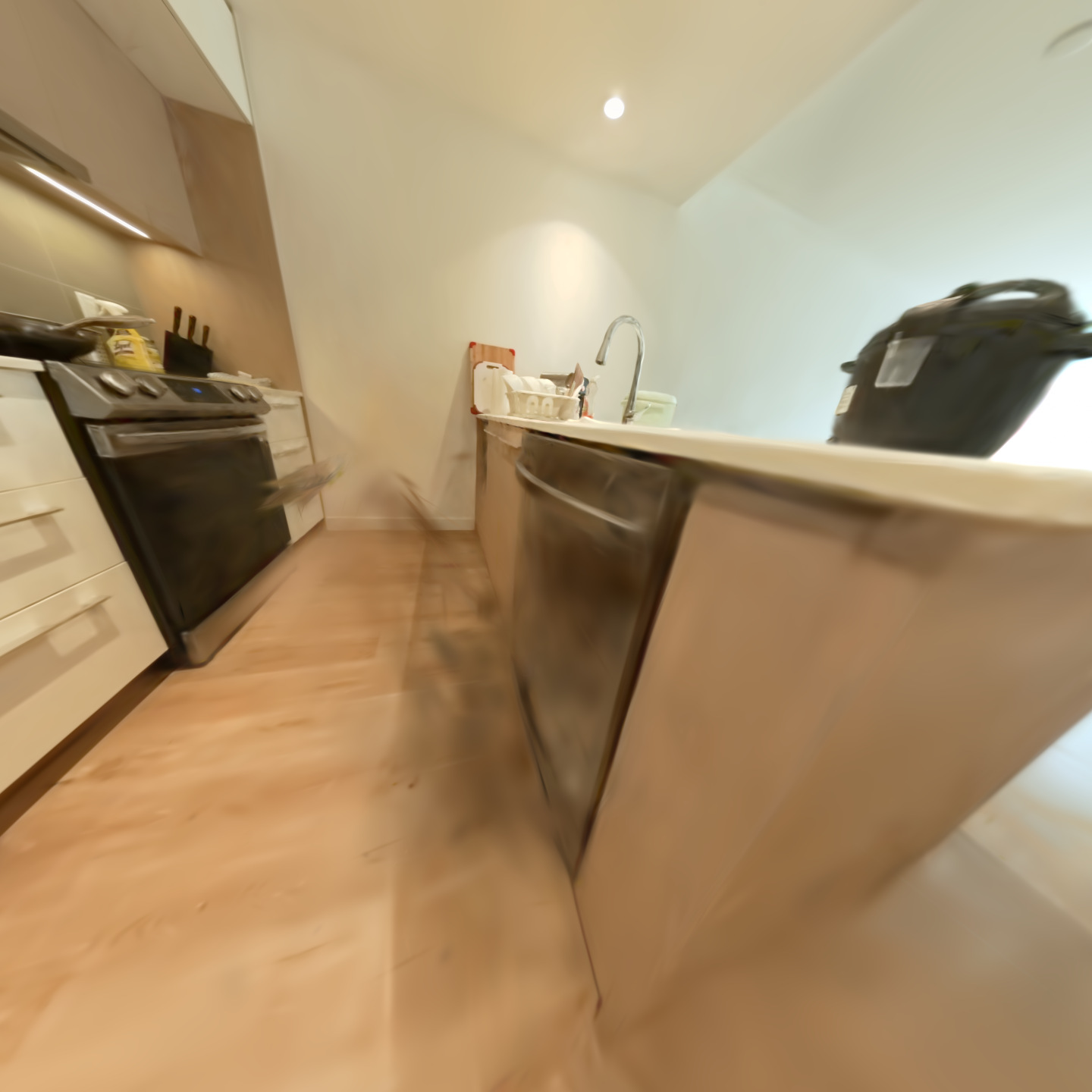} &
        \includegraphics[width=0.18\textwidth]{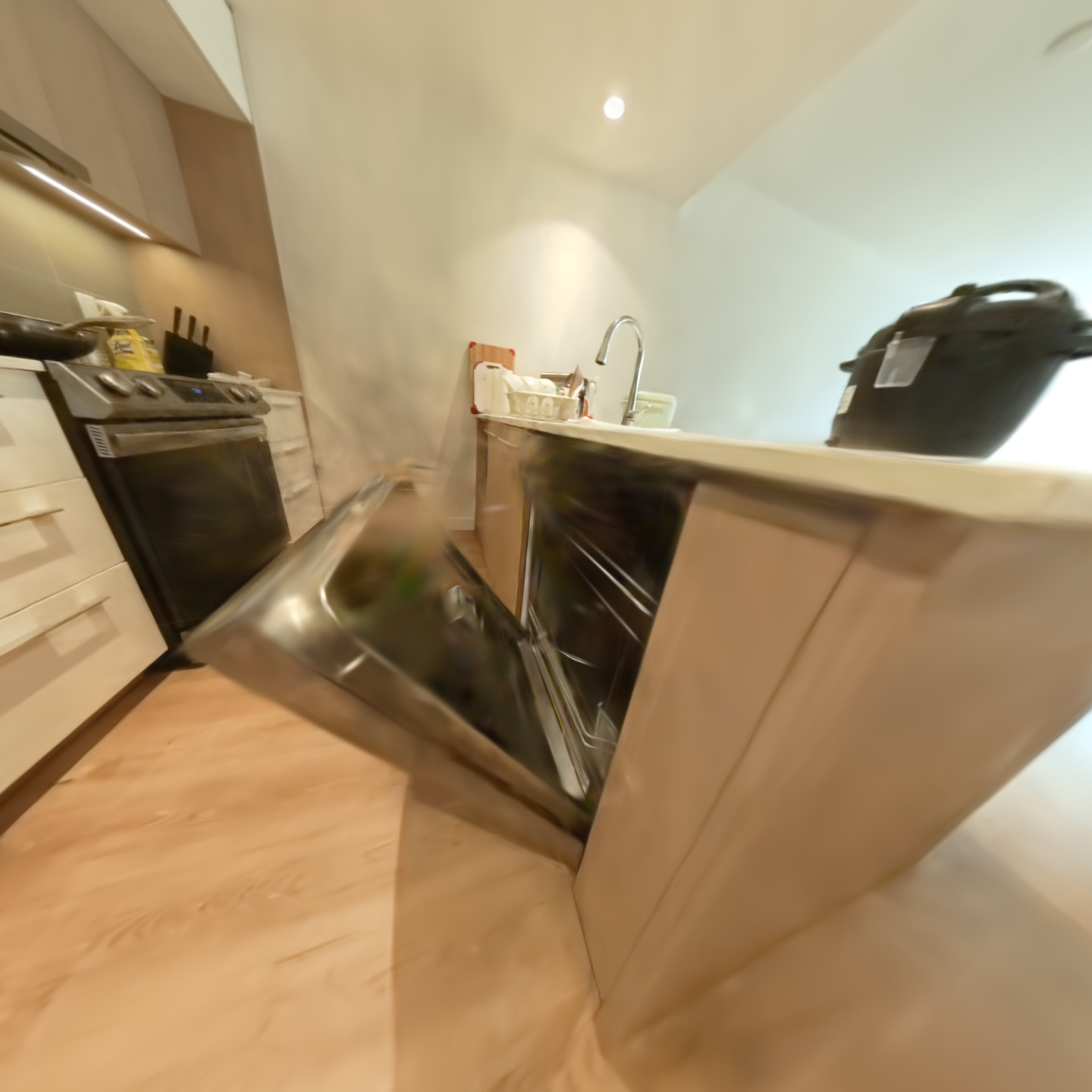} &
        \includegraphics[width=0.18\textwidth]{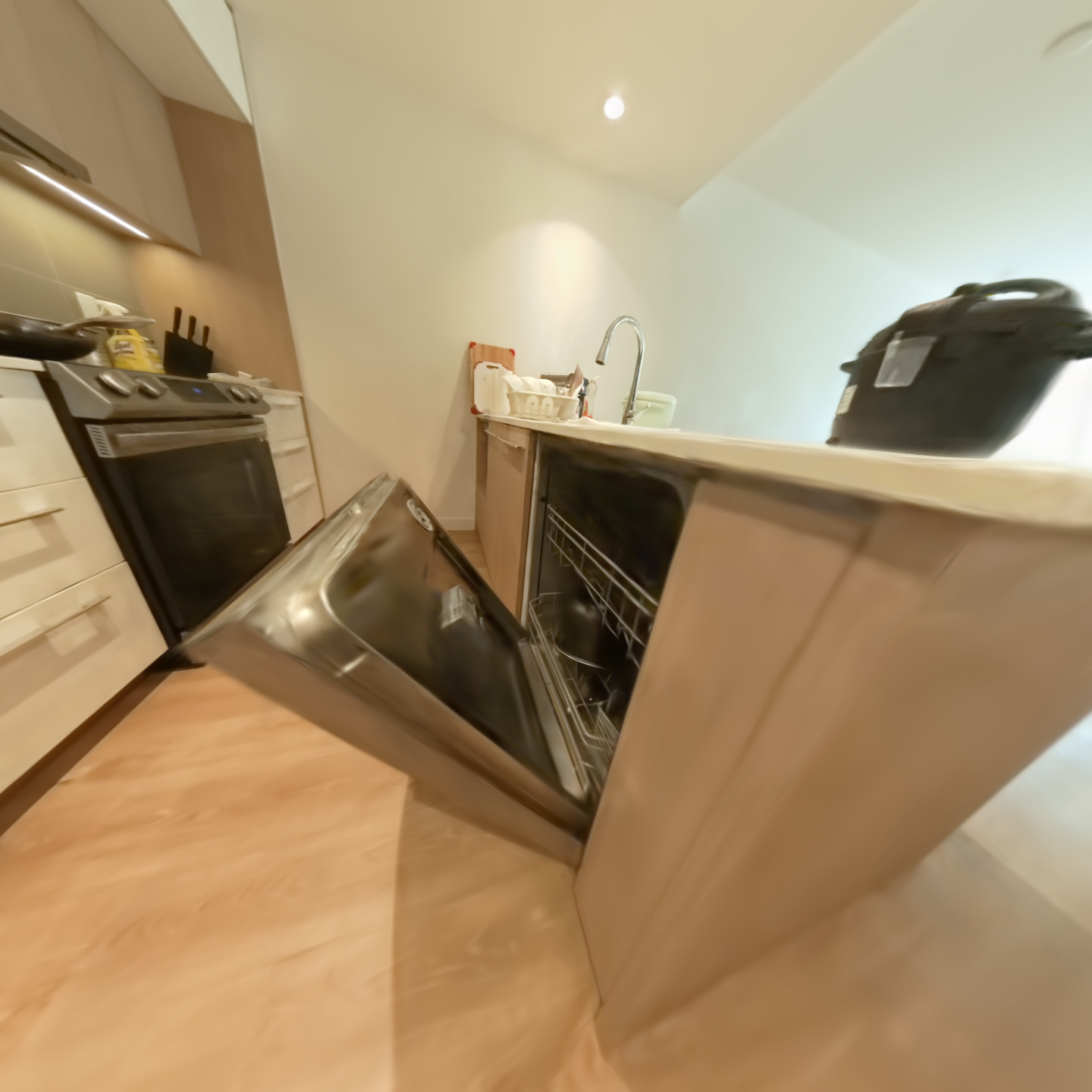} &
        \includegraphics[width=0.18\textwidth]{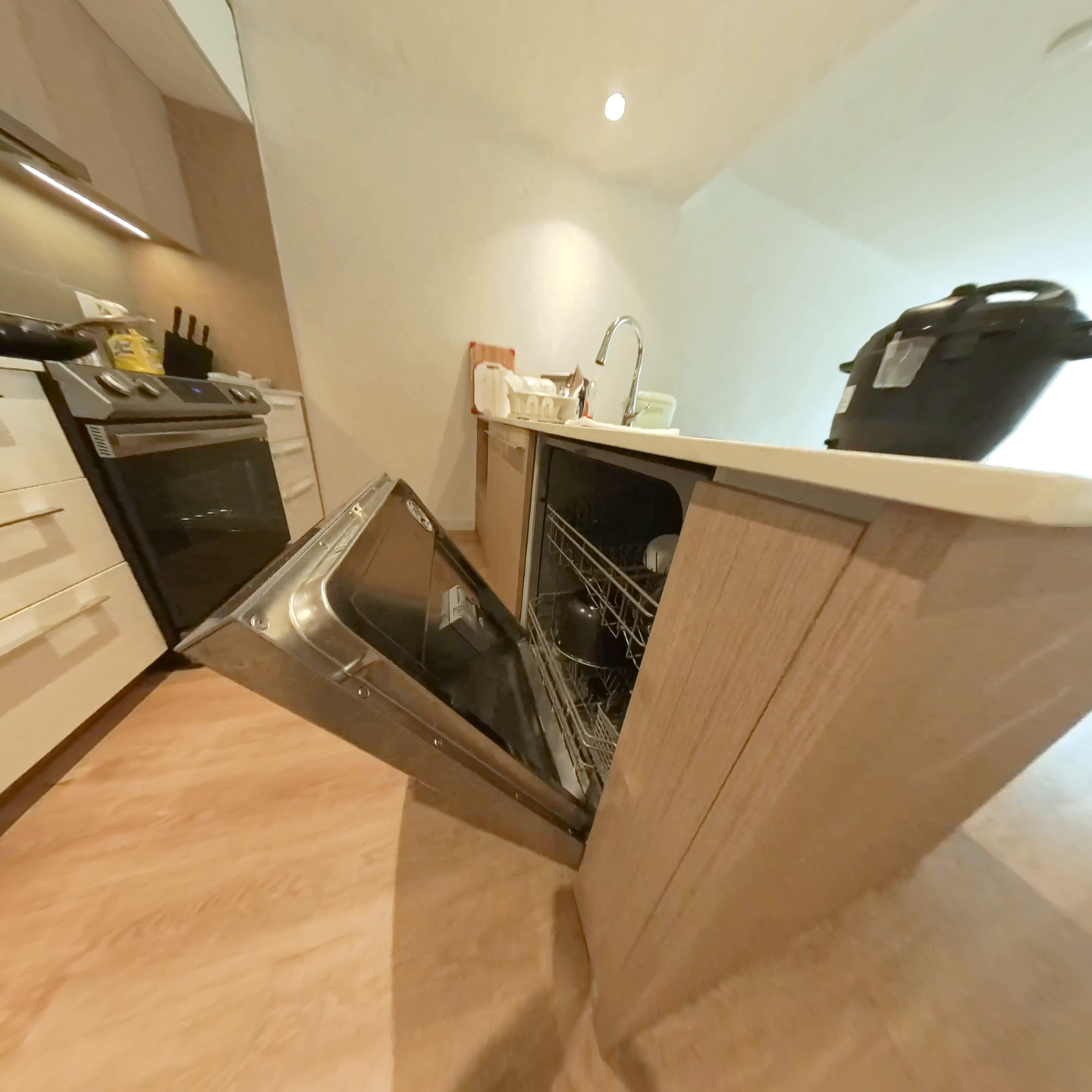} \\[1pt]

        \includegraphics[width=0.18\textwidth]{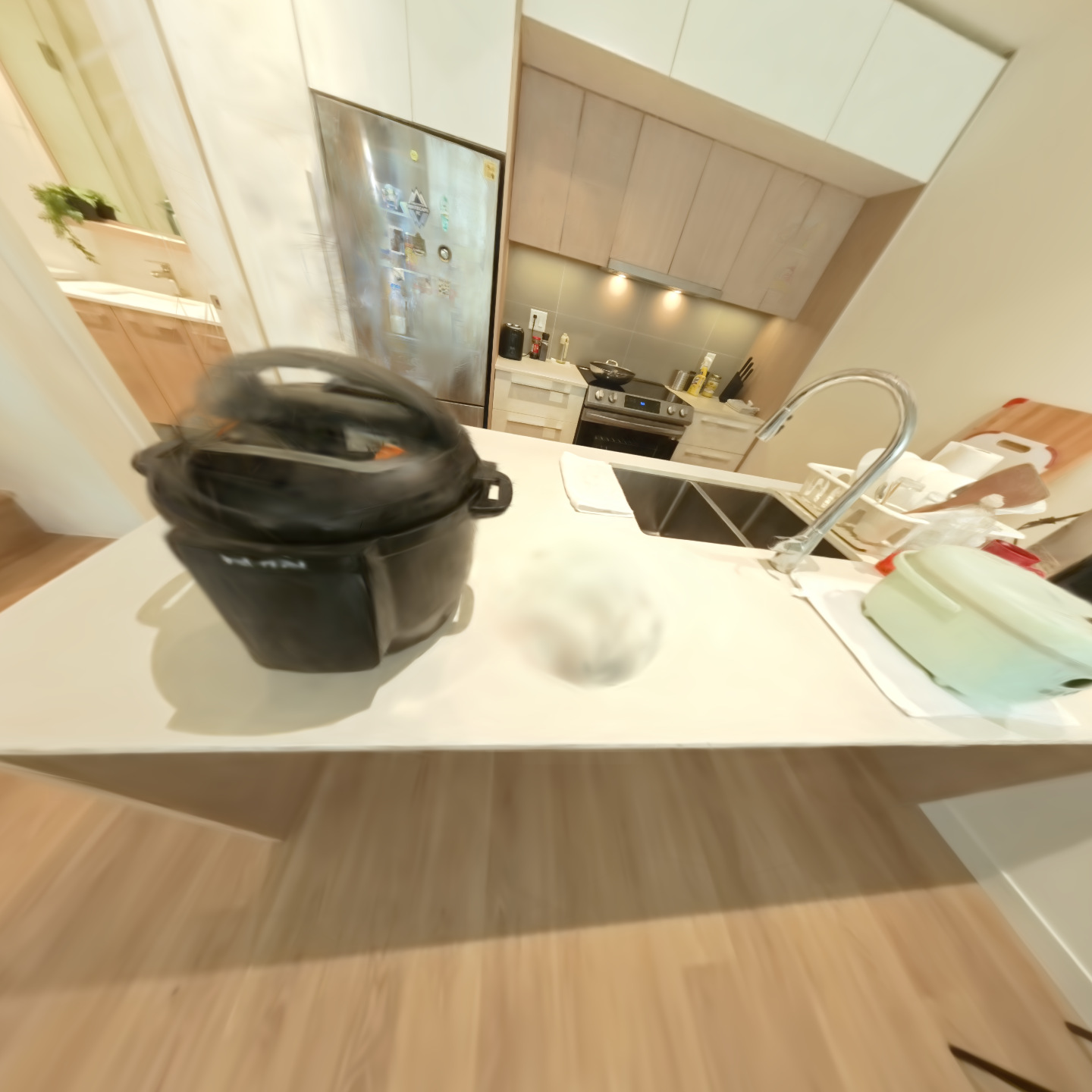} &
        \includegraphics[width=0.18\textwidth]{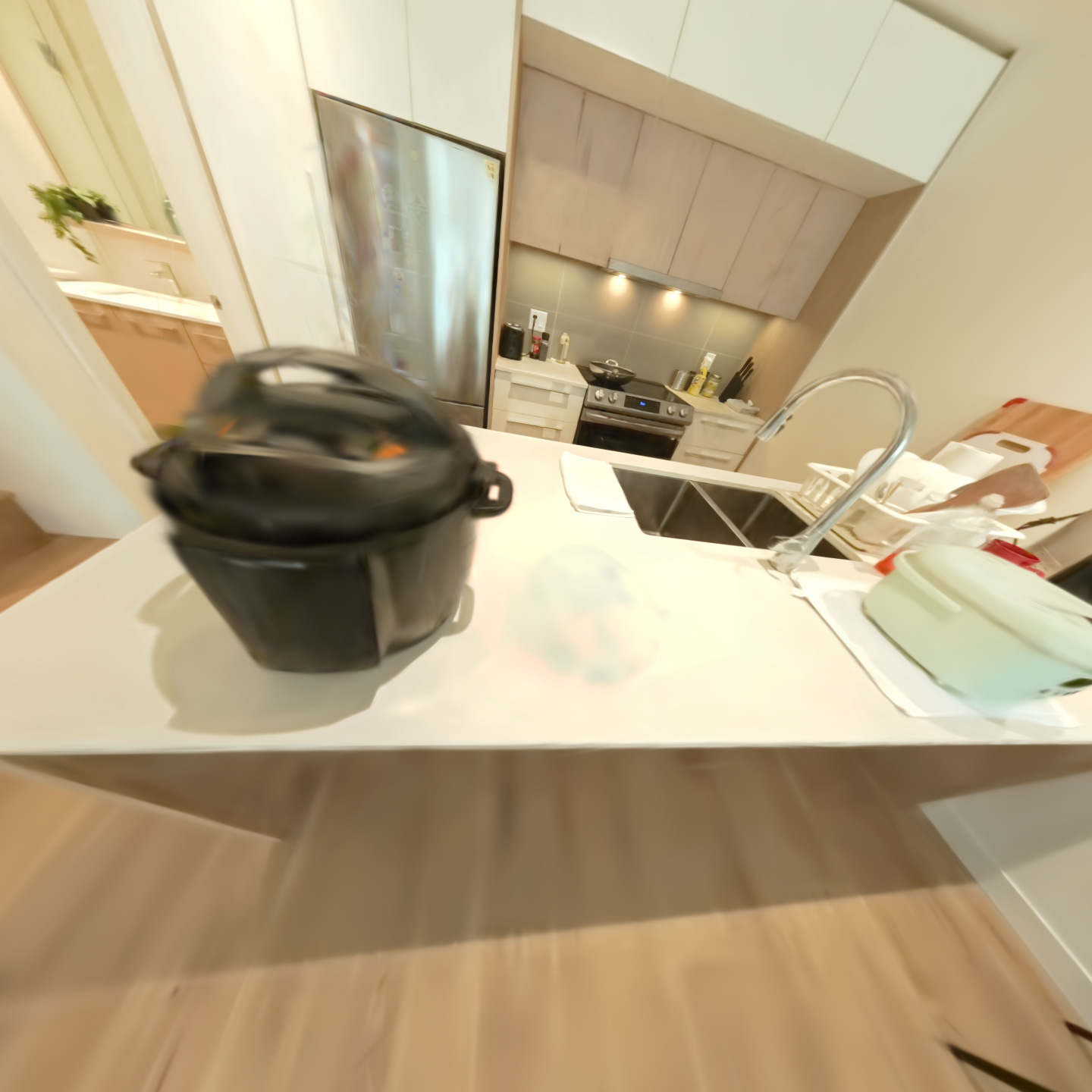} &
        \includegraphics[width=0.18\textwidth]{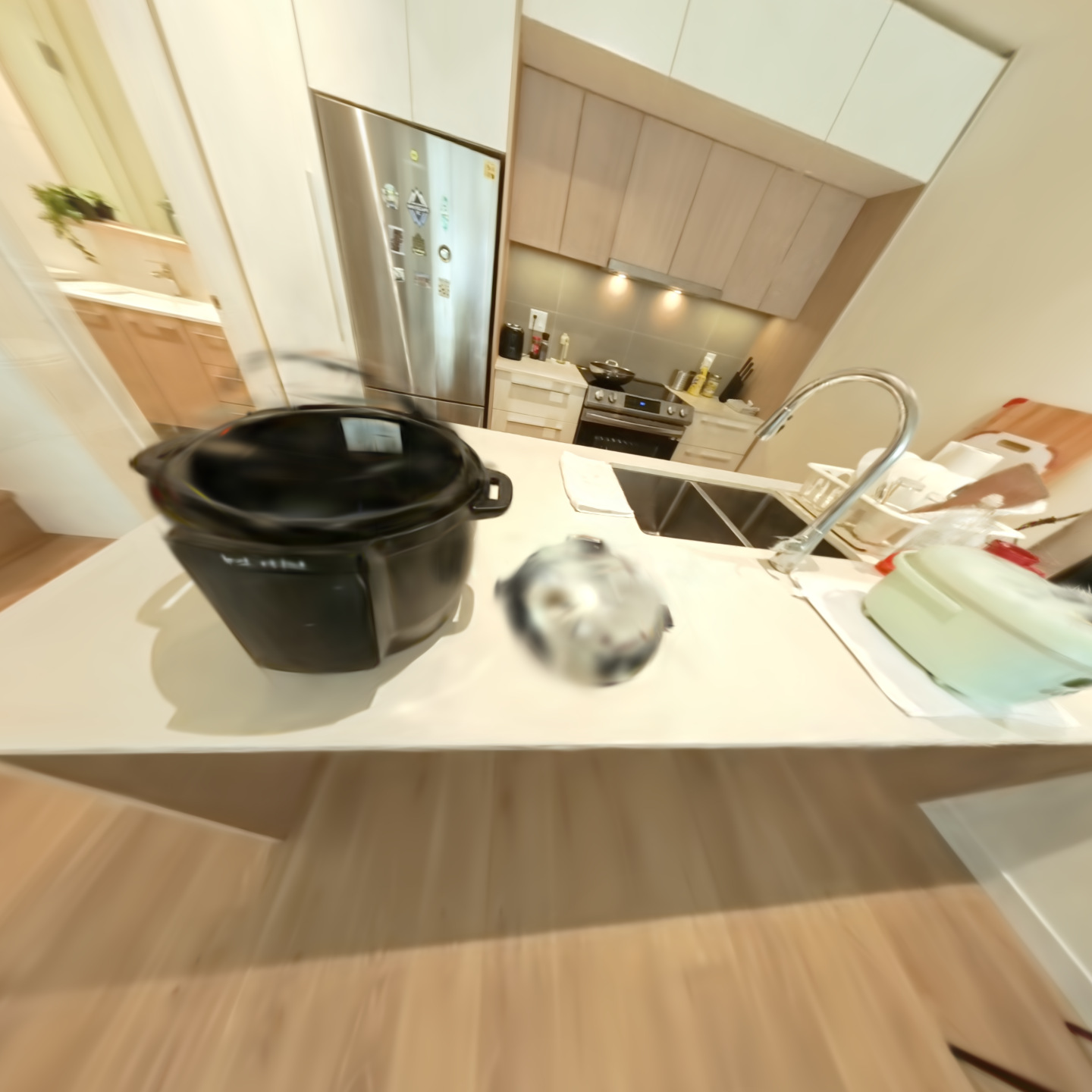} &
        \includegraphics[width=0.18\textwidth]{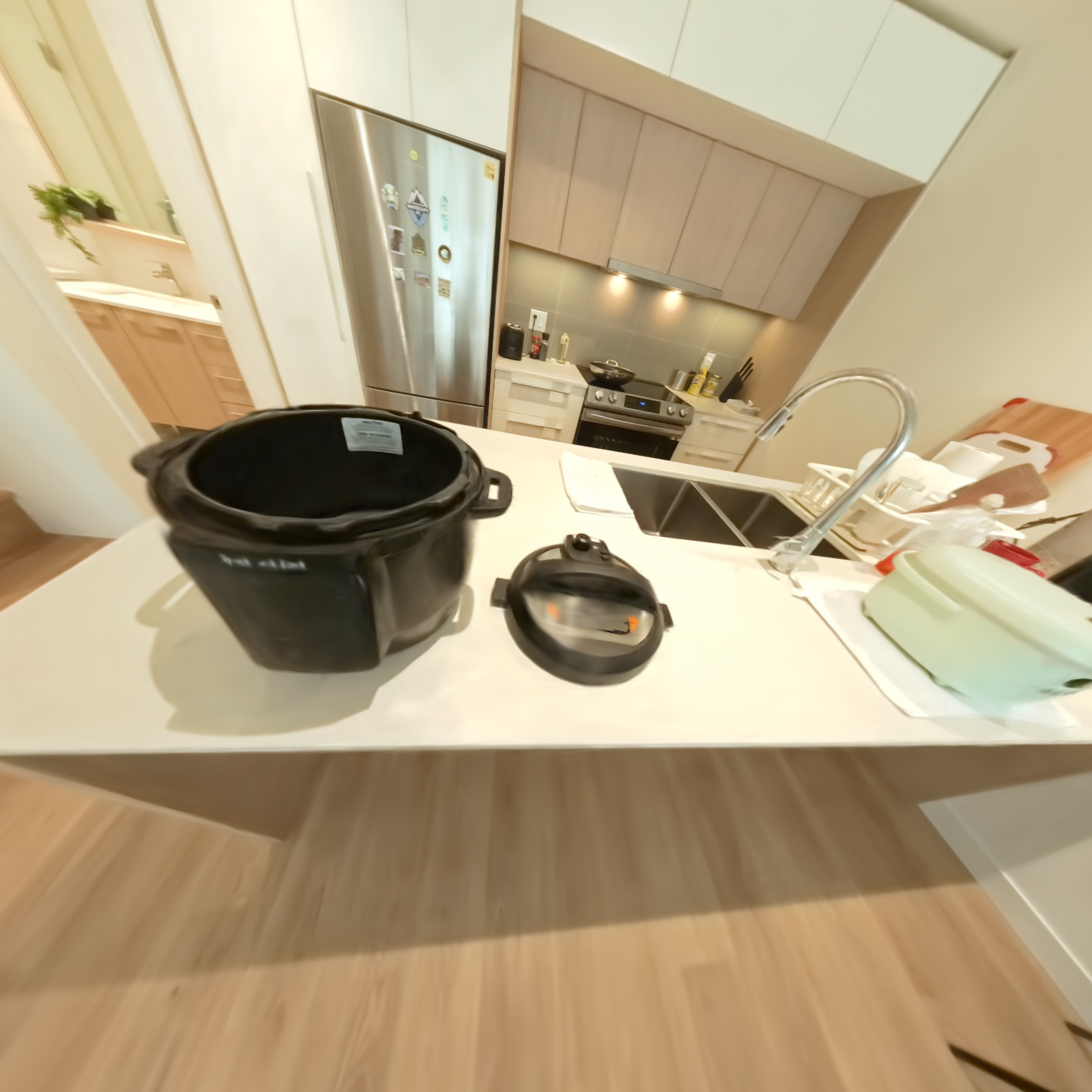} &
        \includegraphics[width=0.18\textwidth]{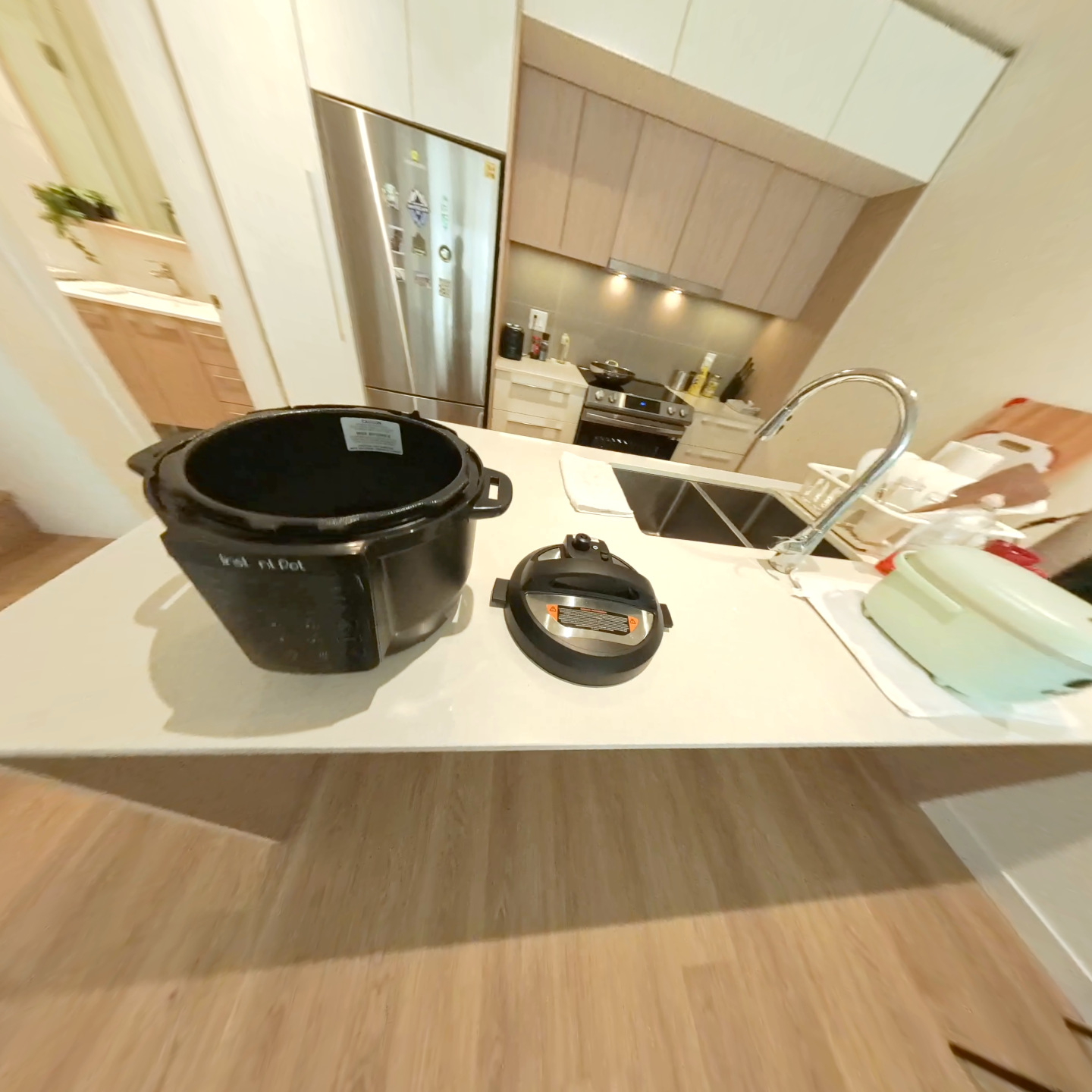} \\[1pt]

        \includegraphics[width=0.18\textwidth]{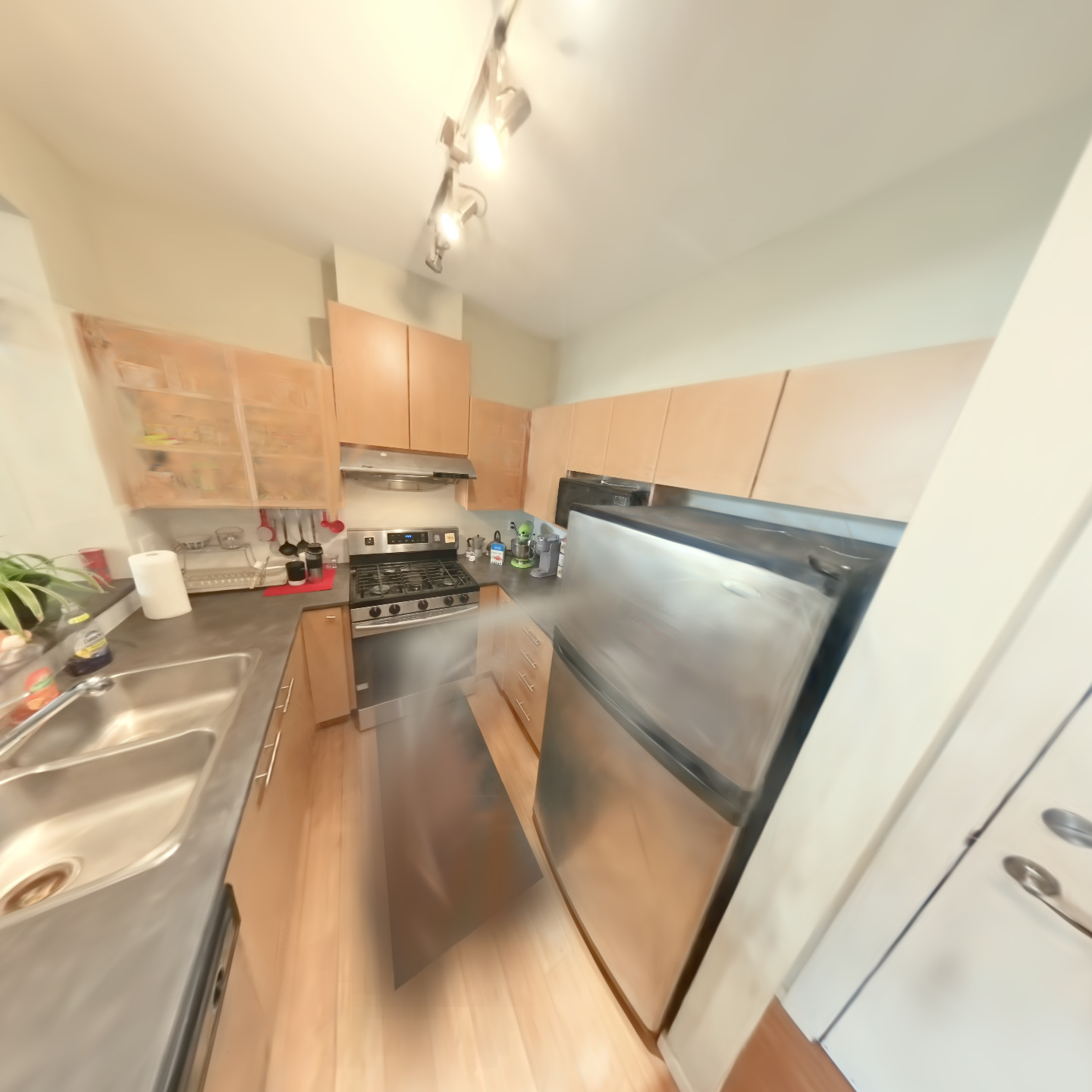} &
        \includegraphics[width=0.18\textwidth]{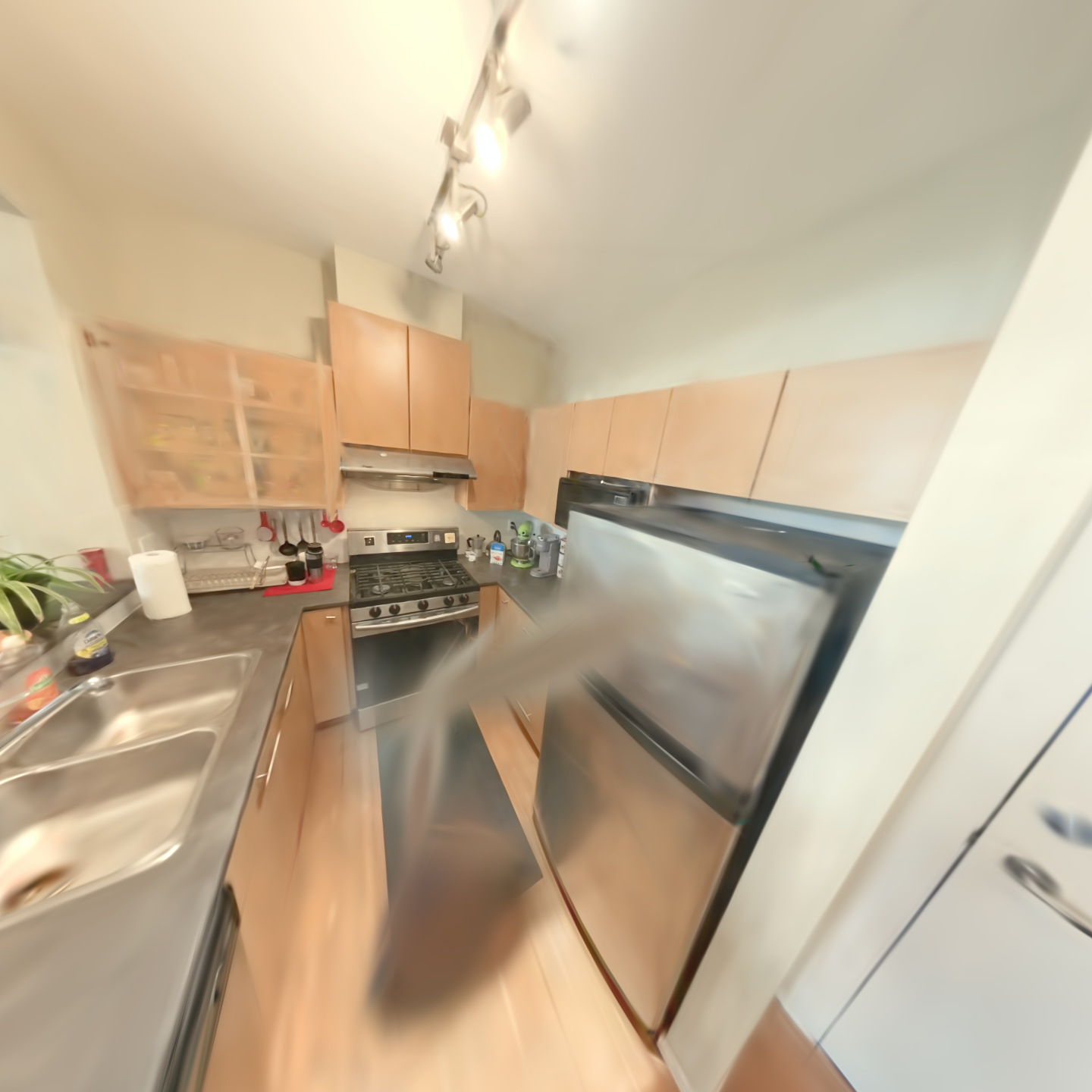} &
        \includegraphics[width=0.18\textwidth]{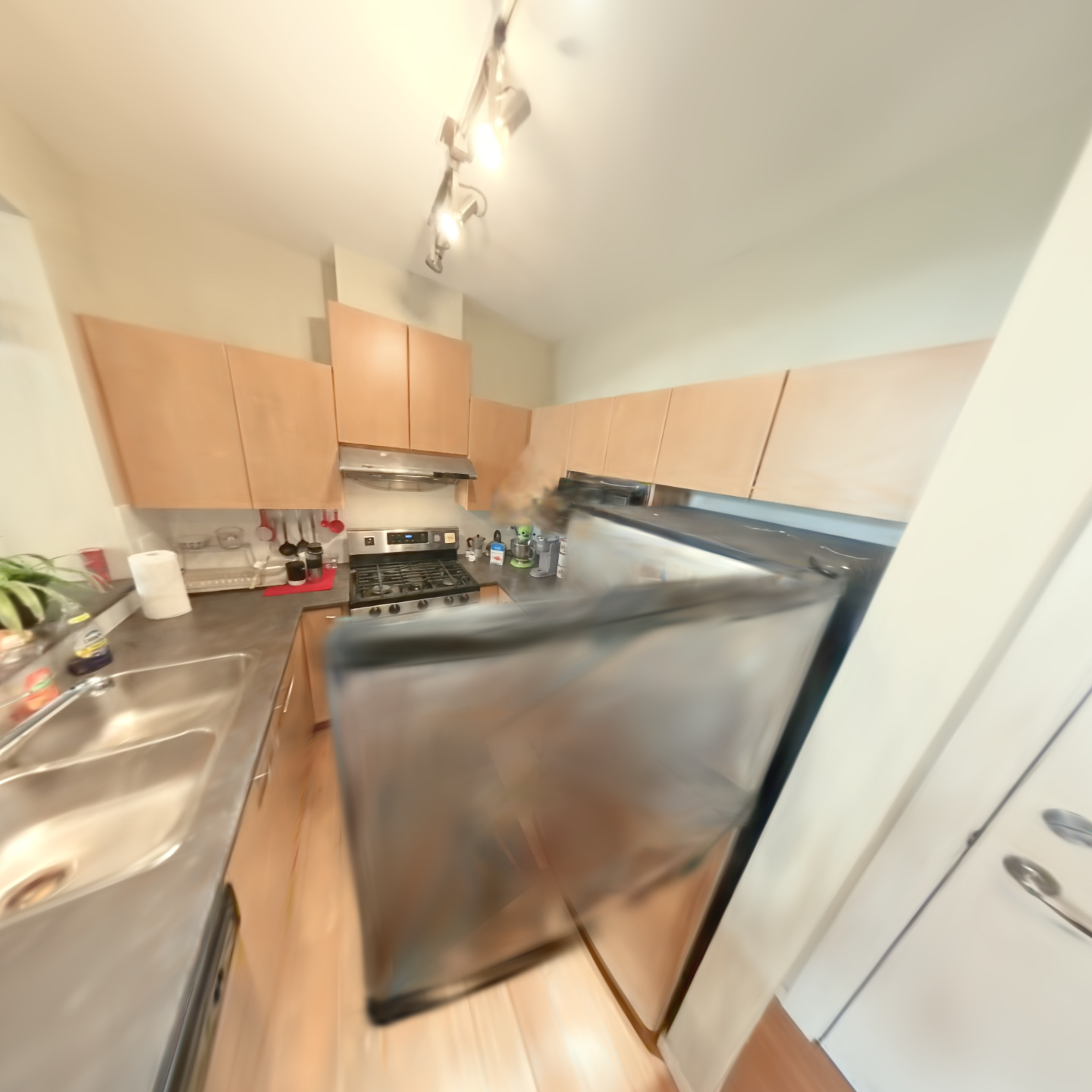} &
        \includegraphics[width=0.18\textwidth]{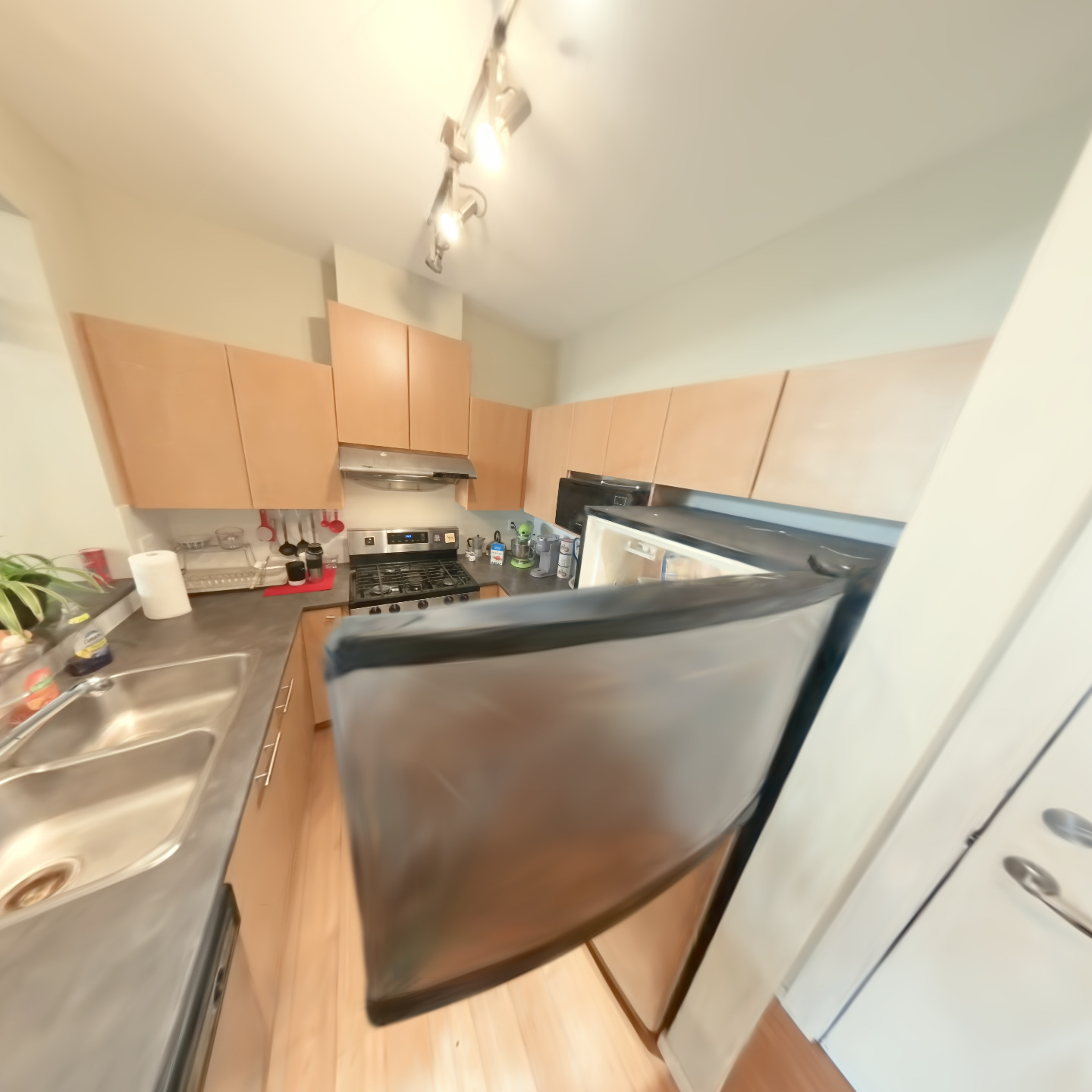} &
        \includegraphics[width=0.18\textwidth]{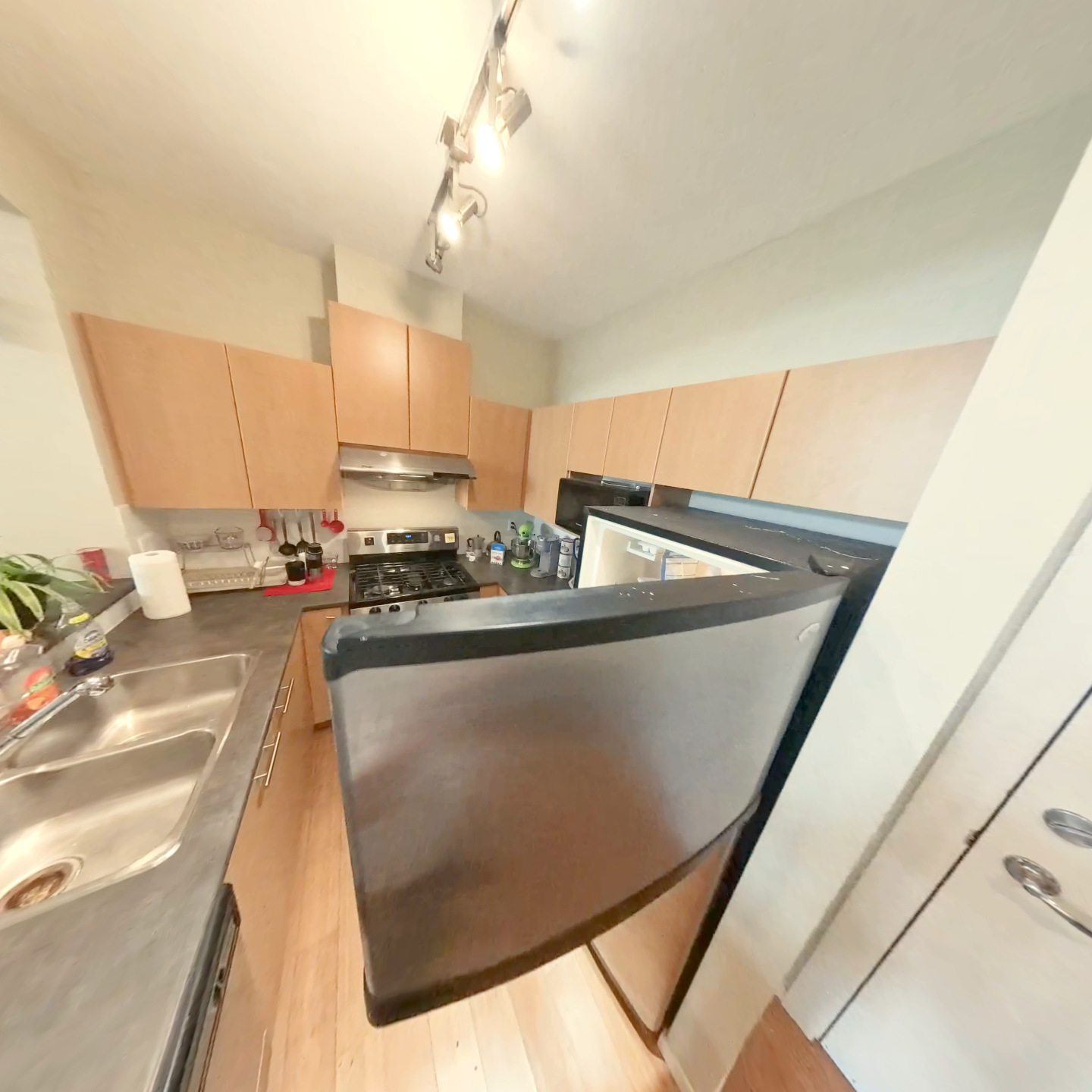}
    \end{tabular}

    \caption{ \textbf{Qualitative comparison.}
We compare FullCircle, DeGauss, 4DGS, and SwitchSplat against ground-truth images from standard held-out test views.
    Scenes are shown from top to bottom: Car, Chair, Flat2, Kitchen 1, Kitchen 1, Flat 3.
    }
    \label{fig:comparison3}
\end{figure*}

\begin{figure*}[p]
    \centering
    \setlength{\tabcolsep}{1pt}\renewcommand{\arraystretch}{1.0}

    \begin{tabular}{@{}ccccc@{}}
        \footnotesize FullCircle &
        \footnotesize DeGauss &
        \footnotesize 4DGS &
        \footnotesize SwitchSplat &
        \footnotesize Ground Truth \\[2pt]

        \includegraphics[width=0.18\textwidth]{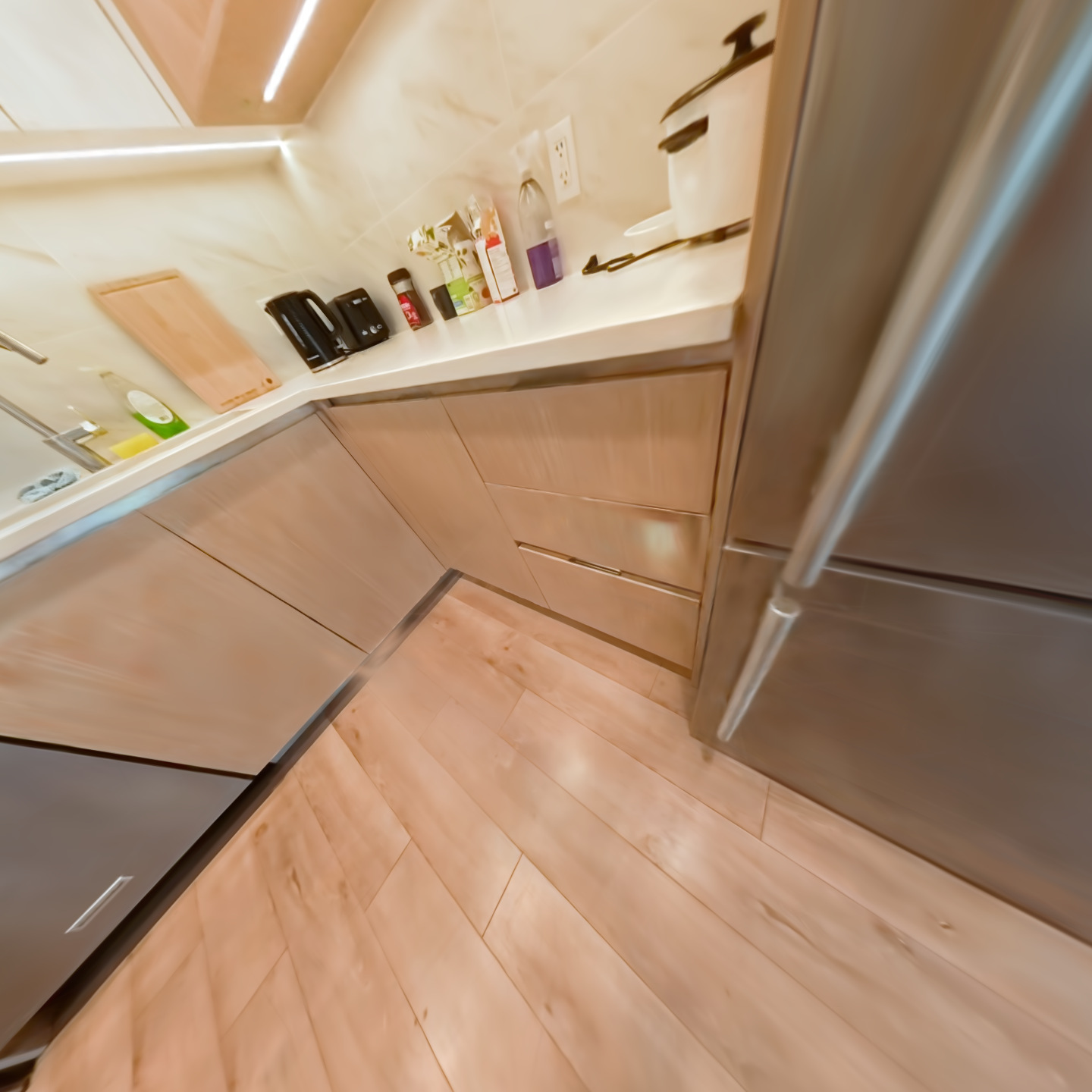} &
        \includegraphics[width=0.18\textwidth]{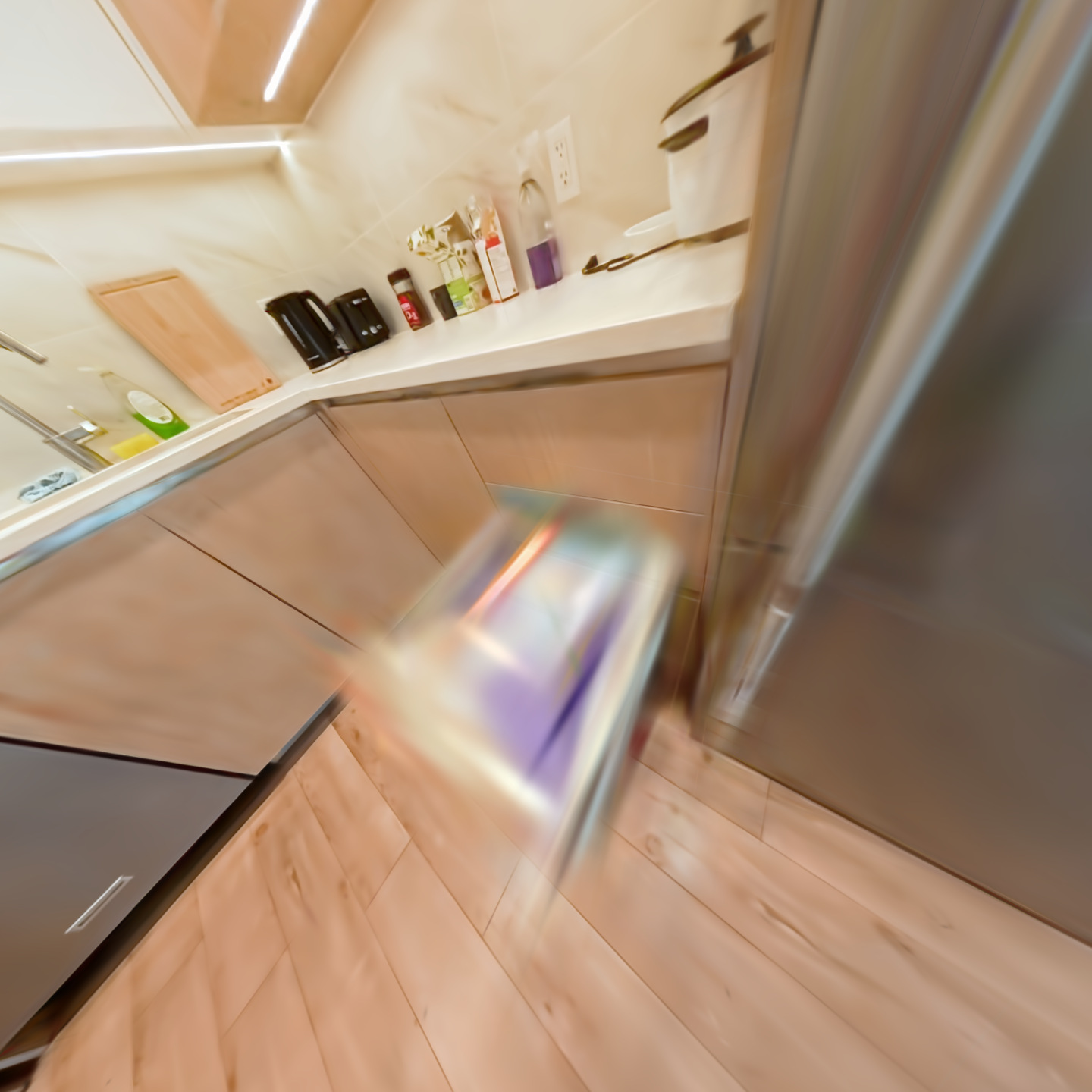} &
        \includegraphics[width=0.18\textwidth]{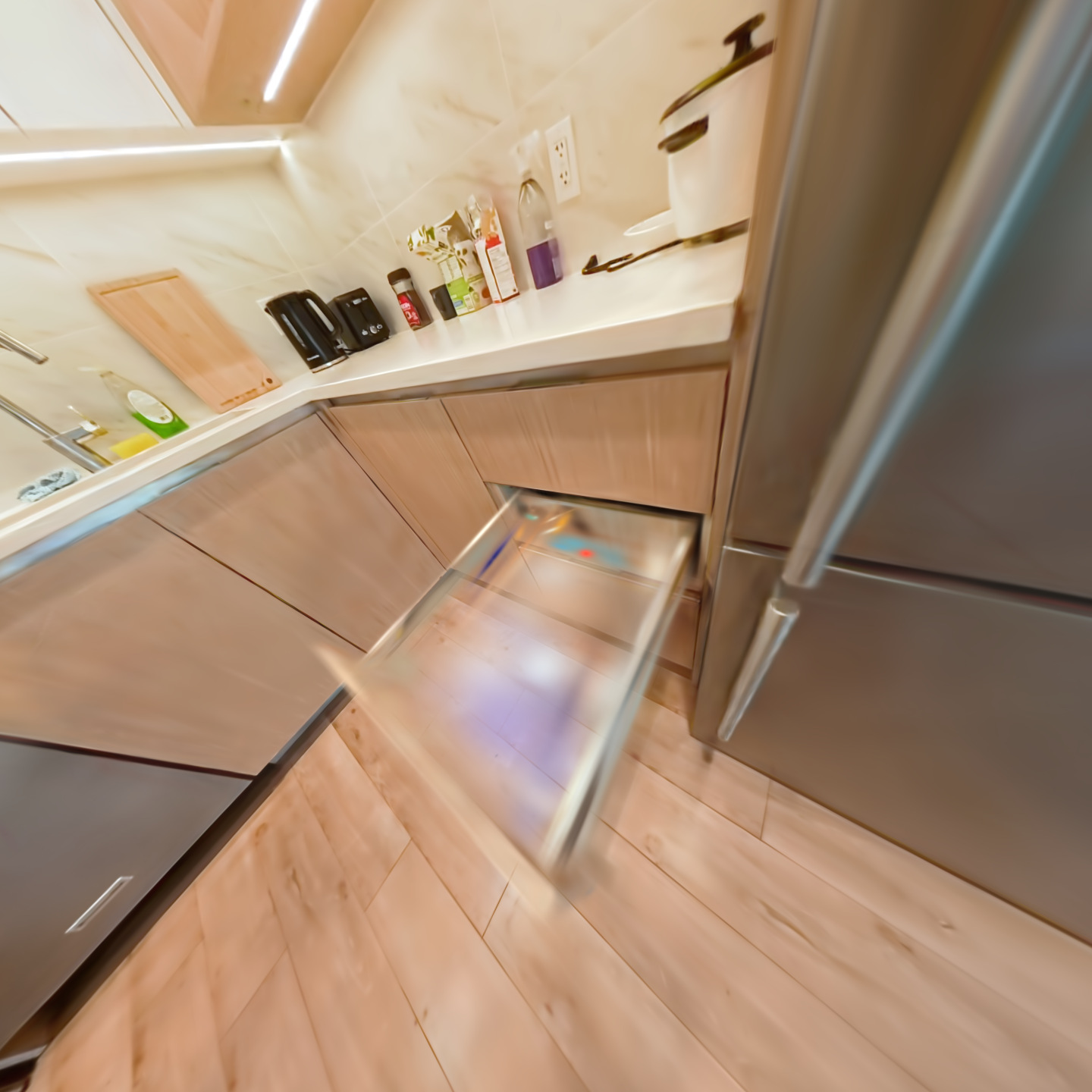} &
        \includegraphics[width=0.18\textwidth]{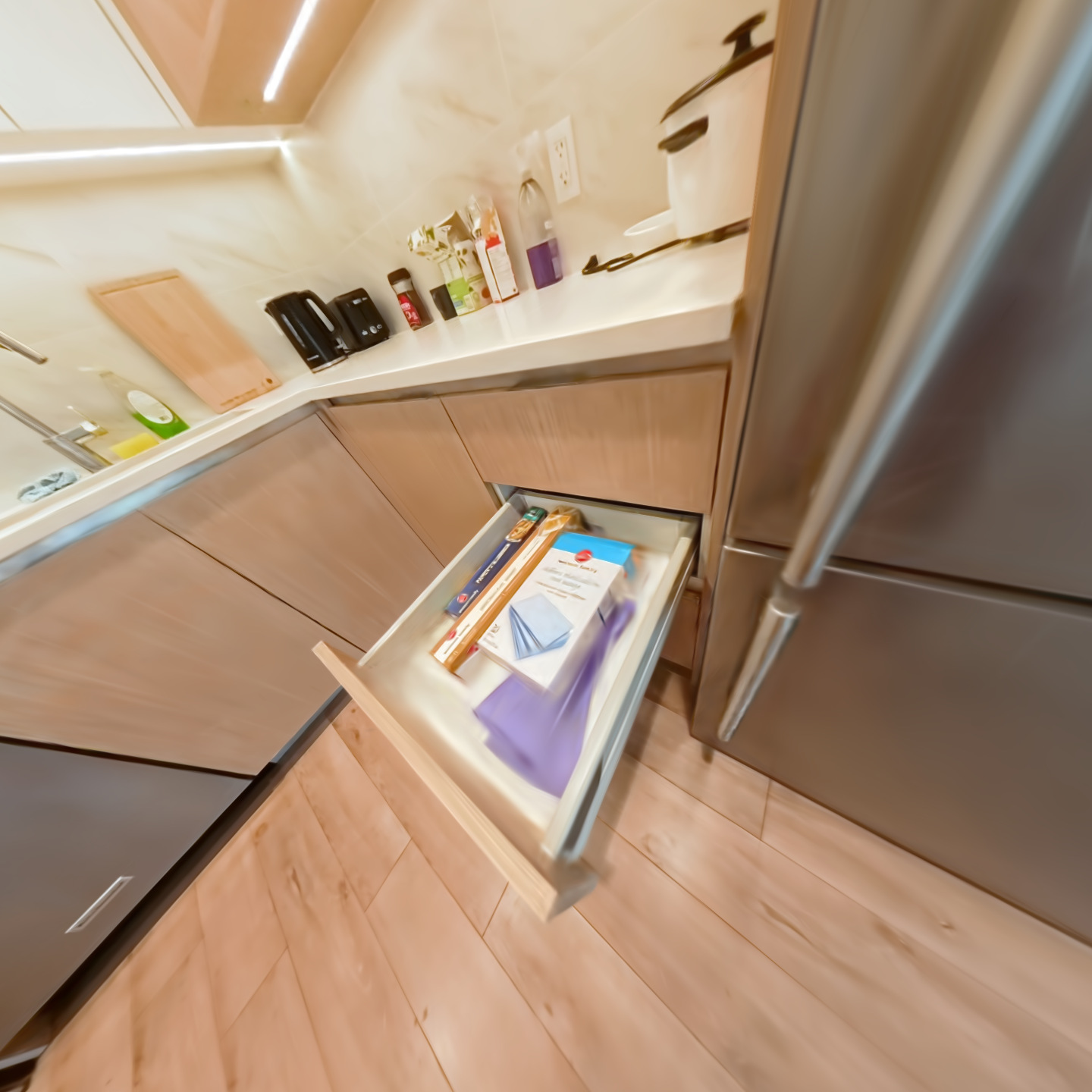} &
        \includegraphics[width=0.18\textwidth]{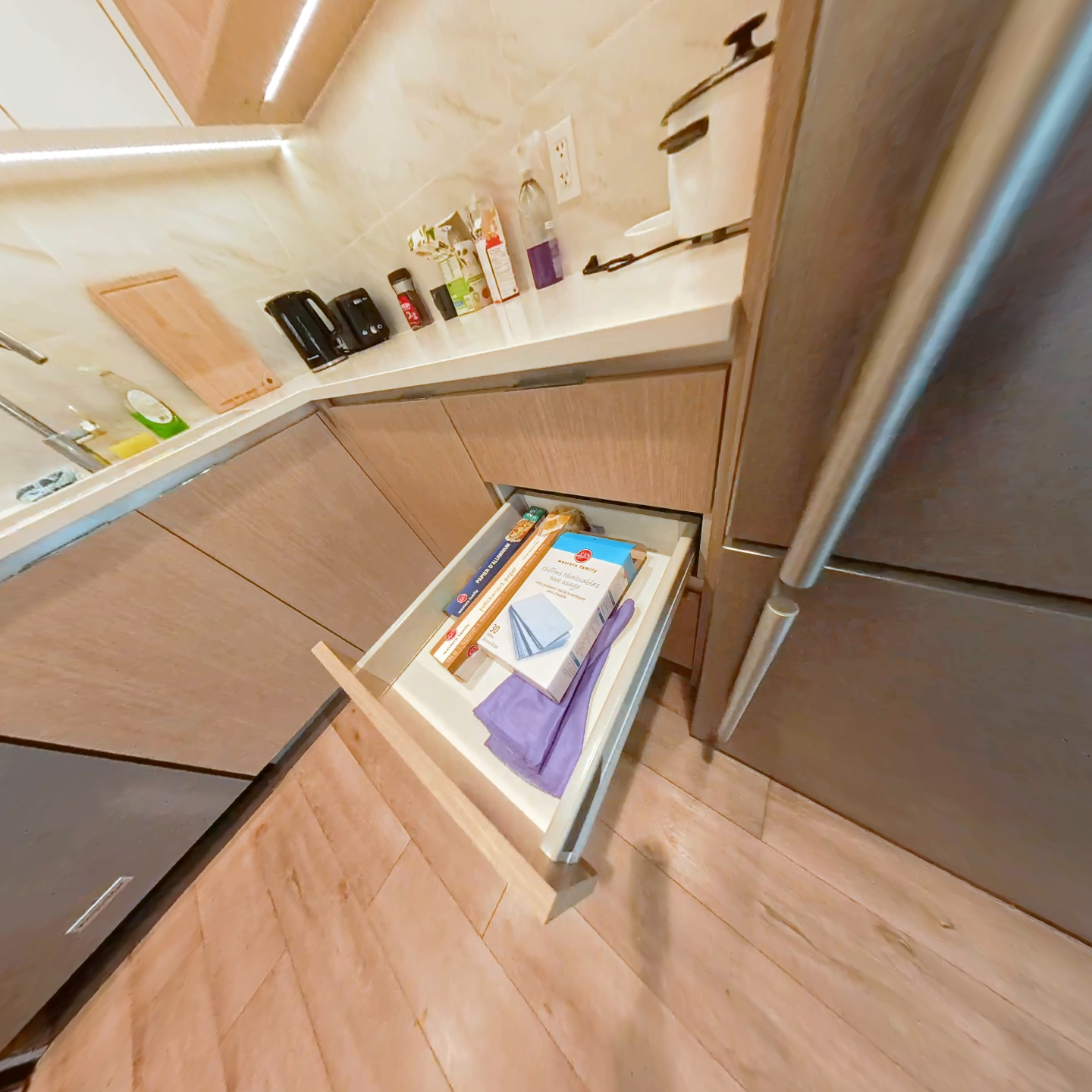} \\[1pt]

        \includegraphics[width=0.18\textwidth]{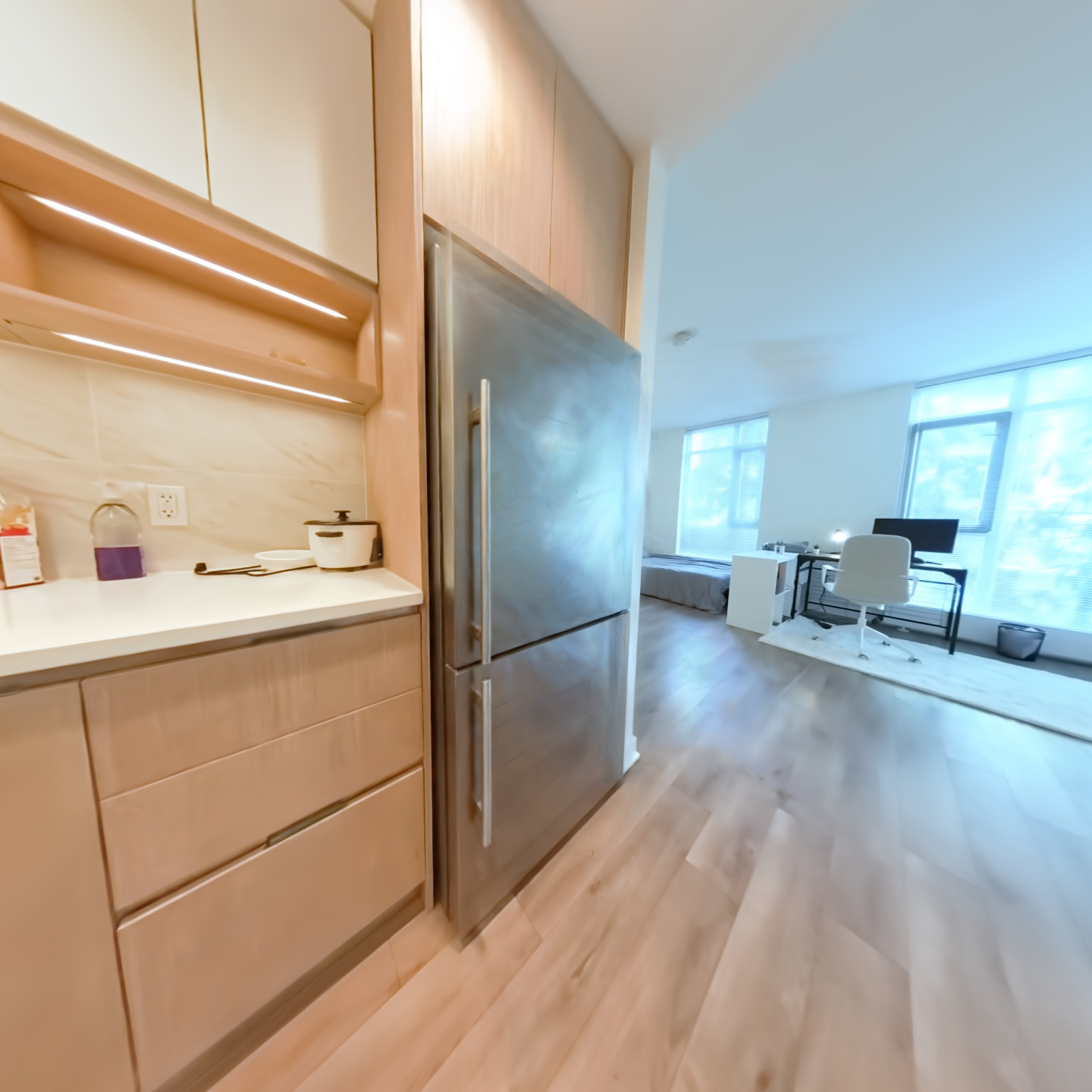} &
        \includegraphics[width=0.18\textwidth]{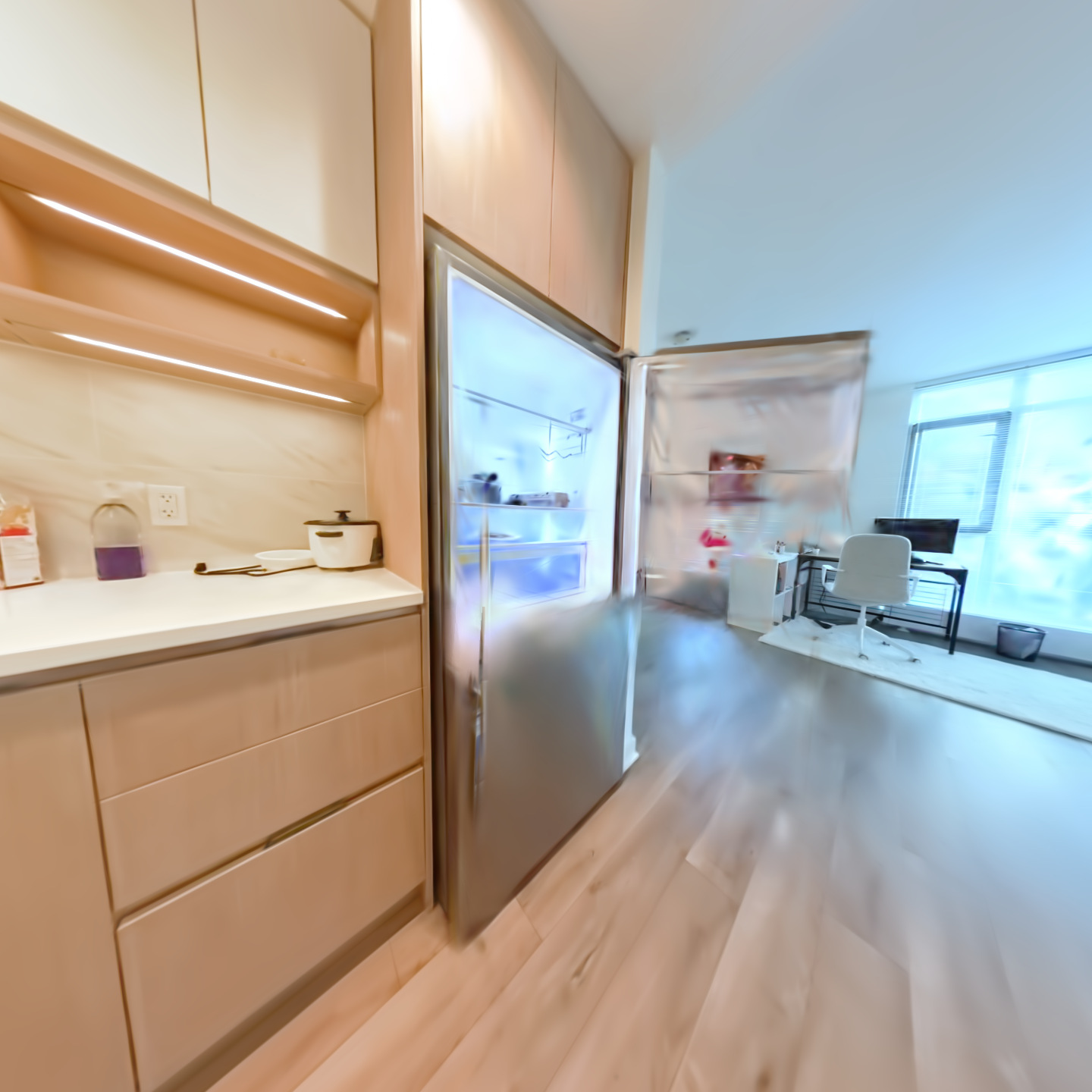} &
        \includegraphics[width=0.18\textwidth]{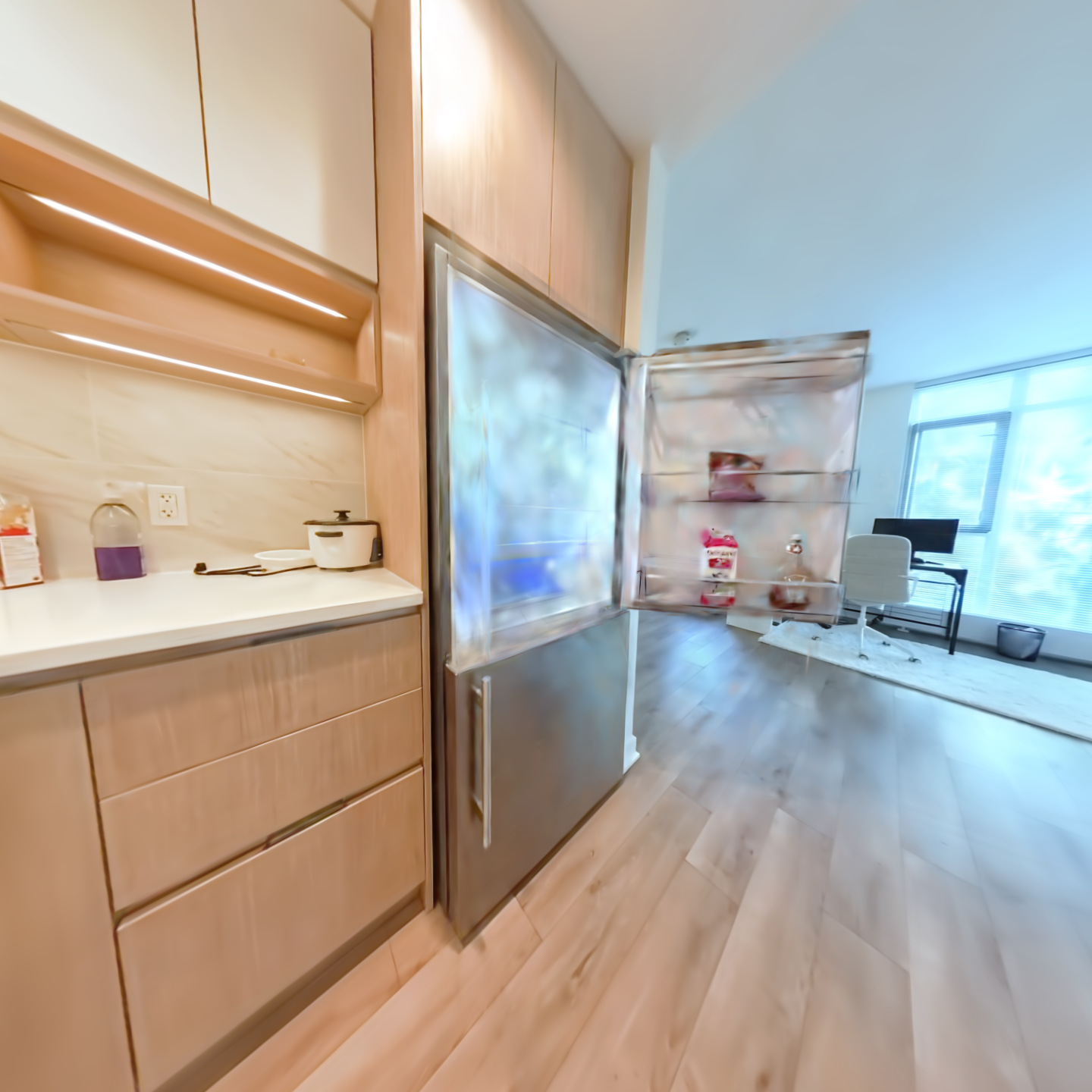} &
        \includegraphics[width=0.18\textwidth]{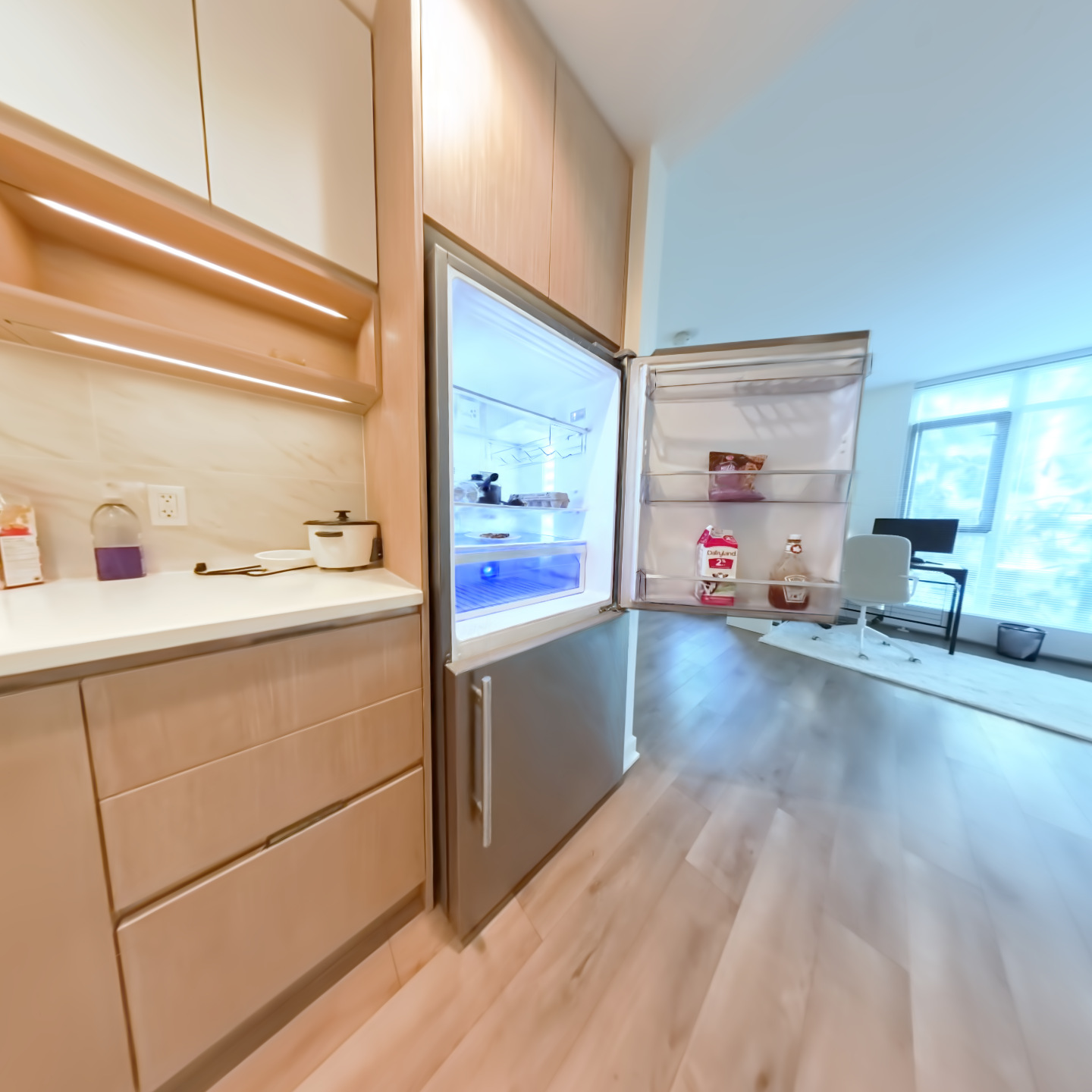} &
        \includegraphics[width=0.18\textwidth]{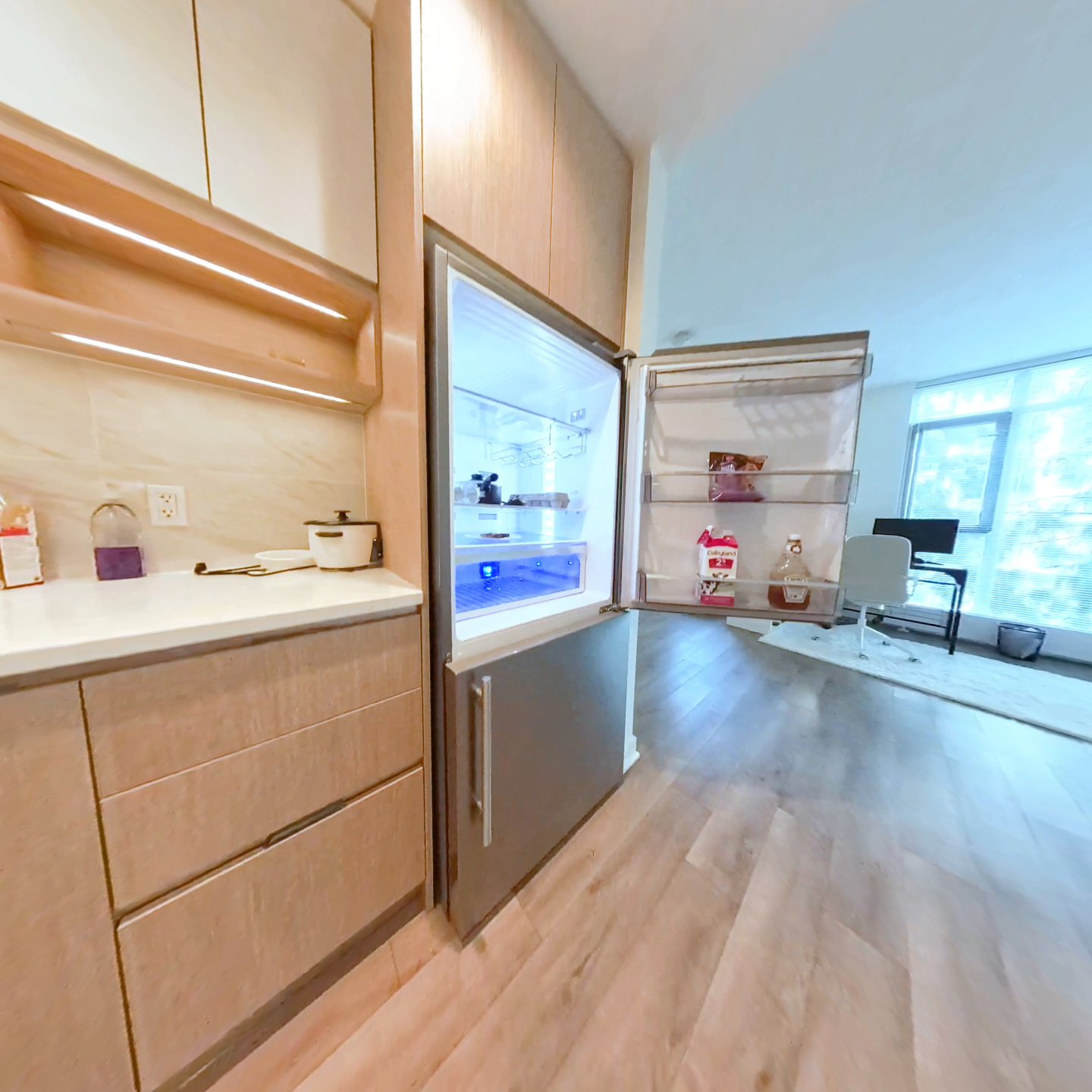} \\[1pt]

        \includegraphics[width=0.18\textwidth]{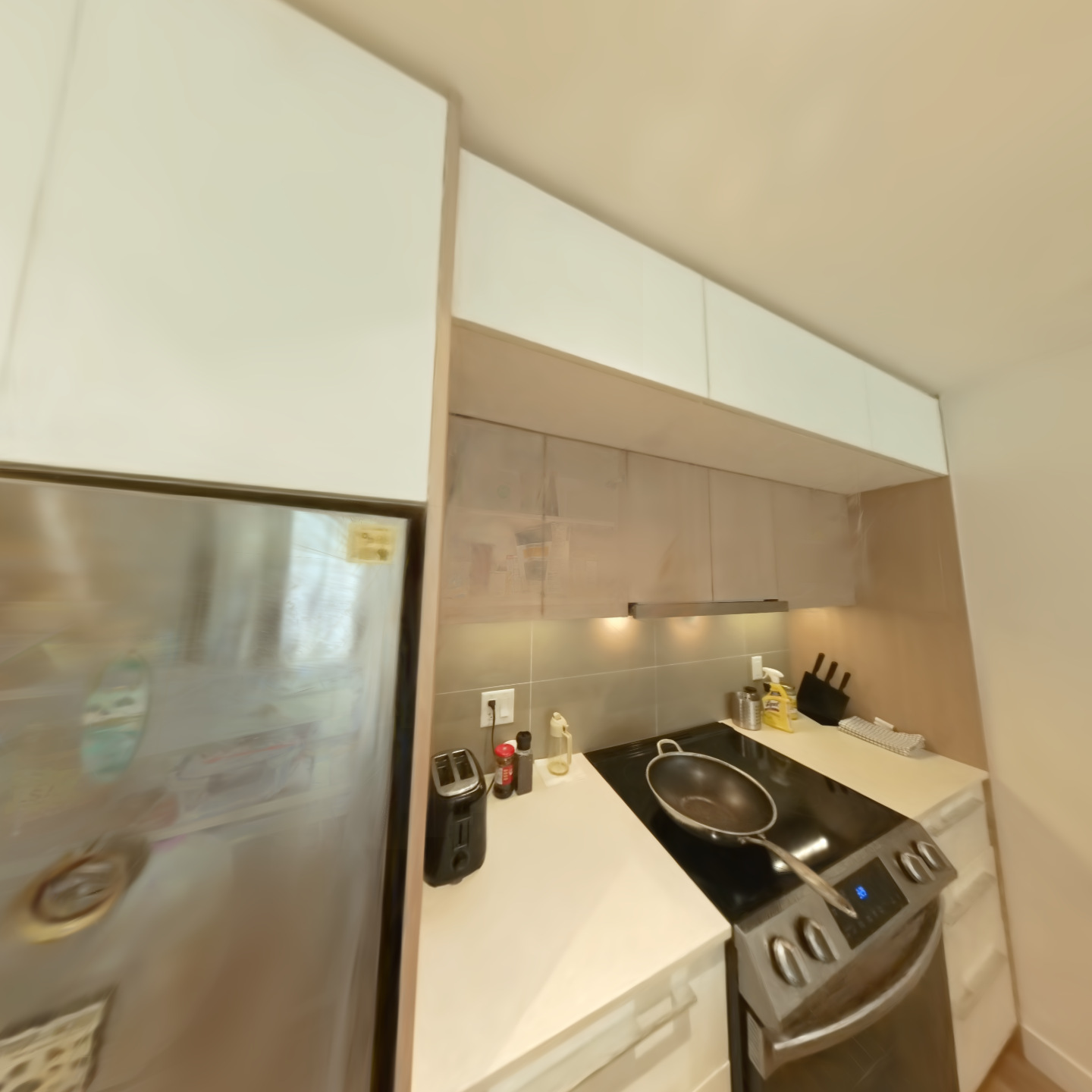} &
        \includegraphics[width=0.18\textwidth]{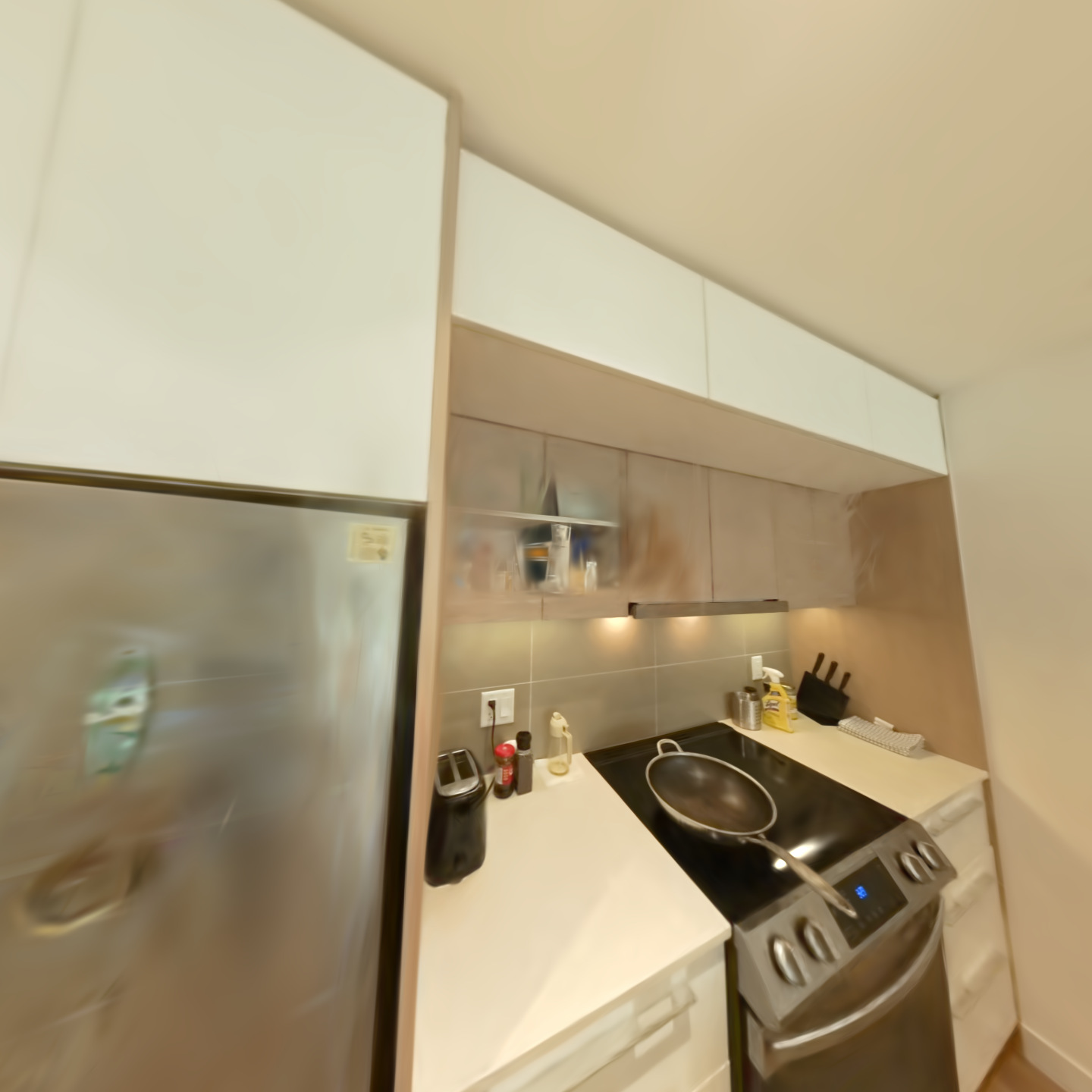} &
        \includegraphics[width=0.18\textwidth]{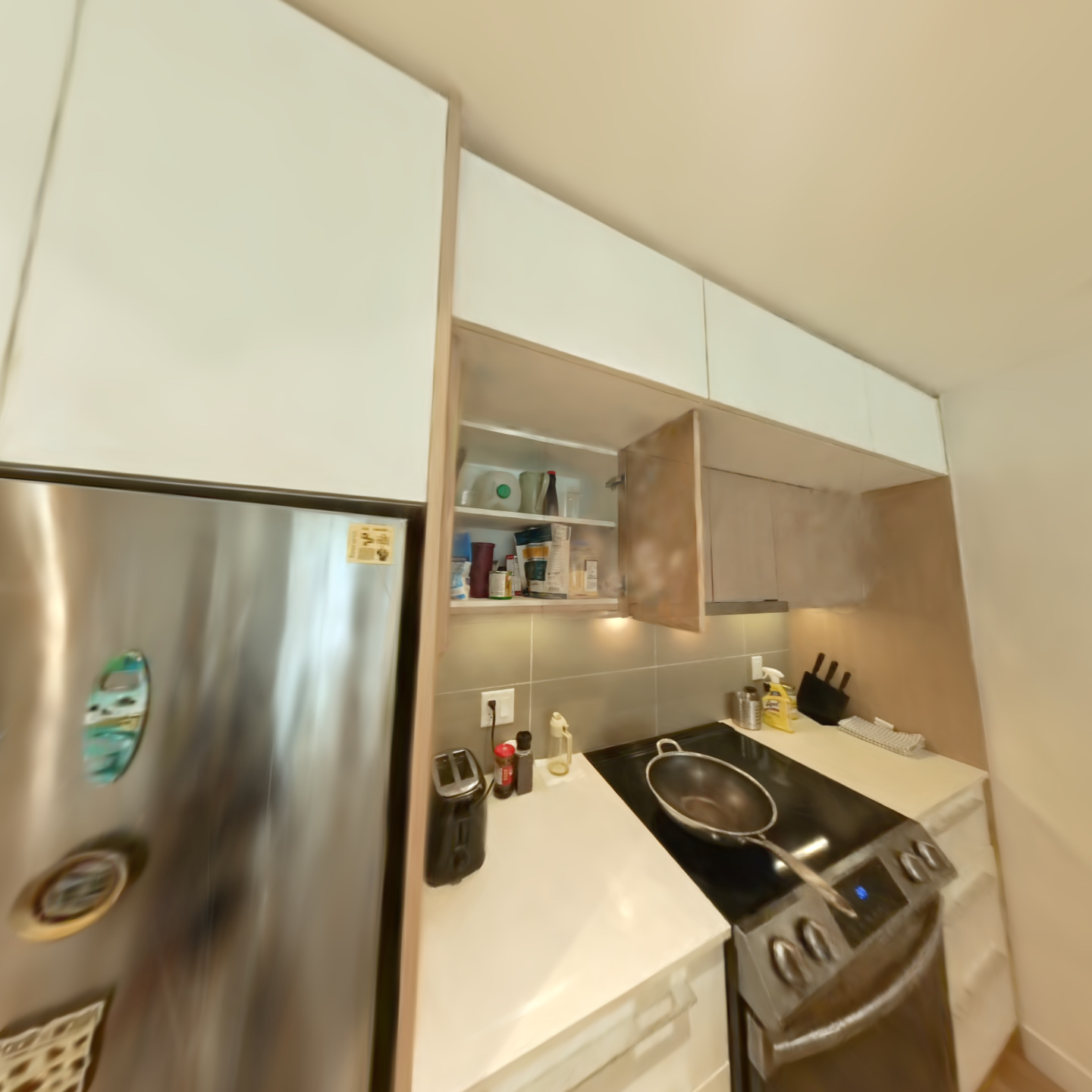} &
        \includegraphics[width=0.18\textwidth]{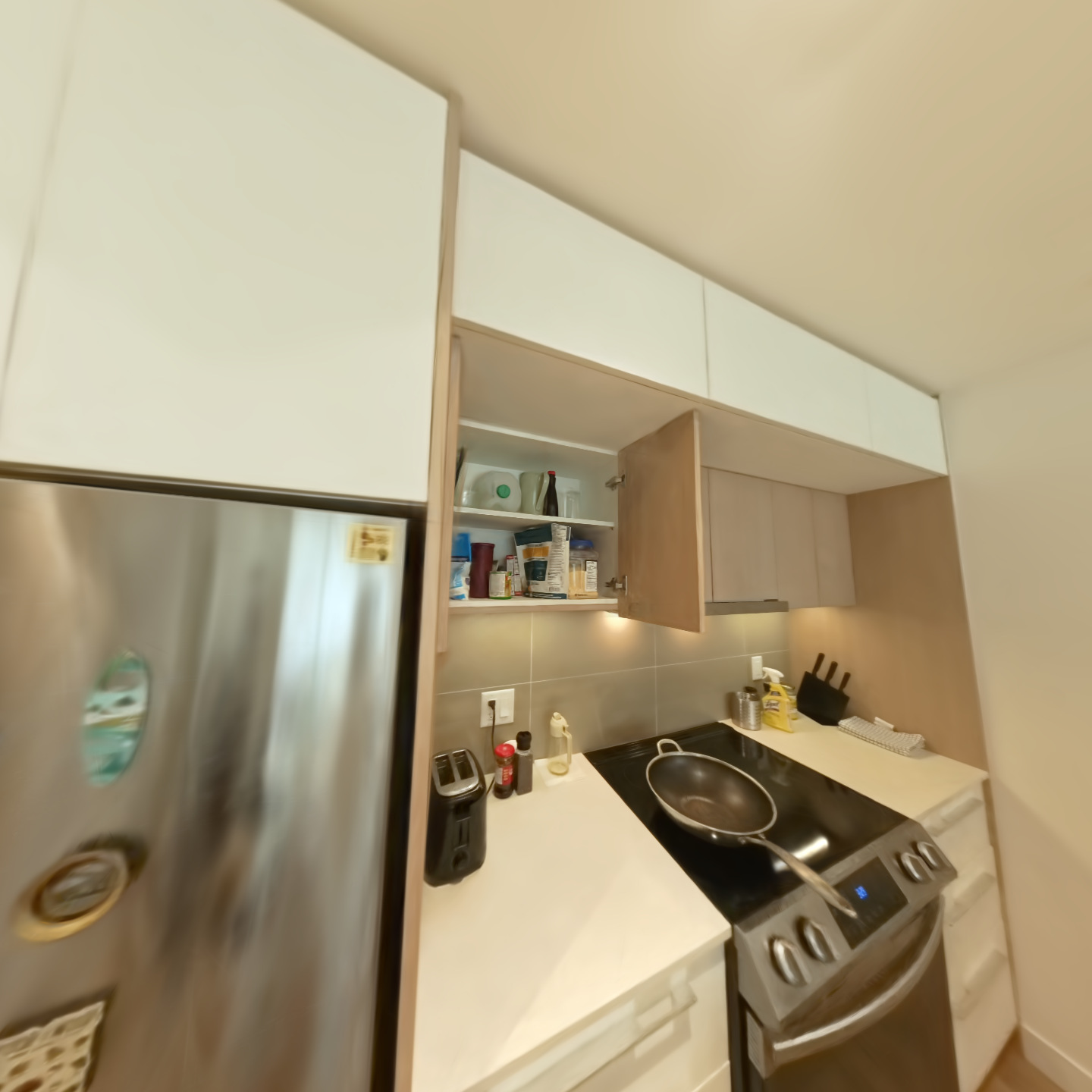} &
        \includegraphics[width=0.18\textwidth]{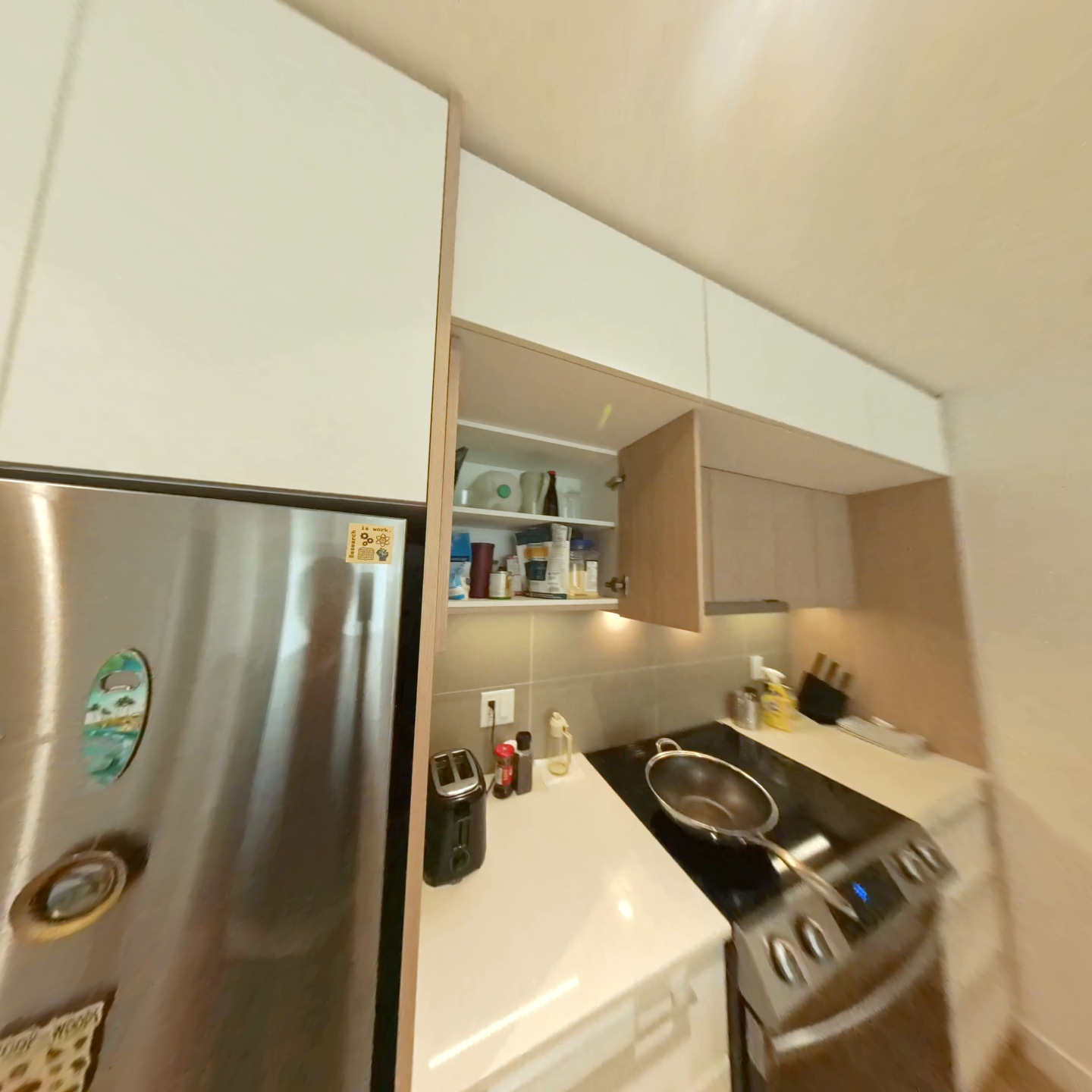} \\[1pt]

        \includegraphics[width=0.18\textwidth]{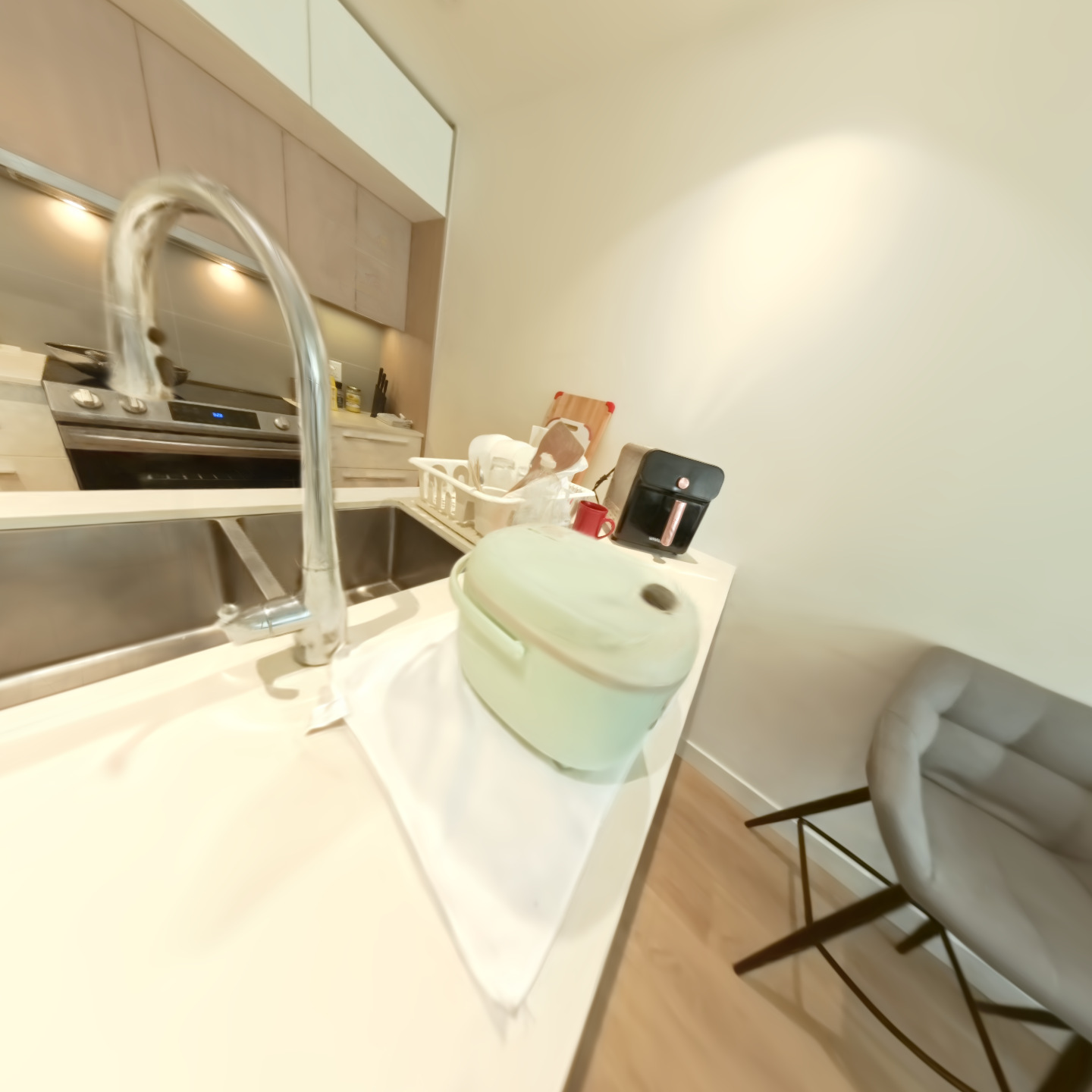} &
        \includegraphics[width=0.18\textwidth]{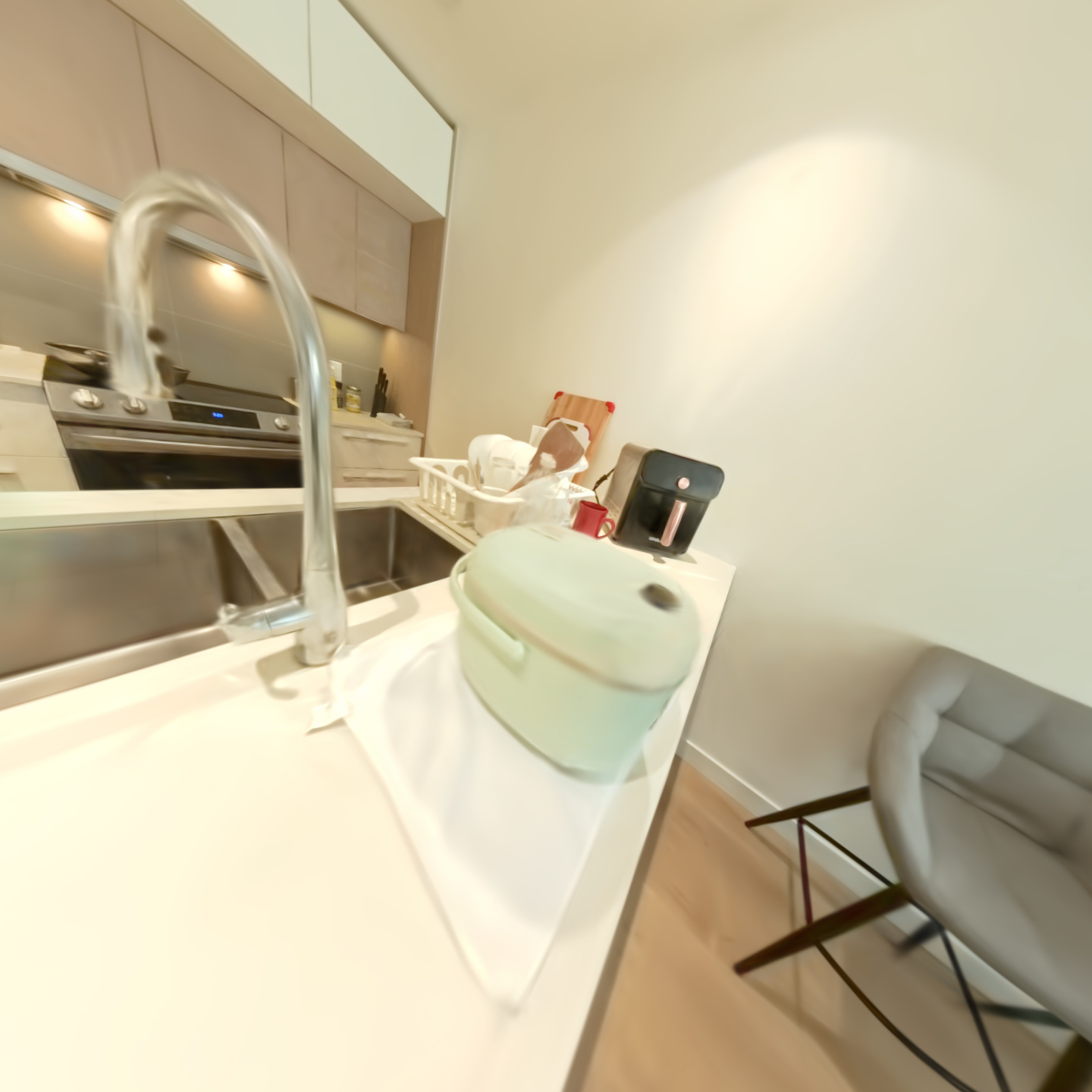} &
        \includegraphics[width=0.18\textwidth]{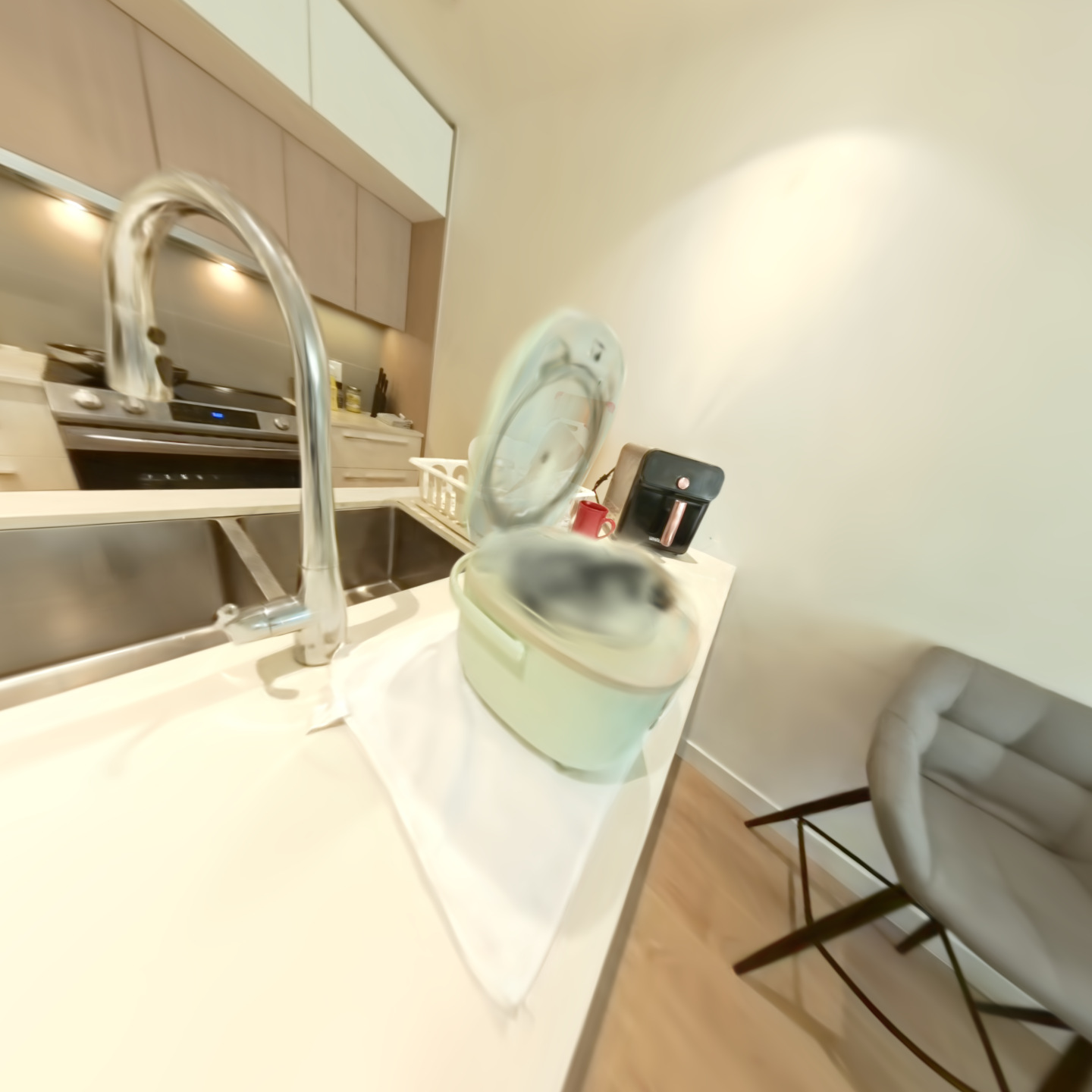} &
        \includegraphics[width=0.18\textwidth]{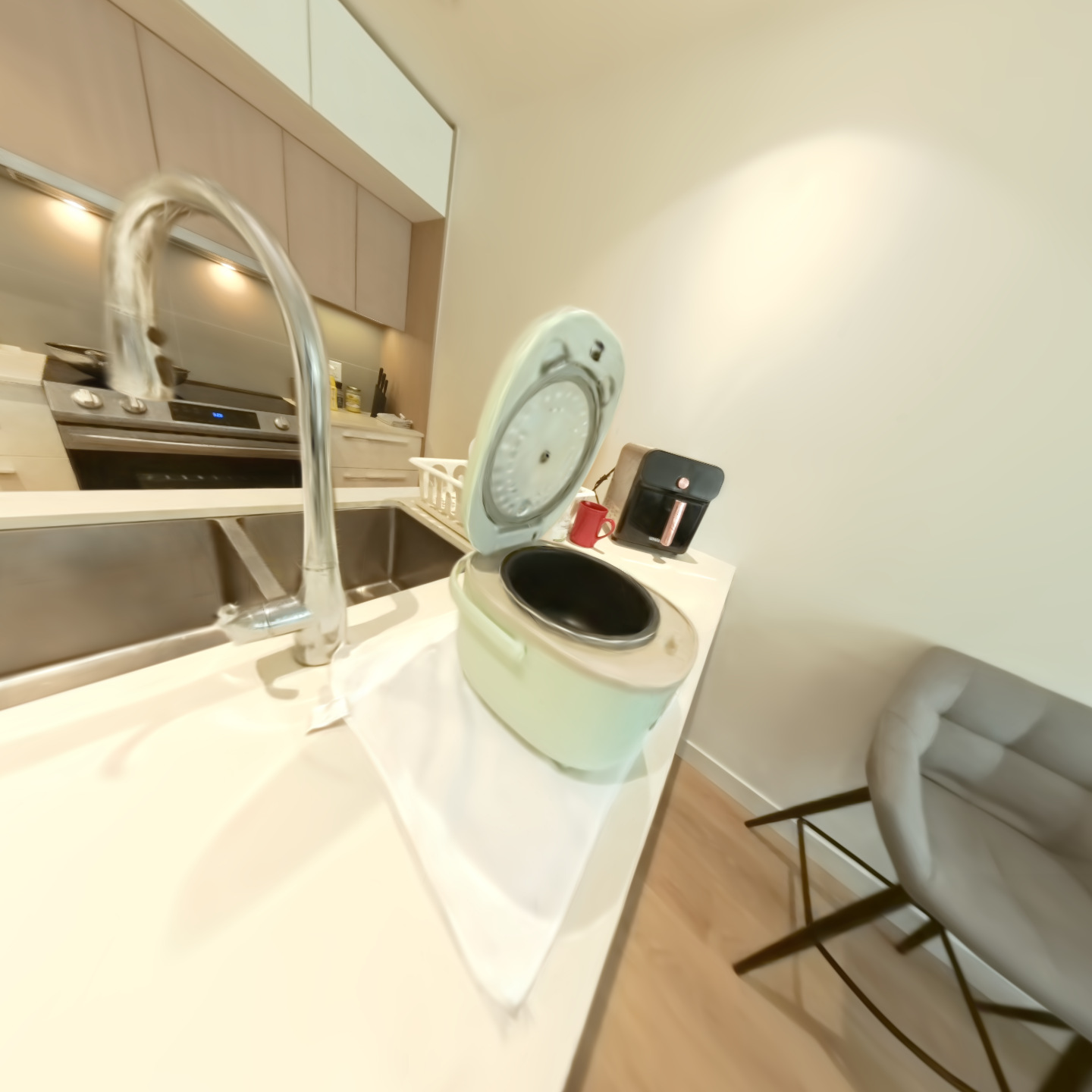} &
        \includegraphics[width=0.18\textwidth]{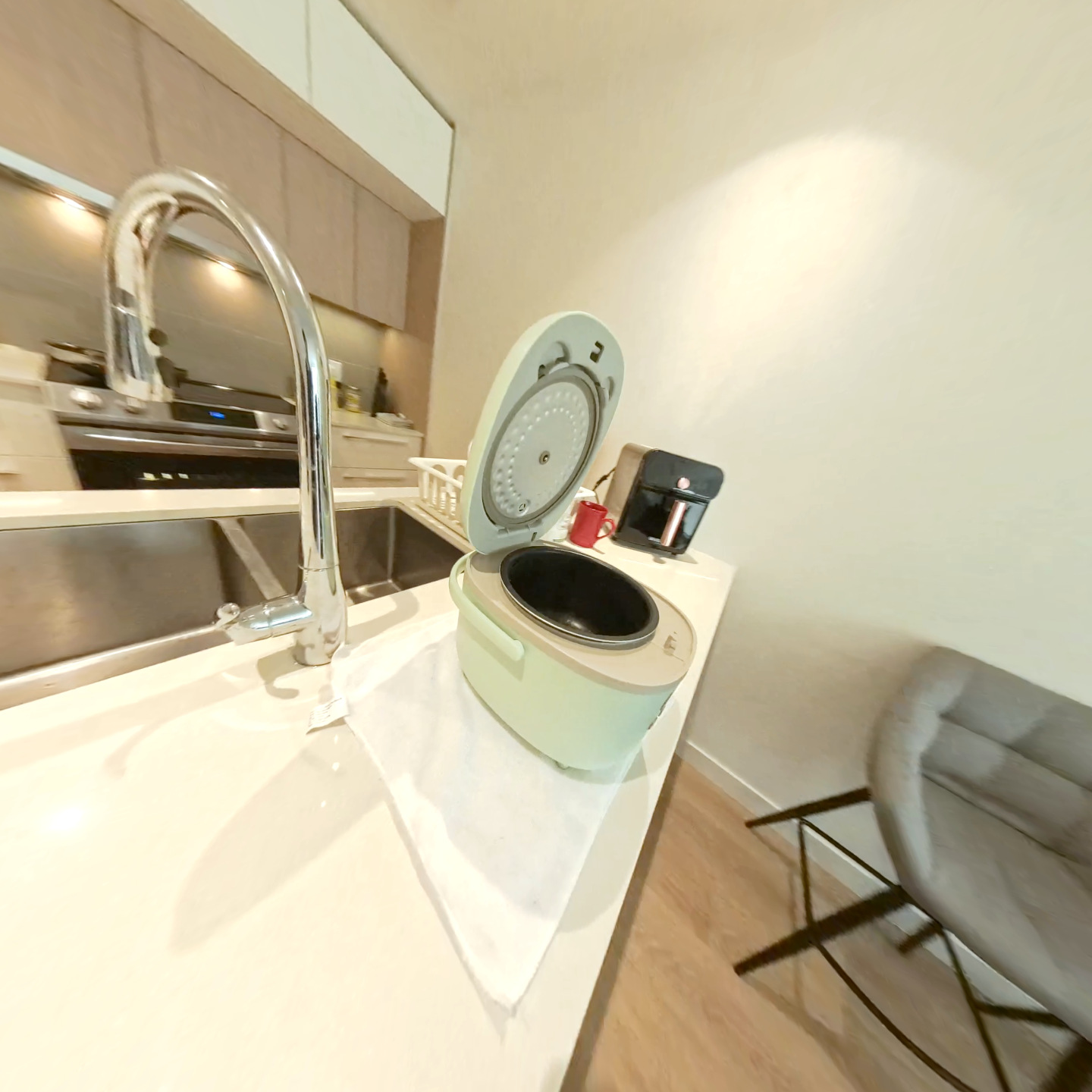} \\[1pt]

        \includegraphics[width=0.18\textwidth]{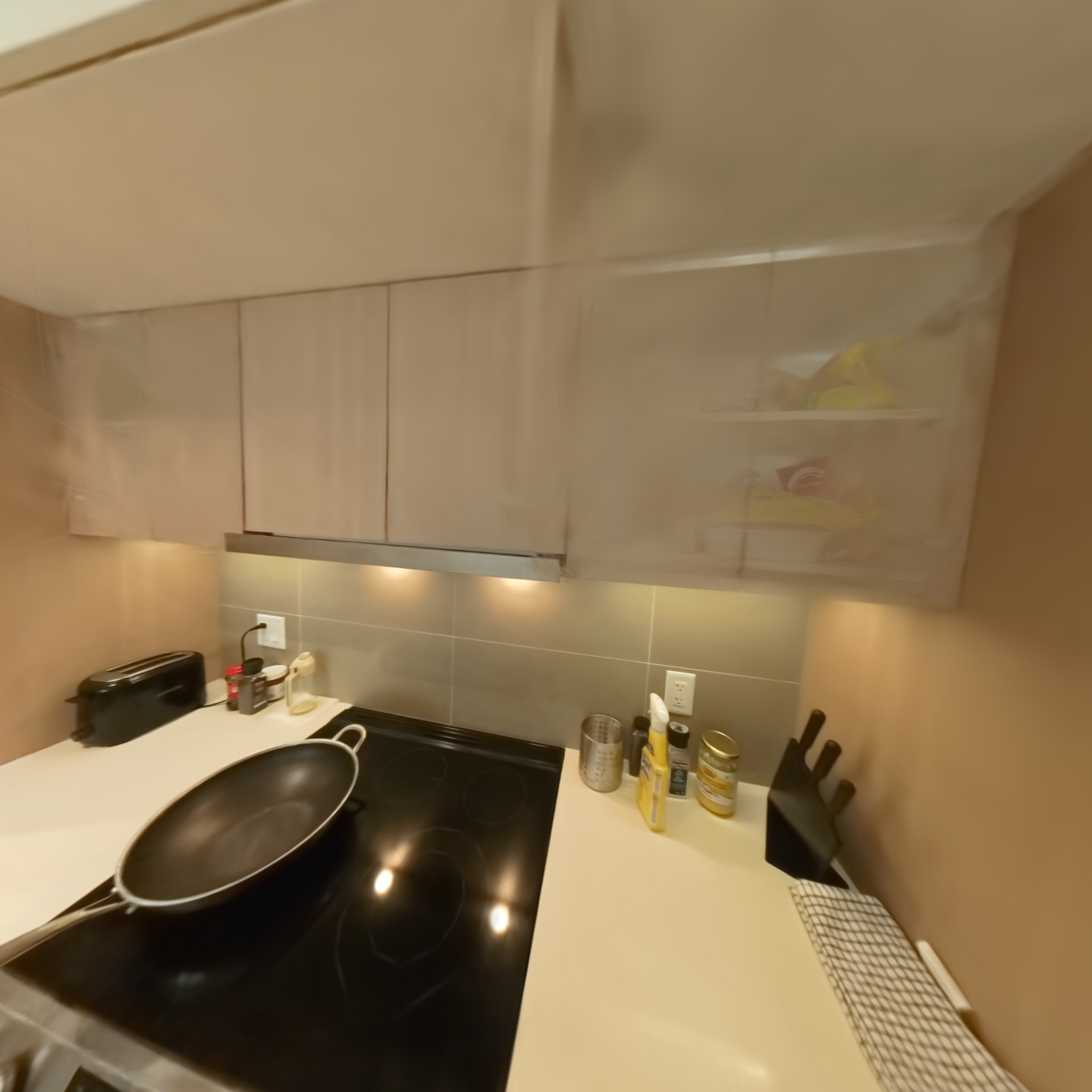} &
        \includegraphics[width=0.18\textwidth]{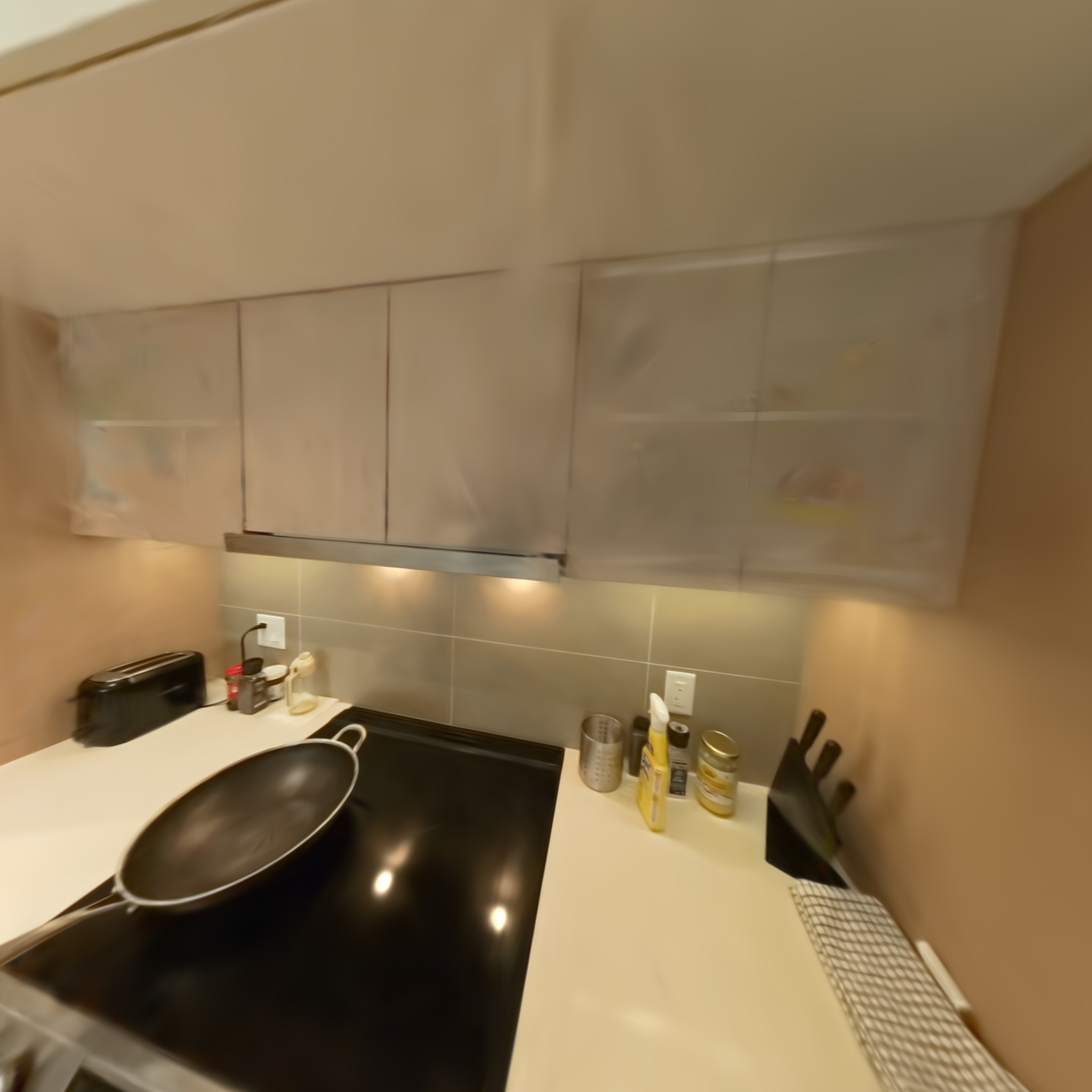} &
        \includegraphics[width=0.18\textwidth]{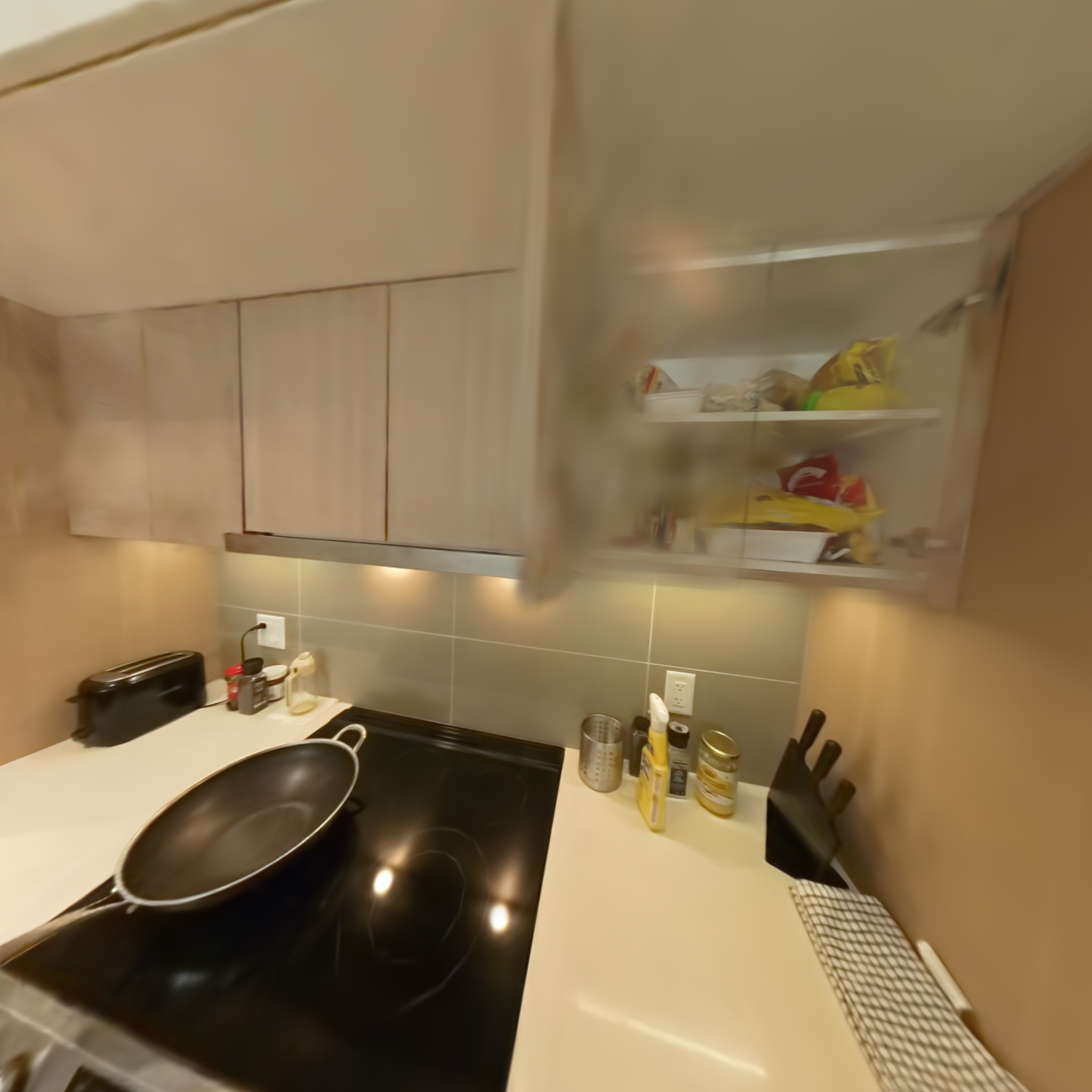} &
        \includegraphics[width=0.18\textwidth]{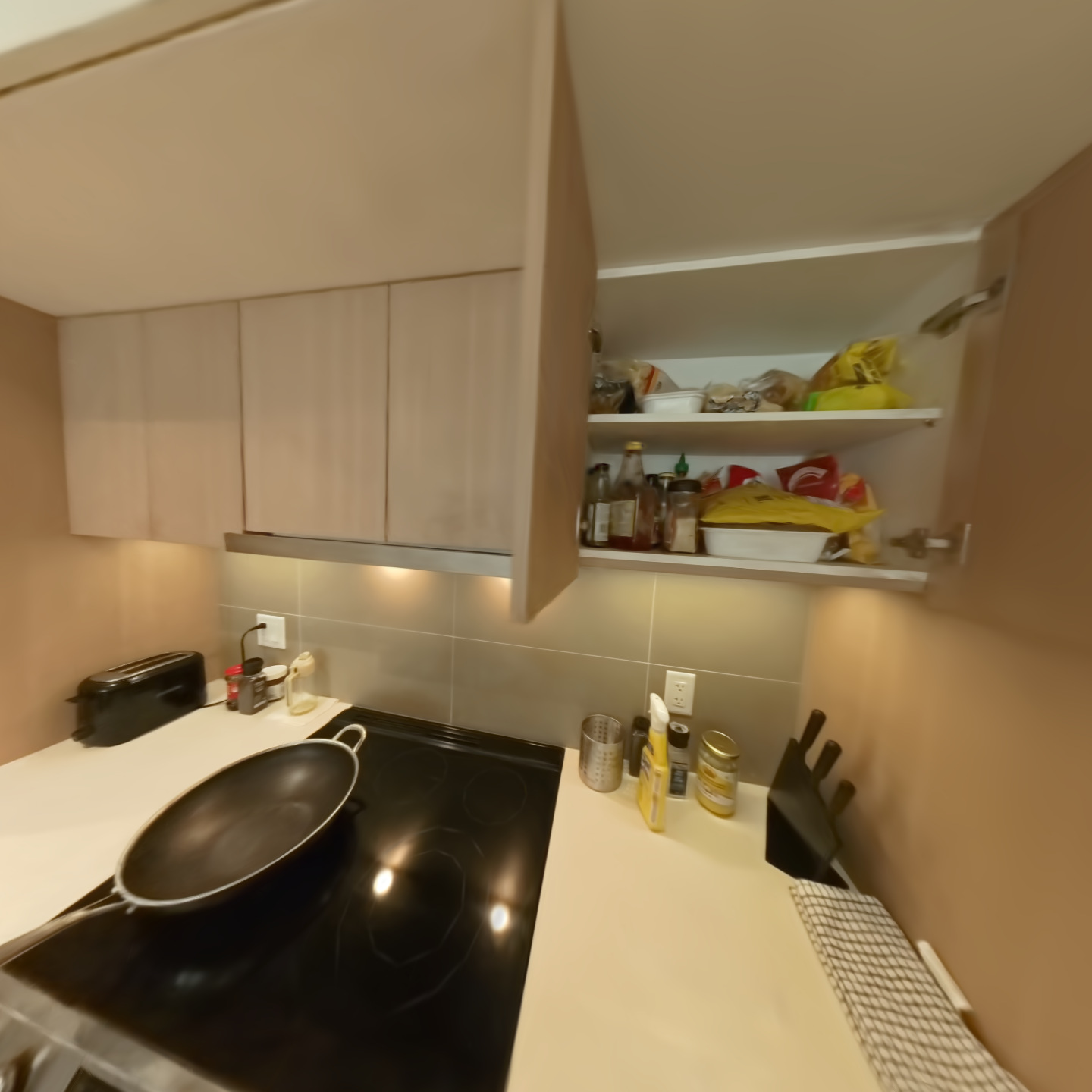} &
        \includegraphics[width=0.18\textwidth]{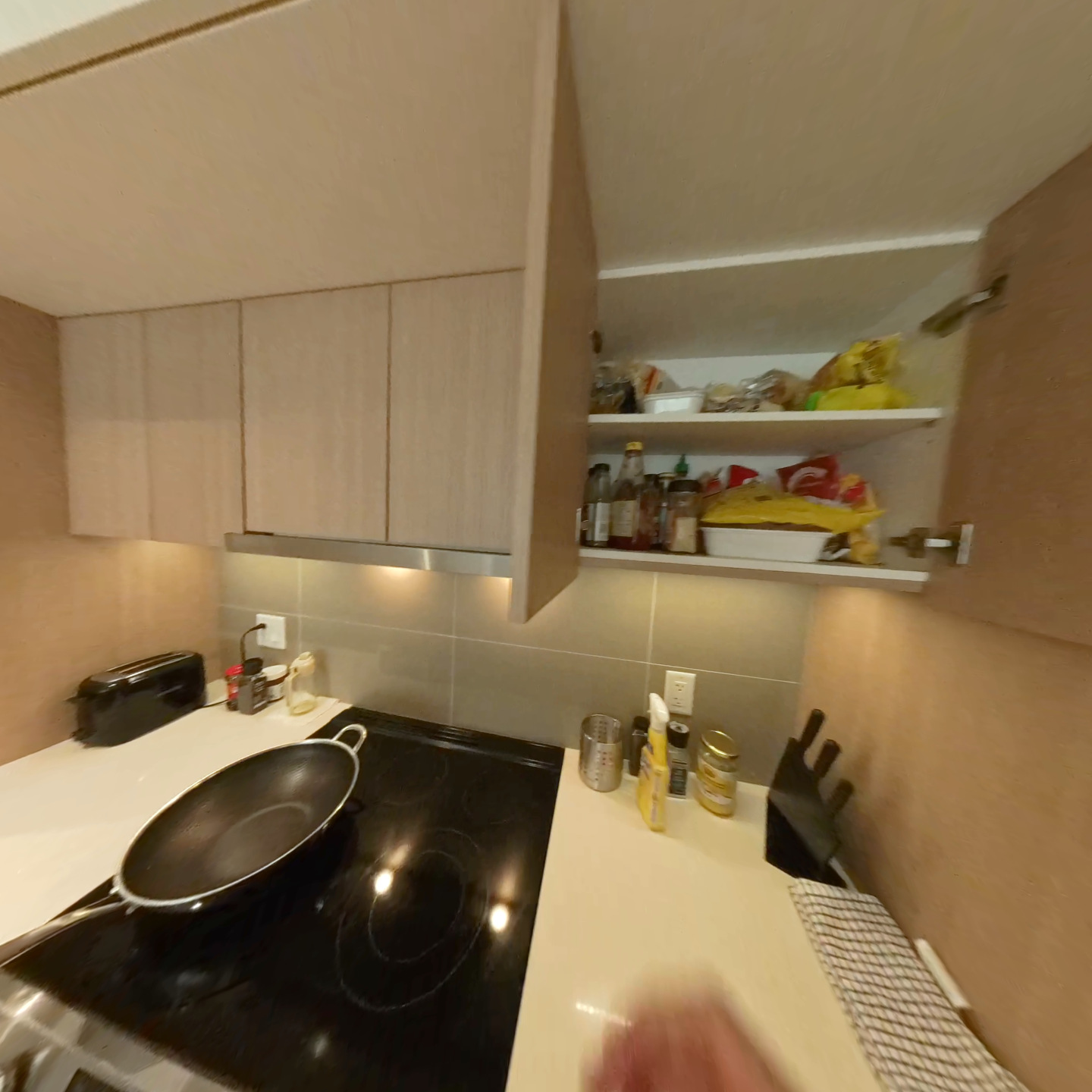} \\[1pt]

        \includegraphics[width=0.18\textwidth]{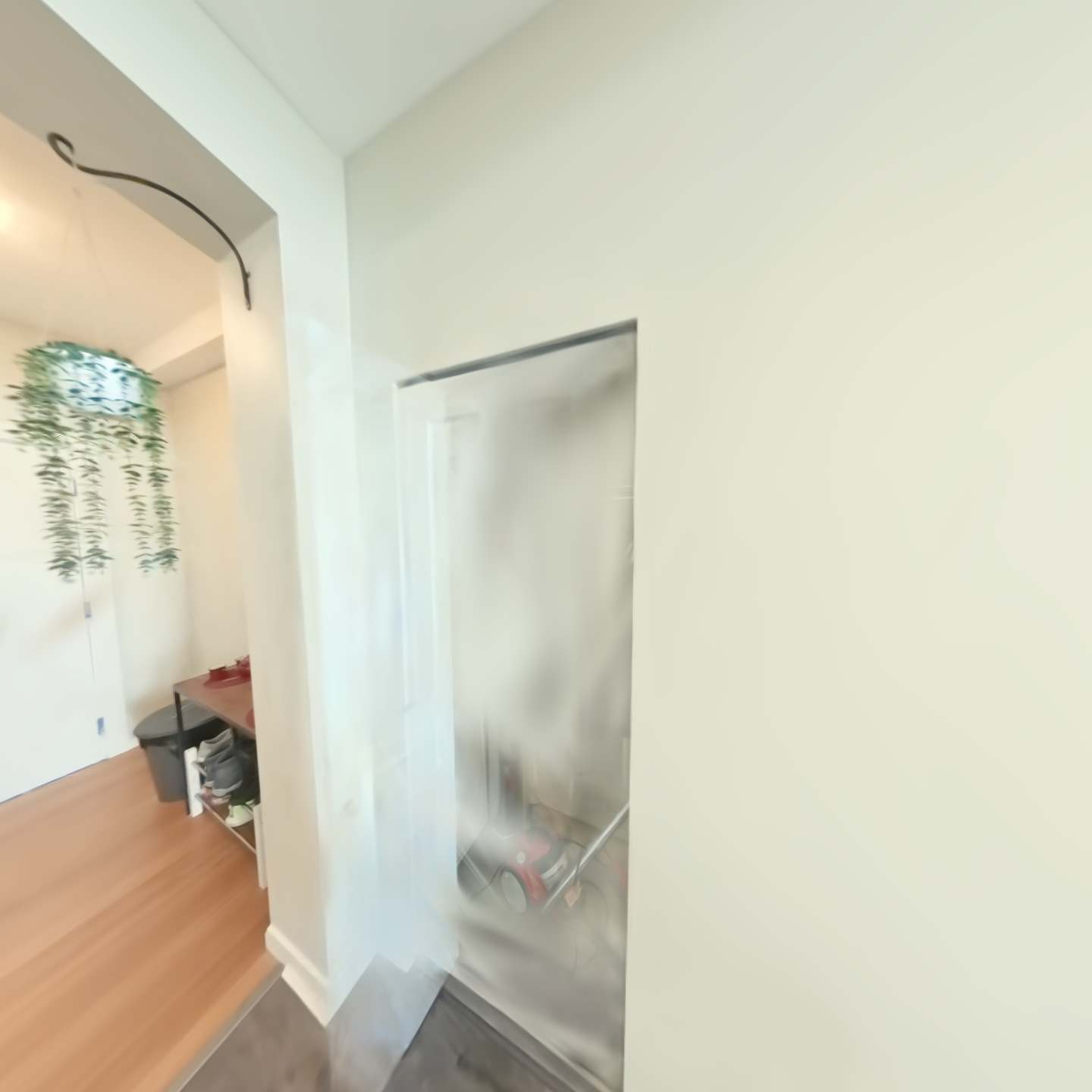} &
        \includegraphics[width=0.18\textwidth]{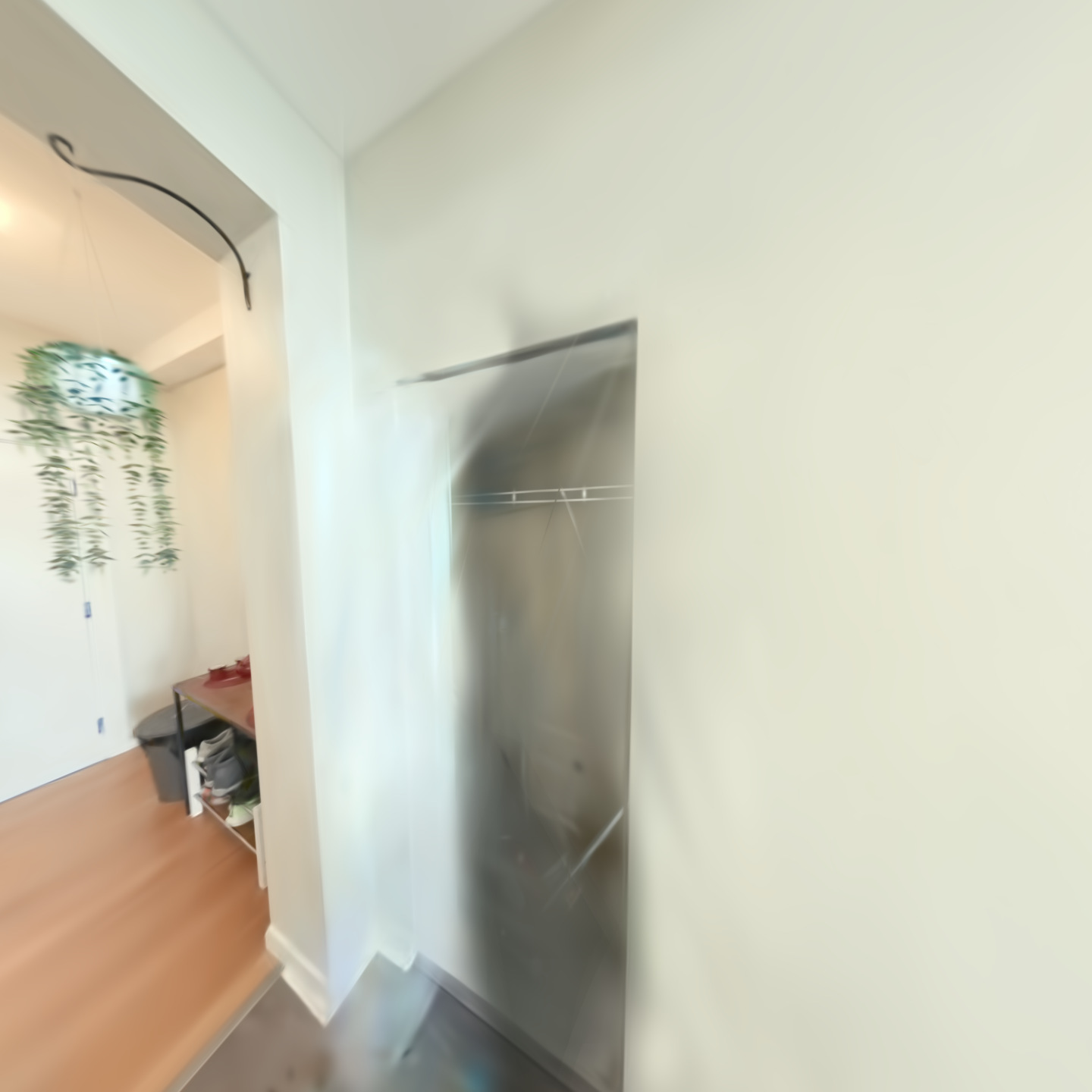} &
        \includegraphics[width=0.18\textwidth]{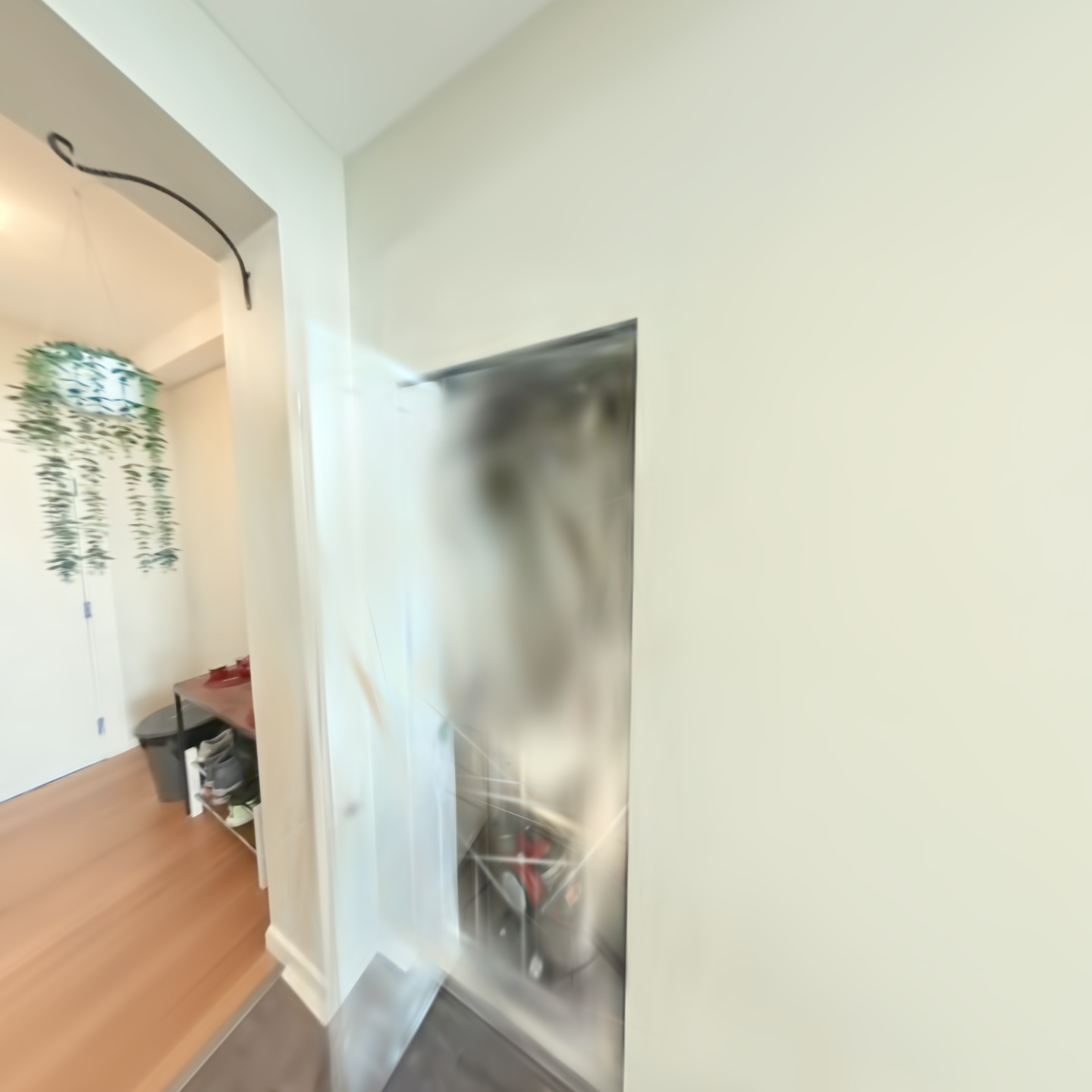} &
        \includegraphics[width=0.18\textwidth]{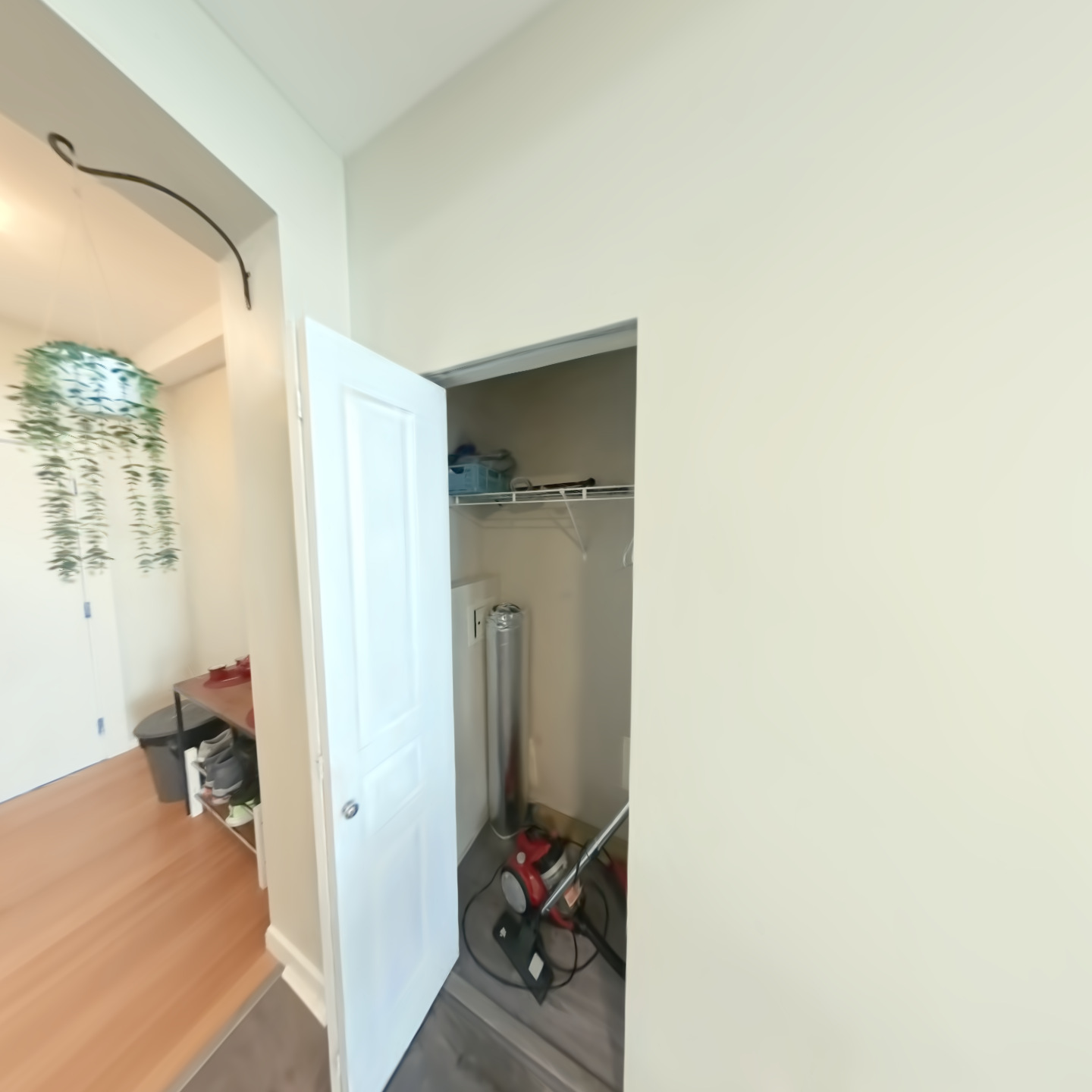} &
        \includegraphics[width=0.18\textwidth]{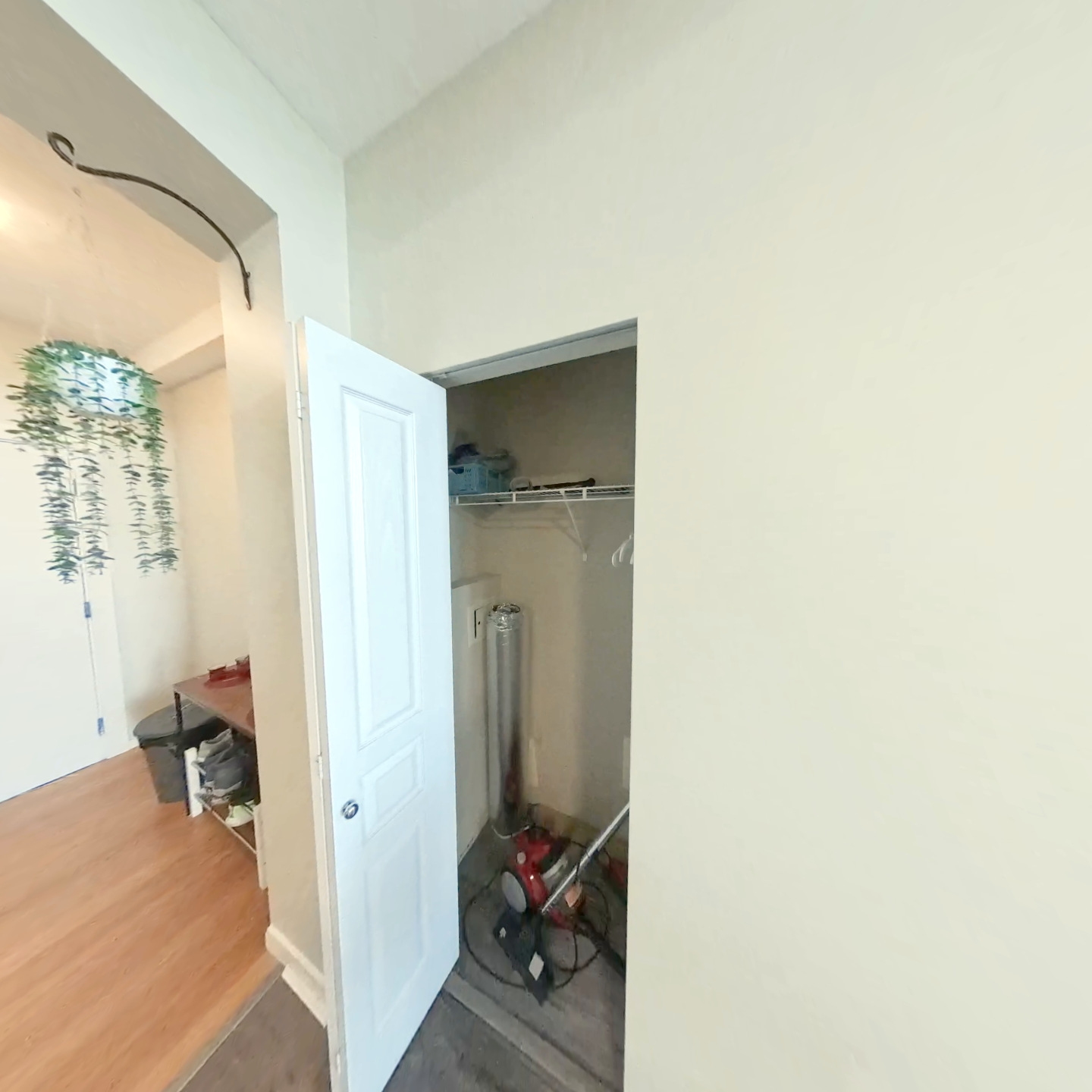}
    \end{tabular}

    \caption{ \textbf{Qualitative comparison.}
We compare FullCircle, DeGauss, 4DGS, and SwitchSplat against ground-truth images from standard held-out test views. Scenes are shown from top to bottom: Flat2, Flat2, Kitchen 1, Kitchen 1, Kitchen 1, Flat 3.}
    \label{fig:comparison5}
\end{figure*}

 \end{document}